\def\paperlanguage{} 

\documentclass{ieeeaccess}

\usepackage{tabularx}

\usepackage{hyperref}
\usepackage{breakurl}
\hypersetup{
    colorlinks=true,
    linkcolor=black,
    citecolor=black,
    urlcolor=black,
    bookmarks=true,
    bookmarksnumbered=true
}

\newcommand{\secref}[1]{Section~\ref{#1}}

\newcommand{\figref}[1]{{Fig.~\ref{#1}}}

\newcommand{\switchlanguage}[2]{%
  \ifx\paperlanguage\empty%
  #1%
  \else%
  #2%
  \fi%
}

\usepackage{amsthm}
\newtheorem{definition}{Definition}[section]

\newcommand{\ctext}[1]{\raise0.2ex\hbox{\textcircled{\scriptsize{#1}}}}

\graphicspath{{figs/}{./figs/}{figures/}{./figures/}}

\usepackage{cite}
\usepackage{amsmath,amssymb,amsfonts}
\usepackage{algorithmic}
\usepackage{graphicx}
\usepackage{textcomp}

\usepackage{bm}
\makeatletter
\AtBeginDocument{\DeclareMathVersion{bold}
\SetSymbolFont{operators}{bold}{T1}{times}{b}{n}
\SetSymbolFont{NewLetters}{bold}{T1}{times}{b}{it}
\SetMathAlphabet{\mathrm}{bold}{T1}{times}{b}{n}
\SetMathAlphabet{\mathit}{bold}{T1}{times}{b}{it}
\SetMathAlphabet{\mathbf}{bold}{T1}{times}{b}{n}
\SetMathAlphabet{\mathtt}{bold}{OT1}{pcr}{b}{n}
\SetSymbolFont{symbols}{bold}{OMS}{cmsy}{b}{n}
\renewcommand\boldmath{\@nomath\boldmath\mathversion{bold}}}
\makeatother

\def\BibTeX{{\rm B\kern-.05em{\sc i\kern-.025em b}\kern-.08em
    T\kern-.1667em\lower.7ex\hbox{E}\kern-.125emX}}

\begin{document}
\history{Date of publication xxxx 00, 0000, date of current version xxxx 00, 0000.}
\doi{10.1109/ACCESS.2024.0429000}
\bstctlcite{BSTcontrol}

\title{Vision-Language-Action Models for Robotics: A Review Towards Real-World Applications}

\author{
  Kento Kawaharazuka\authorrefmark{1},
  Jihoon Oh\authorrefmark{1},
  Jun Yamada\authorrefmark{2},
  Ingmar Posner\authorrefmark{2}\authorrefmark{*},
  and Yuke Zhu\authorrefmark{3}\authorrefmark{*}
}

\address[1]{Department of Mechano-Informatics, The University of Tokyo, Bunkyo-ku, Tokyo 113-8656 JAPAN (e-mail: [kawaharazuka, oh]@jsk.imi.i.u-tokyo.ac.jp)}
\address[2]{Department of Engineering Science, University of Oxford, Parks Road, Oxford OX13PJ UK (e-mail: [jyamada, ingmar]@robots.ox.ac.uk)}
\address[3]{Department of Computer Science, The University of Texas at Austin, Austin, TX 78712 USA (e-mai: yukez@cs.utexas.edu)}
\address[*]{Equal advising}
\tfootnote{This research was partially supported by JST CRONOS under Grant Number JPMJCS24K6.}

\markboth
{K. Kawaharazuka \headeretal: Vision-Language-Action Models for Robotics}
{K. Kawaharazuka \headeretal: Vision-Language-Action Models for Robotics}

\corresp{Corresponding author: Kento Kawaharazuka (e-mail: kawaharazuka@jsk.imi.i.u-tokyo.ac.jp).}

\begin{abstract}
\switchlanguage%
{%
  Amid growing efforts to leverage advances in large language models (LLMs) and vision-language models (VLMs) for robotics, Vision-Language-Action (VLA) models have recently gained significant attention.
  By unifying vision, language, and action data at scale, which have traditionally been studied separately, VLA models aim to learn policies that generalise across diverse tasks, objects, embodiments, and environments.
  This generalisation capability is expected to enable robots to solve novel downstream tasks with minimal or no additional task-specific data, facilitating more flexible and scalable real-world deployment.
  Unlike previous surveys that focus narrowly on action representations or high-level model architectures, this work offers a comprehensive, full-stack review, integrating both software and hardware components of VLA systems.
  In particular, this paper provides a systematic review of VLAs, covering their strategy and architectural transition, architectures and building blocks, modality-specific processing techniques, and learning paradigms.
  In addition, to support the deployment of VLAs in real-world robotic applications, we also review commonly used robot platforms, data collection strategies, publicly available datasets, data augmentation methods, and evaluation benchmarks.
  Throughout this comprehensive survey, this paper aims to offer practical guidance for the robotics community in applying VLAs to real-world robotic systems.
  All references categorized by training approach, evaluation method, modality, and dataset are available in the table on our project website: \href{https://vla-survey.github.io}{\textcolor{magenta}{https://vla-survey.github.io}}.

}%
{%
  大規模言語モデルや大規模視覚言語モデルの発展をロボットに活用した取り組みとして, 現在Vision-Language-Action Model (VLA)が注目を集めている．
  これまで別々に研究されてきた視覚・言語・ロボットのアクションの全てを統合的に学習することで, より汎用的なロボットのポリシーを構築することが期待されている．
  本サーベイでは, このVLAについて, その歴史, アーキテクチャ, 各モダリティの扱い, 学習方法について体系的にまとめる.
  また, より実ロボットへの応用を見据え, 用いられているロボットやデータ収集方法, 現在のデータセット, データ拡張, 評価方法についてもまとめる.
  本サーベイがVLAを実ロボットに応用する際の一つの指針となることを期待する.
}%
\end{abstract}



\titlepgskip=-21pt

\maketitle

\section{Introduction}\label{sec:introduction}
The recent success in developing a variety of large language models (LLMs)~\cite{touvron2023llama2, openai2024gpt4technicalreport} and large vision-language models (VLMs)~\cite{li2023blip2, openai2024gpt4ocard} has catalised remarkable advances in natural language processing and computer vision, fundamentally transforming both fields.
These advancements have also been extended to the field of robotics, where LLMs and VLMs are leveraged to interpret multimodal inputs, reason about tasks, and perform context-aware actions, thereby laying the groundwork for more generalisable and scalable robotic systems~\cite{hu2023survey, kawaharazuka2024foundation, firoozi2025survey}.

Earlier works decouple LLMs and VLMs from the underlying robot policies responsible for action generation~\cite{ichter2023saycan, liang2023codeaspolicies}.
While effective for a limited set of predefined tasks, such systems typically rely on selecting from fixed motion primitives or on policies learned through imitation learning, which limits their ability to generalise to a broader range of tasks.
Learning policies that can generalise from current observations and instructions to unseen tasks remains a significant challenge.

To overcome these limitations, a growing body of research focuses on Vision-Language-Action (VLA) models~\cite{zitkovich2023rt2}.
By jointly learning visual, linguistic, and action modalities in an end-to-end framework, VLAs aim to enable robots to perform a wider range of tasks.
The hope is that the resulting generalist policies aim to achieve generalization across diverse tasks and facilitate effective transfer across varying robotic embodiments.
This approach reduces the need for extensive task-specific data collection and training, significantly lowering the cost of real-world deployment.
As such, VLAs offer a promising path toward more scalable and accessible robotic systems.


\begin{figure*}[t]
  \centering
  \includegraphics[width=0.8\textwidth]{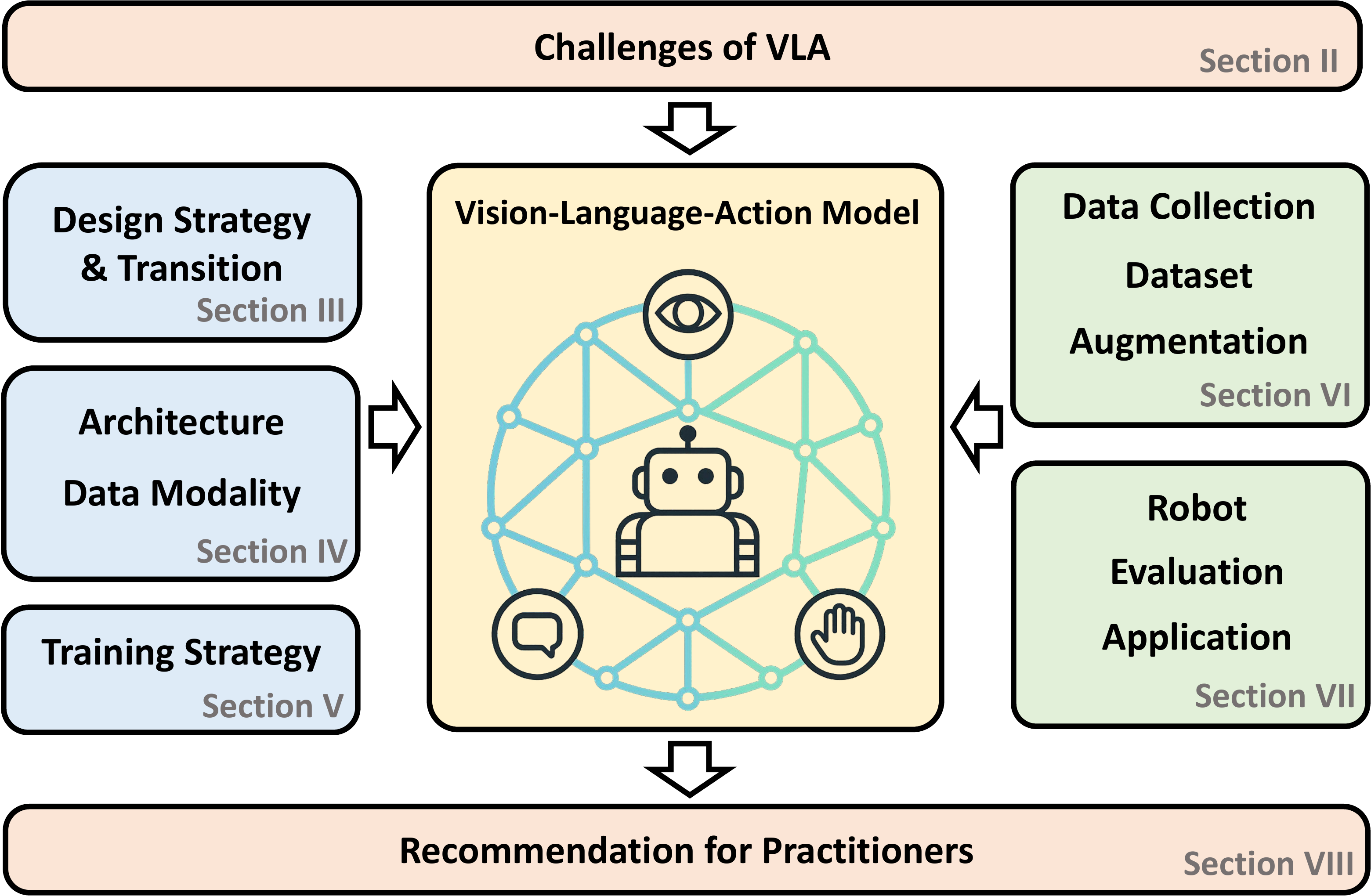}
  \caption{\textbf{Structure of this survey.} \secref{sec:challenges} outlines the key challenges in developing Vision-Language-Action (VLA) models. \secref{sec:vla_strategy} and \secref{sec:architectures_and_building_blocks} review the evolution of VLA strategies, architectures, and modality-specific design choices. \secref{sec:training} categorizes training strategies and practical implementation considerations. \secref{sec:datasets} discusses the data collection methologies, publicly available dataset, and data augmentation. \secref{sec:real_world} discusses the types of robots used, evaluation benchmarks, and the applications of VLA models in real-world robot systems. Guidance for practitioners is presented in~\secref{sec:recommendation}, based on the findings of the systematic review.}
  \label{figure:overview}
\end{figure*}

Despite growing interest, research on VLAs remains in its early stages. Architectural and training methodologies are not yet standardized, making it difficult to form a cohesive understanding of the field.
This survey provides a systematic overview of the current landscape of VLAs, including their historical development, model architectures, modality integration strategies, and learning paradigms.
While several previous surveys~\cite{ma2024survey, sapkota2025survey, yifan2025survey} have focused primarily on either action tokenization or general architectural advancements, this survey provides a comprehensive, full-stack overview, covering both software and hardware components. 
Specifically, beyond architecture and the development of VLAs, it includes robot platforms, data collection strategies, publicly available datasets, data augmentation techniques, and evaluation benchmarks.
We also introduce a taxonomy of existing VLA models and analyze representative models within each category.
This survey is intended to serve as a practical guide for researchers aiming to apply VLA models to real-world robotic systems.

In this review, to clarify the scope, we define VLA models as systems that take visual observations and natural language instructions as core inputs and produce robot actions by directly generating control commands (see Def.~\ref{def:vla}). While additional modalities (e.g., proprioception or depth) may be included, the integration of vision and language is essential. We exclude approaches that use vision and language solely for high-level reasoning or task planning without grounding them in action execution, such as those that select from a set of pre-trained skills using a high-level policy.

\begin{definition}[Vision-Language-Action (VLA) Model]\label{def:vla}
\textbf{A Vision-Language-Action (VLA) model is a system that takes visual observations and natural language instructions as required inputs and may incorporate additional sensory modalities. It produces robot actions by directly generating control commands. Thus, models in which a high-level policy (e.g., a vision-language model backbone) merely selects an index from a set of pre-trained skills or control primitives are excluded from this definition.}
\end{definition}

The overall structure of this survey is illustrated in \figref{figure:overview}.
First, \secref{sec:challenges} outlines the key challenges addressed in VLA research.
\secref{sec:vla_strategy} reviews major strategies and the architectural transition of VLA models.
\secref{sec:architectures_and_building_blocks} introduces core architectural components and building blocks, including modality-specific processing modules.
\secref{sec:training} discusses key training strategies and practical implementation considerations.
\secref{sec:datasets} summarises data collection methodologies, publicly available datasets, and data augmentation.
Then, \secref{sec:real_world} provides guidance for real-world deployment, covering commonly used robot platforms, evaluation protocols, and current real-world applications.
Based on the findings of the systematic review, we present recommendations for practitioners in~\secref{sec:recommendation}.
Finally, \secref{sec:discussion} discusses open challenges and future directions, and \secref{sec:conclusion} presents our concluding remarks.

\section{Challenges}\label{sec:challenges}
The integration of visual, linguistic, and motor modalities presents a promising pathway toward the development of generalist robot policies. 
However, the advancement of robust and deployable VLA models is still constrained by several fundamental challenges.
These limitations span across data availability, embodiment mismatches, and computational constraints, each imposing critical design trade-offs in model architecture, training strategy, and deployment feasibility.

\subsection{Data Requirements and Scarcity}
Training VLA models require large-scale, diverse, and well-annotated data that aligns visual observations with natural language instructions and corresponding actions.
However, datasets satisfying all three modalities, vision, language, and action, are limited in both scale and diversity.
While vision-language datasets such as COCO Captions~\cite{chen2015cococaptions} or web-scale corpora offer broad linguistic grounding, they lack the action grounding necessary for robotics.
Conversely, robot demonstration datasets often contain limited linguistic variability or are confined to narrow task distribution.

This mismatch leads to two data-related bottlenecks.
First, models pre-trained on large-scale web or video datasets may not transfer effectively to robotic tasks due to a lack of motor grounding or a discrepancy in the domain.
Second, high-quality robot demonstrations, often collected via teleoperation are expensive and difficult to scale.
Such an issue is further exacerbated when the number of modalities increases, such as adding tactile, acoustic, and 3D information.

\subsection{Embodiment Transfer}
\switchlanguage{
Robots exhibit a wide range of embodiments. Some are equipped solely with arms, while others incorporate wheels, legs, or other mobility mechanisms. Their joint configurations, link structures, sensor types and placements, and even physical appearances vary significantly.
While VLA models are increasingly trained on data from diverse robot embodiments, transferring policies across embodiments remains a major challenge. Each robot typically operates in a distinct action space and proprioceptive observation space, reflecting differences in degrees of freedom, sensor modalities, and kinematic structure. 


A related challenge lies in leveraging human motion data for training. Given the high cost of collecting large-scale robot data, human demonstrations offer a promising alternative. However, such data generally lack explicit action labels, and even when actions are inferred, they differ substantially from robot actions in both form and semantics. As with robot-to-robot transfer, mapping human demonstrations into robot-executable actions is highly non-trivial.

These embodiment-related challenges raise fundamental questions for VLA development: What kinds of data best support cross-embodiment generalization? How should morphological and sensory differences be represented? And how can models be trained to ensure robust grounding of vision and language across diverse robotic and human embodiments?
}{
ロボットは多様な身体構造で満ち溢れている.
腕だけのロボットもいれば, 車輪がついたロボット, 脚がついたロボットもいる.
各ロボットの関節配置やリンク構造は異なり, ついているセンサの種類や位置も, 見た目さえも異なる.
そのため, 基本的にある一つのロボットを使ってデータ収集をし, 学習したところで, そのポリシーを他のロボットに転移することはできない. 
これを解決するためには, 多様な身体性におけるデータを収集, 学習し, 様々な身体に対応可能なVLAを学習するという方法が考えられる.
しかし, 各身体性によってネットワークの入出力は異なるうえ, 身体性間の内挿ができるほどの多様な身体性におけるデータを収集することは難しい.

もう一点, 人間とロボットの間の身体性の変換という問題がある.
ロボットにおけるデータ収集が非常にコストが高いため, 人間の動作データを収集し, これをロボットに転移するというのは自然な発想である.
もしこれが可能であれば, 使うことのできるデータが大幅に増え, より汎用的なVLAが学習できるはずである.
しかし, 人間の動作データにはそもそもアクションがなく, そのままではVLAに活用することができない.
また, アクションが得られたとしても, それはロボットとは大きく異なるものであり, ロボット間の転移と同じく, 一筋縄ではいかない.

これら問題は, そもそもどのようなロボットを使うべきか, どのようなデータを収集すべきか, それらをどう統一的にVLAを学習させるべきかについて大きな課題を提示している.
}

\subsection{Computational and Training Cost}
Training VLA models entails a considerable amount of computational demands due to the high-dimensional and multi-modal nature of their input, typically including vision, language, and actions.
While many recent approaches leverage pre-trained VLM as a backbone, these models are typically adapted for robotics domain via large-scale robot demonstrations or simulated data.
Most practitioners are expected to build upon such pre-trained models and further fine-tune them for downstream tasks using task-specific, high-quality expert demonstrations, rather than training end-to-end from scratch.
Nonetheless, both the adaptation and fine-tuning stages remain computationally intensive, especially when processing long temporal sequences, high-resolution images, or additional modalities such as 3D point clouds or proprioception. %
Transformer-based architectures, which dominate current VLA designs, also scale poorly with respect to sequence length and input dimensionality, further amplifying memory and compute costs. %
At inference time, running these models in real-world settings, particularly on resource-constrained robotic platforms, poses additional challenges related to latency and memory usage. %
These computational burdens limit the accessibility and deployability of VLA systems, motivating ongoing research into efficient model architectures and distillation methods that can reduce resource requirements without significantly degrading performance.

\begin{figure*}[ht]
  \centering
  \includegraphics[width=0.85\textwidth]{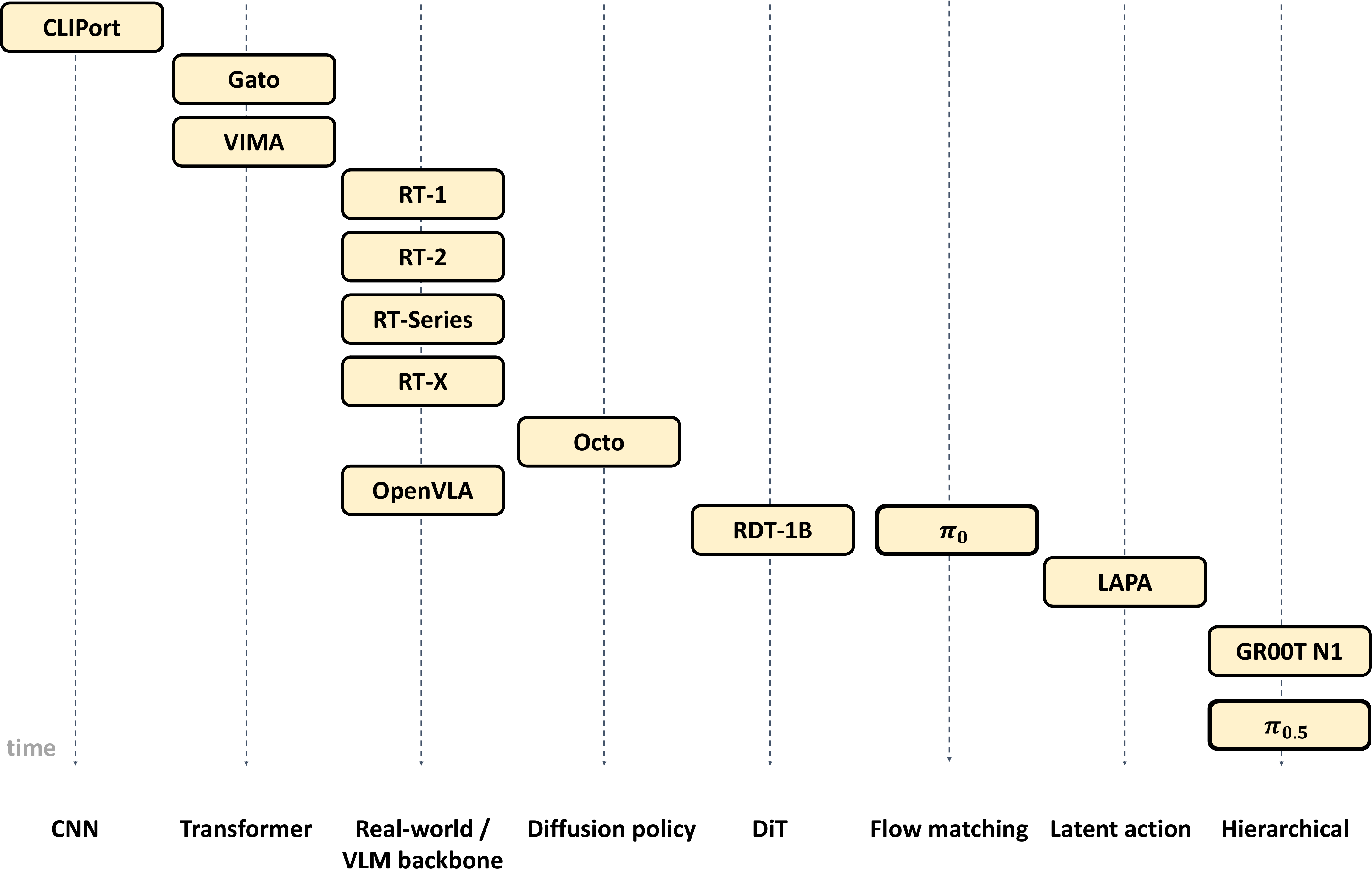}
  \caption{\textbf{Timeline of major Vision–Language–Action (VLA) models.} 
This figure summarizes the historical progression of representative VLA models shown in \secref{sec:vla_strategy}: from early CNN-based models (e.g., \textit{CLIPort}~\cite{shridhar2021cliport}), to real-world scalable policies leveraging pre-trained VLM backbones (e.g., \textit{RT-1}, \textit{RT-2}, \textit{RT-X}, \textit{OpenVLA}~\cite{brohan2023rt1, zitkovich2023rt2, oneill2024openxembodiment, kim2024openvla}), followed by models integrating diffusion and flow matching techniques (e.g., \textit{Octo}, \textit{RDT-1B}, \textit{$\pi_0$}~\cite{octoteam2024octo, liu2025rdt, Pi-0}), and more recent approaches focusing on latent action inference and hierarchical control (e.g., \textit{LAPA}, \textit{$\pi_{0.5}$}, \textit{GR00T N1}~\cite{ye2025lapa, Pi-0.5, Gr00t-N1}).
}
  \label{figure:history}
\end{figure*}

\section{VLA Design Strategy and Transition}\label{sec:vla_strategy}
\switchlanguage%
{%
This section categorizes major interface strategies for transforming vision and language inputs into robot actions, following the historical progression of VLA architectures (see~\figref{figure:history}). 
Each architectural category corresponds to a distinct generation of VLA systems, characterized by how multimodal representations are aligned with control. 
The discussion spans from early CNN-based models to transformer-based architectures, diffusion-based policies, and finally, hierarchical control frameworks.

\textbf{Early CNN-based end-to-end architectures.}
  A foundational approach to end-to-end VLAs is CLIPort~\cite{shridhar2021cliport}, one of the earliest models to integrate CLIP~\cite{radford2021clip} for extracting visual and linguistic features. 
  It combines these modalities with the Transporter Network~\cite{zeng21transporternetworks} to learn object manipulation tasks in an end-to-end manner, identifying which object to move and where to place it.
  CLIPort demonstrated the feasibility of jointly training vision, language, and action by leveraging CLIP~\cite{radford2021clip} as a pre-trained VLM.
  However, approaches based on Convolutional Neural Networks (CNNs) and Multi-Layer Perceptrons (MLPs) face challenges in unifying diverse modalities and also struggle to scale effectively.

\textbf{Transformer-based sequence models.}
  To address these limitations, Google DeepMind released Gato~\cite{reed2022gato}, a generalist agent and precursor to the Robotics Transformer (RT) series. 
  Gato performs a wide range of tasks, such as text chatting, visual question answering, image captioning, gameplay, and robot control, using a single transformer~\cite{vaswani2017transformer} model.
  It tokenizes language instructions using SentencePiece~\cite{kudo2018sentencepiece} and encodes images using Vision Transformer (ViT)~\cite{dosovitskiy2021vit}.
  A decoder-only transformer is then used to autoregressively generate actions based on the combined input sequence.
  While Gato enables multiple tasks with a single network, its repertoire of robotic skills remains limited to a narrow set, such as block stacking with a robotic arm.
  Similarly, VIMA~\cite{jiang2023vima} is an encoder-decoder transformer model that enables robots to follow general task instructions provided through a combination of text and goal images.
  Objects are first detected using Mask R-CNN~\cite{he2017maskrcnn}, after which each detected object's image is tokenized using ViT. 
  Bounding box coordinates are separately embedded as tokens, and textual instructions are tokenized using the T5 tokenizer~\cite{raffel2020t5}.
  A frozen T5 encoder and a transformer decoder are then used to autoregressively generate discrete action tokens.
  While VIMA demonstrates the ability to perform a wide range of robotic tasks, all experiments were limited to simulation environments.

\textbf{Unified real-world policies with pre-trained VLMs.}
  To enable scalable real-world applications, Robotics Transformer-1 (RT-1)~\cite{brohan2023rt1} has been introduced as a real-time, general-purpose control model capable of performing a wide range of real-world tasks.
  RT-1 processes a sequence of images using EfficientNet~\cite{tan2019efficientnet}, and performs FiLM conditioning~\cite{perez2018film} with language features encoded by the Universal Sentence Encoder (USE)~\cite{cer2018use}, enabling early fusion of visual and linguistic modalities.
  The extracted tokens are compressed via TokenLearner~\cite{ryoo2021tokenlearner} and then passed through a decoder-only transformer, which outputs discretized action tokens nonautoregressively (see~\secref{subsec:sensorimotor}).
  Trained on a large-scale dataset comprising 700 tasks and 130,000 episodes, RT-1 is regarded as the first VLA that unifies a broad range of robotic tasks.
  Subsequently, RT-2~\cite{zitkovich2023rt2} has been introduced as the successor to RT-1.
  It builds on a Vision-Language Model (VLM) backbone such as PaLM-E~\cite{driess2023palme} or PaLI-X~\cite{chen2024palix}, pre-trained on large-scale internet data.
  RT-2 is jointly fine-tuned on both internet-scale vision-language tasks and robotic data from RT-1, resulting in strong generalization to novel environments.
  This VLM-based design has since become the standard architecture for VLAs.
  In contrast, RT-X~\cite{oneill2024openxembodiment} has been introduced to demonstrate that training on datasets collected from multiple robots enables the development of more general-purpose VLAs, moving beyond the single-robot training paradigm of RT-1 and RT-2.
  
  The RT series has been extended into several variations, including RT-Sketch, which takes sketch images as input; RT-Trajectory, which takes motion trajectories as input; and others such as RT-H, Sara-RT, and AutoRT~\cite{sundaresan2025rtsketch, gu2024rttrajectory, belkhale2024rth, leal2024sarart, AutoRT}.
  Among these, RT-H~\cite{belkhale2024rth} is particularly notable for introducing a hierarchical policy structure. 
  Built on the RT-2 architecture, RT-H incorporates a high-level policy that predicts an intermediate representation known as language motion, and a low-level policy that generates actions based on it.
  By modifying the input prompt, the model can flexibly alternate between generating high-level actions expressed in language and producing low-level robot actions directly. 
  By sequentially switching between high-level and low-level policies, RT-H demonstrates improved performance, particularly in long-horizon tasks.
  %
  Such hierarchical VLA architectures have since become a recurring design pattern in subsequent models.
  Building upon the RT-series, OpenVLA~\cite{kim2024openvla} is introduced as an open-source VLA framework that closely mirrors the architecture of RT-2, leveraging a pre-trained VLM as its backbone.
  Specifically, it employs Prismatic VLM~\cite{karamcheti2024prismaticvlm}, based on LLaMa 2 (7B)~\cite{touvron2023llama2}, and encodes image inputs using DINOv2~\cite{oquab2023dinov2} and SigLIP~\cite{zhai2023siglip}.
  Through full fine-tuning on the Open-X Embodiment (OXE) dataset~\cite{oneill2024openxembodiment}, OpenVLA outperforms both RT-2 and Octo, and has since emerged as a mainstream architecture for VLA.

  \textbf{Diffusion policy.}
  Octo~\cite{octoteam2024octo}, introduced after the RT series, is the first VLA to leverage Diffusion Policy~\cite{chi2023diffusionpolicy}, and also gained attention for its fully open-source implementation.
  Octo supports flexible goal specification, which can include a language instruction and a goal image, processed by a T5 encoder and a CNN, respectively.
  For input observations, it similarly uses a CNN to encode images and a lightweight multilayer perceptron (MLP) to embed proprioceptive signals.
  All tokens are concatenated into a single sequence, augmented with modality-specific learnable tokens, and passed into a transformer.
  Finally, a diffusion policy generates continuous actions, conditioned on the output readout tokens.

\textbf{Diffusion transformer architectures.}
  RDT-1B~\cite{liu2025rdt} has been proposed as a large-scale diffusion transformer for robotics. 
  In contrast to prior approaches, where the diffusion process is applied only at the action head, RDT-1B employs a Diffusion Transformer (DiT)~\cite{peebles2023dit} as its backbone, integrating the diffusion process directly into the transformer decoder to generate actions.
  In RDT-1B, language inputs are tokenized using the T5 encoder, while visual inputs are encoded using SigLIP. 
  A diffusion model is then trained using a diffusion transformer with cross-attention, conditioned on both visual and textual tokens. 
  To facilitate multimodal conditioning and avoid overfitting, Alternating Condition Injection is proposed, in which image and text tokens are alternately used as queries at each transformer layer.

\textbf{Flow matching policy architectures.}
  Recently, inspired by Transfusion~\cite{zhoutransfusion}, $\pi_0$ builds on PaliGemma~\cite{beyer2024paligemma} and introduces a custom action output module, the action expert, which enables a multimodal model to handle both discrete and continuous data. 
  The action expert leverages flow-matching~\cite{lipman2023flow} to generate actions at rates up to $50$Hz. 
  It receives proprioceptive input from the robot and the readout token from the transformer, producing actions through a reverse diffusion process. 
  Rather than generating tokens autoregressively, it outputs entire action chunks in parallel, enabling smooth and consistent real-time control.

\textbf{Latent action learning from video.}
 Another notable approach is LAPA~\cite{ye2025lapa}, which leverages unlabeled video data for pre-training to learn latent actions for use in VLA models. 
  This enables policies to effectively utilize human demonstrations, making them robust to changes in embodiment and well-suited for real-world deployment. 
  The method applies patch embeddings, a spatial transformer, and a causal temporal transformer to images $\bm{x}_{t}$ and $\bm{x}_{t+H}$, then computes their difference. 
  VQ-VAE~\cite{van2017vqvae} is applied to this difference, generating a discrete token $\bm{z}_{t}$ which, together with $\bm{x}_{t}$, is used to reconstruct $\bm{x}_{t+H}$. 
  This entire network is trained jointly, forming a Latent Quantization Network. 
  Building on LWM-Chat-1M (7B)~\cite{liu2023world}, the vision and text encoders are kept frozen, and the resulting readout token is processed through an MLP trained to predict $\bm{z}_{t}$.
  Finally, only the MLP component is replaced by a separate network trained to directly output robot control commands. 

\textbf{Hierarchical policy architectures.}
  The most recent generation of VLAs adopts hierarchical policies to bridge high-level language understanding with low-level motor execution. 
  RT-H~\cite{belkhale2024rth} exemplifies this design by introducing a high-level controller that predicts intermediate ``language motion'' plans, followed by a low-level controller that refines these into concrete actions. 
  The system can dynamically switch between generating symbolic actions and executing detailed control sequences, improving performance in long-horizon, multi-step tasks. 

  This design is extended in $\pi_{0.5}$~\cite{Pi-0-FAST}, which combines high-level action token generation (using FAST tokens) with a low-level controller trained via flow matching.
  Pre-training aligns symbolic actions with language, while post-training ensures smooth execution via continuous action decoding.
  GR00T N1~\cite{Gr00t-N1} integrates multiple elements: latent actions from LAPA, diffusion-based generation from RDT-1B, and flow-matching controllers from $\pi_0$, unified into a multi-stage policy that generalizes across robots and tasks. 
  Hierarchical architectures now represent a state-of-the-art approach for scalable and adaptable VLA models, balancing the abstraction of language grounding with the precision of motor control.
}%
{%
  まず, 主要なVLAモデルについてその歴史的な流れを示す.
  その全体像を\figref{figure:history}に示す.
  ここでは, 以下の主要なVLAモデルの流れについてまとめる.
  \begin{itemize}
    \item CLIPort~\cite{shridhar2021cliport}
    \item Gato, VIMA \cite{reed2022gato, jiang2023vima}
    \item RT-1, RT-2, RT-X \cite{brohan2023rt1, zitkovich2023rt2, oneill2024openxembodiment}
    \item RT-Sketch, RT-Trajectory, RT-H, Sara-RT, AutoRT \cite{sundaresan2025rtsketch, gu2024rttrajectory, belkhale2024rth, leal2024sarart, AutoRT}
    \item Octo, OpenVLA \cite{octoteam2024octo, kim2024openvla}
    \item RDT-1B, LAPA \cite{liu2025rdt, ye2025lapa}
    \item $\pi_0$, $\pi_{0.5}$, GR00T N1 \cite{Pi-0, Pi-0.5, Gr00t-N1}
  \end{itemize}

  End-to-EndなVLAでもっとも原始的な方法はCLIPort \cite{shridhar2021cliport}である.
  CLIPortはCLIP \cite{radford2021clip}を用いて画像と言語の特徴量を抽出し, これとTransporter Network \cite{zeng21transporternetworks}を合わせることで, どの物体をどこに移動させるかをEnd-to-Endに学習することができる初期のモデルである.
  このCLIPortにより, CLIPのような視覚言語モデルを用いることで, 視覚・言語・アクションを統合的に学習することが可能であることが示された.
  その一方で, Convolutional Neural Network (CNN)やMulti-Layer Perceptron (MLP)によるネットワークでは多様なモダリティをを統一的に扱うことが難しく, また規模がスケーリングしにくいという問題があった.

  これを受けて, Google DeepMindから汎用エージェントGato \cite{reed2022gato}が発表された.
  Gatoは単一のtransformer \cite{vaswani2017transformer}モデルによってテキストチャット・質問応答・画像キャプショニング・ゲームプレイ・ロボット制御など, 様々なタスクを実行できる, 後継のRobotics Transformer (RT)シリーズの先駆けとなったモデルである.
  言語指示をSentencePiece \cite{kudo2018sentencepiece}でトークン化, 画像をViT \cite{dosovitskiy2021vit}でトークン化し, Decoder-onlyなtransformerを用いて, 次に取るべきActionトークンを自己回帰的に生成している.
  Gatoは多様なタスクを単一のネットワークで可能な一方で, ロボットスキルについては, ロボットアームによるブロック積みという限られたスキルしか学習していない.
  これに似た方法としてVIMA \cite{jiang2023vima}がある.
  これは, ゴール画像やテキストでロボットに汎用的なタスクを指示できるEncoder-Decoder型のtransformerモデルである.
  Mask R-CNN \cite{he2017maskrcnn}でオブジェクトを検出, 各オブジェクト画像をViTによりトークン化, バウンディングボックスの情報もトークン化, テキストはT5 Tokenizer \cite{raffel2020t5}でトークン化し, 凍結したT5エンコーダとtransformer decoderによって離散アクショントークンを自己回帰的に出力する.
  こちらは多様なロボットタスクを実行できるものの, 全てシミュレーション上での実験にとどまっていた.

  そこで, ロボットに特化し, リアルタイムかつ多様な実世界タスクを実行可能な汎用ロボット制御モデルとして開発されたのがRobotics Transformer-1 (RT-1) \cite{brohan2023rt1}である.
  RT-1は画像の時系列をEfficientNet \cite{tan2019efficientnet}に入れ, Universal Sentence Encoder (USE) \cite{cer2018use}により変換された言語特徴量でFiLM Conditioning \cite{perez2018film}を行うことで, 画像と言語の特徴量を早期に統合する.
  得られたトークンをTokenLearner \cite{ryoo2021tokenlearner}により圧縮し, Decoder-onlyのtransformerを通して離散化されたAction Tokenを出力するというものである.
  700タスク, 130,000エピソードという大規模なデータセットで学習されており, 多様なロボットタスクを統合した初めてのVLAといえるだろう.
  その後継として開発されたのがRT-2 \cite{zitkovich2023rt2}である.
  RT-2は, 大規模なインターネットスケールのデータで学習されたVision-Language Model (VLM)であるPaLM-E \cite{driess2023palme}またはPaLI-X \cite{chen2024palix}をbackboneとして, インターネットスケールの視覚言語タスクとRT-1由来のロボットデータを同時にfine-tuningに使用することで, 未知環境に対しても高い汎用性を持つVLAとなっている.
  この, VLMをbackboneとするVLAという考え方は, その後のVLAアーキテクチャの主流となっている.
  また, RT-1やRT-2では単一のロボットにおけるデータセットを用いて学習していたが, その後に発表されたRT-Xでは, 複数のロボットにおけるデータセットを用いることで, より汎用的なVLAを学習できることが明らかになっている.

  このRTシリーズには, スケッチ画像を入力とするRT-Sketchや動作軌道を入力とするRT-Trajectory, その他にもRT-H, Sara-RT, AutoRTなどのバリエーションがある \cite{sundaresan2025rtsketch, belkhale2024rth, leal2024sarart, AutoRT}.
  その中でも階層構造に着目した重要なモデルがRT-H \cite{belkhale2024rth}である.
  RT-HはRT-2のアーキテクチャをベースにしながら, 中間表現であるlanguage motionを予測する高レベルポリシーと, language motionからアクションを生成する低レベルポリシーへの分割を導入したモデルである.
  同一のモデルに対して, promptを変化させることでlanguage motionを予測するかactionを予測するかを切り替えることができ, 高レベルポリシーと低レベルポリシーを順に適用することで, 性能を上げることに成功している.
  このような階層構造を持つVLAは, その後のVLAアーキテクチャにも頻繁に現れている.

  これらRTシリーズの後に開発されたのがOctoである.
  Octo \cite{octoteam2024octo}はDiffusion Policy \cite{chi2023diffusionpolicy}の考え方を取り入れた最初のVLAであり, 全てのコードをオープンソースとして公開することで大きな注目を集めた.
  タスクトークンとしては, 言語をT5に準拠してトークン化した埋め込みベクトル, ゴール画像をCNNで変換した埋め込み表現を利用する.
  観測トークンとしては, 同様に画像をCNNで, プロプリオセプションを小さなMLPでベクトル化し利用する.
  その後, モダリティごとのlearnable tokenを加算しながら全トークンを一列に並べ, transformerに入力し, readoutトークン出力を条件としたdiffusion policyを適用する.
  また, Octoと同様にオープンソースとして開発されたのがOpenVLA \cite{kim2024openvla}である.
  OpenVLAのアーキテクチャはRT-2と非常に似ており, 事前学習済みのVLMをbackboneとして使っている.
  特に, LLaMa 2 (7B) \cite{touvron2023llama2}をベースとして, 画像入力をDINOv2 \cite{oquab2023dinov2}とSigLIP \cite{zhai2023siglip}により変換して入力する, Prismatic VLM \cite{karamcheti2024prismaticvlm}と呼ばれるVLMが使われている.
  これをRT-Xのデータセットを用いてfull fine-tuningすることで, RT-2やOctoよりも高い性能を達成しており, その後のVLAのアーキテクチャの主流となっている.

  次に, RDT-1B \cite{liu2025rdt}と呼ばれる大規模なロボット用のdiffusion transformerも開発されている.
  action headとしてdiffusion policyを用いるのではなく, transformerのDecoderを用いて拡散過程を表現する.
  RDT-1Bでは, テキストをT5のencoderでトークン化, 画像はSigLIPによってトークン化し, Diffusion Transformer (DiT) \cite{peebles2023dit}とCross Attentionを用いて, 画像とテキストのトークンを条件とした拡散モデルを学習する.
  このとき, transformerの各層で画像とテキストを交互にクエリとして用いるAlternating Condition Injectionを行っている.
  また, もう一つ注目を浴びている方法として, LAPA \cite{ye2025lapa}がある.
  これは, アクションラベルなしの動画から事前学習を行い, Latent Actionを学習してVLAに利用するという手法である.
  これにより, 人間のデモンストレーションが活用でき, embodimentの変化に頑健で, 実環境に強いポリシーを学習することができる.
  $\bm{x}_{t}$と$\bm{x}_{t+H}$の画像にpatch embedding, spatial transformer, causal temporal transformerを適用し, その差分を計算する.
  これに対してVQ-VAE \cite{van2017vqvae}を適用, 得られる離散トークン$\bm{z}_{t}$と$\bm{x}_{t}$から$\bm{x}_{t+H}$を復元するという形のnetworkを一気に学習させる(Latent Quantization Network).
  LWM-Chat-1M (7B) \cite{liu2023world}をベースに, Vision/Text encoderは固定で, 得られたreadout tokenをMLPで変換し, さきほどの$\bm{z}_{t}$が出力されるように学習を行う.
  最後にMLP部分のみを別Networkに挿げ替え, ロボットの制御入力を直接出力するように学習する.
  これら, RDT-1Bにおける大規模なdiffusion transformerやLAPAにおける動画からの行動表現の学習は, 現在のVLAのアーキテクチャにおいても重要な要素となっている.

  最後に, Physical Intelligenceが開発した$\pi_0$ \cite{Pi-0}と$\pi_{0.5}$ \cite{Pi-0.5}, NVIDIAが開発したGR00T N1 \cite{Gr00t-N1}について述べる.
  $\pi_{0}$は, PaliGemma \cite{beyer2024paligemma}をベースとして, 独自の動作出力モジュールであるAction Expertを追加し, flow matching \cite{lipman2023flow}により連続的アクションを最大50 Hzで出力可能なモデルである.
  Action Expertはロボットのproprioceptionと, transformerから得られたreadout tokenを入力として, 理想的なアクションへの復元ベクトルを学習する.
  このとき, 自己回帰的に1ステップずつトークンを出力するのではなく, 一括でアクションチャンクを出力している点が重要である.
  これにより, 滑らかで一貫した, リアルタイムなアクション出力が可能である.
  その後継である$\pi_{0.5}$は, $\pi_{0}$のアーキテクチャをベースにしながら, RT-HのようにHigh-Level PolicyとLow-Level Policyの2つの階層を統合することでより性能を高めたモデルである.
  $\pi_{0}$と同じくPaliGemmaをベースとしたモデルを, multimodalなデータとtask promptを入力として, subtaskのpromptやFAST \cite{Pi-0-FAST}と呼ばれる離散アクショントークン, captioningなどを出力するように, 多様なロボットやwebデータを使い学習する.
  なお, FASTは一般的な離散アクショントークンを圧縮し推論スピードを高めたトークンである(詳しくは\secref{subsubsec:action}で述べる).
  先に学習したpolicyをhigh-level policyとして, post-trainingではsubtaskのpromptを入力し, action expertがflow matchingによりaction出力を学習するlow-level policyを構築する.
  FASTのような離散アクショントークンの方が言語と統一して学習しやすいが, 最終的には滑らかでリアルタイムなactionを出したいという考え方を結合して pre-trainingとpost-trainingに分けた形を取っている.
  さらに, このような階層型の考え方, flow matching, RDT-1Bに見られるdiffusion transformer, LAPAなどを統合したのが, NVIDIAが開発したGR00T N1である.
  VLMから出力されたトークンを条件としたクロスアテンションをdiffusion transformerに適用し, flow matchingにより連続的なアクションを出力する.
  また, LAPAを活用しながら人のデモンストレーションを含む多様なembodimentのデータセットで事前学習を行い, その後個別のロボットデータでpost-trainingを行っている.
}%

\begin{figure*}[t]
  \centering
  \includegraphics[width=1.0\textwidth]{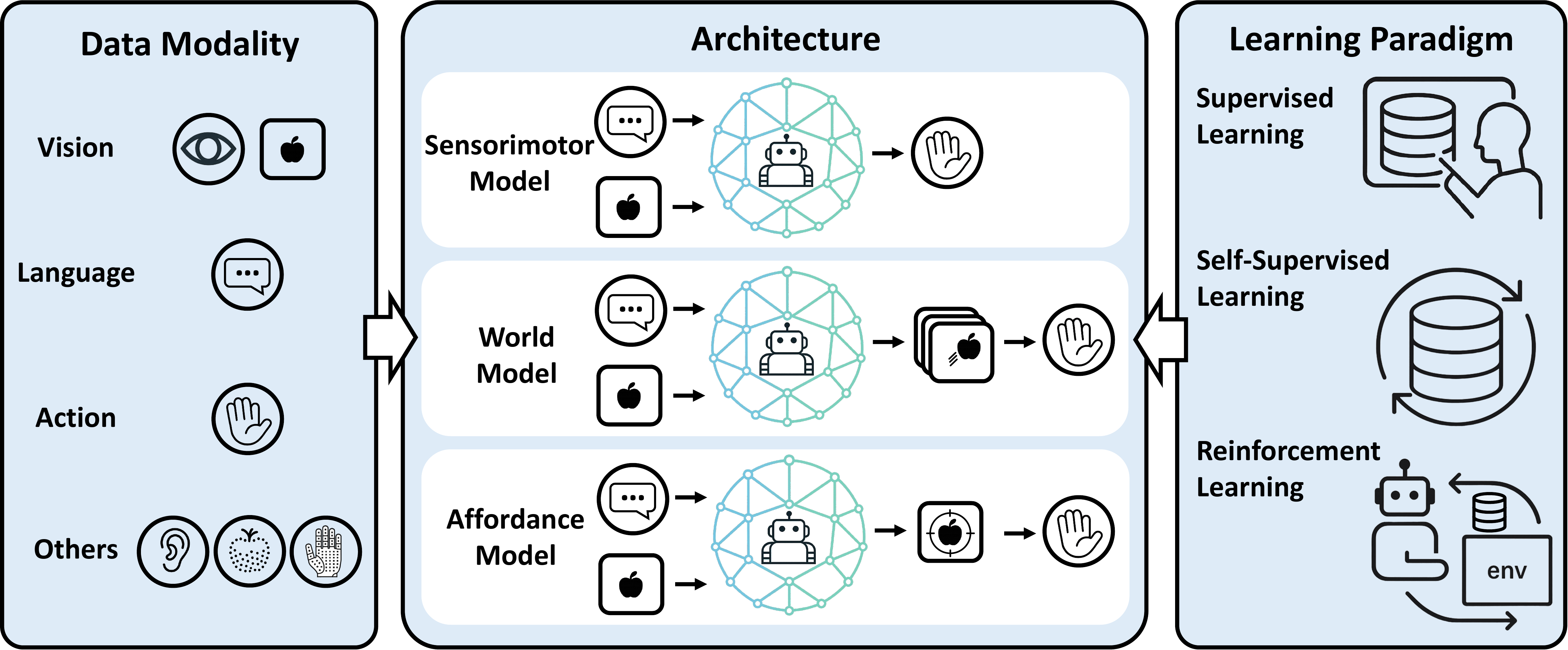}
  \caption{\textbf{Structure of \secref{sec:architectures_and_building_blocks} and \secref{sec:training}.} The figure summarizes key components of VLA models. The center illustrates core architectural types, including sensorimotor models, world models, and affordance-based models. The left side depicts the input and output modalities—vision, language, action, and other auxiliary modalities. The right side presents training strategies, including supervised learning, self-supervised learning, and reinforcement learning, along with practical implementation considerations.}
  \label{figure:method}
\end{figure*}

\section{Architectures and Building Blocks}\label{sec:architectures_and_building_blocks}
\switchlanguage%
{%
Vision-Language-Action (VLA) models encompass a wide range of architectural designs, reflecting diverse strategies for integrating perception, instruction, and control. 
A widely adopted approach is the sensorimotor model, which jointly learns visual, linguistic, and action representations. 
These models take images and language as input and directly output actions, and can adopt either a flat or hierarchical structure with varying backbone architectures.
While sensorimotor models form a foundational class of VLA systems, several alternative architectures have been proposed. World models predict the future evolution of sensory modalities, typically visual, conditioned on language input, and use these predictions to guide action generation. 
Affordance-based models are another variant that predict action-relevant visual affordances based on language, and then generate actions accordingly.
%
}%
{%
  VLAと一口に言っても, そのアーキテクチャは様々である.
  その中で, 最も単純で効果的なものは, 視覚・言語・アクションを統合的に学習するSensorimotor Modelである.
  これは, 画像と言語を入力として, 直接アクションを出力するモデルであり, フラットな構造や階層構造があり, そのバックボーンにも様々なバリエーションがある.
  このSensorimotor ModelがVLAの基本アーキテクチャであるが, 一部の研究ではこれとは異なる形のアーキテクチャが提案されている.
  その一つがWorld Modelである.
  言語から視覚情報を基本とした様々なモダリティの変化を予測するモデルであり, この予測した情報を用いてアクションを生成する.
  また, Affordance-based Modelは, 言語から視覚におけるアフォーダンスを予測するモデルであり, このアフォーダンスに従ってアクションを生成する.
  最後に, VLAの文脈からは多少逸れるが, ロボティクスへの適用を前提に構築されてきたVLMについても紹介する.
}%

\begin{figure*}[t]
  \centering
  \includegraphics[width=0.9\textwidth]{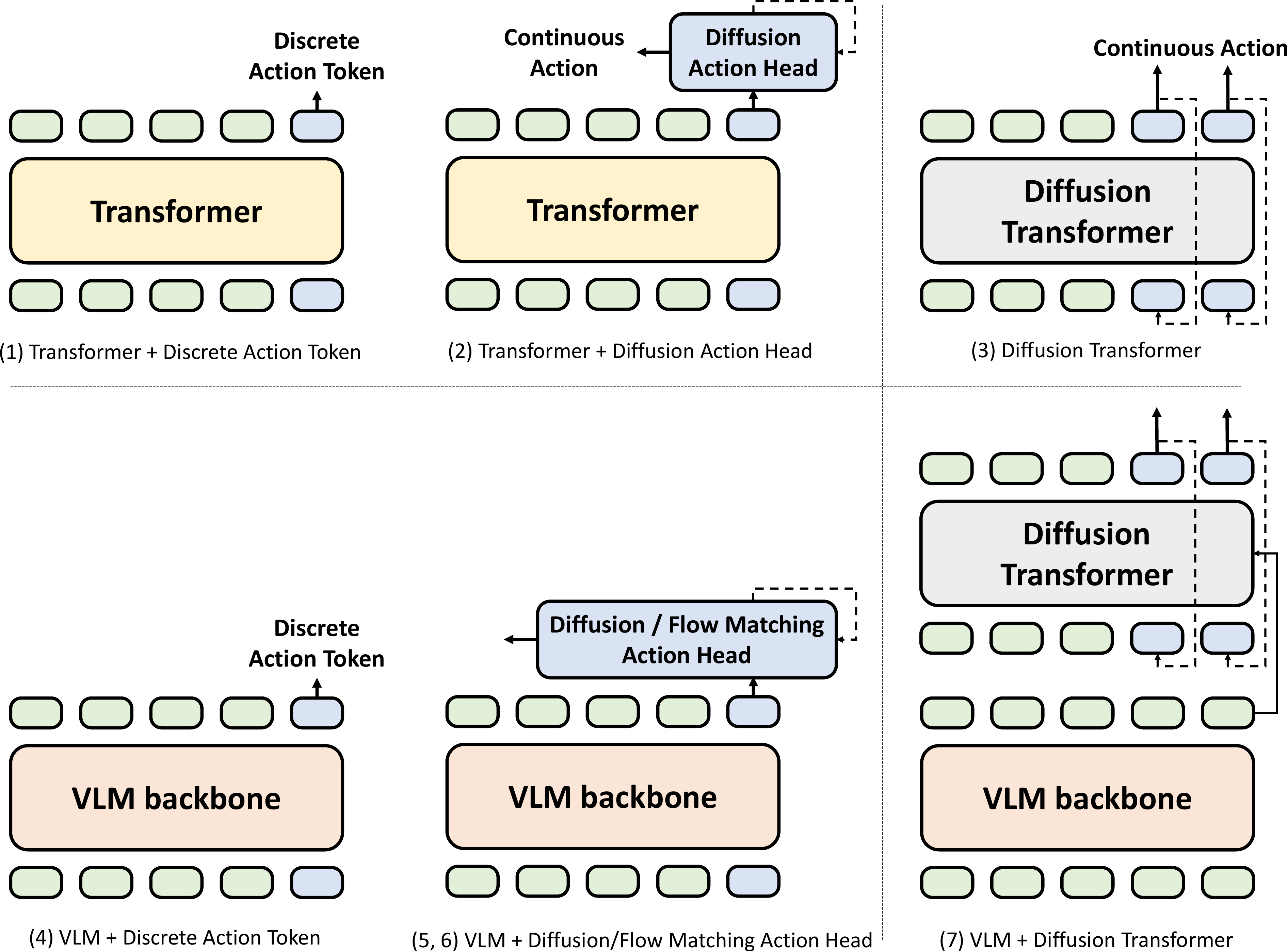}
  \caption{\textbf{Architecture of sensorimotor models for VLA.} This figure categorizes seven representative architectures used in recent VLA research. (1) \textit{Transformer + Discrete Action Token}: A standard transformer processes tokenized inputs to predict discrete actions. (2) \textit{Transformer + Diffusion Action Head}: A diffusion model is appended to the transformer for generating smooth, continuous actions. (3) \textit{Diffusion Transformer}: The diffusion process is integrated directly within the transformer architecture. (4) \textit{VLM + Discrete Action Token}: Vision-language models (VLMs) replace transformers to leverage pre-trained knowledge while predicting discrete actions. (5) \textit{VLM + Diffusion Action Head}: Combines VLMs with diffusion heads for continuous control. (6) \textit{VLM + Flow Matching Action Head}: Substitutes diffusion with flow matching to enhance real-time control. (7) \textit{VLM + Diffusion Transformer}: Employs a VLM as a backbone and a diffusion transformer as a low-level policy for end-to-end continuous action generation.}
  \label{figure:sensorimotor}
\end{figure*}

\subsection{Sensorimotor Model}\label{subsec:sensorimotor}
\switchlanguage%
{%
There are currently seven architectural variations of the sensorimotor models, as illustrated in \figref{figure:sensorimotor}.

\textbf{(1) Transformer + Discrete Action Token.}
This architecture represents both images and language as tokens, which are fed into a transformer to predict the next action, typically in the form of discrete tokens (see \figref{figure:sensorimotor} (1)). 
This category also includes models that use CLS tokens and generate continuous actions through an MLP. 
Representative examples include VIMA~\cite{VIMA} and Gato~\cite{reed2022gato}, which tokenize multiple modalities using language tokenizers, vision transformers, MLPs, and other components, and output discretized actions such as binned values.
VIMA employs an encoder-decoder transformer conditioned on diverse task modalities, whereas Gato uses a decoder-only transformer that autoregressively processes all tokens in a single sequence.

In contrast to VIMA and Gato, which generate action tokens autoregressively, RT-1~\cite{brohan2023rt1} adopts a different approach by compressing inputs using TokenLearner~\cite{ryoo2021tokenlearner} and employing a decoder-only transformer to predict all action tokens non-autoregressively.
In practice, $48$ tokens are fed into the transformer, and the final $11$ tokens are extracted as action outputs.
This architecture has been adopted by several approaches, such as MOO~\cite{stone2023moo}, RT-Sketch~\cite{RT-Sketch}, and RT-Trajectory~\cite{RT-Trajectory}.
It has also become a common design choice in other VLA models such as Robocat~\cite{bousmalis2024robocat}, RoboFlamingo~\cite{RoboFlamingo}, and many others~\cite{lu2024unifiedio2, shi2024yay, haldar2024baku, Actra, pang2024di2, doshi2025crossformer}, due to its simplicity and scalability.

\textbf{(2) Transformer + Diffusion Action Head.}
This architecture builds upon the structure in (1) by incorporating a diffusion policy as the action head following the transformer.
While discrete action tokens often lack real-time responsiveness and smoothness, these models achieve continuous and stable action outputs using diffusion models~\cite{ho2020ddpm}.
Representative examples include Octo~\cite{octoteam2024octo} and NoMAD~\cite{sridhar2024nomad}. 
Octo processes image and language tokens as a single sequence through a transformer, then applies a diffusion action head conditioned on the readout token. 
In contrast, NoMAD replaces the language input with a goal image, compresses the transformer output via average pooling, and uses the resulting vector to condition the diffusion model.
TinyVLA~\cite{wen2024tinyvla}, RoboBERT~\cite{RoboBERT}, and VidBot~\cite{chen2025vidbot} also adopt this architecture.
  
\textbf{(3) Diffusion Transformer.}
The diffusion transformer model shown in \figref{figure:sensorimotor} (3) integrates the transformer and diffusion action head, executing the diffusion process directly within the transformer.
This enables the model to perform the diffusion process conditioned directly on image and language tokens.
For example, RDT-1B~\cite{liu2025rdt}, built on this architecture, generates a sequence of action tokens via cross-attention with a vision and language query, which are subsequently mapped to executable robot actions through an MLP.
Similarly, Large Behavior Models (LBMs) also adopt the diffusion transformer architecture and emphasize the importance of large-scale and diverse pre-training.
In addition, StructDiffusion, MDT, DexGraspVLA, UVA, FP3, PPL, PPI, and Dita~\cite{liu2023structdiffusion, reuss2024mdt, DexGraspVLA, UnifiedVideoActionModel, FP3, yao2025ppl, PPI, Dita} uses this architecture.

\textbf{(4) VLM + Discrete Action Token.}
VLM + Discrete Action Token models, as illustrated in \figref{figure:sensorimotor} (4), improve generalization by replacing the transformer in (1) with a Vision-Language Model (VLM) pre-trained on large-scale internet data. 
Leveraging a VLM allows these models to incorporate human commonsense knowledge and benefit from in-context learning capabilities.
For example, RT-2 uses large-scale VLMs such as PaLM-E or PaLI-X as the backbone, which processes image and language tokens as input and outputs the next action as discrete tokens.
Furthermore, LEO, GR-1, RT-H, RoboMamba, QUAR-VLA, OpenVLA, LLARA, ECoT, 3D-VLA, RoboUniView, and CoVLA~\cite{huang2024leo, GR-1, belkhale2024rth, liu2024robomamba, ding2024quarvla, kim2024openvla, li2025llara, zawalski2025ecot, 3D-VLA, RoboUniView, arai2025covla} adopt this architecture.

\textbf{(5) VLM + Diffusion Action Head.}
VLM + Diffusion Action Head models, as shown in \figref{figure:sensorimotor} (5), build on (2) by replacing the transformer with a VLM. 
This architecture combines VLMs, which enable better generalization, with diffusion models that generate smooth, continuous robot action commands.
For example, Diffusion-VLA, DexVLA, ChatVLA, ObjectVLA, GO-1 (AgiBot World Colosseo), PointVLA, MoLe-VLA, Fis-VLA, and CronusVLA~\cite{DiffusionVLA, DexVLA, ChatVLA, ObjectVLA, AgiBotWorldColosseo, PointVLA, MoLe-VLA, Fast-in-Slow, CronusVLA} adopt this architecture.
HybridVLA~\cite{HybridVLA} further combines (4) and (5) to both autoregressively generate discrete tokens as well as use a diffusion action head to generate continuous actions within a single model.

\textbf{(6) VLM + Flow Matching Action Head.}
VLM + Flow Matching Action Head models, as shown in \figref{figure:sensorimotor} (6), replace the diffusion model in (5) with a flow matching action head~\cite{lipman2023flow}, improving real-time responsiveness while maintaining smooth, continuous control. 
A representative example is $\pi_0$, based on PaliGemma~\cite{beyer2024paligemma}, which achieves control rates of up to $50$ Hz. 
Other examples include GraspVLA, OneTwoVLA, Hume, and SwitchVLA~\cite{GraspVLA, OneTwoVLA, Hume, SwitchVLA}.
$\pi_{0.5}$~\cite{Pi-0.5} integrates the architectures of (4) and (6), supporting both discrete tokens and flow matching within a unified framework.

\textbf{(7) VLM + Diffusion Transformer.}
VLM + Diffusion Transformer models, shown in \figref{figure:sensorimotor} (7), combine a VLM with a diffusion transformer described in (3). 
The VLM typically serves as a high-level policy (system 2), while the diffusion transformer acts as a low-level policy (system 1). 
The diffusion transformer may be implemented using either diffusion or flow matching. 
A representative model is GR00T N1~\cite{Gr00t-N1}, which applies cross-attention from the diffusion transformer to VLM tokens and generates continuous actions via flow matching.
This design is also used in CogACT, TrackVLA, SmolVLA, and MinD~\cite{CogAct, TrackVLA, SmolVLA, MinD}.
}%
{%
  このSensorimotor Modelは, \figref{figure:sensorimotor}に示すような, 以下の7つのバリエーションがある(CLIPortのような原始的な構造は除いている).
  \begin{enumerate}
    \item Transformer + Discrete Token
    \item Transformer + Diffusion Action Head
    \item Diffusion Transformer
    \item Vision-Language Model + Discrete Token
    \item Vision-Language Model + Diffusion Action Head
    \item Vision-Language Model + Flow Matching Action Head
    \item Vision-Language Model + Diffusion Transformer
  \end{enumerate}

  まず, (1)にあるようなTransformer + Discrete Tokenのモデルは, 画像と言語をそれぞれトークン化し, これらをTransformerに入力して, 次に取るべきアクションを離散的なトークンとして出力するモデルである.
  ここには, CLSトークンなどを利用してMLPにより連続的なアクションを出力するものも含まれる.
  その代表的なモデルはVIMA \cite{jiang2023vima}やGato \cite{reed2022gato}である.
  どちらも言語用のTokenizerやViT, MLPなどを使って様々なモダリティをトークン化し, それらをTransformerに入力して, ビンに分割された離散的なアクションを表すトークンを出力する.
  この際, VIMAはモダリティを識別するための明示的なModality Type Embeddingを用いているのに対して, Gatoはモダリティごとに異なるembeddingのlookup tableを用意して対応している点が異なる.
  また, VIMAは多様なモダリティによるタスク指示を条件としてアクションを生成するためEncoder-Decoder型のTransformerを用いているのに対し, Gatoは全てのトークンを一列に並べたDecoder-onlyのTransformerを用いている点も異なる.
  VIMAとGatoはどちらも自己回帰的にアクショントークンを生成するモデルであるが, その一方でRT-1 \cite{brohan2023rt1}はTokenLearner \cite{ryoo2021tokenlearner}を用いてトークンを圧縮し, その後にDecoder-onlyのtransformerを用いてアクショントークンを非自己回帰的に全て出力するという形を取っている.
  実際には, 得られた48トークンをtransformerに入れ, 最後の11トークンをアクショントークンとして出力している.
  RT-1を発展させた手法であるMOO \cite{stone2023moo}やRT-Sketch \cite{RT-Sketch}, RT-Trajectory \cite{RT-Trajectory}などはどれも同様のアーキテクチャである.
  この他にも, Robocat \cite{bousmalis2024robocat}やRoboFlamingo \cite{RoboFlamingo}, その他多くのVLAにおいて, この構造は最もシンプルで使いやすいことから, 現在も非常に多く採用されている\cite{lu2024unifiedio2, shi2024yay, haldar2024baku, Actra, pang2024di2, doshi2025crossformer}.

  次に, (2)にあるようなtransformer + Diffusion Action Headのモデルは, (1)と同じようなアーキテクチャでありながら, transformerの後段にDiffusion Policyを採用したAction Headを追加した形である.
  離散的なトークンではどうしてもリアルタイム性や滑らかさが欠けるが, 拡散モデル \cite{ho2020ddpm}を使うことで連続的でより安定したアクションを出力することに成功している.
  その代表的なモデルはOcto \cite{octoteam2024octo}とNoMAD \cite{sridhar2024nomad}がある.
  Octoはtransformerに画像や言語のトークンを一列に入力し, readoutトークン出力を条件としたdiffusion action headを採用している.
  NoMADは言語ではなくゴール画像を入力としているが, transformerから得られた7$\times$256次元のトークンを平均プーリングにより1$\times$256次元に圧縮し, これを条件として拡散モデルを適用している.
  この他にも, TinyVLA, RoboBERT, VidBotなどがこのアーキテクチャを採用している \cite{wen2024tinyvla, RoboBERT, chen2025vidbot}.

  (3)のDiffusion Transformerモデルは, (2)におけるtransformerとDiffusion Action Headを統合し, transformerの中で拡散過程を実行するモデルである.
  これにより, 画像や言語のトークンを直接条件として, 拡散過程を実行することができる.
  この代表的な例がRDT-1B \cite{liu2025rdt}である.
  画像トークンと言語トークンを条件としたtransformer型の拡散モデルを学習することで, 連続的なアクショントークンを出力することができ, これをMLPにより変換することで実際のアクションとしている.
  また, Large Behavior Models (LBMs)も同様のdiffusion transformerを用いたモデルであり, これらのモデルにおける量と多様性を持った事前学習の重要性を説いている. 
  この他にも, StructDiffusion, MDT, DexGraspVLA, UVA, FP3, PPL, PPI, Ditaなどがこのアーキテクチャを採用している \cite{liu2023structdiffusion, reuss2024mdt, DexGraspVLA, UnifiedVideoActionModel, FP3, yao2025ppl, PPI, Dita}.

  (4)のVLM + Discrete Tokenのモデルは, (1)のtransformerをインターネットスケールの大規模なデータセットによって学習されたVLMに置き換えることで, より汎用性を高めたモデルである.
  VLMにより, 人間の常識や知識を取り込むことができ, VLMが持ち合わせているin-context learningの能力を活用することもできる.
  この代表的な例がRT-2である.
  PaLM-EやPaLI-Xといった大規模なVLMをbackboneとして用い, 画像と言語のトークンを入力として, 次に取るべきアクションを離散的なトークンとして出力する.
  LEO, GR-1, RT-H, RoboMamba, QUAR-VLA, OpenVLA, LLARA, ECoT, 3D-VLA, RoboUniView, CoVLAなどがこのアーキテクチャを採用している \cite{huang2024leo, GR-1, belkhale2024rth, liu2024robomamba, ding2024quarvla, kim2024openvla, li2025llara, zawalski2025ecot, 3D-VLA, RoboUniView, arai2025covla}.

  (5)のVLM + Diffusion Action Headのモデルは, (2)のtransformerをVLMに置き換えたモデルである.
  VLMによる汎化性能と拡散モデルによる連続的なアクション出力の両方を活用することができる.
  Diffusion-VLA, DexVLA, ChatVLA, ObjectVLA, GO-1 (AgiBot World Colosseo), PointVLA, MoLe-VLA, Fis-VLA, CronusVLAなどがこのアーキテクチャを採用している \cite{DiffusionVLA, DexVLA, ChatVLA, ObjectVLA, AgiBotWorldColosseo, PointVLA, MoLe-VLA, Fast-in-Slow, CronusVLA}.
  また, HybridVLA \cite{HybridVLA}のように, 同じネットワークで離散トークンとDiffusion Action Headの両者を扱う, (4)と(5)を合体させたようなモデルも登場している.

  (6)のVLM + Flow Matching Action Headのモデルは, (5)における拡散モデルの代わりに, flow matching \cite{lipman2023flow}を用いたAction Headを追加したモデルである.
  flow matchingを用いることで, よりリアルタイム性の高い連続的なアクションを出力することができる.
  この代表的な例が$\pi_0$である.
  $\pi_0$はPaliGemma \cite{beyer2024paligemma}をベースとし, flow matchingにより連続的なアクションを最大50 Hzで出力することができる.
  GraspVLA, OneTwoVLA, Hume, SwitchVLAなどがこのアーキテクチャを採用している \cite{GraspVLA, OneTwoVLA, Hume, SwitchVLA}.
  また, $\pi_{0.5}$ \cite{Pi-0.5}のように, 同じネットワークで離散トークンとflow matching Action Headの両者を扱う, (4)と(6)を合体させたようなモデルも登場している.

  (7)のVLM + Diffusion Transformerのモデルは, VLMと(3)におけるdiffusion transformerを組み合わせたモデルである.
  この場合, VLM側をSystem 2 (High-Level Policy), diffusion transformer側をSystem 1 (Low-Level Policy)と呼ぶことがある.
  なお, diffusion transformerについては, 拡散モデルとして定式化するものと, flow matchingとして定式化するものがある.
  この代表的な例がGR00T N1 \cite{Gr00t-N1}である.
  GR00T N1は, VLMから出力されたトークンを条件として, diffusion transformerにクロスアテンションを適用し, flow matchingにより連続的なアクションを出力する.
  CogACT, TrackVLA, SmolVLA, MinDなどは, このアーキテクチャを採用している \cite{CogAct, TrackVLA, SmolVLA, MinD}.

  この7種類が主なVLAのアーキテクチャであるが, ここでは他のいくつかの重要な取り組みについて深堀りする.
  \begin{itemize}
    \item 階層型アーキテクチャ
    \item Chain-of-Thought (CoT)
  \end{itemize}

  まずは階層型のアーキテクチャである.
  最も原始的な方法は, Atomic Skill \cite{li2025atomic}やLMM-3DP \cite{LMM-3DP}のように, 既存のVLMを利用してtask instructionをsubtaskに分解してからVLAに入力するような方法である.
  これにより, 難しいtask instructionではなく, より明確で分かりやすい言語指示を元に動作を行うことができる.
  既存のVLMではなく自前で学習させたものとしてはHi Robot \cite{HiRobot}もある.
  また, NAVILA \cite{cheng2024navila}やHumanoidVLA \cite{Humanoid-VLA}のように, 下位の制御を強化学習に預ける方法もある.
  そして, この上位と下位の両者を同一のネットワークで学習する方法が, RT-H \cite{belkhale2024rth}やLoHoVLA \cite{LoHoVLA}で提案されている.
  プロンプトを変更することで, task instructionをsubtaskに分解するか, subtaskをactionに変換するかを切り替えることができる.
  これは現在, $\pi_{0.5}$ \cite{Pi-0.5}のように, subtaskへの分解, 離散アクショントークンの生成, 連続アクションの生成を同一のネットワークで学習する形へと発展している.
  難しいtask instructionの単純なsubtask instructionへの分解とVLAの融合は, 一つの重要なアプローチとなっている.
  また, 中間表現を明示的なsubtaskにせず, 潜在的な空間で高レベルポリシーと低レベルポリシーを繋げるFiS-VLA \cite{Fast-in-Slow}やOpenHelix \cite{OpenHelix}, DP-VLA \cite{DP-VLA}なども提案されている.
  加えて, 現在画像とタスク指示から未来の状態を予測するStable Video Diffusion, 視覚と言語から環境とタスク目標を理解するVLM, それらのトークンにクロスアテンションを適用して動作シーケンスを出力するPolicy Moduleという, 3つの階層を組み合わせたTri-VLAも存在する\cite{TriVLA}.

  これと似ているが少し異なるアプローチとして, CoTというreasoningの仕組みがある.
  これをVLAに取り込んだのがECoT \cite{zawalski2025ecot}とCoT-VLA \cite{zhao2025cotvla}である.
  ECoTは, VLAが観測と命令から直接アクションを出力してしまうため中間的な思考プロセスがなく, 計画や推論能力が弱いという課題に着目した取り組みである.
  そこで, 観測とタスク指示から, 自己回帰的にタスク記述やサブタスク, 物体の位置などの言語情報を予測していき, 最後にアクション系列を予測することで, VLAの性能を向上させることに成功している.
  また, CoT-VLA \cite{zhao2025cotvla}ではタスク記述などの代わりにサブゴール画像を生成することで, より視覚的タスクへの成功率を向上させた.
  さらにECoT-Lite \cite{ECoT-Lite}では, 学習時に一部のreasoning skillをスキップすることで, 実際のアクション生成時におけるreasoningによる遅延を軽減することに成功している.
  また, Fast ECoT \cite{FastECoT}では, 中間のreasoningを再利用し, アクション生成とreasoningを並列に行うことで, より高速なアクション生成を実現している.
}%

\begin{figure*}[t]
  \centering
  \includegraphics[width=0.9\textwidth]{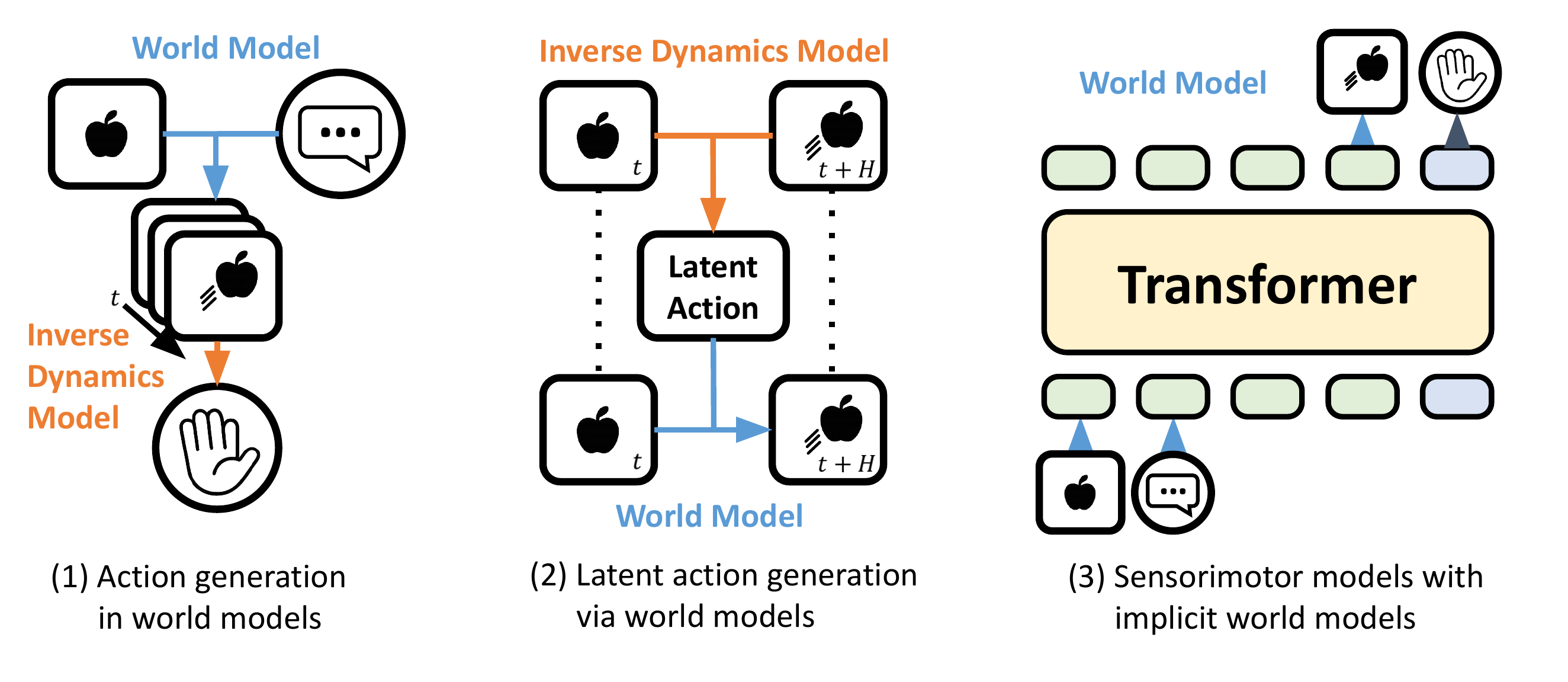}
  \caption{\textbf{Design patterns for incorporating world models in VLA.}  
(1) Using world models in conjunction with inverse dynamics models to generate actions.  
(2) Leveraging world models to learn latent action representations, particularly from human videos; the resulting latent tokens are then used for VLA training to incorporate human video datasets.  
(3) Generating future observations in addition to actions, enabling predictive planning and multimodal reasoning.}
  \label{figure:world}
\end{figure*}

\subsection{World Model}\label{subsec:world}
\switchlanguage%
{%
World models are capable of anticipating future observations or latent representations based on the current inputs.
Their forward predictive capabilities have made them increasingly central to VLA systems, where they support planning, reasoning, and control.
In this section, we group these approaches into three types, as illustrated in~\figref{figure:world}.

\textbf{(1) Action generation in world models.}
  In contrast to models that directly generate actions, world models generate future visual observations, such as images or video sequences, which are then used to guide action generation.
  For example, UniPi~\cite{du2023unipi} employs a diffusion model inspired by Video U-Net~\cite{ho2022videodiffusionmodel} to generate video sequences from an initial observation image and task instruction. 
  Then, an inverse dynamics model (IDM) translates the predicted image sequence into low-level actions. 
  This combination of visual prediction and IDM-based control is a common design pattern in model-based VLAs.
  Similarly, DreamGen~\cite{DreamGen} and GeVRM~\cite{zhang2025gevrm} predict future visual representations for action generation.
  HiP~\cite{ajay2023hip} extends this idea by incorporating subtask decomposition with a LLM, enabling the execution of longer-horizon behaviors.
  Dreamitate~\cite{liang2025dreamitate} finetunes Stable Video Diffusion~\cite{blattmann2023stablevideodiffusion} to synthesize a video of human using a tool for manipulation tasks.
  Then, given the generated video, MegaPose~\cite{labbe2022megapose} estimates the 6-DoF pose of the tool so that the robot can follow the estimated tool poses.
  In contrast to generating full video sequences, SuSIE~\cite{black2024susie} predicts abstract subgoal images by using InstructPix2Pix~\cite{brooks2023instructpix2pix} to generate intermediate goal images from the initial observation and task instruction, which are then used to condition a diffusion policy.
  CoT-VLA employs a similar approach for chain-of-thought reasoning (see \secref{subsec:sensorimotor} for further details.).
  LUMOS~\cite{nematollahi25lumos} also generates a goal image, but does so using a world model that takes low-level action commands as input.
  In LUMOS, a policy is trained to imitate expert demonstrations by interacting with the learned world model.
  
  In addition to video and image generation, many recent works leverage optical flow or feature point tracking.
  Because optical flow and feature tracking are agnostic to robot embodiment, they offer a more generalizable way to leverage human demonstrations.
  AVDC~\cite{ko2024avdc}, similar to UniPi, generates video sequences and computes optical flow for each frame using GMFlow~\cite{xu2022gmflow}. 
  It then formulates the estimation of SE(3) rigid body transformations for target objects as an optimization problem.
  ATM~\cite{wen2024atm} predicts future trajectories of arbitrary feature points (using CoTracker~\cite{karaev2024cotracker} during training), and trains a transformer that generates actions guided by these trajectories.
  Track2Act~\cite{bharadhwaj2024track2act} predicts feature point trajectories between an initial and goal image, optimizes for 3D rigid body transformations, and learns a residual policy to refine the motion.
  LangToMo~\cite{LangToMo} predicts future optical flow from an initial image and task instruction, using RAFT~\cite{teed2020raft} for optical flow supervision, and maps this prediction to robot actions.
  MinD~\cite{MinD} adopts an end-to-end approach that jointly learns video and action prediction.
  In particular, MinD combines a low-frequency video generator, which predicts future visual observations in a latent space from initial images and instructions, with DiffMatcher, which transforms these predictions into time-series features that the high-frequency action policy then uses to efficiently generate an action sequence.
  PPI~\cite{PPI} takes visual and language inputs to predict gripper poses and object displacements (Pointflow) at each keyframe. These are then used as intermediate conditions for action generation.

\textbf{(2) Latent action generation via world models.}
  This category of VLAs leverages world models to learn latent action representations from human demonstrations. 
  For example, LAPA (Latent Action Pre-training from Videos)~\cite{ye2025lapa} (see \secref{sec:vla_strategy} for details) jointly learns to predict action representations from tuples of current and future images, as well as to generate future frames conditioned on the current image and the latent action.
  This dual objective enables training on datasets without explicit action labels, such as human videos.
  Once latent actions are learned, a VLA policy is trained using these tokens. 
  The action head is then replaced and fine-tuned to output robot actions.
  LAPA has been used for pre-training in GR00T N1~\cite{Gr00t-N1} and DreamGen~\cite{DreamGen}.  
  Moreover, GO-1~\cite{AgiBotWorldColosseo} and Moto~\cite{Moto} employ a similar approach.
  UniVLA~\cite{UniVLA} augments the latent space of DINOv2~\cite{oquab2023dinov2} with language inputs and uses a two-stage training process to disentangle task-independent and task-dependent latent action tokens.
  UniSkill~\cite{UniSkill} employs image editing based approach to extract latent actions from RGB-D images and uses them as conditions for a diffusion policy.

\textbf{(3) Sensorimotor models with implicit world models.}
  This category refers to VLAs that jointly output both actions and predictions of future observations to improve performance.
  GR-1~\cite{GR-1} integrates a pre-trained MAE-ViT encoder~\cite{he2022mae}, CLIP text encoder~\cite{radford2021clip}, and a transformer, and is trained on the Ego4D dataset~\cite{grauman2022ego4d} to predict future observation images. 
  It is then fine-tuned to jointly predict both actions and future frames from image, language, and proprioceptive inputs. 
  By incorporating observation prediction, akin to a video prediction model, into a standard VLA framework, GR-1 demonstrates improved task success.
  GR-2~\cite{GR-2} builds on GR-1 by scaling up the training dataset and incorporating architectural improvements, including VQGAN-based image tokenization~\cite{esser2021vqgan} and a conditional VAE~\cite{kingma2014cvae} for action generation.
  GR-MG~\cite{li2025grmg} generates intermediate goal images using an InstructPix2Pix-based model~\cite{brooks2023instructpix2pix} and embedding them within a GR-1-style framework.
  Furthermore, GR-3~\cite{cheang2025gr3technicalreport} implements a hierarchical structure by integrating VLM (Qwen2.5-VL~\cite{bai2025qwen25vl} and diffusion transformer with flow matching for action.
  3D-VLA~\cite{3D-VLA} extends this line of work by predicting RGB-D images with Stable Diffusion~\cite{rombach2022stablediffusion} and point clouds using Point-E~\cite{nichol2022pointe}.
  Several other models incorporate full video prediction into sensorimotor models, including FLARE~\cite{zheng2025flare}, UVA~\cite{UnifiedVideoActionModel}, WorldVLA~\cite{WorldVLA}, and ViSA-Flow~\cite{visaflow}.
}%
{%
  World Modelは, ロボットが将来の状態や観測を予測するためのモデルである.
  このWorld Modelの考え方を用いたVLAが現在多く提案されてきている.
  ここでは, それらを\figref{figure:world}に示すような以下の3つのカテゴリに分類して紹介する.
  \begin{enumerate}
    \item 世界モデル構築と予測に基づくアクション生成
    \item 世界モデルを用いた潜在的なアクション表現生成
    \item sensorimotorモデルと世界モデルの統合
  \end{enumerate}

  (1)は, 主に画像の変化を予測する世界モデルを構築し, その予測をアクションへと変換するようなVLAを指す.
  このようなVLAの初期の例がUniPi \cite{du2023unipi}である.
  UniPiはまず, Video U-Net \cite{ho2022videodiffusionmodel}をベースとしたDiffusion Modelにより, 初期の観測画像とタスク指示から, 将来の行動を示す動画を予測する.
  また, 生成された画像列からロボットの制御信号を計算するInverse Dynamics Model (IDM)をCNNとMLPにより学習する.
  これにより, 初期画像とタスク指示に基づいたロボットのアクション生成が可能である.
  このような, 画像に関する世界モデルとIDMの組み合わせは, 世界モデルを用いたVLAの最も一般的なアプローチであり, 同様の方法にはDreamGen \cite{DreamGen}やGEVRM \cite{zhang2025gevrm}が存在する.
  HiP \cite{ajay2023hip}はこれに, LLMによるサブタスクへの分解を追加しており, より長期的な動作を実行することができる.
  Dreamitate \cite{liang2025dreamitate}はStable Video Diffusion \cite{blattmann2023stablevideodiffusion}をベースとして人間の道具操作に関する世界モデルを学習, 生成された動画と道具のCADモデル用いてMegaPose \cite{labbe2022megapose}でツールの6DOF姿勢を推定, ロボットはその動きを実現する.
  また, 動画ではなく未来のサブゴールという, より抽象的な目標を世界モデルで生成するのがSuSIE \cite{black2024susie}である.
  画像編集モデルであるInstructPix2Pix \cite{brooks2023instructpix2pix}によって初期画像とタスク指示から中間的な目標画像を生成し, それをDiffusion Policyの条件として活用する.
  同様の手法には, LUMOS \cite{nematollahi25lumos}や\secref{subsubsec:sensorimotor}の最後でも説明したCoT-VLA \cite{zhao2025cotvla}などが含まれる.
  このような動画や画像の生成だけでなく, 現在はオプティカルフローや特徴点トラッキングを活用した世界モデル型のVLAが多く提案されている.
  オプティカルフローや特徴点トラッキングはロボット固有の身体性に依存せず, 人間のデモンストレーションなども効果的に扱えることから, より汎用的なアプローチとして注目されている.
  AVDC \cite{ko2024avdc}はUniPiと同様に動画を生成し, 各フレームに対してGMFlow \cite{xu2022gmflow}を用いてオプティカルフローを計算, これを満たすような対象オブジェクトの3D剛体変換を最適化問題として解くことで, 直接的にSE3のアクションを出力する.
  ATM \cite{wen2024atm}は動画ではなく任意の特徴点の将来軌道を予測し(学習時はCoTracker \cite{karaev2024cotracker}を用いる), これをガイドとしてアクションを生成するtransformerを学習する.
  Track2Act \cite{bharadhwaj2024track2act}は初期画像とゴール画像を入力として特徴点の将来軌道を予測, 最適化により対象オブジェクトの3D剛体変換を計算, さらにこれを微調整するresidual policyも学習する.
  LangToMo \cite{LangToMo}はRAFT \cite{teed2020raft}により生成したオプティカルフローを教師として, 初期画像とタスク指示から将来のオプティカルフローを予測するモデルを構築, これをロボットのアクションに結びつける.
  MinD \cite{MinD}は世界モデルとIDMを統合してEnd-to-Endに学習するアプローチであり, LoDiffモデルが初期画像とタスク指示から低頻度で未来の視覚観測の時系列潜在空間を予測, DiffMatcherがそれらを時系列特徴量に変換, HiDiffモデルがそれを条件に高速にアクション系列を生成する.
  PPI \cite{PPI}は画像や言語から各キーフレームにおけるグリッパの姿勢とオブジェクトの位置変化(Pointflow)を予測し, これを中間条件としてアクションを生成する.
  現在はVeo3 \cite{googledeepmind2025veo3}のように性能の高い動画生成モデルが登場していることから, この世界モデルベースのVLAは今後さらなる発展が期待される.

  (2)は, 世界モデルの考え方を用いて, 人間のデモンストレーションからその潜在的なアクション表現を学習し, 利用するVLAを指す.
  この代表的な方法は, \secref{subsec:history}でも述べたLAPA (Latent Action Pre-training from Videos) \cite{ye2025lapa}である.
  現在と将来の画像からのアクション予測と, 現在の画像とアクションからの将来予測を同時に学習することで, 人間のデモンストレーションのような明示的なアクションが無いデータセットもVLAに活用できるようになる.
  ここで得られた潜在的なアクショントークン用いて, 人間のデモンストレーションからVLAを学習した後に, Action Headを置き換えて実際のロボットのアクションが出るような微調整を行う.
  LAPAはGR00T N1 \cite{Gr00t-N1}やDreamGen \cite{DreamGen}の事前学習に利用されている.
  この他にも, GO-1 \cite{AgiBotWorldColosseo}で使用されているLAM (Latent Action Model)やMoto (Latent Motion Tokenizer) \cite{Moto}も似た考え方を用いている.
  また, DINOv2 \cite{oquab2023dinov2}の潜在空間内で言語入力も追加して構築し, かつ2段階の学習によってタスク非依存とタスク依存のアクショントークンに分けて学習する方法がUniVLA \cite{UniVLA}で提案されている.
  さらに, UniSkill \cite{UniSkill}では深度画像の利用, アクションの連続値化, アクションによる画像編集としての定式化によって, Diffusion Policyの条件として直接使うことが可能な潜在的アクションを抽出している.

  (3)は, sensorimotor modelの出力として, アクションのみならず観測の予測を含めることで性能を向上させるようなVLAを指す.
  このようなVLAの代表的な例がGR-1 \cite{GR-1}である.
  GR-1は, 事前学習済みのMAE-ViT \cite{he2022mae}のEncoderとCLIP \cite{radford2021clip}のText Encoder, transformerによって構成されたモデルについて, Ego4D \cite{grauman2022ego4d}のデータセットから将来の観測画像を予測するように学習する.
  その後, 画像と言語, proprioceptionから, actionと未来画像を同時に予測できるようにfine tuningを行う.
  通常のVLAに世界モデルのような観測の予測を組み込むことで, VLAの性能を向上させることができる.
  GR-2 \cite{GR-2}はGR-1からデータ規模を大幅に拡大し, VQGAN \cite{esser2021vqgan}による画像のトークン化やアクション生成時のconditional VAE \cite{kingma2014cvae}の利用などの改善を施している.
  GR-MG \cite{li2025grmg}は(1)の考え方を組み合わせているが, InstructPix2Pixベースのモデルにより中間のゴール画像を生成し, これをGR-1に似たアーキテクチャと統合している.
  GR-3はさらに, flow matchingの導入と, VLM (Qwen2.5-VL)とdiffusion transformerによる階層化を行っている.
  3D-VLA \cite{3D-VLA}はVLAに対してStable Diffusion \cite{rombach2022stablediffusion}によるRGBD画像の予測とPoint-E \cite{nichol2022pointe}によるポイントクラウドの予測を追加している.
  同様の形で, 動画生成を組み込んだFLARE \cite{zheng2025flare}やUVA, WorldVLA \cite{WorldVLA}, ViSA-Flow (Video Semantic Action Flow) \cite{visaflow}を組み込んだモデルなどが開発されている.
}%

\begin{figure*}[t]
  \centering
  \includegraphics[width=0.9\textwidth]{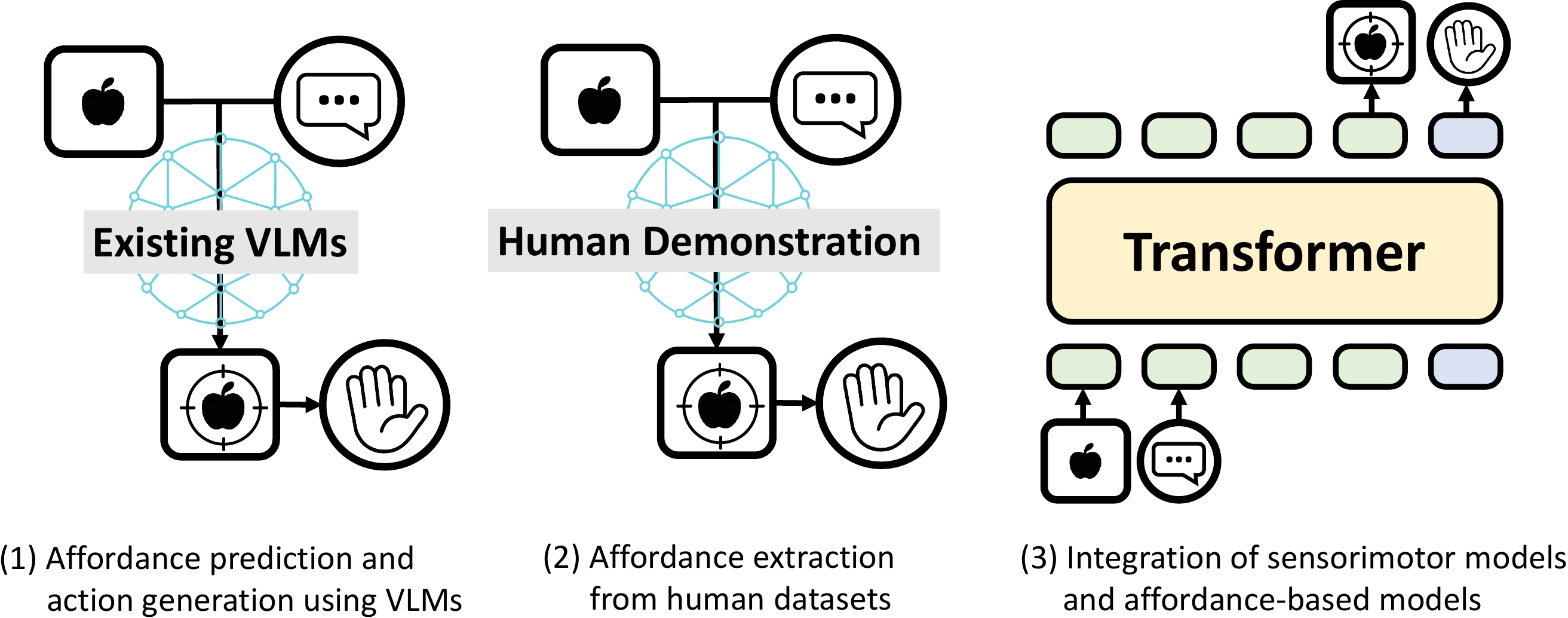}
  \caption{
    \textbf{Design patterns for incorporating affordance-based models in VLA.} (1) Predicting affordances and subsequently generating actions conditioned on the predicted affordances;
    (2) Extracting affordances from human demonstration videos and learning latent representations to guide action generation;
    (3) Integrating affordance prediction modules directly into the VLA architecture.}
  \label{figure:affordance}
\end{figure*}

\subsection{Affordance-based Model}\label{subsec:affordance}
\switchlanguage%
{%
Affordances~\cite{gibson} refer to the action possibilities that an environment offers an agent, relative to its physical and perceptual capabilities. In robotics, this concept is often adapted to denote the actionable properties of objects or scenes, specifically, what actions are possible given the robot's embodiment and the spatial or functional cues present.
VLAs based on affordance prediction can be currently categorized into three types, as illustrated in \figref{figure:affordance}.

\textbf{(1) Affordance prediction and action generation using VLMs.}
Pre-trained VLMs are often used to estimate affordances and generate corresponding actions.
For example, VoxPoser~\cite{huang2023voxposer} uses GPT-4~\cite{openai2024gpt4technicalreport}, OWL-ViT~\cite{minderer2022owlvit}, and Segment Anything~\cite{kirillov2023sam} to generate Affordance and Constraint Maps from language instructions, which are then used to guide action generation via Model Predictive Control (MPC).
KAGI~\cite{lee2024kagi} employs GPT-4o~\cite{openai2024gpt4ocard} to infer a sequence of target keypoints from top-down and side-view images with overlaid grid lines, providing guidance for RL.
LERF-TOGO~\cite{rashid2023lerftogo} builds a 3D scene using a NeRF~\cite{mildenhall2020nerf} trained on visual features extracted from CLIP and DINO~\cite{radford2021clip, caron2021dino} (LERF~\cite{kerr2023lerf}). 
CLIP's text encoder is used to compute similarity between language instructions and visual features, and high-activation regions are converted into 3D point clouds, which are then processed by GraspNet~\cite{fang2020graspnet} to rank grasp poses.
Splat-MOVER~\cite{shorinwa2024splatmover} replaces NeRF with Gaussian Splatting~\cite{kerbl2023gaussiansplatting} for faster scene construction and incorporates affordance heatmaps from the VRB model~\cite{bahl2023vrb}, improving both efficiency and performance.

\textbf{(2) Affordance extraction from human datasets.}
This line of work focuses on extracting affordances from human motion videos, often without annotations, to enable scalable learning for robotic action generation.
VRB~\cite{bahl2023vrb} learns contact points and hand trajectories from demonstration videos in the EPIC-KITCHENS datasets~\cite{damen2018epickitchens, damen2022epickitchens100}. 
In VRB, Hand-Object Detector (HOD)~\cite{shan2020hod} is used to identify hand positions and contact states, then tracks subsequent hand movements on the image plane to automatically construct a training dataset. 
The extracted data are projected into 3D and used to generate robot actions.
HRP~\cite{srirama2024hrp} extracts hand, contact, and object affordance labels from the Ego4D dataset~\cite{grauman2022ego4d}, trains a ViT model to predict these labels, and uses its latent representations for imitation learning.
VidBot~\cite{VidBot} extends 2D affordance representations to 3D, aiming to support zero-shot deployment on robots.

\textbf{(3) Integration of sensorimotor models and affordance-based models.}
This approach incorporates affordance prediction into VLA.
CLIPort~\cite{shridhar2021cliport} predicts affordances of objects and the environment from visual and language inputs, and generates actions based on these affordances.
RoboPoint~\cite{yuan2025robopoint} builds a vision-language model that identifies affordance points, specific locations in an image where the robot should act, which are then projected into 3D to generate corresponding actions.
RoboGround~\cite{huang2025roboground} predicts masks for the target object and placement area in pick-and-place tasks given image and language inputs; RT-Affordance~\cite{RT-Affordance} outputs key end-effector poses at critical moments; $A_0$\cite{A_0} predicts object contact point trajectories; and RoboBrain~\cite{ji2025robobrain} identifies affordance regions as bounding boxes. 
Collectively, these models leverage affordance information as conditioning input for action generation.
Chain-of-Affordance~\cite{Chain-of-Affordance}, inspired by Chain-of-Thought reasoning (see \secref{subsec:sensorimotor}), predicts a sequence of affordances such as object positions, grasp points, and placement locations in an autoregressive manner, and then generates actions, leading to improved performance.  
}%
{%
  アフォーダンスとは, ある環境や物体が, ロボットに対して提供する行為の可能性のことである.
  このようなアフォーダンス予測に基づくVLAは, \figref{figure:affordance}に示すように, 以下の3つのカテゴリに分類される.
  \begin{enumerate}
    \item 既存のVLMを用いたアフォーダンス予測とアクション生成
    \item 人間のデータセットからのアフォーダンス抽出
    \item sensorimotorモデルとアフォーダンスモデルの統合
  \end{enumerate}
  これら(1)--(3)は, \secref{subsubsec:world}の(1)--(3)に対応している.

  (1)は, 既存のVLMを用いてアフォーダンスを予測し, そのアフォーダンスに基づいてアクションを生成するVLAである.
  その代表的な例がVoxPoser \cite{huang2023voxposer}である.
  VoxPoserは既存のLLM (GPT-4 \cite{openai2024gpt4technicalreport})やVLM (OWL-ViT \cite{minderer2022owlvit}やSegment Anything \cite{kirillov2023sam})を活用して言語指示からAffordance MapとConstraint Mapを構築, これを条件としてモデル予測制御によりアクションを生成する.
  また, KAGI \cite{lee2024kagi}ははGPT-4o \cite{openai2024gpt4ocard}を用いて, グリッド線の引かれたトップダウンな画像とサイド画像上のどこを順番に訪れるべきか推定し, これをガイダンスとして強化学習を行うことで, ロボットのアクションを生成する.
  このように, 既存のLLMやVLMを用いることでアフォーダンスを推定することが可能になってきており, これをアクション生成に活用することができている.
  これらと違ったアプローチには, LERF-TOGO \cite{rashid2023lerftogo}とSplat-MOVER \cite{shorinwa2024splatmover}などがある.
  LERF-TOGOはNeRF \cite{mildenhall2020nerf}を訓練し, 3DシーンとCLIP/DINO \cite{radford2021clip, caron2021dino}に基づく視覚特徴を構築する(LERF \cite{kerr2023lerf}).
  次にCLIPのTex Encoderを用いて言語指示と視覚特徴量の類似度を計算し, 高活性点を取得, NeRFから3D点群を得る.
  この点群に対してGraspNet \cite{fang2020graspnet}を適用し, ランキング付けして把持姿勢を計算する.
  さらに, Splat-MOVERはNeRFの代わりにGaussian Splatting \cite{kerbl2023gaussiansplatting}を用いて高速にシーンを構築, 後述するアフォーダンス抽出が可能なVRBのaffordance heatmapも埋め込み, さらなる性能向上を実現している.

  (2)は人間の動作データセットから自動的にアフォーダンスを抽出しモデル構築に用いる.
  その代表的な例がVRB \cite{bahl2023vrb}である.
  VRBは人間のデモンストレーション動画から環境や物体への接触点と接触後の手の軌道を学習し, それをロボットの行動生成に活かす取り組みである.
  EPIC-KITCHENSデータセット \cite{damen2018epickitchens, damen2022epickitchens100}について, 手の位置と接触状態を検出可能なHODモデル (Hand-Object Detector) \cite{shan2020hod}により接触点を抽出, その後の画像平面上での手の軌道を収集することで, 自動的にデータセットが構築でき, これを用いてアフォーダンスモデルを学習する.
  得られた接触点と軌道を3次元空間に射影し, ロボットのアクションに変換することができる.
  また, HRP \cite{srirama2024hrp}はEgo4Dデータセット \cite{grauman2022ego4d}から手・接触点・対象物のアフォーダンスラベルを抽出, これらを推論するViTを学習し, その潜在表現を用いて摸倣学習を行う.
  VidBot \cite{VidBot}はこれまでの画像上の2次元のアフォーダンスを3次元に拡張し, ゼロショットでロボットに展開できるようにする取り組みである.
  どれも人間の単純なデモンストレーション動画から, アノテーション無しにアフォーダンスモデルを構築することができており, スケールしやすいというメリットがある.

  (3)はsensorimotorモデルにアフォーダンスモデルを組み込むことで, アクションを生成する, またはVLAの性能を向上させるアプローチである.
  この代表的な例がCLIPort \cite{shridhar2021cliport}である.
  画像と言語から, 直接のアクションではなく物体と環境のアフォーダンスを出力し, これを元にアクションを生成している.
  また, RobotPoint \cite{yuan2025robopoint}は画像と言語から画像内でどこに動くべきかというアフォーダンスポイントを予測するVLMを構築し, これを3次元に投影してロボットのアクションを生成する.
  RoboGround \cite{huang2025roboground}は画像と言語からピックアンドプレイスタスクにおける対象物とその設置場所のマスクを生成, RT-Affordance \cite{RT-Affordance}は重要なタイミングでのエンドエフェクタのキーポーズを予測, $A_0$ \cite{A_0}は物体の接触点軌道を予測, RoboBrain \cite{ji2025robobrain}はバウンディングボックス形式のアフォーダンス領域を予測し, これらのアフォーダンスを条件としたアクションポリシーを学習している.
  さらに, Chain-of-Affordance \cite{Chain-of-Affordance}は\secref{subsubsec:sensorimotor}のChain-of-Thought (CoT)を用いたVLAと似ており, 対象物の位置や掴む部位, 置く位置などのアフォーダンスを自己回帰的に順番に予測していき, その後にアクションを生成することで, VLAの性能を向上させることができる.
}%

\subsection{Data Modalities}\label{subsec:modality}
\switchlanguage%
{%
VLAs process multiple modalities simultaneously, including vision, language, and action. This section summarizes how each modality is handled in state-of-the-art systems.
}%
{%
  VLAでは, 画像・言語・アクション・その他のモダリティを扱う.
  ここでは, それぞれのモダリティを現在のVLAがどのように扱っているかについてまとめる.
}%

\subsubsection{Vision}\label{subsubsec:vision}
\switchlanguage%
{%
The most common approach for visual feature extraction in VLAs is to use ResNet~\cite{he2016resnet} or Vision Transformer (ViT)~\cite{dosovitskiy2021vit}. 
These models are typically pre-trained on large-scale datasets such as ImageNet~\cite{deng2009imagenet, russakovsky2015imagenet1k} or LAION~\cite{schuhmann2021laion400m, schuhmann2022laion5b}, although ResNet is often trained from scratch.
Some methods apply ResNet directly to the image and convert the output into tokens using an MLP, while others first divide the image into patches before applying the encoder.
Furthermore, ViT pre-trained with MAE~\cite{he2022mae} and EfficientNet~\cite{tan2019efficientnet} are also commonly used.

Vision-language models such as CLIP~\cite{radford2021clip} and SigLIP~\cite{zhai2023siglip} are also widely used. CLIP learns joint visual and textual representations via contrastive learning, while SigLIP improves upon it by removing the softmax constraint and reducing sensitivity to batch size. 
These models are often used alongside DINOv2~\cite{oquab2023dinov2}, a self-supervised vision model that learns image features without requiring paired text or contrastive objectives.
%
%
While CLIP was initially the dominant choice, SigLIP and DINOv2 have emerged as the preferred models for visual feature extraction in VLAs. 
OpenCLIP~\cite{cherti2023openclip} and EVA-CLIP~\cite{sun2023evaclip} are also adopted in several prior works.

In addition, VQ-GAN~\cite{esser2021vqgan} and VQ-VAE~\cite{van2017vqvae} are commonly used for discretizing images into token sequences.
Unlike ViT or CLIP, which produce continuous embeddings, these models generate discrete tokens that are more naturally aligned with the input format of LLMs.
%
%
The resulting visual tokens are often further processed to integrate with other modalities or to reduce token length.
A well-known example is the Perceiver Resampler from Flamingo~\cite{alayrac2022flamingo}, which compresses visual information using a fixed-length set of learnable latent tokens via cross-attention.
Building on this idea, Q-Former in BLIP-2~\cite{li2023blip2} combines cross-attention and self-attention to extract task-relevant information, while QT-Former~\cite{fu2025orion} incorporates temporal structure into the process.
TokenLearner~\cite{ryoo2021tokenlearner} takes a different approach by performing spatial summarization to reduce token count.
These compression and integration techniques are widely used in VLAs.

Several works in VLA adopt object-centric features, such as bounding box coordinates or cropped region embeddings, instead of relying solely on continuous feature maps. These features are typically extracted using object detection, segmentation, or tracking models, including Mask R-CNN~\cite{he2017maskrcnn}, OWL-ViT~\cite{minderer2022owlvit}, SAM~\cite{kirillov2023sam}, GroundingDINO~\cite{liu2024groundingdino}, Detic~\cite{zhou2022detic}, and Cutie~\cite{cheng2024cutie}.
}%
{%
  最も一般的に用いられている画像の特徴量抽出はResNet \cite{he2016resnet}またはVision Transformer (ViT) \cite{dosovitskiy2021vit}を用いたものである.
  これらは, ImageNet \cite{deng2009imagenet, russakovsky2015imagenet1k}やLAION \cite{schuhmann2021laion400m, schuhmann2022laion5b}などの大規模な画像データセットで事前学習されている場合もあれば, 特にResNetはスクラッチから学習される場合も多い.
  画像をそのままResNetに入れてMLPによりトークン成形する場合や, パッチに分割した後にResNetを適用する場合などのバラエティもある.
  派生としては, MAE \cite{he2022mae}により学習されたViT, EfficientNet \cite{tan2019efficientnet}などがある.

  また, 事前学習済みのVLMであるCLIP \cite{radford2021clip}やSigLIP \cite{zhai2023siglip}も頻繁に用いられる.
  CLIP \cite{radford2021clip}は画像とテキストの特徴量をContrastive Learningにより学習したモデルであり, SigLIP \cite{SigLIP}はCLIPのSoftmax制約やバッチ依存性を改良したものである.
  これに併用されて, DINOv2 \cite{oquab2023dinov2}が用いられている.
  DINOv2は画像単体における知識蒸留による自己教師あり学習により, 画像の特徴量を計算するモデルである.
  初期はCLIPが頻繁に用いられていたが, 現在はSigLIPとDINOv2が主流となっている.
  これらの派生としては, OpenCLIP \cite{cherti2023openclip}やEVA-CLIP \cite{sun2023evaclip}などが用いられている.

  また, 画像の離散的なトークン化にはVQ-GAN \cite{esser2021vqgan}やVQ-VAE \cite{van2017vqvae}が用いられている.
  これらはViTやCLIPなどと違い離散的なトークン列を出せるため, LLMと相性が良い場合がある.
  また, 画像のトークン化だけでなく, 画像の生成にも用いることができる点も重要である.

  これらの得られたトークンは, 他のモダリティと統合したり, トークン数を圧縮するために, 別の処理を施す場合がある.
  その代表的な例はFlamingo \cite{alayrac2022flamingo}で提案されたPerceiver Resamplerである.
  固定長のlearnable latent tokenを用いてクロスアテンションを行い, 視覚トークン情報を圧縮している.
  この派生系として様々な方法が行われており, BLIP2 \cite{li2023blip2}で提案されたQ-Formerはクロスアテンションとセルフアテンションを通して, 必要な情報を抽出している.
  さらに時系列を考慮したQT-Former \cite{fu2025orion}や, 空間的な要約を行いトークン数を圧縮するTokenLearner \cite{ryoo2021tokenlearner}などが提案されており, これらも頻繁に用いられている.

  研究によっては, 特徴量を直接取り出すのではなく, 物体認識におけるバウンディングボックスの情報を用いたり, 切り出された画像の特徴量を用いる場合もある.
  これらの場合は, Mask R-CNNやOWL-ViT, SAM, GroundingDINO, Detic, Cutieなどの物体検出モデルやセグメンテーションモデル, 追跡モデルが用いられる \cite{he2017maskrcnn, minderer2022owlvit, kirillov2023sam, liu2024groundingdino, zhou2022detic, cheng2024cutie}.
}%

\subsubsection{Language}\label{subsubsec:language}
\switchlanguage%
{%
For language tokenization, VLAs typically inherit the tokenizer from their underlying LLM backbone, such as the T5 tokenizer~\cite{raffel2020t5} or LLaMA tokenizer~\cite{touvron2023llama2}. 
When the base model is not a pre-trained LLM, tokenization is typically performed using subword algorithms such as Byte-Pair Encoding (BPE) or tools like SentencePiece~\cite{kudo2018sentencepiece}, which implements BPE as well as other algorithms.
For language encoding, VLAs employ various text encoders to embed natural language instructions into vector representations, including the Universal Sentence Encoder (USE)~\cite{cer2018use}, CLIP Text Encoder~\cite{radford2021clip}, Sentence-BERT~\cite{reimers2019sentencebert}, and DistilBERT~\cite{sanh2020distilbert}. 
These language embeddings are frequently used to condition visual features via techniques such as FiLM conditioning.
In architectures that use VLMs as backbones, visual information is directly integrated into the LLM component, with popular choices including LLaMA 2~\cite{touvron2023llama2}, Vicuna~\cite{chiang2023vicuna}, Gemma~\cite{gemmateam2024gemma}, Qwen2~\cite{yang2024qwen2}, Phi-2~\cite{javaheripi2023phi}, SmolLM2~\cite{allal2025smollm2}, GPT-NeoX~\cite{black2022gptneox20b}, and Pythia~\cite{biderman2023pythia}.
}%
{%
  まず言語のtokenizerについては, 基本的にT5 Tokenizer~\cite{raffel2020t5}やLLaMA Tokenizer \cite{touvron2023llama2}など, 後段のLLMに合わせたものが用いられる.
  後段がLLMなどの事前学習されたものでない場合は, BPE (Byte-pair Encoding)などのアルゴリズムやSentencePiece \cite{kudo2018sentencepiece}などのツールを用いたトークン化が行われる.
  この他にも, テキスト指示をベクトル化するために, Universal Sentence Encoder (USE) \cite{cer2018use}やCLIP \cite{radford2021clip}のText Encoder, Sentence-BERT \cite{reimers2019sentencebert}, DistilBERT \cite{sanh2020distilbert}などが頻繁に用いられている.
  これらは, FiLM conditioningを通して画像と条件づけられる場合が多い.
  また, この後に説明するVision-Language Modelがbackboneとして用いられる場合と, LLaMA2やVicuna, Gemma, Qwen2, Phi-2, SmolLM2, GPT-NeoX, PythiaのようなLLMに, 自分で画像情報を埋め込む場合がある \cite{touvron2023llama2, chiang2023vicuna, gemmateam2024gemma, yang2024qwen2, javaheripi2023phi, allal2025smollm2, black2022gptneox20b, biderman2023pythia}.
}%

\subsubsection{Action}\label{subsubsec:action}
\switchlanguage%
{%
Action representation in end-to-end VLA models can be categorized into several primary approaches. 
This classification excludes specialized architectures such as affordance-based or world model-based methods.

\textbf{Discretized action tokens obtained via binning.}
  The most common approach to representing actions in VLAs is to discretize each dimension of the action space into bins (typically 256), with each bin ID treated as a discrete token. 
  For example, RT-2 with PaLI-X~\cite{zitkovich2023rt2, chen2024palix} directly outputs numeric tokens as actions; and RT-2 with PaLM-E~\cite{driess2023palme} and OpenVLA~\cite{kim2024openvla} reserve the 256 least frequent tokens in the vocabulary for action representation.
  These models are typically trained using cross-entropy loss and adopt autoregressive decoding, similar to LLMs.  
  Several models instead use non-autoregressive decoding, by inserting a readout token to enable parallel generation of all action tokens~\cite{zhao2025cotvla}, or by treating the final few output tokens as discretized arm and base action (as in RT-1).
  A known drawback of standard binning is the increase in token length, which can limit control frequency. To mitigate this, FAST~\cite{Pi-0-FAST} applies the Discrete Cosine Transform (DCT) along the temporal axis, quantizes the frequency components, and compresses them using Byte-Pair Encoding (BPE). This significantly reduces token length and enables faster inference compared to conventional binning.

\textbf{Decoding tokens into continuous actions.}
  In this approach, tokens generated by a transformer are mapped to continuous actions via a multilayer perceptron (MLP), typically trained with an L2 or L1 loss. 
  For binary outputs such as gripper open/close, binary cross-entropy is often used.
  OpenVLA-OFT~\cite{OpenVLA-OFT} suggests that L1 loss may yield better performance.
  The MLP decoder can be replaced by alternative modules, such as an LSTM~\cite{hochreiter1997lstm} to incorporate temporal context, or a Gaussian Mixture Model (GMM) to model stochasticity in the action space.
  Proprioceptive or force signals are often incorporated into the decoding module, such as an MLP or LSTM. 
  Non-autoregressive variants commonly apply pooling operations (e.g., average or max pooling) to compress multiple tokens into a single action representation, as seen in RoboFlamingo~\cite{RoboFlamingo}. 
  OpenVLA-OFT~\cite{OpenVLA-OFT} extends this by predicting multi-step action chunks, resulting in smoother and more temporally coherent trajectories.

\textbf{Continuous action modeling via diffusion or flow matching.}
  Diffusion models and flow matching have become prominent approaches for generating continuous actions in VLAs, as seen in Octo~\cite{octoteam2024octo} and $\pi_0$~\cite{Pi-0}. 
  These models generate actions non-autoregressively, enabling smoother and more scalable control.
  Flow matching is particularly suitable for real-time applications, as it requires fewer inference steps than traditional diffusion. While some models implement diffusion as an external action head after the transformer, recent designs increasingly embed the process within the transformer itself, for example, in diffusion transformer architectures.
  Training and inference are commonly based on DDPM~\cite{ho2020ddpm} and DDIM~\cite{song2021ddim}, with improved performance in stability and efficiency offered by methods such as TUDP~\cite{TUDP}, which ensure denoising consistency at every time step.

\textbf{Learning latent action representations from web-scale data.}
  This approach utilizes world modeling to obtain latent action representations when explicit actions are unavailable, such as in human demonstrations. By leveraging web-scale video data, this method enables training on significantly larger datasets and facilitates learning more generalizable VLAs. LAPA~\cite{ye2025lapa}, Moto~\cite{Moto}, UniVLA~\cite{UniVLA}, and UniSkill~\cite{UniSkill} demonstrate this approach. For additional details, see Section \ref{subsec:world}.

\textbf{Alternative action representation.} 
  SpatialVLA~\cite{SpatialVLA} statistically discretizes the action space and reduces the number of spatial tokens by allocating higher resolution to frequently occurring motions.
  ForceVLA~\cite{ForceVLA} and ChatVLA~\cite{ChatVLA} employ Mixture of Experts (MoE) architectures to dynamically switch action policies based on task phases. 
  iManip~\cite{iManip} enables continual learning by incrementally adding learnable action prompts, preserving prior skills while acquiring new ones.

\textbf{Cross-embodiment action representation.}
  The challenge of embodiment diversity arises when handling robot-specific modalities such as actions and proprioception. 
  Open X-Embodiment Project~\cite{oneill2024openxembodiment} was the first to tackle this embodiment challenge. 
  Building upon the RT-1~\cite{brohan2023rt1} and RT-2~\cite{zitkovich2023rt2} architectures, this work standardized datasets across different robots using a unified format: single camera input, language instructions, and 7-DoF actions (position, orientation, and gripper open/close). 
  This approach demonstrates a key insight that integrating data from robots with diverse embodiments leads to significantly improved VLA model performance compared to training on a single embodiment.
  Moreover, another prior work~\cite{yang2024extreme} has proposed to normalize and align actions and observations from heterogeneous embodiments into a shared first-person perspective, thereby enabling unified control of various robots using only observations and goal images~\cite{yang2024extreme}.
  However, such approaches struggle to uniformly handle robots with drastically different observations or control inputs, such as manipulators, mobile robots, and legged robots.
  
  To address this limitation, CrossFormer~\cite{doshi2025crossformer} enables unified processing across diverse embodiments by first tokenizing heterogeneous sensor observations—such as vision, proprioception, and task specifications—using modality-specific tokenizers. 
  All tokens are then assembled into a unified token sequence, with missing modalities masked as needed.
  This sequence is processed by a shared decoder-only transformer, which uses readout tokens to extract task-relevant representations. 
  These are subsequently passed to embodiment-specific action heads (e.g., single-arm, bimanual, navigation, or quadruped) to generate actions tailored to each robot type.
  UniAct~\cite{zheng2025uniact} proposes a Universal Action Space (UAS) implemented as a discrete codebook shared across embodiments. 
  A transformer predicts discrete action tokens from this codebook, which are then converted into continuous actions by embodiment-specific decoders. 
  By explicitly defining a shared atomic action space, UniAct facilitates knowledge transfer and promotes reusability across diverse robot embodiments.
  Furthermore, UniSkill~\cite{UniSkill} incorporates human demonstration knowledge by extracting latent skill representations from unlabeled human video data, in addition to robot data, similar to LAPA~\cite{ye2025lapa}, enabling more generalizable VLA models. 
  Additionally, embodiment-agnostic frameworks such as LangToMo~\cite{LangToMo} and ATM~\cite{wen2024atm} achieve cross-embodiment learning by leveraging intermediate representations, such as optical flow and feature point trajectories, thereby bypassing the need for direct action space alignment.
}%
{%
  アクションの扱い方は主に以下の4つに分類できる.
  なお, ここではAffordanceやWorld Modelのような特殊なケースは除外し, End-to-EndなVLAについてのみ考える.
  \begin{enumerate}
    \item ビン分割された離散的なアクショントークンを用いる
    \item トークンを連続的なアクションにマッピングする
    \item 拡散モデルやフローマッチングを用いた連続的なアクションを用いる
    \item ウェブ上のデータから潜在的なアクション表現を学習する
  \end{enumerate}

  (1)は最も典型的なアクションの扱い方である.
  アクションの各次元を離散的なビン(大抵256個)に分割し, そのビンのIDをトークンとして扱う.
  そのまま数字のトークンをアクションとして出力する方法(RT-2 \cite{zitkovich2023rt2}のPaLI-X \cite{chen2024palix})や, 最も使用頻度の低いトークン256個をアクション用に割り当てる方法(RT-2のPaLM-E \cite{driess2023palme}やOpenVLA \cite{kim2024openvla})が用いられている.
  基本的にクロスエントロピー損失を用いて学習される.
  この方法はLLMやVLMとの相性が良いため, 最も一般的に用いられている.
  LLMと同様に自己回帰的にトークンを生成するのが一般的であるが, 予め入力にreadout tokenを追加しておき非自己回帰的に一括でアクショントークンを出力する方法や, 最後に出力される複数のトークンをアクショントークンとして利用する(RT-1)ような, 非自己回帰的な方法も存在する.
  加えて, 通常のビン分割ではトークン数の増大により高周期の制御が難しいため, これを改善するFAST \cite{Pi-0-FAST}と呼ばれるトークン化方法が提案されている.
  これは, 離散コサイン変換(DCT)を用いて時間方向のアクション信号を周波数領域に変換し, 得られた周波数成分を量子化して, BPEで圧縮してトークン化する方法である.
  ビン分割に比べて圧倒的にトークン数を減らすことができるため, 推論の高速化が可能である.

  (2)はtransformerから出力されたトークンの値をMLPによって連続的なアクションの形に成形し, L2またはL1損失を用いて学習する方法である(グリッパの開閉についてはbinary cross entropyで学習される場合が多い, L1の方が良いと主張する論文もある[OpenVLA-OFT]) \cite{OpenVLA-OFT}.
  MLPの箇所については, LSTM \cite{hochreiter1997lstm}を用いて履歴を考慮する場合や, GMM (Gaussian Mixture Model)を用いて確率的な挙動を考慮する場合もある.
  proprioceptionや力の状態はこれらのMLPやLSTMの入力として再度用いられることもある.
  こちらも(1)と同様に自己回帰的な場合と非自己回帰的な場合が存在している.
  非自己回帰的な場合は, 複数のトークンを統合するために, max poolingやaverage poolingを用いる場合が多い(RoboFlamingo \cite{RoboFlamingo}).
  また, 1ステップ分のアクションだけではなく, アクションチャンクとして複数ステップ分を一括で出力する場合もある(OpenVLA-OFT \cite{OpenVLA-OFT}).
  これにより, より滑らかなアクションを出力することができる.

  (3)の拡散モデルやフローマッチングはOcto \cite{octoteam2024octo}や$\pi_0$ \cite{Pi-0}などで用いられている, 現在主流となりつつある方法である.
  連続値を非自己回帰的に出力することができる.
  特にFlow Matchingは, 拡散モデルに比べてイテレーション回数が少なく, 推論時間が短いことから, よりリアルタイム性の高いアクション出力が可能である.
  これらはtransformerの後段にAction Headとして追加されることが多いが, 現在はdiffusion transformerのように, transformerの中で拡散過程を実行する方法も多く提案されている.
  拡散モデルは基本的にDDPM \cite{ho2020ddpm}やDDIM \cite{song2021ddim}により学習、推論するが, この他にもTUDP \cite{TUDP}のように全時刻を通して一貫したdenoisingを行うことで推論時間の短縮や学習の安定化を図る方法も提案されている.
  また, リアルタイム推論の課題に対処するための非同期なチャンク生成であるReal-Time Chunking \cite{RealTimeChunking}という手法が提案されている.
  遅延により実行されることが確定されている動作を固定してそれに整合するように非同期で残りのアクションを生成し, soft maskingにより過去チャンクとの滑らかな連続性を保ちつつ新しい観測に基づいた補完を実現している.

  (4)は, 人間のデモンストレーションのような, アクションが直接わからない場合に, その潜在的な表現を世界モデルの学習を活用し行う方法である.
  これにより, さらに多くのデータを事前学習に活用することができ, より汎用的なVLAを学習することが可能になる.
  LAPA \cite{ye2025lapa}やMoto \cite{Moto}, UniVLA \cite{UniVLA}, UniSkill \cite{UniSkill}などがその例である.
  なお, 詳細については重複するため\secref{subsubsec:world}の(2)を参照されたい.

  これらの他にも, アクション空間を統計的に離散化し, 空間行動トークンとしてトークン数を削減し高頻度な動きに解像度を集中させる方法(SpatialVLA \cite{SpatialVLA})や, MoE (Mixture of Experts)を用いてタスクフェーズごとにアクションを動的に変化させる方法(ForceVLA \cite{ForceVLA}, ChatVLA \cite{ChatVLA}), learnable action promptの逐次的な追加により既存スキルを維持しながら新しいスキルを学んでいくアクションの学習方法(iManip \cite{iManip})も提案されている.
  また, \cite{szot2024mllm}では, いくつかのアクション空間について比較を行い, Residual Vector Quantizer (RVQ)のパフォーマンスが最も良かったと結論づけている.

  最後に, cross embodimentについても触れておく.
  actionやproprioceptionのような, 各ロボット固有のモダリティについては, 必ず身体性の問題がつきまとう.
  この問題に初めてメスを入れたのがOpen X-Embodiment Project \cite{oneill2024openxembodiment}である.
  単一の身体性だけでなく, 様々な身体性を持つロボットのデータを統合して学習することで, より性能の高いVLAを学習できることを示している.
  ここではRT-1 \cite{brohan2023rt1}とRT-2 \cite{zitkovich2023rt2}のネットワーク構造を利用し, データセットを1カメラ・言語・7自由度のアクション(位置姿勢とグリッドの開閉)に統一して学習を行っていた.
  これに加えて, 異種身体の行動と観測を一人称視点というフレームで正規化・整列し, 観測とゴール画像だけであらゆるロボットの制御を統一するといった方法も提案された\cite{yang2024extreme}
  しかし, これでは限界があり, 観測や制御入力の大きく異なるマニピュレータや移動ロボット, 脚型ロボットなどを統一的に扱うことはできない.
  そこでCrossFormer \cite{doshi2025crossformer}は, 各ロボットのセンサの数や種類が異なっていても, 観測入力をトークン列として並べることで統一的に処理できるようにしている.
  また, readout tokenを入力し, transformerからの出力を, 各身体性に対応した, single-arm, navigation, bimanual, またはquadruped専用のaction headに接続し, それぞれの身体性に対応したアクションを出力できるようにした.
  さらにUniAct \cite{zheng2025uniact}では, Universal Action Space (UAS)と呼ばれる離散コードブックを用意し, これをtransformerが出力, ロボットごとに異なるデコーダによって連続的なアクションに変換する.
  明示的な共通の原子行動空間を用意することで再利用性が高まり, 異なる身体間の知識の転移が容易になった.
  そして, ロボットのみならず, 人間の身体性からの転移まで考慮したのがUniSkill \cite{UniSkill}である.
  ラベルなしの人間の動画データから, LAPA \cite{ye2025lapa}に似た形で潜在的なスキル表現を取り出し, 人間の動作データを事前学習に活用することで, より汎用的な形でVLAを構築することが可能になった.
  この他にも, 身体性に寄らないオプティカルフローや特徴点の動作軌道を中間表現として用いることで, 身体性に依存しないVLAを構築する方法であるLangToMo \cite{LangToMo}やATM~\cite{wen2024atm}などのVLAも提案されている.
}%

\subsubsection{Miscellaneous Modalities}\label{subsubsec:other}
\switchlanguage%
{%
In addition to vision, language, and action, modern VLA models increasingly incorporate additional modalities to enhance perception and interaction capabilities.
In this section, we describe three additional sensing modalities relevant to VLA systems: audio, tactile sensing, and 3D spatial information.
  
\textbf{Audio.}
  Several prior works such as Unified-IO 2~\cite{lu2024unifiedio2}, SOLAMI~\cite{jiang2025solami}, FuSe~\cite{jones24fuse}, VLAS~\cite{zhao2025vlas}, and MultiGen~\cite{MultiGen} leverage audio information as input.
  Audio encoders typically take spectrograms or mel-spectrogram images as input, which are then converted into audio tokens using models like ResNet or ViT-VQGAN.
  RVQ-VAE-based SpeechTokenizer~\cite{zhang2024speechtokenizer}, Audio Spectrogram Transformer (AST)~\cite{gong2021ast}, or the Whisper encoder~\cite{radford2023whisper} are also frequently used as pre-trained models.
  These encoders enable the system to leverage rich audio information that may not be readily transcribed into text for robotic decision-making.
  SoundStorm~\cite{borsos2023soundstorm} or the decoder of VQGAN are often employed for decoding.
  A common and straightforward approach, as employed in RoboNurse-VLA~\cite{RoboNurse-VLA}, is to convert audio into text using standard automatic speech recognition (ASR) systems.

\textbf{Tactile sensors.}
  FuSe~\cite{jones24fuse}, TLA~\cite{TLA}, VTLA~\cite{VTLA}, and Tactile-VLA~\cite{Tactile-VLA} incorporate tactile information as part of inputs.
  Tactile sensors such as DIGIT~\cite{lambeta2020digit} and GelStereo 2.0~\cite{zhang2024gelstereo2}, which produce image-based outputs, are commonly used.
  These tactile images are either encoded using a Vision Transformer (ViT) or tokenized via a pre-trained Touch-Vision-Language (TVL) model~\cite{fu2024tvl}.
  This enables the integration of visual and tactile information for learning fine-grained manipulation skills in contact-rich tasks, such as peg insertion.
  Although not tactile sensors in the strict sense, ForceVLA~\cite{ForceVLA} incorporates general $6$-axis force-torque sensors. 
  In particular, a force-aware Mixture-of-Experts fusion module integrates force tokens derived from $6$-axis force–torque sensor data with visual-language features extracted by a pre-trained VLM, and generates actions through an action head.

\textbf{3D information.} 
  Incorporating 3D information enables robots to more accurately perceive their environment and plan actions accordingly.
  In 3D perception, we specifically introduce (a) depth images, (b) multi-view images, (c) voxel representations, and (d) point clouds below.

\textbf{(a) Depth images.}
  A common strategy for incorporating depth information involves tokenizing depth images using standard visual backbones, such as Vision Transformers (ViTs) or ResNets, similar to the processing of RGB images. 
  In scenarios where direct depth sensing is not available, monocular depth estimation models such as Depth Anything~\cite{yang2024depthanything} and ZoeDepth~\cite{bhat2023zoedepth} are frequently utilized to predict depth from RGB inputs.
  SpatialVLA~\cite{SpatialVLA} is a representative method that utilizes depth images by introducing Ego3D Position Encoding. In this framework, depth maps are first estimated from RGB inputs using ZoeDepth, and the corresponding 3D coordinates for each pixel are computed via the camera's intrinsic parameters. 
  The 3D coordinates are first processed using sinusoidal positional encoding and an MLP, and the resulting features are added to the 2D visual features extracted by SigLIP~\cite{zhai2023siglip}. 
  This combined representation is used as the Ego3D positional encoding and provided as input to the LLM.
  Additionally, HAMSTER~\cite{li2025hamster}, RationalVLA~\cite{RationalVLA}, and OpenHelix~\cite{OpenHelix} incorporate a 3D Diffuser Actor~\cite{ke2025threeddiffuseractor}, a diffusion-based action head that operates in 3D space and processes RGB-D inputs to generate actions.

\textbf{(b) Multi-view images.}
  Several works attempt to extract 3D information from multi-view images.
  For example, GO-1~\cite{AgiBotWorldColosseo} simply takes as input multi-view RGB-D images, encouraging implicit understanding of 3D structure.
  3D-VLA~\cite{3D-VLA} extends Q-Former (described in \secref{subsubsec:vision}) to handle RGB-D and multi-view inputs.
  Evo-0~\cite{Evo-0} employs Visual Geometry Grounded Transformer (VGGT)~\cite{wang2025vggt} to extract implicit 3D geometric information from multi-view RGB images.
  RoboUniView~\cite{RoboUniView} and RoboMM~\cite{RoboMM} utilize UVFormer, a pre-trained model that takes multi-view RGB-D images and corresponding camera parameters as input and outputs a 3D occupancy grid. The encoder's output features are then used as tokens for downstream processing.
  Furthermore, SAM2Act~\cite{fang2025samact} and HAMSTER~\cite{li2025hamster} use Robotic View Transformer-2 (RVT-2)~\cite{goyal2024rvt2} to reproject point cloud or depth information into a virtual view (often using orthographic projection to generate three images), and each image is tokenized by ViT.
  Similar approaches are also used in OG-VLA~\cite{OG-VLA} and BridgeVLA~\cite{BridgeVLA}.
  Overall, two main approaches have emerged: integrating information from multiple viewpoints, and projecting 3D data into orthographic images to facilitate easier processing.

\textbf{(c) Voxel representations.}
  Voxel-based representations are another widely adopted approach for encoding 3D information.
  OccLLaMA~\cite{OccLLaMA} and OpenDriveVLA~\cite{OpenDriveVLA} convert 3D occupancy grids into 2D Bird's Eye View (BEV) feature maps, which are then tokenized using VQ-VAE.
  Several approaches operate directly on three-dimensional voxel grids, such as iManip~\cite{iManip}, which extracts features using a 3D U-Net~\cite{peng2020unet3d}, and VidBot~\cite{chen2025vidbot}, which first converts voxel grids into Truncated Signed Distance Fields (TSDFs) and then processes them using a 3D U-Net.
  Because voxel representations resemble image structures and are compatible with convolutional processing, they have been widely adopted across various studies.

\textbf{(d) Point clouds.}
  A common approach involves tokenizing point clouds using pre-trained point-based transformers such as PointNet~\cite{qi2017pointnet}, PointNet++~\cite{qi2017pointnetpp}, PointNext~\cite{qian2022pointnext}, and Uni3D ViT~\cite{zhou2023uni3d}. 
  These backbones are widely adopted in models such as SOFAR~\cite{SoFar}, LEO~\cite{huang2024leo}, PPI~\cite{PPI}, LMM-3DP~\cite{LMM-3DP}, GeneralFlow~\cite{yuan2025generalflow}, FP3~\cite{FP3}, and DexTOG~\cite{DexTOG}.
  In contrast, some methods opt for task-specific training: StructDiffusion~\cite{liu2023structdiffusion} uses the Point Cloud Transformer (PCT)~\cite{guo2021pct}, and PointVLA~\cite{PointVLA} employs PointCNN~\cite{li2018pointcnn}, with both models trained from scratch for their respective tasks.
  Additionally, although less common, LERF-TOGO~\cite{rashid2023lerftogo} and Splat-MOVER~\cite{shorinwa2024splatmover} integrate point clouds reconstructed using Neural Radiance Fields (NeRF) or Gaussian Splatting with semantic features extracted from CLIP~\cite{radford2021clip}. 
  These enriched representations are then used in conjunction with GraspNet~\cite{fang2020graspnet} to generate grasping plans.

  Beyond the primary modalities discussed above, several VLA models have been proposed to incorporate additional forms of information.
  ARM4R~\cite{niu2025arm4r}, for example, integrates 3D tracking data to enhance motion understanding.
  SOLAMI~\cite{jiang2025solami} introduces a Motion Tokenizer that applies VQ-VAE to discretize the joint angles of SMPL-X~\cite{pavlakos2019smplx} on a per-body-part basis, following the approach introduced in motionGPT~\cite{jiang2023motiongpt}.
  Additionally, PPL~\cite{yao2025ppl} and LangToMo~\cite{LangToMo} incorporate motion dynamics by using RAFT~\cite{teed2020raft} to estimate optical flow from pairs of images, enabling fine-grained temporal reasoning.
}%
{%
  画像・言語・アクション以外にも, 現在は様々なモダリティがVLAに用いられるようになってきている.
  それらには, 主に以下のようなものがある.
  \begin{enumerate}
    \item 音声
    \item 接触センサ
    \item 3次元情報
  \end{enumerate}

  (1)の音声は, Unified-IO 2やSOLAMI, FuSe, VLAS, MultiGenなどのVLAで用いられている \cite{lu2024unifiedio2, jiang2025solami, jones24fuse, zhao2025vlas, MultiGen}.
  音声情報のエンコーダは大抵スペクトログラムまたはメルスペクトログラム画像を入力としており, ResNetやViT-VQGANなどにより音声トークンに変換される.
  事前学習済みでRVQ-VAEベースのSpeechTokenizer \cite{zhang2024speechtokenizer}やAudio Spectrogram Transformer (AST) \cite{gong2021ast}, Whisper \cite{radford2023whisper}のEncoderなどが用いられる場合も多い.
  これにより, テキストにならないような音声情報もロボットの動作に活用することができる.
  また, デコードにはSoundStorm \cite{borsos2023soundstorm}やVQGANのデコーダが用いられている.
  もちろん, RoboNurse-VLA \cite{RoboNurse-VLA}のように, 単純に音声を一般的な自動音声認識でテキストに変換して用いる方法も非常に多い.

  (2)の接触センサについては, FuSe \cite{jones24fuse}や, TLA \cite{TLA}, VTLA \cite{VTLA}, Tactile-VLA \cite{Tactile-VLA}などのVLAで用いられている.
  センサとしてはDIGITやGelStereo 2.0 \cite{zhang2024gelstereo2}, などの, 画像を出力する接触センサが用いられることが多い.
  これらの画像はViTにより変換, または事前学習済みのTVL (Touch-Vision-Language) Modelを用いてトークン化されている.
  これにより, ペグ挿入のような接触を伴うタスクにおいて, 画像と接触センサの情報を統合してより繊細な動作を学習することが可能になる.
  また接触センサではないが, 一般的な6軸トルクセンサを用いたForceVLA \cite{ForceVLA}なども存在している.
  このトルクセンサの値はAction HeadにおいてForce-Aware Vision-Language Mixture Modelとして, 力覚トークンを含むMoEの形で利用されている.

  (3)の3次元情報を扱うことで, ロボットはより詳細に環境を理解し動作を計画することが可能になる.
  この三次元情報を扱う方法は主に以下の4つである.
  \begin{itemize}
    \item 深度画像
    \item multi-view画像
    \item ボクセル表現
    \item 点群
  \end{itemize}

  最も単純な方法は, 画像と同様に深度画像をViTやResNetによりトークン化する方法である.
  深度画像が直接得られない場合でも, Depth Anything \cite{yang2024depthanything}やZoeDepth \cite{bhat2023zoedepth}などを用いてRGBから深度画像を推定することができ, これを入力する場合も少なくない.
  深度画像を活用した方法として, SpatialVLA \cite{SpatialVLA}はEgo3D Position Encodingと呼ばれる方法を構築している.
  これは, RGBからZoeDepthを用いて深度画像を推定し, これとカメラの内部パラメータから各ピクセルの三次元座標を抽出する.
  SigLIP \cite{zhai2023siglip}により抽出された2D視覚特徴量に, 3次元座標にsinusoidal positional encodingとMLPを適用したものを加算し, これをEgo3D位置表現としてLLMへの入力に用いている.
  この他にも, HAMSTER \cite{li2025hamster}やRationalVLA \cite{RationalVLA}, OpenHelix \cite{OpenHelix}などは, 3D Diffuser Actorと呼ばれる, RGBD画像を入力とした3D空間awareなdiffusion action headを用いている.

  次に, multi-view画像を用いて三次元情報を取り出す取り組みがある.
  GO-1 \cite{AgiBotWorldColosseo}では単純にmulti-viewなRGBD画像を入力して暗黙的に三次元情報の理解を促しており, 3D-VLA \cite{3D-VLA}では\secref{subsubsec:vision}で説明したQ-FormerをRGB-Dかつmulti-viewに拡張, Evo-0 \cite{Evo-0}ではVGGT (Visual Geometry Grounded Transformer) \cite{wang2025vggt}を用いてmulti-viewのRGB画像から暗黙的な三次元幾何情報を取り出している.
  RoboUniView \cite{RoboUniView}やRoboMM \cite{RoboMM}で用いられているUVFormerでは, multi-viewのRGBD画像とカメラパラメータを入力として, 3次元のoccupancy grid mapを出力するように事前学習し, そのエンコーダの出力をトークンとして用いている.
  また, SAM2Act \cite{fang2025samact}やHAMSTER \cite{li2025hamster}では, RVT-2 (Robotic View Transformer-2) \cite{goyal2024rvt2}を用いて, 点群や深度情報から空間を仮想視点に再投影し(多くはorthogonal投影で3枚の画像が得られる), 各画像をViTによりトークン化している.
  同様の方法は, OG-VLA \cite{OG-VLA}やBridgeVLA \cite{BridgeVLA}にも用いられている.
  複数視点からの情報を統合する取り組みと, 3次元情報を扱いやすい正投影画像に変換する取り組みの2種類があることが見て取れる.

  次に, ボクセル表現を用いた方法も多く使われている.
  OccLLaMA \cite{OccLLaMA}やOpenDriveVLA \cite{OpenDriveVLA}は3D occupancy grid mapを, 上空から見下ろしたBird's Eye View (BEV)の2D特徴量マップとして表現し, これをVQ-VAEなどによりトークン化している.
  また, 構築した3D voxel mapに対して, iManip \cite{iManip}は3D U-Netにより特徴量抽出しており, VidBot \cite{chen2025vidbot}はTSDFに変換してから3D U-Net \cite{peng2020unet3d}により特徴量抽出している.
  ボクセル表現は画像と同様に扱えることから, 多様な研究で活用されている.

  最後に, 三次元情報として点群を扱う方法を紹介する.
  大抵は, PointNet \cite{qi2017pointnet}やPointNet++ \cite{qi2017pointnetpp}, PointNext \cite{qian2022pointnext}, Uni3D ViT \cite{zhou2023uni3d}のような事前学習済みの点群用transformerを用いたトークン化が行われている(SOFAR, LEO, PPI, LMM-3DP, GeneralFlow, FP3, DexTOG) \cite{SoFar, huang2024leo, PPI, LMM-3DP, yuan2025generalflow, FP3, DexTOG}.
  これに対して, StructDiffusion \cite{liu2023structdiffusion}はPoint Cloud Transformer (PCT) \cite{guo2021pct}を, PointVLA \cite{PointVLA}はPointCNNを, タスクごとにスクラッチで学習している.
  加えて, 数は少ないが, LERF-TOGO \cite{rashid2023lerftogo}やSplat-MOVER \cite{shorinwa2024splatmover}は, NeRFやGaussian Splattingを用いて構築した点群とCLIP \cite{radford2021clip}によるsemanticsを統合し, ここにGraspNet \cite{fang2020graspnet}を適用することで把持計画を出力している.

  これらの他にも, 3D tracking情報を活用したARM4R \cite{niu2025arm4r}, motionGPT \cite{jiang2023motiongpt}で提案されたSMLPL-X \cite{pavlakos2019smplx}の関節角度を部位ごとにVQ-VAEしてトークン化するmotionTokenizerを利用するSOLAMI \cite{jiang2025solami}, 2枚の画像からoptical flowを推定するRAFT (Recurrent All-Pairs Field Transforms) \cite{teed2020raft}を活用するPPL \cite{yao2025ppl}やLangToMo \cite{LangToMo}など, 多様なモダリティを扱うVLAが登場してきている.
}%

\subsection{Emerging Techniques}
Recent advances in VLA research highlight two emerging directions: \textit{hierarchical architectures} and \textit{Chain-of-Thought (CoT) reasoning}. %
Both approaches introduce structured intermediate representations between language instructions and low-level actions, enabling more robust planning, decomposition, and reasoning.
Further details of these approaches are provided below.

\textbf{Hierarchical architectures.}
The most foundational approach is Atomic Skill~\cite{li2025atomic} and LMM-3DP~\cite{LMM-3DP}, which use existing VLMs as high-level policies to decompose task instructions into subtasks. 
These subtask descriptions are then passed to a VLA acting as the low-level policy. 
Since the low-level policy receives cleaner and more concise language inputs, it can execute actions more reliably than when processing complex, unstructured instructions directly.
On the other hand, Hi Robot~\cite{HiRobot} trains a custom high-level policy instead of relying on existing VLMs.
NAVILA~\cite{cheng2024navila} and HumanoidVLA~\cite{Humanoid-VLA} employ low-level policies trained using reinforcement learning (RL) to achieve fine-grained motor control.
RT-H~\cite{belkhale2024rth} and LoHoVLA~\cite{LoHoVLA} take a more integrated approach by jointly training both high-level and low-level policies within a single network.
By switching the input prompt, these models can flexibly alternate between decomposing a task instruction into subtasks and converting a subtask into a corresponding action.
This approach has been further extended to $\pi_{0.5}$~\cite{Pi-0.5}, which unifies subtask decomposition, discrete action token generation, and continuous action generation within the same network.
The integration of task decomposition with VLA models is emerging as a promising approach for enabling more flexible and scalable robot behavior.
Additionally, FiS-VLA~\cite{Fast-in-Slow}, OpenHelix~\cite{OpenHelix}, and DP-VLA~\cite{DP-VLA} propose connecting high-level and low-level policies through latent spaces, without explicitly defining intermediate representations as subtasks.
Tri-VLA~\cite{TriVLA} integrates a pre-trained vision-language model for scene understanding with Stable Video Diffusion, which produces visual representations capturing both static observations and future dynamics. These representations are then used as input to a diffusion transformer, which generates actions via cross-attention.

\textbf{Chain-of-Thought (CoT) reasoning.}
Chain-of-Thought (CoT) reasoning, while conceptually similar to hierarchical approaches, introduces a distinct mechanism that has been integrated into VLA models such as ECoT~\cite{zawalski2025ecot} and CoT-VLA~\cite{zhao2025cotvla}. 
ECoT addresses a key limitation of typical VLAs, which is their lack of intermediate reasoning, by introducing a step-by-step process between observations, instructions, and action generation, thereby enhancing planning and inference capabilities. 
In particular, ECoT achieves this by autoregressively predicting intermediate representations, such as task descriptions, subtasks, and object positions, before generating the final action sequence.
On the other hand, CoT-VLA~\cite{zhao2025cotvla} generates subgoal images, thereby improving success rates on more visually grounded tasks.
ECoT-Lite~\cite{ECoT-Lite} reduces inference latency caused by reasoning by selectively dropping certain reasoning components during training.
Fast ECoT~\cite{FastECoT} takes this further by reusing intermediate reasoning outputs and parallelizing reasoning and action generation, resulting in faster action execution.

\section{Training Strategy and Implementation}\label{sec:training}
\switchlanguage%
{%
  We categorize the training approaches of Vision-Language-Action (VLA) models into supervised learning, self-supervised learning, and reinforcement learning.
  Below, we summarize the core characteristics and representative methods of each approach.
}%
{%
  VLAの学習方法は, Supervised Learning, Self-Supervised Learning, Reinforcement Learningの3つに分類される.
  ここでは, それぞれの学習方法についてまとめる.
}%

\subsection{Supervised Learning}\label{subsec:supervised}
\switchlanguage%
{%
Most VLA models are trained using supervised learning on datasets consisting of pairs of images, language, and actions.
Since many VLAs are built on LLMs, training is often formulated as a next-token prediction task.
The choice of action loss function depends on the architecture of the action head, such as MLPs, diffusion models, or flow matching networks, ensuring appropriate supervision for each model type.  
  
VLA training generally consists of two stages: pre-training and post-training.
In many cases, a LLM or VLM pre-trained on web-scale data is first used as the initial backbone for training.
While some models are trained from scratch, it is more common to initialize training with a pre-trained VLM that has already acquired commonsense knowledge, in order to enhance generalization.
Pre-training is typically performed using datasets such as human demonstrations, heterogeneous robot demonstrations, or VQA datasets related to robotic planning.
Similar to LLMs, data scale plays a crucial role in VLA pre-training. 
Leveraging large and diverse datasets enables the development of VLA models with stronger generalization across tasks and embodiments.
In the pre-training stage, the pre-trained VLM is typically fully fine-tuned to adapt to robotics-related domains.
For further details about pre-training, see~\secref{subsubsec:pre-training}.

After pre-training, post-training is performed using task- or robot-specific datasets.
In this stage, data quality tends to be more important than quantity, and the datasets are often smaller to those used in pre-training.
Finetuning strategies differ across implementations. 
In some cases, the entire model undergoes full finetuning, whereas in others, adaptation is limited to the action head. 

Moreover, in-context learning, a technique originally developed for LLMs, has also been adapted for use in VLA systems.
Rather than explicitly fine-tuning on demonstration data, in-context VLA models condition on a small number of human teleoperation trajectories at test time to infer appropriate actions. For instance, ICRT~\cite{ICRT} introduces a framework in which 1--3 teleoperated demonstrations are provided as prompts, enabling the model to generate corresponding robot actions in a zero-shot manner.
}%
{%
  VLAはほとんどの場合, 画像・言語・アクションなどのペアで構成されるデータセットありきの, 教師あり学習によって学習される.
  ほとんどのVLAはLLMをベースとしており, next token predictionの形で行われている.
  またアクションについては, Action HeadがMLPか, diffusion modelか, flow matching modelに応じて, 適切な損失関数が用いられる.

  VLAの学習には, 事前学習と事後学習の2つの段階がある.
  なお, 多くのケースで, ウェブデータから事前学習済みのLLMやVLMから学習はスタートしている.
  もちろん, フルスクラッチでネットワークを学習する場合もあるが, 汎化性能の問題から近年は既存のVLMをスタート地点として学習することが多い.
  事前学習は人間のデモンストレーションや異種ロボットのデモンストレーション, ロボットの動作計画などを扱うVQA等のデータセットを用いて行われる.
  LLMと同様に事前学習ではデータの量が重要であり, 大規模なデータセットを用いることで汎用的なVLAを学習することができる.
  ここでは基本的に既存のVLMをfull finetuningする形で学習される.
  次に, 事後学習を行う.
  事後学習は, ある決まったタスクやロボットに特化したデータセットを用いて行われる.
  ここでは, データの量よりも室が重視され, 事前学習に比べるとより少ないデータセットで学習されることが多い.
  full finetuningを行う場合もあれば, action headのみをfine tuningする場合, LoRAを用いて軽量に学習する場合もある.
  他にも, 事前学習では離散アクショントークンを用いて, 事後学習ではflow matchingによる連続アクションで学習する, action headの勾配がVLMバックボーンに伝播しないようにしたほうが良い\cite{driess2025insulating}, 破壊的忘却を防ぐために視覚エンコーダの重みを段階的に事前学習の重みに戻していく\cite{RevLA}, excess lossに基づき訓練時の各データセットの重みを調整する\cite{hejna2025remix}など, データセットごとにいくつかの知見が共有されている.

  また, LLMに特有のIn-context learningをVLAで活用した例もある.
  ICRT \cite{ICRT}は, 人間による1-3のテレオペレーションによるデモンストレーションをプロンプトとして入力すると, それを元にロボットの動作を生成することができる構造を提案している.
}%

\subsection{Self-Supervised Learning}\label{subsec:selfsupervised}
\switchlanguage%
{%
Self-supervised learning is occasionally incorporated into the training of Vision-Language-Action (VLA) models, serving three primary purposes.
 
\textbf{Modality alignment} focuses on learning temporal and task-level consistency across modalities in VLA models.
For instance, TRA~\cite{TRA} uses contrastive learning to align representations of current and future states within a shared latent space, achieving temporal alignment.
Similarly, task alignment is achieved by aligning language instruction embeddings with those of goal images through contrastive objectives.

\textbf{Visual representation learning} aims to extract visual features from images or videos using techniques such as masked autoencoding (e.g., MAE~\cite{he2022mae}), contrastive learning (e.g., CLIP~\cite{radford2021clip}), and self-distillation (e.g., DINOv2~\cite{oquab2023dinov2}). These pre-trained models are widely adopted in VLAs as foundational visual encoders.

\textbf{Latent action representation learning} leverages self-supervised techniques to learn action embeddings, as discussed in \secref{subsec:world} and \secref{subsubsec:action}.
By extracting a latent action from the initial and goal images, and reconstructing the goal image using the initial image and the extracted latent action, the model learns meaningful action representations without requiring explicit labels.
This approach is highly scalable and well-suited for large, unannotated datasets.
}%
{%
  VLAの学習の一部に自己教師あり学習が用いられることがある.
  自己教師あり学習は以下の3つの目的で用いられる.
  \begin{enumerate}
    \item visual representation learning
    \item latent action representation learning
    \item modality alignment
  \end{enumerate}

  (1)は, 画像や動画から視覚的な特徴量を学習するために用いられる.
  これはMasked Autoencoder \cite{he2022mae}のような形, CLIP \cite{radford2021clip}のようなcontrastive learningのような形, DINOv2 \cite{oquab2023dinov2}のようなself-distillationの形で行われる.
  これらのモデルはVLAでは非常に重要な役割を果たしており, 画像や動画の特徴量を抽出するために広く用いられている.

  (2)は, \secref{subsubsec:world}や\secref{subsubsec:action}でも登場した, アクションの潜在表現を学習するために用いられる.
  初期画像とゴール画像からの潜在的なアクション表現抽出と, 初期画像とアクション表現からのゴール画像の再構成を行うことで, アクションの潜在表現を自己教師あり学習により獲得することができる.
  世界モデルの学習は基本的にアノテーションが必要なく, スケールしやすいというメリットがある.

  (3)は, VLAの時間的な整合性やタスクの整合性を学習するために用いられる.
  例えばTRA \cite{TRA}では, 現在状態と将来状態の表現が一貫した表現空間に配置されるように, 同じタスク内の異なる時間ステップ間での整合性を高めるcontrastive learningを行っている(temporal alignment).
  また同様に, 言語指示と目標状態画像の埋め込みを整合させるようなcontrastive learningも行っている(task alignment).
}%

\begin{figure}[t]
  \centering
  \includegraphics[width=1.0\columnwidth]{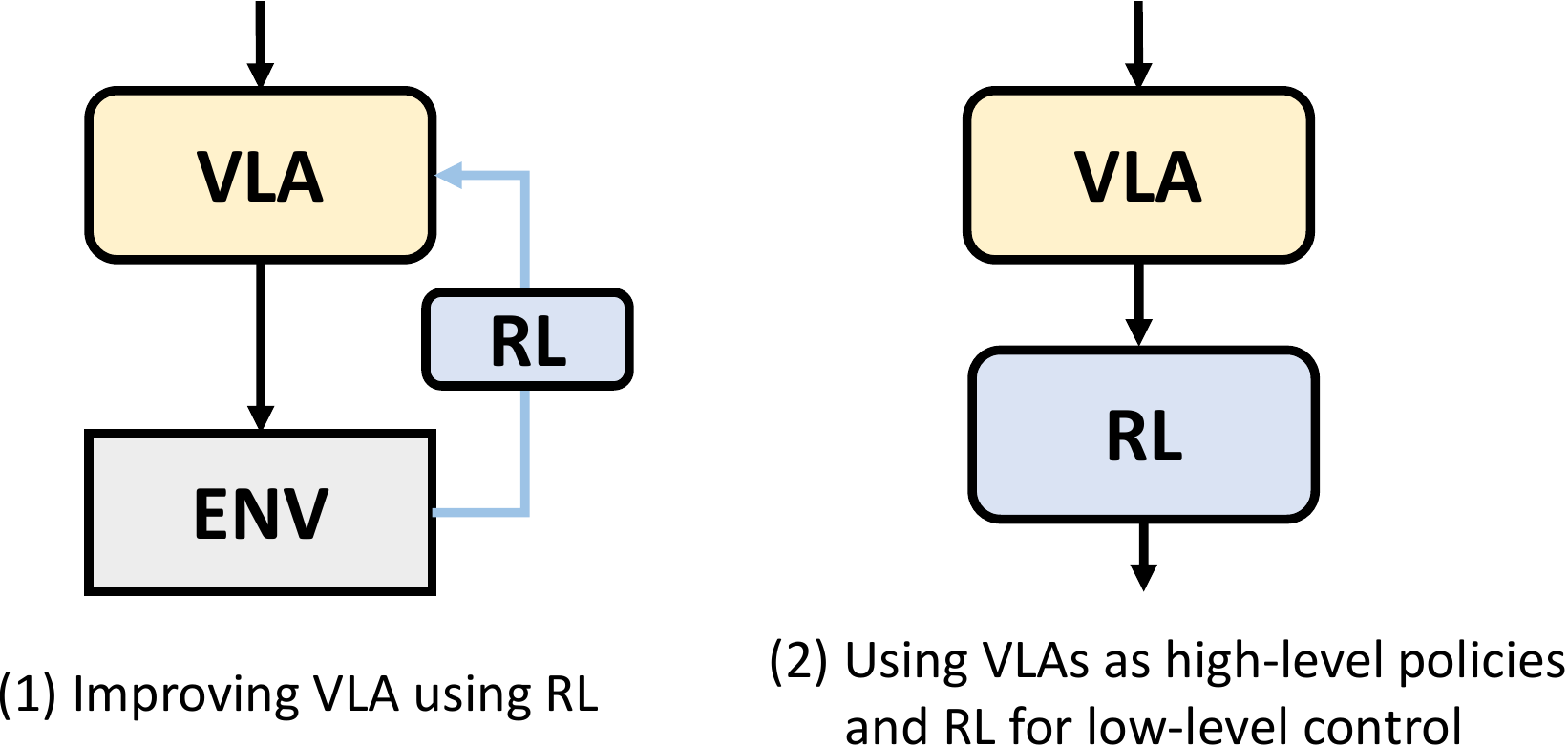}
  \caption{\textbf{Approaches to integrating RL with VLA models.} (1) RL is used to fine-tune VLA models to enhance their performance. (2) VLA models serve as high-level policies, while RL policies handle low-level control.}
  \label{figure:reinforcement}
\end{figure}

\subsection{Reinforcement Learning}\label{subsec:reinforcement}
\switchlanguage%
{%
While VLA is trained via imitation learning in general, imitation learning alone faces challenges such as the inability to handle novel behaviors and the requirement for sufficiently large and high-quality expert demonstrations.
To address these issues, several prior arts have explored finetuning VLA or training low-level policies using reinforcement learning (RL), such as PPO~\cite{schulman2017ppo} and SAC~\cite{haarnoja2018sac}.
These approaches can be broadly categorized into the following two types, as shown in \figref{figure:reinforcement}.

\textbf{(1) Improving VLA using RL.}
  Recent work leverages RL to improve the robustness, adaptability, and real-world performance of VLA models. 
  Several approaches fine-tune VLAs using RL with task success or failure as the reward signal.
  iRe-VLA~\cite{iRe-VLA} achieves high performance by repeatedly combining supervised fine-tuning (SFT) on expert data, online RL on the action head using success and failure rewards, and subsequent SFT using both expert data and successful trajectories collected during online learning.
  ConRFT~\cite{ConRFT} applies imitation learning on a small set of demonstrations, performs offline RL to learn a Q-function, and subsequently fine-tunes the policy online through human interventions. 
  This approach is inspired by prior frameworks such as SERL~\cite{luo2024serl} and HIL-SERL~\cite{HIL-SERL}, which are reset-free~\cite{han2015resetfree, gupta2021resetfree2}, off-policy RL methods~\cite{ball2023rlpd} designed for real-world robot learning.
  VLA-RL~\cite{VLA-RL} introduces the Robotic Process Reward Model (RPRM), which replaces sparse binary rewards with dense pseudo-rewards derived from gripper actions and task progress, enabling more stable PPO-based training.
  RLDG~\cite{RLDG} fine-tunes large VLA models such as OpenVLA~\cite{kim2024openvla} and Octo~\cite{octoteam2024octo} using successful trajectories collected via HIL-SERL, allowing integration of multiple expert policies into a unified VLA.
  MoRE introduces a Mixture of Experts (MoE) structure into the VLA, enabling token-wise expert selection and refinement via RL.
  RLRC~\cite{RLRC} compresses OpenVLA by pruning up to $90\%$ of its parameters, recovers performance via SFT, and then applies RL for final fine-tuning using task-level feedback.
  These studies demonstrate that RL, especially when combined with expert demonstrations or human interventions, can significantly improve the flexibility and reliability of VLA models in real-world settings.
  More recently, to address the potential instability associated with backpropagation through diffusion chains, DSRL~\cite{wagenmaker2025steeringdiffusionpolicylatent} proposes applying RL in the latent noise space of the diffusion policy. This approach avoids updating the parameters of the underlying VLA model during RL fine-tuning. Instead, it learns a distribution over the latent noise, allowing the model to sample informative initial noise vectors rather than from a standard Gaussian.
  Notably, DSRL demonstrates that the success rate of $\pi_{0}$ can be improved from approximately 20\% to nearly 100\% using only $10$K samples.

\textbf{(2) Using VLAs as high-level policies and RL for low-level control.}
  This class of approaches delegates high-level decision-making to the VLA, while low-level control is handled by policies trained with RL.
  Humanoid-VLA~\cite{Humanoid-VLA} uses a VLA to generate high-level commands, which are executed by a whole-body controller trained via RL for humanoid robots.
  NaVILA~\cite{cheng2024navila} adopts a similar strategy, applying RL to convert velocity commands from the VLA into torque control for a legged robot.
  A more advanced system, SLIM~\cite{SLIM}, targets a mobile manipulator comprising a quadruped base and robotic arm. It first trains a teacher policy using RL with privileged inputs, such as footstep plans, object placements, and subtask identifiers, to generate base and arm trajectories. 
  This policy is then distilled into a student VLA via imitation learning, enabling end-to-end mapping from images and language to actions.
  RPD~\cite{julg2025rpd} takes a complementary approach, using a pre-trained VLA to guide exploration during RL. Here, the VLA acts as a teacher, shaping the learning process rather than serving as a high-level controller.

  In addition, LUMOS performs imitation learning in the latent space of a world model by employing reinforcement learning guided by an intrinsic reward that quantifies the deviation from expert trajectories within the latent space.
  DexTOG~\cite{DexTOG} generates a diverse set of grasp poses using a diffusion model and employs reinforcement learning to evaluate whether each candidate pose leads to task success. Through iterative fine-tuning with successful trajectories, the diffusion model learns object-specific grasp poses that are well-suited for subsequent tasks.

  Despite the growing number of VLA methods incorporating RL, most prior work remains limited to simulation or simplified real-world setups, due to sample inefficiency, unsafe exploration, and computational inefficiency.
}%
{%
  VLAの学習は大抵の場合摸倣学習により行われている.
  その一方で, 摸倣学習単体では未知の動作に対応できない, 十分で高品質なデータセットが必要であるなどの問題点がある.
  そこで, 強化学習, 特にPPO~\cite{schulman2017ppo}やSAC~\cite{haarnoja2018sac}を用いて高い性能のVLAを学習する研究が複数行われている.
  これらは主に, 以下の3つに分類できる.
  \begin{enumerate}
    \item 既存のVLAのパフォーマンスを強化学習により改善する.
    \item LLMやVLMを活用して一からポリシーを強化学習する.
    \item VLAを上位ポリシーとして, 強化学習により得られた下位ポリシーを利用する.
  \end{enumerate}

  (1)には, 特に実機強化学習に長けた方法として, Sample-Efficient Robotic Reinforcement Learning (SERL)~\cite{luo2024serl}がある.
  SERLは実ロボット向けのサンプリング効率の良いリセットフリーかつオフポリシー強化学習であり, さらに人間の介入を取り入れ, 失敗しやすい状態からの回復動作を学習可能なHuman-in-the-Loop SERL (HIL-SERL) \cite{HIL-SERL}も提案されている.
  これらは, VLAから得られたポリシーを調節してよりロバストにするために活用できる.
  また, タスクの失敗成功を報酬として, 強化学習によりVLAを調整する手法が多数開発されている.
  iRe-VLA \cite{iRe-VLA}はExpertデータによるVLAのSFT (Supervised Fine Tuning), 成功失敗報酬によるaction headのみのオンライン強化学習, オンライン強化学習時の成功軌跡とExpertデータによる再度のSFTを繰り返すことで, VLAに高い性能を与えることができる.
  ConRFT \cite{ConRFT}は, 摸倣学習により得られたVLAを少数のデモンストレーションからオフライン強化学習し, そのQ関数を引き継いで人間の介入を取り入れつつオンライン強化学習を行うことで, より高い性能を得ることができる.
  Mixture of Robotic Experts (MoRE)は, VLAにMoE構造を組み込むことでトークンごとに専門化されたエキスパートを動的に選択できるようにし, 強化学習によって失敗成功を報酬としてVLAを調整している.
  文脈は異なるが, RLRC \cite{RLRC}では, block-wise pruningにより90\%までOpenVLA \cite{kim2024openvla}のパラメータを削減し, これをSFTにより性能回復, さらに成功失敗を報酬として強化学習することで, VLAを圧縮しつつその性能を保っている.
  VLA-RL \cite{VLA-RL}では, 成功失敗のようなスパースな報酬ではなく, グリッパの開閉情報とプロセスの進行具合からアクションの擬似的な報酬を計算するRobotic Process Reward Model (RPRM)を構築し, これをもとにOpenVLA \cite{kim2024openvla}をPPOで強化学習することで, VLAの性能を改善する方法を提案している.
  加えて, RLDG \cite{RLDG}では, OpenVLA \cite{kim2024openvla}やOcto \cite{octoteam2024octo}をベースモデルとして, HIL-SERLでの成功軌道を用いてモデルをファインチューニングすることで, 複数の専門ポリシーの能力を単一のVLAに統合することに成功している.

  (2)には, 初期の例として, VLM-RL \cite{VLM-RL}とRoboCLIP \cite{sontakke2023roboclip}が挙げられる.
  VLM-RL \cite{VLM-RL}では, CLIP \cite{radford2021clip}を活用して観測画像とタスク指示の類似度を計算し, これを活用することでポリシーを学習していた.
  RoboCLIP \cite{sontakke2023roboclip}では, 与えられた動画とタスク指示の類似度を測定するモデルを構築し, これを報酬とした.
  この他にも, RL-VLM-F \cite{RL-VLM-F}は既存のVLMに2つの画像とタスク指示を入力し, どちらの画像がよりタスクのゴールに近いかを判定させ, 画像からの報酬モデルを学習, これをもとに強化学習を行っている.
  このような画像とタスク指示の類似度を報酬とする方法以外にも, そもそも報酬設計をLLMやVLMによって行う手法が開発されてきている.
  Eureka \cite{ma2024eureka}は, あるタスクに対して, 環境の定義コードやtask descriptionを与えることで, LLMによる報酬の反復設計と強化学習により適切なポリシーを獲得する.
  DrEureka \cite{ma2024dreureka}はそれをsafety instructionとdomain randomizationをもとに, シミュレーションのみならず実機にも適用できる形とした.
  Video2Policy \cite{Video2Policy}はvideoからタスクを定義, そこからIsaacGym \cite{makoviychuk2021isaacgym}のtaskコードを生成, 学習されたポリシーの実行軌跡をもとtaskコードを反復的に修正することで, より汎用性が高いポリシー学習に成功している.
  また違うアプローチとして, KAGI \cite{lee2024kagi}はGPT-4o \cite{openai2024gpt4ocard}を用いて画像上での手先の目標軌道を生成し, これを満たすように言語と画像に基づくオフライン強化学習を実行することでVLAを構築している.

  (3)には, 例えばHumanoid-VLA \cite{Humanoid-VLA}やNaVILA \cite{NaVILA}が挙げられる.
  Humanoid-VLA \cite{Humanoid-VLA}はVLAの後段として, 強化学習により構築されたヒューマノイドのWhole-Body Controllerを使用している.
  NaVILA \cite{cheng2024navila}も, VLAの後段として, 得られた速度ベクトルを脚型ロボットのトルク制御に変換する強化学習を行っている.
  SLIM \cite{SLIM}はより高度であり, 四足歩行ロボットにマニピュレータがついたMobile Manipulatorについて, 脚歩行の低レベルポリシーだけでなく, 物体の配置やサブタスクのIDなどの特権情報を入力して速度コマンドと腕の動きを出すネットワークを強化学習し(Teacher Policy), これを摸倣学習によって画像と言語から動作が出るVLAに蒸留している(Student Policy).
  RPD \cite{julg2025rpd}は特殊であり, VLAの後段として強化学習を用いるのではなく, VLAを用いて強化学習の行動探索を誘導, つまり, VLAを教師としてRL Policyを学習している.

  これらの他にも, 単純に摸倣学習の代わりに強化学習を用いたり, タスク成功するかどうかの判断に強化学習を用いたりする場合がある.
  LUMOS \cite{nematollahi25lumos}は, World Model内での摸倣学習を行うが, World Modelの潜在空間とゴール状態を条件としたpolicyの学習を, 摸倣データとの分布が近くなるような報酬に基づく強化学習で行っている.
  また, DexTOG \cite{DexTOG}は, タスクに応じた多様なグリップをdiffusion modelから出力し, それを初期姿勢としてその後のタスクが成功するかどうかを強化学習によって学習し判断する.
  成功したデータを用いてdiffusion modelを再学習することを繰り返すことで, タスクに応じた物体の適切な把持姿勢を学習している.

  なお, 現状強化学習を用いたVLAに関する研究のうち(1)と(2)は, シミュレーション上で行われているものが多い.
  特に(2)については, そのほとんどがシミュレーション上での検証にとどまっている.
}%

\subsection{Training Stages}
Training Vision-Language-Action (VLA) models typically involves multiple stages, each targeting a specific aspect of learning. The pre-training aims to acquire general capabilities and promote transferability across diverse robotic embodiments.
When a pre-trained Vision-Language Model (VLM) is used as the backbone of a VLA model, it must be adapted to the robotics domain to effectively ground language and visual understanding in action.
This is followed by a post-training, in which the model is further refined using high-quality robot demonstration data to improve performance on specific downstream tasks.
This section provides a stage-wise overview of representative training strategies, highlighting common data sources, model backbones, and adaptation techniques used in recent VLA systems.

\subsubsection{Pre-training}
\label{subsubsec:pre-training}
Pre-training plays a pivotal role in shaping the generalization ability and semantic grounding of VLA models.
This subsection outlines key strategies and design choices in recent pre-training pipelines, highlighting how large-scale multimodal data, powerful VLM backbones, and training stabilization techniques contribute to effective policy initialization.

\textbf{Data scale and source.}
The scale and heterogeneity of training data significantly impact the generalization ability of VLA models across diverse scenes, objects, and tasks. %
Recent models increasingly leverage large-scale datasets that combine robot demonstrations, web-scale vision-language corpora, and structured annotations to improve semantic understanding and visuomotor grounding.

$\pi_{0}$\cite{Pi-0} is trained on millions of real-world trajectories collected across varied embodiments and tasks. %
Its successor, $\pi_{0.5}$\cite{Pi-0.5}, extends this approach by incorporating not only robotic data but also large-scale vision-language datasets commonly used for object detection and visual reasoning (e.g., COCO\cite{lin2014coco}, VQA~\cite{antol2015VQA}). 
The model is trained with auxiliary cross-entropy losses for multiple tasks, including bounding box prediction, image captioning, subtask language generation, and discrete action prediction.

Similarly, Gr00T N1~\cite{Gr00t-N1} incorporates an auxiliary bounding box loss to improve spatial localization and affordance detection. %
These bounding box labels are obtained using OWL-ViT~\cite{minderer2022owlvit}, allowing the model to learn from weakly supervised visual data. %
Gr00T N1 further leverages egocentric human videos, from which latent action representations are extracted to supervise the VLA model. %
Additionally, it introduces diverse synthetic trajectories generated in simulation, which are transformed into realistic visual observations using the COSMOS world model~\cite{nvidia2025cosmosworldfoundationmodel}, enhancing the model's capacity to learn long-horizon, multi-stage behaviors.

These approaches demonstrate a growing trend toward enriching VLA training data not only in scale but also in structure and modality. %
By jointly training on action, grounding, and reasoning tasks, modern VLAs acquire richer representations that support robust policy learning and generalization.

\textbf{VLM backbones.}
\switchlanguage%
{%
A common practice in recent VLA models is to leverage vision-language models (VLMs) that have been pre-trained on large-scale web data. This strategy enables models to inherit broad visual and linguistic priors, including common sense knowledge, semantic grounding, and reasoning capabilities.
By decoupling low-level perceptual grounding from action policy learning, pre-trained VLMs provide a flexible foundation that can be adapted to various robotic tasks with limited additional supervision.
We now introduce a selection of representative VLM backbones that have been employed in VLA models.
\begin{itemize}[leftmargin=2em]
  \item \textbf{PaLM-E}~\cite{driess2023palme}, developed by Google, has been used—along with PaLI-X~\cite{chen2024palix}—as the backbone for RT-2 and its successor VLA models.
  \item \textbf{PaliGemma}~\cite{beyer2024paligemma} combines Gemma~\cite{gemmateam2024gemma} with SigLIP~\cite{zhai2023siglip}, and is used in $\pi_{0}$~\cite{Pi-0} and $\pi_{0.5}$~\cite{Pi-0.5} developed by Physical Intelligence.
  \item \textbf{PrismaticVLM}~\cite{karamcheti2024prismaticvlm} is based on LLaMA 2~\cite{touvron2023llama2} and combines it with DINOv2~\cite{oquab2023dinov2} and SigLIP~\cite{zhai2023siglip}.
  It is widely used in current VLA models, including OpenVLA~\cite{kim2024openvla} and CogACT~\cite{CogAct}.
  \item \textbf{Qwen2.5-VL}~\cite{bai2025qwen25vl}, developed by Alibaba, combines Qwen2.5 LLM~\cite{qwen2025qwen25} with a ViT-based vision encoder.
  It is used in a variety of VLA models such as NORA~\cite{NORA}, Interleave-VLA~\cite{Interleave-VLA}, and CombatVLA~\cite{chen2025combatvla}.
  \item \textbf{LLaVA}~\cite{liu2023llava} integrates the LLaMA-based LLM Vicuna~\cite{chiang2023vicuna} with the vision encoder from CLIP~\cite{radford2021clip} via an MLP.
  It has been widely adopted in models such as OpenHelix~\cite{OpenHelix}, OE-VLA~\cite{OE-VLA}, and RationalVLA~\cite{RationalVLA}.
  \item \textbf{Gemini 2.0}~\cite{hassabis2024gemini2}, developed by Google, includes variants such as Gemini Robotics-ER for robotic question answering and Gemini Robotics, which extends its capabilities to VLA applications~\cite{GeminiRobotics}.
    \item \textbf{Fuyu-8B}~\cite{bavishi2023fuyu8b}: QUAR-VLA~\cite{QUAR-VLA} and MoRE~\cite{MoRE},
  \item \textbf{OpenFlamingo}~\cite{RoboFlamingo}: RoboFlamingo~\cite{RoboFlamingo}, DeeR-VLA~\cite{yue2024deervla}, and RoboMM~\cite{RoboMM},
  \item \textbf{BLIP-2}~\cite{li2023blip2}: 3D-VLA~\cite{3D-VLA}, 
  \item \textbf{LLaMA3.2}~\cite{grattafiori2024llama3}: FOREWARN~\cite{FOREWARN},
  \item \textbf{AnyGPT}~\cite{zhan2024anygpt}: SOLAMI~\cite{jiang2025solami},
  \item \textbf{Phi}~\cite{javaheripi2023phi}: TraceVLA~\cite{zheng2025tracevla}, UP-VLA~\cite{zhang2025upvla}, and HybridVLA~\cite{HybridVLA},
  \item \textbf{Molmo}~\cite{deitke2025molmo}: UAV-VLA~\cite{UAV-VLA},
  \item \textbf{VILA}~\cite{lin2024vila}: NaVILA~\cite{cheng2024navila} and HAMSTER~\cite{li2025hamster},
  \item \textbf{InternVL}~\cite{chen2025internvl} GO-1~\cite{AgiBotWorldColosseo},
  \item \textbf{Eagle-2}~\cite{li2025eagle2}: GR00T N1~\cite{Gr00t-N1},
  \item \textbf{Chameleon}~\cite{chameleonteam2025chameleon}: WorldVLA~\cite{WorldVLA}.
\end{itemize}
This demonstrates the extensive diversity in VLM backbones currently employed across the VLA landscape.

\textbf{Gradient insulation.}
An emerging trend in training VLA models involves preventing gradient flow from the action head into the vision-language backbone~\cite{driess2025insulating}. 
Allowing gradients from a randomly initialized action head to propagate can compromise pre-trained representations, resulting in unstable and inefficient training. 
Prior work demonstrates that this form of gradient insulation significantly improves both training stability and efficiency~\cite{driess2025insulating}.
GR00T N1.5~\cite{Gr00t-N1} also freezes the VLA model entirely, likely for similar reasons.
Similarly, RevLA~\cite{RevLA} also addresses catastrophic forgetting by gradually reversing the backbone model weights, inspired by model merging.

\textbf{Stability and efficiency heuristics.}
Re-Mix~\cite{hejna2025remix} adjusts the sampling weights of individual datasets based on excess loss, which quantifies the remaining potential for policy improvement within each domain.

}%
{%
  Vision and Language Model (VLM)はVLAのバックボーンとして頻繁に用いられている.
  代表的なVLMには, PaLM-E, PaliGemma, OpenFlamingo, PrismaticVLM, Qwen-VL, LLaVA, Gemini 2.0が挙げられる~\cite{PaLM-E, beyer2024paligemma, awadalla2023openflamingo, karamcheti2024prismaticvlm, bai2023qwenvl, liu2023llava, hassabis2024gemini2}.
  PaLM-E \cite{driess2023palme}はGoogleが開発したVLMであり, PaLI-X \cite{chen2024palix}とともにRT-2とその後継のVLAに用いられている.
  PaliGemma \cite{beyer2024paligemma}はGemma \cite{gemmateam2024gemma}とSigLIP \cite{zhai2023siglip}を組み合わせたVLMであり, Physical IntelligenceのPi-0 \cite{Pi-0}やPi-0.5 \cite{Pi-0.5}などに活用されている.
  PrismaticVLM \cite{karamcheti2024prismaticvlm}はLLaMA 2 \cite{touvron2023llama2}をベースに, DINOv2 \cite{oquab2023dinov2}とSigLIP \cite{zhai2023siglip}を組み合わせたVLMであり, OpenVLA \cite{kim2024openvla}やCogACT \cite{CogAct}を含め, 現在多くのVLAで用いられている.
  Qwen2.5-VL \cite{bai2025qwen25vl}はAlibabaが開発した, Qwen2.5 LLM \cite{qwen2025qwen25}とViTを組み合わせたVLMであり, NORA \cite{NORA}やInterleave-VLA \cite{Interleave-VLA}, CombatVLA \cite{chen2025combatvla}など, 様々なVLAで用いられる.
  LLaVA \cite{liu2023llava}はLLaMAベースのLLMであるVicuna \cite{chiang2023vicuna}とCLIP \cite{radford2021clip}のVision EncoderをMLPにより結合したVLMであり, OpenHelix \cite{OpenHelix}やOE-VLA \cite{OE-VLA}, RationalVLA \cite{RationalVLA}など, 多くのVLAで用いられている.
  Gemini 2.0 \cite{hassabis2024gemini2}はGoogleが開発したVLMであり, これをロボットにおける質問応答に特化させたGemini Robotics-ER, VLAにまで発展させたGemini Roboticsなどが開発されている \cite{GeminiRobotics}.
  この他にも, QUAR-VLA \cite{QUAR-VLA}やMoRE \cite{MoRE}はFuyu-8B \cite{bavishi2023fuyu8b}を, RoboFlamingo \cite{RoboFlamingo}やDeeR-VLA \cite{yue2024deervla}, RoboMM \cite{RoboMM}などはOpenFlamingo \cite{RoboFlamingo}を, 3D-VLA \cite{3D-VLA}はBLIP2 \cite{li2023blip2}を, FOREWARN \cite{FOREWARN}はLLaMA3.2 \cite{grattafiori2024llama3}を, SOLAMI \cite{jiang2025solami}はAnyGPT \cite{zhan2024anygpt}を, TraceVLA \cite{zheng2025tracevla}やUP-VLA \cite{zhang2025upvla}, HybridVLA \cite{HybridVLA}はPhi \cite{javaheripi2023phi}を, UAV-VLA \cite{UAV-VLA}はMolmo \cite{deitke2025molmo}を, NaVILA \cite{cheng2024navila}やHAMSTER \cite{li2025hamster}はVILA \cite{lin2024vila}を, GO-1 \cite{AgiBotWorldColosseo}はInternVL \cite{chen2025internvl}を, GR00T N1 \cite{Gr00t-N1}はEagle-2 \cite{li2025eagle2}を, WorldVLA \cite{WorldVLA}はChameleon \cite{chameleonteam2025chameleon}をバックボーンとして活用するなど, その幅は非常に広いことがわかる.
}%

\subsubsection{Post-training}
\label{subsubsec:post-training}
In contrast to pre-training, which relies on large-scale and diverse datasets, post-training requires high-quality, robot- and task- specific data.
As full fine-tuning typically demands substantial computational resources, an alternative strategy is to fine-tune only the action head while keeping the backbone weights frozen. Another approach is to use Low-Rank Adaptation (LoRA)~\cite{hu2022lora}, which enables computationally efficient fine-tuning with minimal performance degradation.
%

In addition, BitVLA~\cite{BitVLA} introduces a distillation-based approach to quantize the vision encoder, aiming to enable memory-efficient training. 
Specifically, the vision encoder is compressed to $1.58$ bits by distilling a full-precision encoder into a quantized student model. This strategy achieves substantial memory savings with minimal performance degradation, thereby facilitating efficient deployment on resource-constrained systems.


\textbf{Freezing backbone vs. full fine-tuning.}
When adapting pre-trained VLMs for robotic tasks, a critical design choice is whether to freeze the vision-language backbone or perform full fine-tuning. This decision involves fundamental trade-offs across multiple dimensions.

\textbf{(a) Computational efficiency:} Freezing the backbone requires significantly less GPU memory and training time as gradients only need to be computed for the action head, enabling training on consumer-grade GPUs. In contrast, full fine-tuning demands substantial computational resources, often requiring large GPU clusters and extended training periods, which limits accessibility for many researchers.

\textbf{(b) Domain adaptation:} Full fine-tuning excels by enabling end-to-end optimization that jointly learns perception and control, allowing the model to adjust to robot-specific visual patterns and domain-specific knowledge. Frozen backbones, however, cannot adapt to these domain shifts, potentially creating a gap between pre-trained representations and robotic perception requirements.

\textbf{(c) Performance-resource trade-off:} 
Full fine-tuning of VLA models often yields the highest task-specific performance when sufficient data and compute are available, but it incurs substantial computational cost. To mitigate this, parameter-efficient adaptation methods such as Low-Rank Adaptation (LoRA)\cite{hu2022lora} offer a compelling alternative. For instance, OpenVLA\cite{kim2024openvla} demonstrates that LoRA can achieve competitive performance while significantly reducing memory and compute requirements, enabling training on consumer-grade GPUs rather than large-scale clusters. Recent work has also explored intermediate strategies, such as staged unfreezing or selective fine-tuning of specific layers, to strike a balance between adaptation capability and efficiency.

\textbf{(d) Knowledge preservation:} Frozen backbones maintain the rich visual and linguistic representations learned from web-scale data, preventing catastrophic forgetting of general vision-language capabilities. Full fine-tuning, while allowing the model to specialize for robotic visual features and action-grounded language, risks degrading these pre-trained representations, potentially losing valuable general knowledge that could benefit zero-shot generalization.

\subsection{Inference}
\switchlanguage%
{%
To address latency during real-world execution, Real-Time Chunking (RTC)~\cite{RealTimeChunking} introduces an asynchronous action generation strategy. 
RTC mitigates delays by fixing previously executed actions while generating subsequent actions in the sequence.
This method uses soft masking to maintain temporal consistency with past trajectories while enabling dynamic replanning based on updated sensory inputs.

Furthermore, DeeR-VLA~\cite{Deer-VLA} is trained to enable action prediction at each layer of the transformer. 
If the difference between actions predicted from two consecutive layers is small, the remaining layers are skipped to accelerate inference. 
VLA-Cache~\cite{VLA-Cache} improves inference speed by identifying static tokens and reusing previously computed features from earlier steps.
}%
{%
事前学習が多様な身体性における大量のデータを扱うのに対して, 事後学習はより質を重視した, 扱うロボット・タスクに特化したデータを用いて学習を行う.
この際, ネットワーク全体をfine tuningするか, action headだけをfine tuningするか, またはLoRAによる軽量な学習を行うかという選択肢が存在する.
事後学習自体は, 事前学習と違う部分はデータの質と量が大半を占める.
その一方で, bit-VLAのようにdistillationによってvision encoderを量子化するような, 軽量化の技術も開発されている.
}%
%
%

\begin{figure*}[t]
  \centering
  \includegraphics[width=1.0\textwidth]{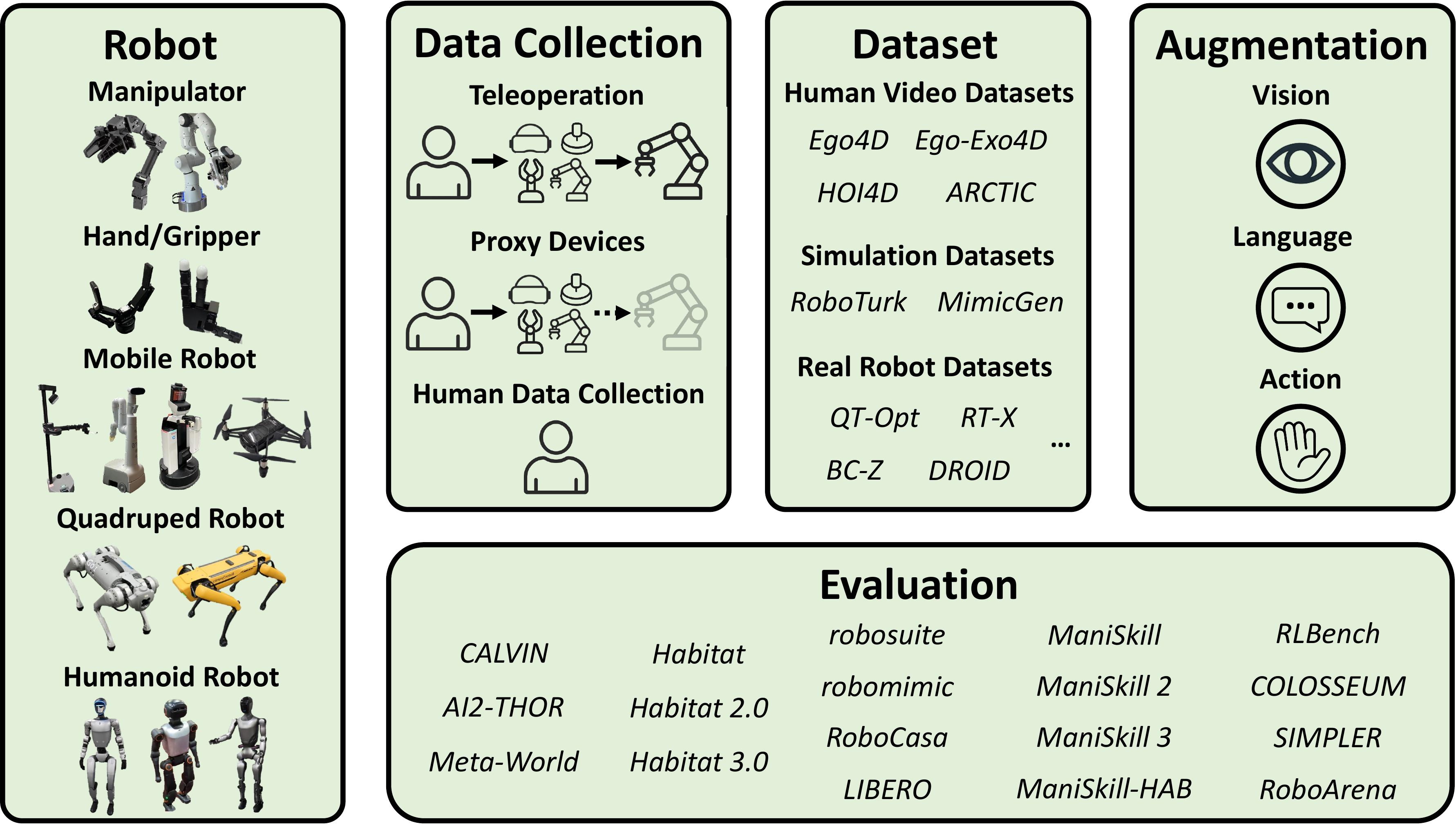}
  \caption{\textbf{Structure of \secref{sec:datasets} and \secref{sec:real_world}:} robots used VLA research --- including manipulator, hand/gripper, mobile robot, quadruped robot, and humanoid robot; data collection methods --- including teleoperation, proxy devices, and human data collection; publicly available dataset --- including human egocentric data, simulation data, and real-world robot data; augmentation for vision, language, and action; various evaluation benchmarks.}
  \label{figure:data}
\end{figure*}

\section{Datasets}\label{sec:datasets}
\subsection{Data Collection for VLA}\label{subsec:datacollection}
\switchlanguage%
{%
Training VLA models requires access to large volumes of high-quality data. 
This section outlines the primary data collection strategies employed in VLA research. 
Note that data collection via simulation is discussed in \secref{subsec:dataset}; here we focus on methods based on real devices.

\textbf{Teleoperation.}
  In this approach, demonstrations are recorded in real time while a human operator directly controls the robot, enabling the collection of high-quality trajectories. 
  This method forms the basis of many VLA datasets.
  For example, ALOHA~\cite{zhao2023aloha} employs a unilateral teleoperation setup consisting of a dual-arm WidowX 250 as the leader and a dual-arm ViperX-300 as the follower. 
  The follower robot mimics the leader's motions, allowing precise manipulation data to be captured.
  Mobile ALOHA~\cite{fu2024mobilealoha} extends this framework by mounting the system on a mobile base, enabling the collection of mobile manipulation demonstrations. 
  The ALOHA framework has evolved through multiple iterations. 
  ALOHA 2 introduces refined hardware components, such as upgraded grippers and gravity compensation mechanisms, along with open-source hardware and simulation environments~\cite{aloha2team2024aloha2}. Building on this upgraded platform, ALOHA Unleashed investigates large-scale imitation learning ~\cite{zhao25alohaunleashed}.
  Furthermore, Bi-ACT~\cite{buamanee2024biact} introduces bilateral control to enable more responsive interaction between the leader and follower robots, while GELLO~\cite{wu2024gello} adapts the system by employing a scaled-down follower robot with proportionally adjusted link lengths.

  In contrast to leader-follower approaches, which require robots on both the leader and follower sides, many prior works have proposed methods to reduce both the burden on the human operator and the overall cost of the teleoperation system.
  For instance, AnyTeleop~\cite{qin2023anyteleop} estimates the position and orientation of the human hands from a single RGB camera using MediaPipe~\cite{zhang2020mediapipehands}, and retargets this information to the robot via CuRobo~\cite{sundaralingam2023curobo} for teleoperation. 
  ACE~\cite{yang2025ace} combines precise wrist tracking using an exoskeleton device with hand pose estimation from Mediapipe to facilitate accurate teleoperation. 
  Aiming for applications in humanoid robotics, Open-Television~\cite{cheng2025opentelevision} utilizes hand and head pose estimation via the Apple Vision Pro to enable both teleoperation and active-vision-based manipulation. 
  Bunny-VisionPro~\cite{ding2024bunnyvisionpro} also employs the Apple Vision Pro, with greater emphasis on haptic feedback and real-time system integration.

  In addition to these approaches, data collection can also be performed through more direct control methods such as 3D mice or game controllers. While these alternatives simplify the setup and eliminate the need for wearable or vision-based pose estimation systems, they may offer lower fidelity in replicating natural human motions.

\textbf{Data collection using proxy devices.}
  Controlling a physical robot directly poses significant challenges for scaling data collection.
  By decoupling human motion from physical robot control, recent approaches enable more intuitive, flexible, and scalable data collection through the use of proxy devices.
  For example, UMI~\cite{chi2024umi} is a handheld gripper equipped with a GoPro camera, whose 6-DoF trajectory is estimated using visual SLAM. 
  The collected data can be used to train a policy, and by later mounting UMI as the robot's end-effector, the robot can reproduce the demonstrated motions without being physically involved during data collection.
  Recently, LBM~\cite{trilbmteam2025lbm} leverages UMI to collect $32$ hours of demonstrations.
  DexUMI~\cite{xu2025dexumi} extends this concept to dexterous manipulation by replacing the simple gripper with a five-fingered robotic hand. 
  The human demonstrator wears an exoskeleton glove equipped with the same cameras and tactile sensors as the target robot, allowing the recorded hand motions to be faithfully transferred.
  
  Building on similar principles, Dobb-E~\cite{Dobb-E} uses a rod-shaped device resembling the end-effector of Hello Stretch to capture human demonstrations. 
  RUMs~\cite{RUMs} further enhances this paradigm by increasing the diversity of collected tasks, incorporating failure detection mechanisms, and improving the network architecture. 
  These improvements enable the robot to generalize to a wide range of tasks through pre-training alone.
  DexCap~\cite{wang2024dexcap} is the device for data collection by mounting Realsense T265 cameras on Rokoko EMF gloves for both hands, along with additional Realsense T265 and L515 sensors on the chest, enabling SLAM-based 6-DOF wrist pose estimation and glove-based hand pose tracking. 
  In contrast, DexWild~\cite{tao2025dexwild} addresses the wiring complexity and SLAM calibration challenges of DexCap by using EMF gloves in combination with palm-facing cameras on both hands and ArUco marker tracking via external cameras.

\textbf{Human data collection.}
  This approach involves collecting data by recording natural human behavior without relying on proxy devices that mimic the robot's end effector. The simplest form of this method involves mounting a GoPro camera or microphone on the user's head to capture first-person visual and auditory data, often supplemented with inertial measurement unit (IMU) or gaze information. This technique has been widely adopted in large-scale egocentric datasets such as Ego4D and EPIC-KITCHENS~\cite{grauman2022ego4d, damen2018epickitchens, damen2022epickitchens100}.
  Recent advances in wearable sensing technologies have enabled more naturalistic and scalable data collection using compact smart glasses such as Meta's Project Aria. These devices have facilitated the development of enriched datasets including Ego Exo4D, HOT3D, HD EPIC, and Aria Everyday Activities~\cite{grauman2024egoexo4d, banerjee2024hot3d, perrett2025hdepic, lv2024ariaeverydayactivities}.
  Leveraging these datasets, several prior works have trained robot policies directly from human demonstration data.
  For instance, EgoMimic and EgoZero learn visuomotor control by imitating egocentric human behavior~\cite{kareer2024egomimic, liu2025egozero}. 
  Similarly, other studies use data collected with devices such as the Apple Vision Pro to train humanoid robot policies based on natural human motion~\cite{qiu2025humanoidpolicyhuman}.

\textbf{Data collection pipeline.}
  Data collection plays a pivotal role in training VLA models. These models require large-scale, high-quality datasets, and the data acquisition pipeline must be carefully designed to ensure both efficiency and diversity. 
  In the case of RT-1~\cite{brohan2023rt1}, a large-scale real-robot dataset is collected using a framework that samples instructions and randomized initial states from a curated instruction set. This approach enabled the collection of demonstrations across a broad range of tasks and environments, with human operators executing the sampled instructions to generate diverse and balanced data.
  In RoboTurk~\cite{mandlekar2018roboturk}, a 6-DOF teleoperation interface was developed using an iPhone, enabling the collection of large-scale robot manipulation demonstrations via a crowdsourcing platform~\cite{crowston2012mechanicalturk}.
  
  Furthermore, prior work~\cite{lynch2023languagetable, khazatsky2024droid} demonstrates the effectiveness of annotating pre-collected datasets with natural language. For example, the Language Table dataset~\cite{lynch2023languagetable} collects teleoperated trajectories and subsequently adds language annotations via crowdsourcing, resulting in a large-scale dataset with approximately 600,000 language-labeled trajectories. Similarly, DROID~\cite{khazatsky2024droid} conducts distributed data collection across 18 research institutions, gathering 76,000 trajectories and 350 hours of interaction data over 564 scenes and 86 tasks, which are later annotated with natural language through a crowdsourcing platform.
  
  However, since human annotation is costly, recent trends increasingly leverage foundation models such as VLMs to automate the annotation process. ECoT~\cite{zawalski2025ecot} and EMMA-X~\cite{sun2025emmax} combine object detection and gripper localization using Grounding DINO~\cite{liu2024groundingdino} and SAM\cite{kirillov2023sam}, and high-level plan and subtask generation using Gemini 1.0 to produce automatic annotations. 
  NILS~\cite{blank2024nils} is a framework that segments long-horizon robot videos and generates language annotations without human intervention. 
  It integrates multiple VLMs to detect keystates based on object state changes and gripper motions, and employs LLMs to generate natural language instructions. RoboMIND~\cite{wu2025robomind} also employs an annotation system based on Gemini \cite{geminiteam2024gemini15}, and demonstrates substantial performance improvements through pre-training with a VLA model.
  
  While such methods are more cost-effective and scalable than human post-hoc annotations, they face challenges such as fine-grained scene understanding and hallucinations. Particularly in methods like ECoT that rely solely on text, inconsistencies with actual visual context are more likely to occur. Approaches grounded in visual input, such as EMMA-X, or those integrating multiple perceptual modalities, such as NILS, have proven effective in addressing these issues.

}%
{%
  VLAを学習するためには, 大量の質の良いデータが必要である.
  ここでは, VLA研究で用いられているデータ収集方法についてまとめる.
  その収集方法は, 主に以下のように分類できる.
  \begin{enumerate}
    \item Online Data Collection
    \item Offline Data Collection
    \item Human Data Collection
  \end{enumerate}
  なお, シミュレーションによるデータ収集については, \secref{subsec:dataset}で述べ, ここでは実デバイスに基づくデータ収集方法について述べる.

  (1)は, 人間がロボットをその場で操作してデータを収集する方法である.
  この代表的な例がALOHA \cite{zhao2023aloha}である.
  リーダ側には双腕のWidowX 250, フォロワー側には双腕のViperX-300が用いらたユニラテラル制御を行う.
  人間がリーダ側を操作することで, フォロワー側のロボットでタスクを行い, 質の高いデータセットを得ることができる.
  また, このALOHAに台車型ロボットを組み合わせたMobile ALOHA \cite{fu2024mobilealoha}もあり, mobile manipulationのデータ収集が可能となった.
  このALOHAプロジェクトは, グリッパの改良や重力補償などのハードウェアの改善, ハードウェアやシミュレーションの公開を行ったALOHA 2, このALOHA 2における摸倣学習の可能性を突き詰めたALOHA Unleashedへと繋がっている \cite{aloha2team2024aloha2, zhao25alohaunleashed}.
  加えて, このALOHAはユニラテラル制御であったが, これをバイラテラル制御にしようという試みであるBi-ACT, リーダ側が大きなロボットである場合にフォロワー側をリンク長が相似なより小型のロボットにするGELLOなどの研究が行われている \cite{buamanee2024biact, wu2024gello}.

  これらはリーダがフォロワーと同様にロボットであるシステムだが, コストやリーダ側の負担を減らすための方法が様々提案されている.
  AnyTeleop \cite{qin2023anyteleop}は単一のカメラからMediaPipe \cite{zhang2020mediapipehands}やCuRobo \cite{sundaralingam2023curobo}を用いて手の位置や角度を推定, ロボットへとリターゲティングし, テレオペレーションしている.
  DexCap \cite{wang2024dexcap}はRealsense T265とRokoko EMFグローブを両手に, Realsense T265とL515を胸部に取り付け, SLAMベースの6DOF手首位置姿勢推定とグローブによる手のポーズ推定を行い, ポータブルなテレオペレーションを実現している.
  ACE \cite{yang2025ace}は外骨格型デバイスによる手首位置の正確な把握とMediapipe \cite{zhang2020mediapipehands}によるハンドポーズ推定を組み合わせたテレオペレーションを行っている.
  DexWild \cite{tao2025dexwild}はEMF式のグローブと両手のひらを見るカメラ, 外部カメラを用いたArUcoマーカのトラッキングによりテレオペレーションを行い, DexCapにおける両手への配線とSLAMにおけるキャリブレーションの問題を回避している.
  また, ヒューマノイドロボットへの適用を視野に入れ, Open-Television \cite{cheng2025opentelevision}はApple Vision Proからのハンド・ヘッドポーズ推定により, テレオペレーションのみならず能動的視覚を用いた操作を行うことができる.
  Bunny-VisionPro \cite{ding2024bunnyvisionpro}はOpen-Televisionと類似の研究であるが, 触覚フィードバックやリアルタイム性に重きが置かれたシステム構成をしている.
  これらの他にも, もちろん3Dマウスやゲームコントローラを用いてロボットを操作し, データを収集することもある.

  (2)は, ロボットを直接操作せず, 人間の動きとロボットの動きを切り離した形でデータを収集する方法である.
  これにより, 物理的なロボット無しでより直感的にデータを収集することができる.
  この代表的な例がUMI \cite{chi2024umi}である.
  UMIはGoProが取り付けられた手持ちグリッパであり, Visual SLAMによってグリッパの位置と姿勢の軌道を推定することができる.
  このデータを用いてポリシーを学習し, 今度はUMIをグリッパとしてロボットに取り付けることで, 実ロボットを一切動かさなくても動きを再現することができる.
  これを二指グリッパだけでなく五指ハンドにまで拡張したのがDexUMI \cite{xu2025dexumi}である.
  ロボッ側と同じカメラや接触センサのついた手の外骨格デバイスを人間が装着し, これを操作したときの動きをロボットで再現することができる.
  これらと同様の考え方により, Hello Stretchのスティック状の手先と同じ形のスティックデバイスによりデータを収集するDobb-E \cite{Dobb-E}や, これをもとにデータの多様化や失敗検知, ネットワーク構造の改善により事前学習のみで多様なタスクをこなすことが可能なRUMs \cite{RUMs}が開発されている.

  (3)は, ロボットと同じ手先のデバイスなどは使用せず, 単純に人間の動きをデータを収集する方法である.
  もっとも原始的な方法は, 頭にGoProやマイクロフォンなどを取り付け, 一人称視点の動画や音, 場合によってはIMUや視線データなどを収集する方法であり, Ego4DやEPIC-KITCHENSデータセットに使われている \cite{grauman2022ego4d, damen2018epickitchens, damen2022epickitchens100}.
  近年はより小型なスマートグラス型デバイスとしてProject ARIAが普及し, これを元にEgo-Exo4DやHOT3D, HD-EPIC, Aria Everyday Activitiesなどのデータセットが公開されている \cite{grauman2024egoexo4d, perrett2025hdepic, lv2024ariaeverydayactivities}.
  また, これらのデータを用いてロボットのポリシーを学習するEgoMimicやEgoZeroが提案されている \cite{kareer2024egomimic, liu2025egozero}.
  どうようの形でApple Vision Proを用いてデータを収集しポリシーを学習する研究もある[Humanoid Policy Human Policy] \cite{qiu2025humanoidpolicyhuman}.
}%

\subsection{Datasets for VLA}\label{subsec:dataset}

\begin{table*}[t]
\centering
\caption{\textbf{Recent real‑world robot datasets used in VLA research.} Here, \emph{Skill} denote atomic action primitives (e.g., pick, place, reach), whereas \emph{Task} correspond to instruction‑level goals. All statistics are reported as in the original papers; the table is adapted from prior works~\cite{fang2023rh20t, khazatsky2024droid, wu2025robomind, ma2024survey}.}
\label{tab:real_robot_datasets}
\scriptsize
\begin{tabularx}{\textwidth}{l|r|X|X|X|X|X}
\hline
\textbf{Name} & \textbf{Episodes} & \textbf{Skill} & \textbf{Task} & \textbf{Modality} & \textbf{Embodiment} & \textbf{Collection} \\
\hline
%
%
QT-Opt \cite{kalashnikov2018qtopt} & 580K & 1 (Pick) & NA & RGB & KUKA LBR iiwa & Learned \\
MT-Opt \cite{kalashnikov2021mtopt} & 800K & 2 & 12 & RGB, L & 7 robots & Scripted, Learned \\
RoboNet \cite{dasari2020robonet} & 162K & NA & NA & RGB & 7 robots & Scripted \\
BridgeData \cite{ebert2022bridgedata} & 7.2K & 4 & 71 & RGB, L & WidowX 250 & Teleop \\
BridgeData V2 \cite{walke2023bridgedatav2} & 60.1K & 13 & NA & RGB-D, L & WidowX 250 & Teleop \\
BC-Z \cite{jang2022bcz} & 26.0K & 3 & 100 & RGB, L & Google EDR & Teleop \\
Language Table \cite{lynch2023languagetable} & 413K & 1 (Push) & NA & RGB, L & xArm & Teleop \\
RH20T \cite{fang2023rh20t} & 110K & 42 & 147 & RGB-D, L, F, A & 4 robots & Teleop \\
RT-1 \cite{brohan2023rt1} & 130K & 12 & 700+ & RGB, L & Google EDR & Teleop \\
OXE \cite{oneill2024openxembodiment} & 1.4M & 527 & 160,266 & RGB-D, L & 22 robots & Mixed \\
DROID \cite{khazatsky2024droid} & 76K & 86 & NA & RGB-D, L & Franka & Teleop \\
FuSe \cite{jones24fuse} & 27K & 2 & 3 & RGB, L, T, A,  & WidowX 250 & Teleop \\
RoboMIND \cite{wu2025robomind} & 107K & 38 & 479 & RGB-D, L & 4 robots & Teleop \\
AgiBot World \cite{AgiBotWorldColosseo} & 1M & 87 & 217 & RGB-D, L & AgiBot G1 & Teleop \\
%
%
\hline
\end{tabularx}
\end{table*}


\switchlanguage%
{%
We outline key datasets used in the pre-training of VLA models.
Since the development of VLAs builds upon advances in LLMs and VLMs, a wide range of web-based datasets are leveraged.
In this section, we focus specifically on datasets used for \emph{pre-training} of VLA models, grouped into three main categories.
Datasets used for post-training are typically proprietary or integrated into evaluation benchmarks such as CALVIN~\cite{mees2022calvin} and LIBERO~\cite{liu2023libero}, and are therefore excluded from this summary.

\textbf{Human datasets.}
  Collecting human data is significantly more scalable than collecting robotic data, as it does not require access to physical robots, precise calibration, or safety-critical execution environments. 
  While third-person visual data is still used, first-person data has become particularly important for VLA pre-training because it more closely approximates the perceptual input received by real-world robots, especially those equipped with head-mounted sensors or human-like embodiments. 
  As a result, first-person visual data is now widely adopted as a key resource for pre-training VLA models.
  %
  %
  For example, Ego4D~\cite{grauman2022ego4d} is one of the largest and most comprehensive egocentric video datasets, comprising over 3,000 hours of head-mounted RGB footage collected from more than 800 participants across 74 cities in 9 countries. 
  Other notable examples include EPIC-KITCHENS~\cite{damen2018epickitchens, damen2022epickitchens100}, which documents everyday kitchen activities, and HOI4D~\cite{liu2022hoi4d}, which captures fine-grained human-object interactions.
  Several datasets focus specifically on manipulation tasks. 
  OAKINK2~\cite{zhan2024oakink2} and H2O~\cite{kwon2021h2o} capture bimanual object manipulation using RGB-D sensors and motion capture systems. 
  ARCTIC~\cite{fan2023arctic} centers on interaction with articulated objects through dexterous bimanual manipulation, while EgoPAT3D~\cite{li2022egopat3d} focuses on human action target prediction from egocentric views.

  Moreover, the advent of smart-glass-based recording devices has enabled more naturalistic and unobtrusive egocentric data collection (see \secref{subsec:datacollection}). 
  Notable examples include Aria Everyday Activities~\cite{lv2024ariaeverydayactivities}; Ego-Exo4D~\cite{grauman2024egoexo4d}, which integrates egocentric and exocentric perspectives; HOT3D~\cite{banerjee2024hot3d}, focused on fine-grained hand-object tracking; and HD-EPIC~\cite{perrett2025hdepic}, which extends egocentric cooking data.
  These datasets are frequently used for pre-training VLA models, often via latent action prediction approaches such as LAPA~\cite{ye2025lapa}. 
  Although not egocentric, large-scale video-language datasets like HowTo100M~\cite{miech2019howto100m}, Something-Something V2~\cite{goyal2017something}, and Kinetics-700~\cite{carreira2022kinetics700} are also used for model pre-training and are sometimes adapted for VLA-related tasks.
  As VLA research increasingly employs humanoid robots and systems with human-like sensory configurations, egocentric datasets, particularly those capturing natural, goal-directed behavior, are expected to play an increasingly vital role.

\textbf{Simulation datasets.}
  Simulation environments have long been used to generate robotic datasets in a scalable, safe, and cost-effective manner. 
  They support controlled data collection and flexible manipulation of scene configurations, making them particularly suitable for imitation learning and large-scale model pre-training.
  For example, RoboTurk~\cite{mandlekar2018roboturk} consists of task demonstrations on Sawyer robots within the MuJoCo physics engine~\cite{todorov2012mujoco}, collected via remote human teleoperation over the cloud. 
  However, collecting large-scale demonstration data in simulation, particularly via teleoperation, can still be time-consuming.
  To mitigate this limitation, MimicGen~\cite{mandlekar2023mimicgen} introduces a framework for generating large-scale datasets from a small number of expert demonstrations. 
  It decomposes demonstrations into object-centric subtasks and synthesizes new trajectories by transforming and recomposing them into novel scenes. 
  DexMimicGen~\cite{jiang2024dexmimicen} extends this approach to more complex embodiments, such as dual-arm robots and multi-fingered hands.
  
  In parallel, large-scale video world models such as COSMOS~\cite{nvidia2025cosmosworldfoundationmodel} have been developed to generate diverse imagined trajectories, providing rich and scalable training data for VLA models.
  
  Although simulation played a central role in early VLA research, its dominance has declined with the increasing availability of large-scale real-world robot datasets (see the next category, which covers real robot datasets). 
  Nonetheless, simulation remains a powerful tool for producing diverse, controllable data—particularly when real-world collection is impractical or cost-prohibitive.

\textbf{Real robot datasets.}
  Real-world robot datasets play a crucial role in the development and evaluation of VLA models. Collected on physical robot hardware, these datasets offer diverse embodiments, realistic interactions, and rich sensory inputs that are essential for training models capable of generalizing to real-world tasks.
  MIME~\cite{sharma2018mime} is one of the first large-scale robotic datasets. It contains $8.2$K trajectories across 20 tasks, consisting of paired human demonstrations and kinesthetic teaching of a Baxter robot performed by humans.
  Concurrently, QT-Opt~\cite{kalashnikov2018qtopt} has been introduced, comprising 580,000 grasp attempts collected over four months using seven KUKA LBR iiwa robotic arms.
  MT-Opt~\cite{kalashnikov2021mtopt}, an extension of QT-Opt, expands the task scope beyond grasping to support a wider range of manipulation skills.
  RoboNet~\cite{dasari2020robonet} contains $162,000$ trajectories gathered across seven robot types—Sawyer, Baxter, WidowX, Franka Emika Panda, KUKA LBR iiwa, Fetch, and Google Robot. Although the trajectories are generated using random or rule-based actions rather than expert demonstrations, the dataset supports research on generalization across diverse platforms and environments.
  BridgeData~\cite{ebert2022bridgedata} is collected via VR teleoperation using an Oculus Quest 2 and a WidowX 250 robot. It consists of $7,200$ trajectories across $10$ environments and $71$ tasks. 
  An extension of this work, BridgeData V2~\cite{walke2023bridgedatav2}, scales the dataset to $60,000$ trajectories across 24 diverse environments.
  BC-Z~\cite{jang2022bcz} involves $12$ Google Robots operated by seven human teleoperators performing over 100 manipulation tasks. Additional data are collected through policy executions with human oversight, resulting in 25,900 trajectories.
  Language Table~\cite{lynch2023languagetable} contains $600,000$ block manipulation trajectories ($413$K for real-world and $181$K for simulated trajectories) paired with natural language instructions. The data are collected through long, goal-free demonstrations and annotated via crowdsourcing to support instruction-conditioned training.
  RH20T~\cite{fang2023rh20t} provides multimodal data collected from four robots (Franka Emika Panda, UR5, KUKA LBR iiwa, and Flexiv Rizon) across 147 tasks and seven configurations. Unlike earlier datasets, it includes synchronized RGB-D, 6-axis force-torque, joint torque, and audio signals—supporting multimodal perception and control.
  RT-1~\cite{brohan2023rt1} comprises 130,000 real-world robotic demonstration trajectories collected over 17 months using 13 Google Robots. It serves as the foundation for the RT-series of transformer-based VLA models for real-time, instruction-conditioned behavior.
  Finally, Open-X Embodiment (OXE) dataset~\cite{oneill2024openxembodiment} unifies many of these datasets, including RT-1, BC-Z, BridgeData, and Language Table—into a standardized format using the RLDS schema~\cite{ramos2021rlds}. Developed through a large-scale collaboration involving 21 institutions and 173 authors, OXE dataset represents one of the most comprehensive and widely adopted real-robot VLA datasets to date.

  Several additional real-world robot datasets have been released to further advance VLA research. 
  DROID~\cite{khazatsky2024droid} is a large-scale dataset comprising $76,000$ trajectories collected across $13$ institutions using a standardized hardware setup. Each participating lab used a Franka Emika Panda arm equipped with a Robotiq 2F-85 gripper, two external stereo cameras, and a wrist-mounted camera. Unlike Open X-Embodiment dataset, which aggregates data from heterogeneous robot platforms, DROID ensures consistency across environments and embodiments, making it well-suited for benchmarking.
  FuSe~\cite{jones24fuse} provides $27,000$ multimodal trajectories collected using a WidowX 250 platform. The robot is outfitted with external cameras, a wrist-mounted camera, DIGIT tactile sensors, microphones, and an IMU, enabling rich cross-modal learning for VLA tasks.
  RoboMIND~\cite{wu2025robomind} offers $107,000$ trajectories collected from a diverse set of robot embodiments, including single-arm, dual-arm, humanoid, and dexterous-hand configurations. The dataset emphasizes diversity in morphology and manipulation strategies, supporting research in generalization and transfer.
  AgiBot World Dataset~\cite{AgiBotWorldColosseo} is a massive-scale dataset comprising 1 million trajectories collected using over 100 AgiBot G1 robots. Its unprecedented scale enables training of large VLA models under highly diverse conditions.
  In addition to these major releases, several task-specific or platform-specific datasets have been introduced, including Task-Agnostic Robot Play~\cite{rosete2022tacorl, mees23taco2}, Jaco Play~\cite{dass2023jacoplay}, Cable Routing~\cite{luo2024cablerouting}, Berkeley Autolab UR5~\cite{BerkeleyUR5Website}, TOTO~\cite{zhou2023toto}, and RoboSet~\cite{bharadhwaj2024roboagent}.
  Navigation-focused VLA datasets have also emerged, such as SACSoN~\cite{hirose2023sacson}, SCAND~\cite{karnan2022scand}, RECON~\cite{shah2021recon}, and BDD100K~\cite{yu2020bdd100k}, which support instruction-following and goal-directed behaviors in mobile platforms.
  Finally, specialized datasets such as RoboVQA~\cite{sermanet2024robovqa} target robot-specific question answering, further broadening the scope of VLA applications beyond manipulation and navigation.
}%
{%
  VLA構築に向けた様々なデータセットについて述べる.
  VLAの構築は, LLMやVLMの構築をベースとしているため, 様々なウェブデータセットの元に成り立っている.
  一方ここでは, VLAの事前学習に用いられるような, 以下のの3つのカテゴリのデータセットについてまとめる.
  なお, 事後学習については基本的に自前のデータセットやCALVIN~\cite{mees2022calvin}, LIBERO~\cite{liu2023libero}などのEvaluationに含まれるデータセットを用いるため, これらについては述べない.
  \begin{enumerate}
    \item Human Egocentric Dataset
    \item Simulation Dataset
    \item Real Robot Dataset
  \end{enumerate}

  (1)は人間の一人称視点から得られた画像を含むデータセットである.
  三人称視点に比べて, 実ロボットに近い形の視覚情報が得られるため, 多くの事前学習に用いられている.
  その代表的な例がEgo4D~\cite{grauman2022ego4d}である.
  人間の頭にRGBカメラを取り付け, 3000時間以上, 74都市, 9カ国, 800人超の参加者によって収集された世界最大級の一人称視点動画データセットを構築している.
  これに似たデータセットには, キッチンにおける日常的な活動を対象としたEPIC-KITCHENS, HOI (Human Object Interaction)を対象としたHOI4Dなどが構築されている \cite{damen2018epickitchens, damen2022epickitchens100, liu2022hoi4d}.
  また, モーションキャプチャやRGB-Dカメラを使用した, 両手による物体操作タスクのためのOAKINK2やH2O, 両手で器用に操作されるarticulated objectsに関するARCTIC, 人間の行動目標予測に着目したEgoPAT3Dなどのデータセットもある \cite{zhan2024oakink2, kwon2021h2o, fan2023arctic, li2022egopat3d}.
  現在は\secref{subsec:datacollection}に述べたようなスマートグラスを使い, Aria Everyday Activities, 一人称視点と三人称視点を組み合わせたEgo-Exo4D, 手と物体のトラッキングに主眼を置いたHOT3D, キッチンでの料理に主眼を置いたHD-EPICなどのデータセットが公開されている \cite{lv2024ariaeverydayactivities, grauman2024egoexo4d, banerjee2024hot3d, perrett2025hdepic}.
  これら一人称視点のデータセットは, LAPA \cite{ye2025lapa}のような潜在的アクション予測を行ったうえでVLAを学習するための事前学習に用いられることが多い.
  この他, 一人称視点ではないが, HowTo100MやSomething-Something V2, kinetics-700などの動画データセットが使われる場合もある \cite{miech2019howto100m, goyal2017something, carreira2022kinetics700}.
  今後人間と近い身体性を持つヒューマノイドロボットのVLAが構築されるに従って, これらのデータセットはより重要性を増すだろう.

  (2)はロボットのシミュレーション環境で収集されたデータセットである.
  RoboTurk \cite{mandlekar2018roboturk}は, クラウドを介した人間の操作に基づく, かつSawyerのMujoco \cite{todorov2012mujoco}上でのタスクデータセットである.
  MimicGen \cite{mandlekar2023mimicgen}は, 少数の人間のデモンストレーションから大規模摸倣学習データを自動収集するシステムである.
  人間のデモをオブジェクト中心のサブタスクに分割し, それらを様々な新しいシーンにおいて変換・合成することで多様なデモを合成する.
  DexMimicGen \cite{jiang2024dexmimicen}は, これを両腕・多指ハンドへと拡張したものである.
  また, COSMOS \cite{nvidia2025cosmosworldfoundationmodel}のように, 大規模な映像世界モデルを構築し, これにより多様なシーンを生成することでデータセットを構築する方法もある.
  一方で, 現在は(3)で述べる実機のデータセットが充実してきているため, シミュレーション環境でのデータ収集はそこまで主流ではない.

  (3)はロボット実機において収集されたデータセットである.
  MIME \cite{sharma2018mime}は大規模なロボットデータを集めた初期の例であり, 20のタスクについて, 人間のデモンストレーションとBaxterの人間によるkinesthetic teachingがセットになった, 8.2K軌道のデータセットである.
  QT-Opt \cite{kalashnikov2018qtopt}はスケーラブルならオフポリシー深層強化学習フレームワークであり, 7台のKUKA LBR iiwaで4ヶ月に渡って収集された580kの把持試行データが含まれている.
  QT-Optの派生として, 把持に限らない多様なタスクを扱ったMT-Opt \cite{kalashnikov2021mtopt}もある.
  RoboNet \cite{dasari2020robonet}は複数のロボット・複数の環境にわたる大規模なデータセットであり, 7種類のロボット(Sawyer, Baxter, WidowX 250, Franak Emika Panda, KUKA LBR iiwa, Fetch, Google Robot)で162kの軌道を収集している.
  なお, ランダム行動と単純なルールに基づいているため, その動きが最適化とは限らない.
  Bridge Data \cite{ebert2022bridgedata}はWidowX 250を使用し, Oculus Quest 2によるVRテレオペレーションに基づいて, 10環境, 71タスクの合計7.2k軌道を収集したデータセットである.
  これを拡張したのがBridge Data V2 \cite{walke2023bridgedatav2}であり, さらに多様な24環境での60k軌道を収集している.
  BC-Z \cite{jang2022bcz}は12台のGoogle Robotを用いて7人のオペレータが100種類以上のマニピュレーションタスクを行いデータを収集, これを学習したポリシーを人間の介入ありで実行し, さらにデータを収集することで, 合計25kの軌道データを収集している.
  Language Table \cite{lynch2023languagetable}は言語指示に基づきブロックを操作する600kの軌道データセットであり, 目的のない長期的な人間の遠隔教示を, クラウドソーシングで区切りながら言語ラベリングしている.
  RH20T \cite{fang2023rh20t}は, 4種類のロボット(Franka Emika Panda, UR5, KUKA LBR iiwa, Flexiv Rizon)を用いた7種類の構成において, 147のタスクを実行するデータセットである.
  これまでのデータセットと違い, RGB-D, 力覚, 音声, トルクなどのマルチモーダルな情報が含まれている.
  RT-1 \cite{brohan2023rt1}は, 13台のGoogle Robotを17ヶ月にわって使用し収集した, 130kの軌道ロボットデモンストレーションデータを含んでいる.
  そして, RT-X \cite{oneill2024openxembodiment}はこれまで述べたような多様なデータセットをRLDS \cite{ramos2021rlds}というフォーマットで統合したデータセット集合であり, 21の研究機関, 173人の著者が参加する一大プロジェクトとなった.

  その後もいくつかのデータセットが公開されている.
  DROID \cite{khazatsky2024droid}は13機関において, 18台の共通のロボットプラットフォームを用いて76kの軌道を収集したデータセットである.
  RT-Xのように別々のプラットフォームで収集されたデータではなく, Franak Emika PandaにRobotiq 2F-85グリッパがつき, 2台の外部ステレオカメラと1台の手首カメラがついたプラットフォームを各研究機関が保有することで整備されたデータが収集されている.
  FuSe \cite{jones24fuse}はWidowX 250に外部カメラ, 手首カメラ, DIGITセンサによる触覚, マイクロフォン, IMUが取り付けられたプラットフォームを用いて27kの軌道を収集した, マルチモーダルなVLA向けのデータセットである.
  RoboMIND \cite{wu2025robomind}はsingle arm, dual arm, humanoid, dexterous handにまたがる多様な身体性のロボットによる107kの軌道を収集したデータセットである.
  Agibot World Dataset \cite{AgiBotWorldColosseo}は100台以上のAgiBot G1を用いて収集された1Mの軌道を含む大規模なデータセットである.
  ここには紹介しきれなかったが, この他にもTask Agnostic Robot Play, Jaco Play, Cable Routing, Berkley Autolab UR5, TOTO, RoboSetなど, 様々なロボットデータセットが公開されている \cite{rosete2022tacorl, mees23taco2, dass2023jacoplay, luo2024cablerouting, BerkeleyUR5Website, zhou2023toto, bharadhwaj2024roboagent}.
  また, ナビゲーションにはSACSoN, SCAND, RECON, BDD 100kなどのデータセットが用いられている \cite{hirose2023sacson, karnan2022scand, shah2021recon, yu2020bdd100k}.
  加えて, RoboVQA \cite{sermanet2024robovqa}のようなロボット特有の質問応答データセットも存在する.
}%

\subsection{Data Augmentation for VLA}\label{subsec:augmentation}
\switchlanguage%
{%
Given the high cost of collecting datasets, various data augmentation methods have been developed to expand existing datasets. These approaches span multiple modalities, including vision, language, and action.

\textbf{Vision augmentation.}
  In most computer vision tasks, augmentation techniques such as rotation, cropping, and scaling are commonly used to improve generalization. 
  However, in robotics, where the robot's embodiment and its spatial relationship to the camera are critical, such transformations can distort these relationships and negatively affect performance. 
  To address this, recent methods have proposed using image generation models, such as Stable Diffusion~\cite{rombach2022stablediffusion}, to perform embodiment-aware augmentations.
  CACTI~\cite{mandi2022cacti} leverages Stable Diffusion to modify a specific region of images to augment a small, yet high-quality dataset.
  GenAug~\cite{chen2023genaug} introduces more sophisticated visual augmentation by leveraging Stable Diffusion to apply three types of transformations: altering object textures, inserting task-irrelevant distractors, and modifying backgrounds. 
  These augmentations aim to improve policy robustness by increasing visual diversity while preserving task-relevant semantics.
  ROSIE~\cite{yu2023rosie} builds on CACTI and GenAug by using an LLM, OWL-ViT~\cite{minderer2022owlvit}, and Imagen Editor~\cite{wang2023imageneditor} to automatically identify and modify masked regions based on text prompts, enabling controlled edits to target objects, backgrounds, or the insertion of new objects.
  The augmented data is used to train RT-1~\cite{brohan2023rt1}.
  DreamGen~\cite{DreamGen} utilizes a video world model to generate diverse visual variations, paired with an inverse dynamics model (IDM) to infer the corresponding actions. 
  This combination enables the synthesis of training data, facilitating policy learning in novel environments and enhancing generalization. 
  In contrast, MOO~\cite{stone2023moo} forgoes explicit visual augmentation and instead disentangles object and skill representations using a vision-language model (VLM), allowing policies to generalize to unseen object-skill combinations from limited data. 
  It addresses visual variability implicitly by leveraging the broad generalization capabilities of pre-trained VLMs. 
  Moreover, BYOVLA~\cite{BYOVLA} extracts and inpaints task-irrelevant regions in image observations during runtime, aiming to enhance robustness against visual distractions.

\textbf{Language augmentation.}
  DIAL~\cite{ziao2023dial} starts with a small, manually labeled seed set of trajectory-instruction pairs. 
  A VLM is trained on this seed set to compute similarity between trajectories and instructions. 
  Simultaneously, an LLM generates diverse paraphrases of the seed instructions, forming a large pool of candidates. 
  These are then matched to the remaining unlabeled trajectories using the trained VLM, and the top-k most similar instructions are assigned. The resulting dataset is used to train RT-1~\cite{brohan2023rt1}.

\textbf{Action augmentation.}
  Since actions are directly tied to the robot's physical behavior and embodiment, augmenting action data is generally challenging. 
  A common approach to address this challenge is dataset expansion through interactive methods such as DAgger~\cite{ross11dagger}, which iteratively collects expert actions in states visited by the learned policy. 
  Similarly, CCIL~\cite{ke2024ccil} generates corrective data when a policy encounters out-of-distribution states by learning a locally smooth dynamics model. 
  It synthesizes actions that guide the robot from novel states back to expert-visited ones, and the resulting corrective data is combined with the original demonstrations to refine the policy.
}%
{%
  データセットの収集は非常にコストがかかるため, 収集したデータセットを拡張するデータ拡張手法が提案されている.
  これには, 視覚・言語・アクションの各モダリティに対する拡張手法がある.
  \begin{enumerate}
    \item Vision Augmentation
    \item Language Augmentation
    \item Action Augmentation
  \end{enumerate}

  (1)は, 視覚モダリティに対するデータ拡張手法である.
  通常の画像認識タスクにおいては, 画像の回転や切り抜き, 拡大縮小などのデータ拡張手法が広く用いられている.
  一方で, ロボットには身体性があり, ロボットとカメラの位置関係には意味があるため, 画像の回転や拡大縮小は, それらの身体性を変化させることとなってしまう.
  そのため, 主にStable Diffusion~\cite{rombach2022stablediffusion}のような画像生成モデルを用いたデータ拡張手法がいくつか提案されている.
  CACTI \cite{mandi2022cacti}は少量の質の高いデータセットを収集し, Stable Diffusionにより特定の領域を変化させて拡張, R3M \cite{nair2023r3m}により視覚表現を圧縮して, その後模倣学習を行う手法である.
  GenAug \cite{chen2023genaug}はより高度な画像データ拡張を行っており, Stable Diffusionにより特定の物体に異なるテクスチャを与える, タスクとは無関係なdistractorを加える, 背景を変更するという3つの拡張を行い, よりロバストなポリシーの学習を行う.
  ROSIE \cite{yu2023rosie}はCACTIやGenAugをさらに発展させた手法であり, LLMとOWL-ViT \cite{minderer2022owlvit}, Imagen Editor \cite{wang2023imageneditor}を使い, テキストを用いて自動的にマスクする領域を決定, 適切かつ自動的に操作対象や背景の変更, 新しい物体の追加が可能である.
  得られたデータをもとにRT-1 \cite{brohan2023rt1}を学習している.
  DreamGen \cite{DreamGen}はVideo World Modelを用いて多様な画像変化を生成し, Inverse Dynamics Model (IDM)から画像変化に対するアクションを生成, これらをデータセットとしてポリシーを学習する手法である.
  世界モデルによって, 様々な新しい環境・動作を生成することができ, ポリシーの汎化性能を高めることができる.
  Moo \cite{stone2023moo}は少し特殊であるが, これまでのように拡散モデルを使って画像を拡張するのではなく, VLMを用いて物体情報とスキル情報を区別することで, 少ないデータから様々な物体とスキルの組み合わせに対応できるようなモデルを学習している.
  この手法は明示的なデータ拡張を施しているわけではないが, VLMが持つ広い知識に物体の情報を任せることで, 画像データの拡張を不要にしている.
  また, BYOVLAはVLAの実行時にタスクに無関係な領域を抽出してインペインティングすることで, 視覚ノイズや背景への変化への頑健性を高めている\cite{BYOVLA}.
  一方で, 現在は多くのデータセットが集まってきたことと, DINOv2 \cite{oquab2023dinov2}やSigLIP \cite{zhai2023siglip}のような強力な視覚モデルが登場したことにより, 視覚モダリティに対するデータ拡張手法はあまり提案されていない.

  (2)は, 言語モダリティに対するデータ拡張手法であるが, 明示的に言語を拡張している研究は少ない.
  その代表例としてDIAL \cite{ziao2023dial}が挙げられる.
  DIALは, 大量の軌道データの一部に正確な言語指示のアノテーションを施し(データセットA), これに基づいて残りの大規模な軌道データの全て(データセットB)に自動的に言語指示を複数付与する手法である.
  データセットAを用いて軌道と言語指示の間の類似度が計算可能なVLMを構築すると同時に, LLMを用いてデータセットAの言語指示を様々な形で変化させ, 考えられる言語指示の候補を大量に取得する.
  先ほど学習されたVLMを用いて, この大量の言語指示候補と残りの大規模軌道データセットBの間の類似度を計算し, 例えば類似度の高い言語指示の上位3つ選択して付与する.
  最後に, これらのデータセットを全て合わせてRT-1 \cite{brohan2023rt1}の学習を行う.

  (3)は, アクションモダリティに対するデータ拡張手法である.
  アクションというのは直接ロボットの動作に関わるところであるため, 基本的にはデータ拡張は難しい.
  その一方で, CCIL \cite{ke2024ccil}は, expert dataから得られたpolicyが予想外の状態に入った時に, それを正しい方向に戻すための修正データを自動生成して学習を補う手法を提案している.
  これは, 局所的に滑らかなダイナミクスモデルを学習し, それを用いて正しい軌道に戻すための修正アクションデータを生成, 専門家データと混合して再学習させる.
}%

\section{Review of Real-World Robot Applications}\label{sec:real_world}

\switchlanguage%
{%
  In this section, we summarize key practical aspects of VLA research, including the types of robots used, data collection methodologies, publicly available datasets and augmentation techniques, and the evaluation protocols applied to assess model performance.
}%
{%
  ここでは, これまで述べたVLAではどのようなロボットが用いられ, どのようにデータ収集が行われ, どのようなデータセットが公開され, どのように拡張され, 評価されてきているのかについてまとめる.
}%

\subsection{Robot for VLA}\label{subsec:robot}
In this section, we present an overview of the types of robots commonly employed in VLA research. 
\switchlanguage%
{%


\textbf{Manipulator.} 
  Robotic manipulators are the most commonly used robots in VLA research, encompassing both single-arm and dual-arm configurations.
  Single-arm robots used in the prior works reviewed in this survey include: Franka Emika Panda, Franka Research 3, UR5, UR5e, UR3, UR3e, UR10, Kinova Gen3, Kinova Jaco 2, Sawyer, KUKA LBR iiwa 14, UFactory xArm, DENSO Cobotta, FANUC LR Mate 200iD, Realman RM65-B, Realman RM75-6F, AgileX PiPER, Unitree Z1 Pro, Dobot, Flexiv Rizon, AIRBOT Play, ARX, DLR SARA~\cite{iskandar2020sara}, WidowX 250 6DoF, ViperX 300 6DoF, SO-100/101, and PAMY2~\cite{guist2024safe}.
  These manipulators typically feature 5, 6, or 7 degrees of freedom (DoFs).
  The joint configurations and link lengths vary across these manipulators. 
  PAMY2 uses pneumatic actuation, reflecting the diversity of robotic embodiments.
  In addition, several systems (e.g., AgileX PiPER, ARX, Franka Emika Panda, UFactory xArm, UR5e, AIRBOT Play, ALOHA~\cite{zhao2023aloha}, and ALOHA2~\cite{aloha2team2024aloha2} adopt a bimanual configuration by placing two arms side by side.
  WidowX, ViperX, ALOHA, SO-100/101, and PAMY2 are fully open-source in hardware, allowing researchers to flexibly modify or extend their physical embodiment.
  These manipulators are used to perform a wide range of tasks, including object grasping and relocation, assembly, manipulation of deformable objects, and peg-in-hole insertion. 

\textbf{Hand / Gripper.}
  This category refers to the hands and grippers that serve as end-effectors mounted on the manipulators described above.
  Hands used in prior works in VLA include the ROBOTERA Xhand, PSYONIC Ability Hand, Inspire Robots RH56, Shadow Hand, PsiBot G0-R, Robotiq 2F-85/140, LEAP Hand, and UMI. 
  These vary in design: the LEAP Hand~\cite{shaw2023leaphand} has four fingers; the Robotiq Gripper and UMI~\cite{chi2024umi} are two-fingered; the others are five-fingered.
  Some systems also use suction cups or task-specific grippers, as in Shake-VLA~\cite{Shake-VLA}. Platforms such as ALOHA, ARX, and PiPER typically include two-fingered grippers by default.
  The LEAP Hand and UMI are open-source, allowing easy hardware modification. While two-fingered grippers are suited for grasping, four- and five-fingered hands enable tool use and in-hand manipulation.  

\textbf{Mobile robot.}
  Mobile robots in VLA research include both wheeled platforms and mobile manipulators that combine robotic arms with mobile bases. Jackal and TurtleBot 2 are examples of systems that rely exclusively on wheeled locomotion and do not incorporate manipulation capabilities. 
  In contrast, mobile manipulators exhibit diverse configurations, including single-arm platforms such as Hello Stretch, Google Robot, and LoCoBot, as well as dual-arm systems like Mobile ALOHA, PR2, Fibocom, and AgiBot G1.
  LoCoBot and TurtleBot 2 are also notable for their fully open-source hardware, which facilitates embodiment customization and experimentation. 
  Mobile platforms enable locomotion and environmental interaction capabilities beyond those afforded by stationary arms or grippers, supporting tasks that involve navigation and dynamic scene engagement. 
  Some models, such as RT-1, are capable of performing navigation and manipulation concurrently.

\textbf{Quadruped robot.}
  Quadruped robots, characterized by their animal-like locomotion, have been increasingly considered in VLA research due to their ability to navigate unstructured and uneven environments. 
  Unitree A1, Go1, Go2, B1, Boston Dynamics Spot, and ANYmal are frequently used. 
  These are all commercially available systems capable of traversing complex terrain using RL-based control policies.
  These platforms not only provide locomotion but can also be equipped with manipulators to support a wide range of manipulation tasks.

\textbf{Humanoid robot.}
  Humanoid robots, characterized by body structures resembling those of humans, represent another category of platforms explored in VLA research. 
  In prior works, Fourier GR-1, Unitree G1, Unitree H1, and Booster T1 are often used. 
  These systems typically possess two legs, two arms, and five-fingered hands attached to their end effectors.
  Their human-like morphology makes them well-suited for operating in spaces designed for humans and facilitates compatibility with VLAs trained on human motion datasets.
}%
{%
  VLAに用いられているロボットにはどのような種類があるだろうか.
  主に, VLAに用いられているロボットは以下のように分類できる.
  \begin{enumerate}
    \item Manipulator
    \item Hand / Gripper
    \item Mobile Robot
    \item Quadruped Robot
    \item Humanoid Robot
  \end{enumerate}

  (1)は最も一般的な腕型のロボットであるが, これには単腕と双腕の2種類がある.
  これまで紹介した研究に用いられている単腕のロボットには, Franka Emika Panda, Franka Research 3, UR5, UR5e, UR3, UR3e, UR10, Kinova Gen3, Kinova Jaco 2, Sawyer, KUKA LBR iiwa 14, UFactory xArm, DENSO Cobotta, FANUC LR Mate 200iD, Realman RM65-B, Realman RM75-6F, AgileX PiPER, Unitree Z1 Pro, Dobot, Flexiv Rizon, AIRBOT Play, ARX, DLR SARA, WidowX 250 6DOF, ViperX 300 6DOF, SO-100/101, PAMY2がある.
  この中には, 5自由度, 6自由度, 7自由度のマニピュレータが含まれている.
  それぞれ関節の配置も違えばリンクの長さも異なっており, 一部空気圧によって動くなど(PAMY2), 多様な身体性が含まれている.
  また, 双腕のマニピュレータにはALOHA \cite{zhao2023aloha}やALOHA2 \cite{aloha2team2024aloha2}が含まれる.
  加えて, AgileX PiPERやARX, Franka Emika Panda, UFactory xArm, UR5e, AIRBOT Playなどはそれらを横に2つ並べたbimanualの構成を持つ場合もある.
  これらの中で, WidowX, ViperX, ALOHA, SO-100/101, PAMY2はハードウェアが全て公開されており, 研究者が容易に身体構成を変化させることができる.
  マニピュレータによって行われるタスクは, 物体の把持や移動, 組み立て, 柔軟物操作, peg-in-holeなど多岐にわたる.
  VLAを適用した面白い例としては, 両手でカクテルを作るタスク[Shake-VLA] \cite{Shake-VLA}や医療現場における器具の手渡し自動化[RoboNurse-VLA] \cite{RoboNurse-VLA}を行った例がある.

  (2)は, (1)のマニピュレータのエンドエフェクタとして取り付けられるハンドやグリッパである.
  これまで紹介した研究に用いられているハンドには, ROBOTERA Xhand, PSYONIC Ability Hand, Inspire Robots RH56, Shadow Hand, PsiBot G0-R, Robotiq 2F-85/140 Gripper, LEAP Hand, UMIがある.
  この中で, LEAP Hand \cite{shaw2023leaphand}は4指, Robotiq GripperとUMI \cite{chi2024umi}は2指であり, それ以外は5指ハンドである.
  これ以外にも, エンドエフェクタにSuction Cupを取り付けて物体を吸引するものや, Shake-VLA \cite{Shake-VLA}のようにタスクに合った形のグリッパを専用で設計するものもある.
  もちろん, ALOHAやARX, PiPERのように, 最初から2指のグリッパを持つものも多い.
  また, この中でLEAP HandとUMIはハードウェアが公開されており, 研究者が容易に身体構成を変化させることができる.
  2指のパラレルグリッパでは物体の把持と移動が限界であるが, 4指ハンドや5指ハンドでは, 様々な道具を扱う動きやin-hand manipulationも可能である.

  (3)は, 台車型の移動ロボットである.
  これまで紹介した研究に用いられているmobile robotには, Hello Stretch, Jackal, Mobile ALOHA \cite{fu2024mobilealoha}, Google Robot, PR2, Fibocom, AgiBot G1, LoCoBot, TurtleBot 2がある.
  これらのうち, JackalとTurtleBot 2のみが純粋な台車型ロボットであり, 残りがマニピュレータと台車型ロボットを組み合わせたmobile manipulatorである.
  mobile manipulatorと一口に言っても, 単腕のHello StretchやGoogle Robot, LoCoBot, 双腕のMobile ALOHA, PR2, Fibocom, AgiBot G1など構成は様々である.
  また, 稀にDJI TelloのようなUAVが移動ロボットとして用いられることもある.
  この中で, LoCoBotとTurtleBot 2はハードウェアが公開されており, 研究者が容易に身体構成を変化させることができる.
  これらの移動ロボットは, マニピュレータやハンドには出来なかった移動能力が備わり, ナビゲーションタスクが可能になる\cite{MobilityVLA}.
  また, この移動能力とマニピュレーションの協調を扱ったAC-DiT \cite{AC-DiT}やMoManipVLA \cite{MoManipVLA}も開発されている. 
  RT-1のような一部のVLAは, ナビゲーションとマニピュレーションを同時に行うことができる.
  また, UAVの自律飛行システムを構築するUAV-VLAやRaceVLA, CognitiveDrone, 車の運転を扱うOpenDriveVLA, ORION, CoVLA, OccLLaMAなども提案されており, 非常に幅広い応用が期待されている \cite{UAV-VLA, RaceVLA, CognitiveDrone, OpenDriveVLA, fu2025orion, arai2025covla, OccLLaMA}.

  (4)は, 動物のようにロコモーションが可能な四脚ロボットである.
  これまで紹介した研究に用いられているquadruped robotには, Unitree A1, Unitree Go1, Unitree Go2, Unitree B1, Boston Dynamics Spot, ANYMALが含まれる.
  どれも企業が開発したロボットであり, 強化学習を用いて様々な不整地を踏破することができる.
  また, 四脚ロボットの背中にマニピュレータを取り付けて, マニピュレーションを行うことも可能である.
  TrackVLA, MoRE, NaVILA, Uni-NaVid, CrossFormerなどのVLAはこれら四脚ロボットを用いたナビゲーションを行うことができる \cite{TrackVLA, MoRE, cheng2024navila, Uni-NaVid, doshi2025crossformer}.
  また, Track2Act, VidBotはマニピュレータ付きのSpot, SLIMはWidowX 250を接続したUnitree Go1の脚移動性能とマニピュレーション性能を組み合わせて活用している \cite{TraceVLA, VidBot, SLIM}.

  (5)は, 人間と同じような身体構成を持つヒューマノイドロボットである.
  これまで紹介した研究に用いられるヒューマノイドには, Fourier GR-1, Unitree G1, Unitree H1, Booster T1がある.
  そのどれもが2本の脚と2本の腕を持ち, 手先には五指ハンドが取り付けられている場合が多い.
  ヒューマノイドは人間のような身体構造を持つため, 人間のために作られた環境を上手に利用でき, かつ人間の動作データセットによって学習されたVLAと相性が良い.
  NaVILAやはヒューマノイドのロコモーション能力を, EgoVLAはヒューマノイドのマニピュレーション能力を, GO-1とLeVERBはヒューマノイドのロコマニピュレーション能力を活用したVLAである \cite{cheng2024navila, EgoVLA, AgiBotWorldColosseo, LeVERB}.

  また, ここでは扱わなかったが, ゲームや3Dキャラクターを扱うVLAとして, SOLAMIやCombatVLA, JARVIS-VLAなども提案されている \cite{jiang2025solami, chen2025combatvla, JARVIS-VLA}.
}%

\subsection{Evaluation for VLA}\label{subsec:evaluation}

\begin{table*}[t]
\centering
\caption{\textbf{Benchmarks for VLA evaluation.} This table shows various simulation environments used for evaluating VLA models with their key characteristics. Task types include Navigation (Nav), Manipulation (Manip), and Whole-Body Control (WBC). Observation modalities include RGB-D (RGB + Depth), S (Semantic segmentation), and PC (Point Cloud). The Scenes/Objects column indicates the number of available scenes and objects respectively.}
\label{tab:simulation_benchmarks}
\scriptsize
\begin{tabularx}{\textwidth}{l|X|X|X|l|X|l}
\hline
\textbf{Name} & \textbf{Task} & \textbf{Scenes/Objects} & \textbf{Observation} & \textbf{Physics} & \textbf{Built Upon} & \textbf{Description} \\
\hline
robosuite \cite{zhu2020robosuite} & Manip & NA / 10 & RGB-D, S & MuJoCo & NA & Modular framework, 11 tasks \\
robomimic \cite{mandlekar2021robomimic} & Manip & NA / NA & RGB & MuJoCo & robosuite & Offline learning, 8 tasks \\
RoboCasa \cite{nasiriany2024robocasa} & Manip & 120 / 2.5K & RGB & MuJoCo & robosuite & 100 kitchen tasks, photorealistic \\
LIBERO \cite{liu2023libero} & Manip & NA / NA & RGB & MuJoCo & robosuite & 130 tasks in 4 task suites \\
Meta-World \cite{yu2020metaworld} & Manip & 1 / 80 & Pose & MuJoCo & NA & 50 Manip tasks for Meta-RL \\
LeVERB-Bench \cite{LeVERB} & Nav, WBC & 4 / NA & RGB & PhysX & Isaac Sim & Humanoid control \\
ManiSkill \cite{mu2021maniskill1} & Manip & NA / 162 & RGB-D, PC, S & PhysX & SAPIEN & 4 tasks, 36K demos \\
ManiSkill 2 \cite{gu2023maniskill2} & Manip & NA / 2.1K & RGB-D, PC & PhysX & ManiSkill & Extended task diversity \\
ManiSkill 3 \cite{tao2025maniskill3} & Nav, Manip, WBC & NA / NA & RGB-D, PC, S & PhysX & ManiSkill 2 & GPU-parallelized simulation \\
ManiSkill-HAB \cite{shukla2025maniskillhab} & Manip & 105 / 92 & RGB-D & PhysX & \parbox[t]{2cm}{ManiSkill 3,\\Habitat 2.0} & HAB tasks from Habitat 2.0 \\
RoboTwin \cite{mu2025robotwin1, chen2025robotwin2} & Manip & NA / 731 & RGB-D & PhysX & SAPIEN & Dual-arm tasks \\
Ravens \cite{zeng21transporternetworks} & Manip & NA / NA & RGB-D & PyBullet & NA & 10 tabletop tasks \\
VIMA-BENCH \cite{jiang2023vima} & Manip & NA / 29 & RGB, S & PyBullet & Ravens & 17 multimodal prompt tasks \\
LoHoRavens \cite{zhang2023lohoravens} & Manip & 1 / 3 & RGB-D & PyBullet & Ravens & Long-horizon planning \\
CALVIN \cite{mees2022calvin} & Manip & 4 / 7 & RGB-D & PyBullet & NA & Long-horizon lang-cond tasks \\
Habitat \cite{savva2019habitat1} & Nav & 185 / NA & RGB-D, S & Bullet & NA & Fast, Nav only \\
Habitat 2.0 \cite{szot2021habitat2} & Nav, Manip & 105 / 92 & RGB-D & Bullet & Habitat & Mobile manipulation (HAB) \\
Habitat 3.0 \cite{puig2023habitat3} & Nav, Manip & 211 / 18K & RGB-D & Bullet & Habitat 2.0 & Human avatars support \\
RLBench \cite{james2019rlbench} & Manip & 1 / 28 & RGB-D, S & PyBullet & V-REP & Tiered task difficulty \\
THE COLOSSEUM \cite{pumacay2024colosseum} & Manip & 1 / 107 & RGB-D & PyBullet & RLBench & 20 tasks, 14 env variations \\
AI2-THOR \cite{kolve2017ai2thor} & Nav, Manip & NA / 118 & RGB-D, S & Unity & NA & Object states, task planning \\
CHORES \cite{ehsani2024spoc} & Nav & 191K / 40K & RGB & Unity & AI2-THOR & Shortest-path planning \\
SIMPLER \cite{li2025simpler} & Manip & 4 / 17  & RGB & PhysX & \parbox[t]{2cm}{SAPIEN,\\Isaac Sim} & Real-to-sim evaluation \\
RoboArena \cite{atreya2025roboarena} & Manip & NA / NA & RGB & Real & NA & Distributed real-world evaluation \\
\hline
\end{tabularx}
\end{table*}

\switchlanguage%
{%

Evaluation metrics for VLA models remain poorly defined, particularly in real-world settings. 
Assessing generalization on physical robots is challenging due to differences in embodiment, safety concerns, and limited reproducibility. 
Consequently, most evaluations are conducted in simulation, where standardized environments and benchmarks facilitate fair comparisons across methods. 
Below, we introduce representative simulation environments and their variants commonly used for evaluating and comparing VLA models.


\textbf{MuJoCo.}
  Several simulation environments have been developed on top of MuJoCo~\cite{todorov2012mujoco} to support research in robotic manipulation. 
  For example, robosuite~\cite{zhu2020robosuite}, a modular simulation framework in which robots, arenas, and task objects are composed using MJCF files, provides $11$ manipulation tasks.
  
  Building on robosuite, robomimic~\cite{mandlekar2021robomimic} introduces a systematic benchmark for evaluating learning from demonstrations in robotic manipulation. 
  The robomimic benchmark includes $8$ tasks performed using a Franka Emika Panda robot.
  
  RoboCasa~\cite{nasiriany2024robocasa} further extends robosuite by incorporating large-scale, photorealistic scenes that span $100$ tasks across a variety of robot platforms, enabling broader generalization and transfer learning studies.
  Currently, the most widely used benchmark for evaluating VLA models is LIBERO~\cite{Libero}, which is designed for language-conditioned manipulation tasks. 
  It provides $4$ task suites comprising a total of $130$ tasks, all executed by a Franka Emika Panda robot: LIBERO-SPATIAL focuses on spatial reasoning between objects, LIBERO-OBJECT targets object category recognition, LIBERO-GOAL evaluates understanding of object manipulation goals, and LIBERO-100 integrates the three previous suites to assess compositional generalization.
  Furthermore, Meta-World~\cite{yu2020metaworld} is another simulation environment built on MuJoCo, designed to evaluate multi-task and meta-reinforcement learning. 
  It includes 50 distinct tasks performed using a Sawyer robotic arm, enabling evaluation of generalization across diverse manipulation skills.

\textbf{PhysX.}
  IsaacLab~\cite{mittal2023orbit} is a GPU-accelerated framework built on IsaacSim, which employs PhysX as its underlying physics engine. It provides a comprehensive suite of tools for robot learning, including a diverse set of robots, environments, and sensors, along with photorealistic rendering capabilities. 
  LeVERB-Bench~\cite{LeVERB}, also built on IsaacSim, focuses on full-body humanoid control and includes $154$ vision-language tasks and $460$ language-only tasks.
  
  Moreover, ManiSkill~\cite{mu2021maniskill1, gu2023maniskill2, tao2025maniskill3}, built on the SAPIEN simulation platform~\cite{xiang2020sapien}, whose underlying physics engine is also based on PhysX, serves as a comprehensive benchmark for learning object manipulation skills from 3D visual input.
  It includes a wide range of tasks involving articulated and deformable objects, mobility, and diverse robot embodiments, and provides large-scale demonstration data with support for efficient, high-quality simulation.
  ManiSkill-HAB~\cite{shukla2025maniskillhab} is a benchmark focused on object rearrangement tasks that follow the Home Assistant Benchmark (HAB) introduced in Habitat 2.0~\cite{szot2021habitat2}.
  In addition, several other benchmarks have been developed on SAPIEN, such as RoboCAS~\cite{zheng2024robocas}, which evaluates robotic manipulation in complex object arrangement environments, and DexArt~\cite{bao2023dexart}, which focuses on manipulation of articulated objects using multi-fingered hands.
  More recently, RoboTwin~\cite{mu2025robotwin1, chen2025robotwin2} has been proposed as a benchmark for dual-arm manipulation, offering $50$ tasks, $731$ objects, and $5$ distinct embodiments.

\textbf{Bullet.}
  Ravens~\cite{zeng21transporternetworks} is a benchmark of $10$ tabletop manipulation tasks implemented using PyBullet~\cite{coumans2016pybullet}.
  VIMA-BENCH~\cite{jiang2023vima} extends this benchmark with $17$ tasks that allow multi-modal prompt-based task specification.
  LoHoRavens~\cite{zhang2023lohoravens} is another extension that evaluates long-horizon planning capabilities in tabletop manipulation scenarios.
  Moreover, CALVIN~\cite{mees2022calvin} provides a simulation and benchmark for long-horizon manipulation based on natural language instructions, which includes $34$ manipulation tasks performed by a Franka Emika Panda robot.
  In addition, Habitat~\cite{savva2019habitat1, szot2021habitat2, puig2023habitat3} is a simulation framework primarily developed by Meta.
  Habitat 1.0~\cite{savva2019habitat1} provides a simulation platform specialized for visual navigation tasks.
  Habitat 2.0~\cite{szot2021habitat2} extends this to mobile manipulation tasks and introduces the Home Assistant Benchmark (HAB).
  Further, Habitat 3.0~\cite{puig2023habitat3} expands the framework to support not only robots but also human avatars.

\textbf{V-REP.}
  RLBench~\cite{james2019rlbench} is the first large-scale benchmark for imitation and reinforcement learning, built using V-REP~\cite{rohmer2013vrep} and PyRep~\cite{james2019pyrep}.
  It contains $100$ manipulation tasks using the Panda robot.
  THE COLOSSEUM~\cite{pumacay2024colosseum}, built on top of RLBench, is a benchmark designed to systematically evaluate the generalization capabilities of robotic manipulation policies under environment variations.
  THE COLOSSEUM includes $20$ manipulation tasks with 14 types of environment perturbations.

\textbf{Unity.}
  AI2-THOR  is a photorealistic, interactive 3D simulation environment built on the Unity engine, offering four task suites, such as iTHOR, RoboTHOR, ProcTHOR-10K, and ArchitecTHOR~\cite{kolve2017ai2thor, deitke2020robothor, deitke2022procthor}, that collectively encompass a diverse range of indoor environments.
  Moreover, SPOC~\cite{ehsani2024spoc} introduces CHORES, an extension of AI2-THOR designed as a benchmark for shortest-path planning in navigation tasks.

\textbf{Miscellaneous.}
  While not strictly simulation-based benchmarks, several studies have proposed evaluation protocols to assess the capabilities of VLA models. 
  %
  %
  VLATest~\cite{VLATest} systematically evaluates the impact of various factors on VLA model performance, including the number of confounding objects, lighting conditions, camera poses, unseen objects, and mutations in task instructions.
  Moreover, several works aim to improve robustness against adversarial attacks~\cite{Adversarial, cheng2024vulnerabilities} and enhance interpretability by probing the latent representations of VLA models to uncover symbolic structures corresponding to object properties, spatial relations, and action states~\cite{lu2025ca}.

\textbf{Toward realistic and scalable evaluation for VLA.}
  There is increasing emphasis on evaluation under conditions that closely resemble the real world, leading to the development of both realistic simulation benchmarks and scalable systems for distributed real-world evaluation of VLA models. 
  SIMPLER~\cite{li2025simpler} enables the evaluation of policies trained on real-world data within simulation by minimizing visual and control domain gaps, achieving high correlation between simulation and real-world performance. 
  RoboArena~\cite{atreya2025roboarena} is a distributed framework for large-scale, fair, and reliable evaluation of VLA models in the real world. It conducts pairwise comparisons across a network of robots deployed at seven universities, with results aggregated by a central server to produce global rankings. 
  This system is built on the DROID platform.
}%
{%
  VLAの評価指標は未だ何か明確に定まっているわけではない.
  特に身体性を持つロボットにおいて, ロボット実機での評価は非常に難しい.
  そこで, 評価用のシミュレーションをつくり, その中でデータセットとベンチマークの両方を提供することが多く行われている.
  代表的な例を, その派生系とともにまとめながら紹介する.

  まず, CALVIN~\cite{mees2022calvin}, AI2-THOR \cite{kolve2017ai2thor}, Meta-World \cite{yu2020metaworld}について述べる.
  CALVINは自然言語指示による長期的マニピュレーションのためのシミュレーションとベンチマークを提供している.
  PyBulletをベースとした, Franka Emika Pandaによる34種類のタスクを収録している.
  AI2-THORはフォトリアリスティックな3Dインタラクティブシミュレーションであり, 多様な室内環境のベンチマークであるiTHOR, RoboTHOR, ProcTHOR-10K, ArchitecTHORという4つのtask suiteを提供している \cite{kolve2017ai2thor, deitke2020robothor, deitke2022procthor}.
  また, Meta-Worldはmulti-task RLとmeta-RLのためのMuJoCO \cite{todorov2012mujoco}ベースのベンチマークであり, Sawyerを用いた50種類のタスクを提供している.
  これらはいずれもVLAの評価によく用いられている.

  Habitat \cite{savva2019habitat1, szot2021habitat2, puig2023habitat3}はMetaが中心となって開発しているシミュレーションフレームワークである.
  Habitat 1.0 \cite{savva2019habitat1}は視覚ナビゲーションタスクに特化したシミュレーションプラットフォームの提供している.
  Habitat 2.0 \cite{szot2021habitat2}はこれをモバイルマニピュレーションタスクに拡張し, Home Assitant Benchmark (HAB)を提供した.
  さらにHabitat 3.0 \cite{puig2023habitat3}では, ロボットだけでなく人間のアバターも利用できるできるような拡張が行われている.

  robosuite \cite{zhu2020robosuite}とそこから派生したベンチマークについて述べる.
  robosuiteは, MuJoCo \cite{todorov2012mujoco}をベースとしたモジュール型シミュレーションフレームワークであり, robot, arena, taskオブジェクトからMJCFファイルが構成されている.
  このrobosuiteは9つのタスクにおけるベンチーマークを含んでいた.
  これから派生したのがrobomimic \cite{mandlekar2021robomimic}である.
  人間のデモンストレーションからのオフライン学習が, ロボットマニピュレーションにどれだけ効果的かを調査する体系的なベンチマークであり, Pandaによる8種類のマニピュレーションタスクを収録している.
  また, RoboCasa \cite{nasiriany2024robocasa}は, 大規模かつフォトリアリスティックな拡張を施したrobosuiteベースのシミュレーションフレームワークであり, 様々なロボットにおける100のタスクを含む.
  そして, 現状最も良くVLAの評価に使われているのが, robosuiteをベースとしたLIBERO \cite{liu2023libero}である.
  LIBEROは言語条件付きマニピュレーションタスクのベンチマークで, Franka Emika Pandaによる130のタスクを含む4つのtask suiteを提供している.
  具体的には 物体の空間的関係の認識を問うLIBERO-SPATIAL, 物体カテゴリの認識を問うLIBERO-OBJECT, 物体の操作方法を問うLIBERO-GOAL, 複合された知識を問うLIBERO-100が含まれている.

  SAPIEN \cite{xiang2020sapien}をベースとしたManiSkillシリーズ \cite{mu2021maniskill1, gu2023maniskill2, tao2025maniskill3}について述べる.
  ManiSkill 1 \cite{mu2021maniskill1}は, 3D視覚入力に基づく物体操作スキルのベンチマークである.
  特に, articulated objectに対する操作スキルの汎化を重視しており, Pandaによる4つのタスクに対して, 36000以上のデモンストレーションデータを提供している.
  ManiSkill 2 \cite{gu2023maniskill2}はさらに多様なタスク数, 柔軟物体や移動, 身体性を扱ったベンチマークである.
  ManiSkill 3 \cite{tao2025maniskill3}はこれをGPUにより並列化し, 高速なシミュレーションとレンダリングフレームワークによるリアルな視覚入力を可能としている.
  ManiSkill-HAB \cite{shukla2025maniskillhab}は, 物体再配置に向けたベンチマークであり, Habitat 2.0 \cite{szot2021habitat2}のHABに準拠したタスクにを収録する.
  他にも, 複雑な物体配置環境におけるロボットマニピュレーションのベンチマークをSAPIEN上で構築するRoboCAS \cite{zheng2024robocas}, 多指ハンドによるarticulated objectの操作に特化したDexArt \cite{bao2023dexart}などのベンチマークがある.

  Ravens \cite{zeng21transporternetworks}とその派生系について述べる.
  RavensはPyBullet \cite{coumans2016pybullet}をベースとした10のテーブルトップタスクを集めたベンチマークである.
  VIMA-BENCH \cite{jiang2023vima}はこれを拡張し, マルチモーダルプロンプトによるタスク指定が可能な17種類のタスクを収録している.
  LoHoRavens \cite{zhang2023lohoravens}もRavensの拡張であり, 長期的な計画を必要とするテーブルトップ操作タスクの能力を評価している.

  RLBench \cite{james2019rlbench}とその派生系について述べる.
  RLBenchはV-REP \cite{rohmer2013vrep}とPyRep \cite{james2019pyrep}を使った, 摸倣学習や強化学習に関する初の大規模ベンチマークであり, Pandaによる100のタスクを収録している.
  また, THE COLOSSEUM \cite{pumacay2024colosseum}はロボットのマニピュレーションポリシーの環境変化に対する汎化性能を系統的に評価するためのベンチマークである.
  RLBenchを基盤として, 20種類の操作タスクに14種類の環境摂動を加えて評価している.

  近年はより実世界のロボットに近い形での評価が求められるようになっており, それらに対応したベンチマークも登場している.
  SIMPLER \cite{li2025simpler}は, 実世界で訓練したポリシーをシミュレーションで評価するためのフレームワークである.
  視覚ギャップや制御ギャップの最小化の工夫により, シミュレーション上での成功率と実世界での成功率の相関を高めることに成功している.
  また, RoboArena \cite{atreya2025roboarena}はVLAの現実世界での性能評価を, 大規模かつ公正・信頼性を持って行うための分散型評価フレームワークである.
  複数の拠点(7大学)にまたがる実ロボットネットワークで, 任意のタスク・環境でペアワイズ比較(A/Bテスト)を実施し, 評価結果を中央サーバで収集・統合してグローバルなランキングを生成する.
  ここでは, DROIDのプラットフォームをベースに用いている.

  この他にも, 双腕ロボットによるマニピュレーションタスクのベンチマークであるRoboTwin \cite{mu2025robotwin1, chen2025robotwin2}, ヒューマノイドの全身制御に特化した154の視覚-言語タスク, 460の言語タスクを収録するLeVERB-Bench \cite{LeVERB}, タスクの成否ではなく行動予測能力そのものを評価するMultiNet \cite{guruprasad2024benchmarkingvisionlanguage}, 両手によるツール操作行動のベンチマークであるTACO \cite{liu2024taco}, ナビゲーションにおける最短経路計画のベンチマークであるSPOC \cite{ehsani2024spoc}など, 非常に多様なベンチマークが開発されている.
  
  また, VLATestはManiSkill2上で, 対象物や妨害物, 照明条件, カメラ位置のscene fuzzingによって, 多様なVLAの性能比較を行っている\cite{VLATest}
  VLAのadversarial attackに対する脆弱性を解析・評価した取り組み\cite{Adversarial, cheng2024vulnerabilities}や, VLAの内部表現からのシンボリックな状態の推論能力を検証した取り組みもある\cite{lu2025ca}.
  
}%

\subsection{Real-world Applications}
This section provides concrete examples of how the previously introduced robotic platforms, including manipulators, hands, mobile robots, quadrupeds, and humanoids, are employed in the development and evaluation of VLA models.

\textbf{Manipulator.}
  Manipulators represent the most widely used robotic platforms in VLA research. 
  They are employed across a diverse set of tasks, including object grasping and relocation, assembly, deformable object manipulation, and peg-in-hole insertion. 
  Both single-arm and more complex dual-arm robots are commonly utilized, enabling a broader range of dexterous manipulation tasks. 
  Notable demonstrations in this domain include Shake-VLA~\cite{Shake-VLA}, which performs cocktail mixing using dual-arm coordination, and RoboNurse-VLA~\cite{RoboNurse-VLA}, which automates surgical instrument handovers in clinical environments.

\textbf{Hand / Gripper.}
  Hands and grippers, commonly used as end-effectors on manipulators, enable a wide range of manipulation tasks. 
  Two-fingered grippers are particularly well suited for object grasping, while more dexterous four- and five-fingered robotic hands facilitate tool use and in-hand manipulation.
  For instance, GraspVLA~\cite{GraspVLA} develops a VLA model for object grasping using a two-fingered gripper. 
  In contrast, DexGraspVLA~\cite{DexGraspVLA} leverages a multi-fingered robotic hand to construct a VLA model capable of performing more delicate and precise grasping tasks.

\textbf{Mobile robot.}
  Mobile robots are primarily utilized in VLA models for navigation-related tasks~\cite{MobilityVLA}. 
  Beyond navigation, models such as RT-1~\cite{brohan2023rt1} are capable of generating both arm and base motions for mobile manipulators, robots that integrate a mobile base with a robotic arm. 
  The VLA framework has also been extended to other mobile domains. 
  For instance, aerial robots such as the DJI Tello are used in UAV-based VLA research, with works including UAV-VLA~\cite{UAV-VLA}, RaceVLA~\cite{RaceVLA}, and CognitiveDrone~\cite{CognitiveDrone} focusing on autonomous flight. 
  Similarly, VLA applications in autonomous driving have been explored in OpenDriveVLA~\cite{OpenDriveVLA}, ORION~\cite{fu2025orion}, CoVLA~\cite{arai2025covla}, and OccLLaMA~\cite{OccLLaMA}. These developments demonstrate the adaptability of VLA systems across a diverse range of mobile robotic platforms.

\textbf{Quadruped robot.}
  Quadruped robots enable more diverse and versatile navigation compared to wheeled mobile robots due to their ability to traverse uneven, unstructured, and dynamic terrains.
  Several prior works, including TrackVLA~\cite{TrackVLA, cheng2024navila}, NaVILA~\cite{cheng2024navila}, and CrossFormer~\cite{CrossFormer}, successfully demonstrate robust navigation capabilities, including deployment in the wild.
  Furthermore, Track2Act~\cite{TraceVLA} and VidBot~\cite{VidBot} utilize Boston Dynamics Spot equipped with a manipulator for integrated navigation and manipulation in home environments. 
  SLIM~\cite{SLIM} similarly employs a Unitree Go1 equipped with a mounted WidowX 250 arm to perform multimodal tasks, such as grasping objects from the ground while navigating uneven terrain.

\textbf{Humanoid robot.}
  Humanoids have gained significant attention in VLA research, because their human-like morphology offers practical advantages for real-world deployment, as most environments, tools, and interfaces are designed for human use, making task transfer and embodiment alignment more straightforward.
  NaVILA~\cite{cheng2024navila} demonstrates robust locomotion capabilities in tightly controlled laboratory settings. In contrast, EgoVLA~\cite{EgoVLA} and GO-1~\cite{AgiBotWorldColosseo} focus on manipulation tasks commonly encountered in household environments, including picking, placing, pouring, and folding.
  %

\section{Recommendations for Practitioners}
\label{sec:recommendation}
Drawing on insights from recent VLA research, this section provides actionable recommendations for practitioners seeking to design, train, and deploy VLA models in real-world robotic systems. We highlight practical strategies across data collection, architecture selection, and model adaptation.

\textbf{Prioritize diverse and high-quality datasets.} 
  Robust generalization across tasks, objects, and embodiments relies on training with large-scale, high-quality datasets that encompass vision, language, and action modalities. Practitioners should aim to collect or utilize datasets that offer broad task coverage, environmental variability, and embodiment diversity. Such diversity is essential for improving the robustness and transferability of VLA policies.
  
  Prefer continuous action generation via generative methods.
  While it is increasingly well established in recent literature, generating continuous actions, rather than relying on discretized tokens, remains critical for achieving smooth and precise robot behavior. Practitioners are encouraged to adopt generative approaches such as diffusion or flow matching to enable high-fidelity control in real-world settings.

\textbf{Try gradient insulation during pre-training.}
  Allowing gradients from randomly initialized action heads to propagate into pre-trained VLM backbones can degrade the quality of learned representations that already capture common-sense knowledge. 
  To stabilize training and preserve the semantic knowledge in the backbone, practitioners are encouraged to freeze the backbone or apply gradient insulation mechanisms. This approach has been shown to improve both training efficiency and final performance.

\textbf{Begin with lightweight adaptation methods.} 
  Full fine-tuning of large VLA models is often computationally prohibitive. As a first step, practitioners, who do not have access to a GPU cluster, can fine-tune only the action head while keeping the backbone frozen. Alternatively, methods such as LoRA enable parameter-efficient fine-tuning, offering a favorable trade-off between performance and resource consumption.

\textbf{Incorporate world models or latent action learning for scalability.}
  In scenarios involving humanoid robots, incorporating human video data during pre-training can be particularly advantageous due to the similarity in embodiment. However, as such datasets typically lack explicit action annotations, it is beneficial to learn latent action representations that can be used as surrogate action targets during pre-training.
  In addition, the predictive capabilities of world models can support more effective planning and reasoning, especially in manipulation tasks. By anticipating future observations, world models facilitate better long-horizon control and multimodal grounding, as demonstrated in prior work such as FLARE~\cite{zheng2025flare}.

\textbf{Embrace multi-task learning to enhance representations for action generation.}
While VLMs pre-trained on web-scale data offer strong semantic grounding, their representations are not always directly suited for downstream control. Incorporating auxiliary tasks such as affordance estimation, keypoint detection, future state prediction, and segmentation for a target object encourages the model to learn representations that are better aligned with the requirements of action generation. These tasks support spatial reasoning, temporal prediction, and physical interaction modeling, ultimately improving the model's ability to translate perception into effective control.

\section{Future Research Direction} \label{sec:discussion}
\switchlanguage%
{%

\subsection{Data Modality}
  While several prior works have attempted to integrate additional modalities such as audio, tactile sensing, and 3D point clouds into VLA models, collecting large-scale datasets with such modalities remains a significant challenge. In particular, tactile sensing poses serious difficulties due to the diversity of sensor types, data formats, and hardware configurations. The lack of standardization across robotic platforms further complicates multimodal data collection and integration. Although tactile feedback is likely essential for achieving human-level dexterous manipulation, current tactile sensors vary widely in design and are not yet widely adopted. Therefore, unifying sensor configurations is critical to enabling scalable, multimodal VLA systems.

\subsection{Reasoning}
  Reasoning is a particularly important capability for solving long-horizon tasks in VLA systems. Beyond anticipating future events based on current observations, effective reasoning requires the ability to retain relevant information over time and retrieve it when needed. This involves maintaining a form of memory and selectively attending to key information that supports decision-making across temporally extended tasks.
  For example, in mobile robot manipulation, a typical task may involve first locating a shelf, then navigating to a different location to pick up a cup, and finally returning to place the cup on the shelf. In such cases, the robot must remember the location of the shelf encountered earlier and retrieve that information at the appropriate time. This type of temporal abstraction and memory-based retrieval is essential for robust reasoning and planning in real-world scenarios. Enhancing these capabilities is likely to be a key direction for future research in VLA systems, particularly as tasks grow in complexity and duration.

\subsection{Continual Learning}
  A fundamental limitation of current VLA systems is their inability to learn beyond their initial training phase. Once trained offline, these models are typically frozen and do not adapt to new situations. Unlike humans, who continuously learn from ongoing experience, VLA systems remain fixed, making them vulnerable when faced with novel or out-of-distribution scenarios. In such cases, the robot may fail to act appropriately.
  To overcome this limitation, enabling online or continual learning will be essential. By incrementally updating their internal representations and policies based on new data, VLA systems could better adapt to diverse environments. However, this capability introduces several challenges, including catastrophic forgetting and safety concerns related to deploying untested updates in real-world settings.
  Despite these difficulties, continual learning remains a promising direction for future VLA research. Approaches such as reinforcement learning from human feedback (RLHF) and active learning inspired by cognitive development may offer viable pathways toward building adaptive, lifelong-learning VLA systems capable of operating safely and effectively in the real world.

\subsection{Reinforcement Learning}
  While several prior studies~\cite{ConRFT, VLA-RL, iRe-VLA} have explored the use of RL to fine-tune vision-language-action (VLA) models, these efforts have predominantly focused on evaluation in simulated environments.
  This is largely due to the substantial number of samples required for RL and the risk of unsafe behavior during real-world exploration. As a result, fine-tuning VLA models within a learned world model presents a promising research direction, offering a safer and more sample-efficient alternative.
  In addition, real-to-sim techniques allow the construction of digital twin environments in which VLA models can be fine-tuned using RL. However, challenges remain in accurately identifying physical parameters and reconstructing scenes, the latter of which often requires multi-view observations~\cite{nolte2025singleviewmeshreconstructionready}.
  Overall, we posit that advances in world modeling and real-to-sim transfer may enable scalable and safe fine-tuning of VLA models through RL.

\subsection{Safety}
  While VLA models perform well on manipulation tasks in controlled settings, their deployment in unstructured environments poses significant safety challenges. Current systems often lack mechanisms to detect and avoid unexpected human presence in the workspace, increasing the risk of collisions. Although collecting demonstrations of such edge cases is possible, doing so via teleoperation remains risky, as the robot may not respond safely in real time. This underscores the need to integrate VLA with model-based control approaches, which offer predictive reasoning in safety-critical situations. We argue that improving the safety of VLA systems requires hybrid architectures that combine the generalization capabilities of learned policies with the reliability of model-based controllers~\cite{bastani2021safe, jyamada2025graspmpc}.

\subsection{Failure Detection and Recovery}
  In real-world environments, unexpected failures are often unavoidable. However, most current VLA systems lack mechanisms for detecting such failures or responding appropriately. Failures are typically treated as terminal events, with no recovery or re-planning strategies in place.
  To enable reliable deployment in practical applications, it is essential for VLA systems to detect failures and adapt their behavior accordingly. Several recent works have begun to address this gap. SAFE~\cite{SAFE} leverages intermediate representations within VLA models to identify failure events during execution. Agentic Robot~\cite{AgenticRobot} uses a vision-language model (VLM) to detect failures, execute predefined recovery behaviors, and then re-plan the task.
  A more robust solution is proposed in LoHoVLA~\cite{LoHoVLA}, which employs a hierarchical architecture. Upon detecting a failure, the system regenerates the current action; if the same failure is detected multiple times, it escalates the response by re-generating the higher-level subtask, thus enhancing overall robustness.
  FOREWARN~\cite{FOREWARN} introduces a predictive planning mechanism by sampling a large number of action sequences from the policy, clustering them into six behavioral modes, and using the DreamerV3 world model~\cite{hafner2025dreamerv3} to simulate future states. The most promising behavioral mode is then selected based on these predictions.
  As VLA systems are increasingly applied to long-horizon and open-ended tasks, the ability to detect failures and recover through adaptive re-planning will be critical for achieving robustness and reliability in real-world deployment.

\subsection{Evaluation}
  While various VLAs with different architectures, modalities, and training methods have been proposed, it remains unclear which approaches yield the most effective performance.
  This ambiguity largely stems from the lack of a statistically rigorous evaluation.
  As demonstrated in LBM~\cite{trilbmteam2025lbm}, it is crucial to conduct evaluations under controlled and comparable conditions, with a sufficient number of evaluation trials and appropriate statistical analysis (e.g. confidence intervals) to ensure whether observed performance differences are statistically significant.
  
\subsection{Applications}
  VLA systems have potential applications across a wide range of domains, including healthcare, assistive technologies, industrial automation, and autonomous driving. However, despite this breadth of applicability, VLA models have not yet reached the level of performance or reliability required for practical deployment. Most existing systems operate only within constrained, predefined environments and still fall short of human-level capabilities in terms of robustness and adaptability.
  
  As the field increasingly prioritizes real-world use cases, there will likely be growing attention to issues such as safety, reliability, and operational efficiency, key factors that must be addressed to enable the successful deployment of VLA systems in practical applications.

}%
{%
  ここでは, 現在議論されている, またはこれから議論すべきVLAに関するいくつかのトピックを取り上げ, 深堀りする.
  
  \subsection{Cross Embodiment}
  VLAを学習するにあたって, 単一の身体性のデータだけでは限界がある.
  各機関ごとに使っているロボットは異なり, センサ配置などもバラバラである.
  一方, 確かにロボットは様々な種類があるが, 物理的な現象と言う意味では共通である.
  そのため, cross embodimentについて考えることは有意義である.
  この流れから, Open X-Embodiment~\cite{oneill2024openxembodiment}やCrossFormer \cite{doshi2025crossformer}などを含めた様々な研究が行われてきた\cite{yang2024extreme}.
  また, ロボット同士だけでなく, ロボットと人間の間にも大きな相関がある.
  特に現在流行しているヒューマノイドロボットはまさにその良い例だ.
  人間には右手・左手・頭, 右脚, 左脚があるように, ヒューマノイドも同様の構成をしている.
  つまり, 人間のデータとヒューマノイドの相性はかなり良いはずである.
  そのため, 今後人間の動画データはより大きな意味を持ってくると考えている.
  LAPA \cite{ye2025lapa}をはじめとした動画からのアクション抽出は, これまでにないほど多様なデータを活用可能にすることができる.
  この流れは, 今後の大きなトピックとなっていくであろう.

  \subsection{Data Modality}
  VLAに用いられているmodalityには様々なものがあった.
  画像・言語・アクションに加えて, 音声や接触センサ, 3次元点群など, そのモダリティは幅広い.
  一方で, モダリティが増えるにしたがって, VLAの汎化にはデータセットの問題が付きまとう.
  少ないデータ数しかないモダリティは簡単には統合できず, 十分にその能力を活用できない.
  ロボットには現状決まった規格がなく, 用いられるモダリティもバラバラである.
  特に深刻なのは接触センサで, これは非常に多種多様な形態が存在していて, データ構造もバラバラである.
  その一方, 人間のような繊細なマニピュレーションを行うためには, 使わざるを得ないだろう.
  今後ヒューマノイドロボットにセンサ構成の統一化が起きれば, これらの問題は解決されていくだろう.
  しかし, 現状満足のいくセンサ自体がないめた, その道のりは長いだろう.

  \subsection{Reasoning}
  VLAにおけるreasoningは一つの大きなトピックとなっている.
  image captioningや物体ラベルを予測するものから, 画像の遷移, affordanceを予測するものまで様々である.
  このようなreasoningを追加することで, ロボットはより正確に, 賢く動くことができるようになる.
  これらはLLMにおけるChaing-of-Thoughtのような形でVLAに追加されている.
  先にreasoningを行うことで, その後のactionがより良いものになる.
  VLAの進化はLLMの進化から少し遅れて, LLMやVLMでの技術を取り込むことで発展していっている.
  現在in-context learningやgrokkingについて検証するVLAは数少ないが, 今後これらも増えていくことだろう.

  また, 短く簡単な言語指示だけでなく, より複雑で長期的な言語指示を扱えることは, 今後のVLAにとって重要な課題である.
  現在では, LLMで言語指示をsubtaskに分解し, それらをVLAで実行するケースや, 同じモデルでpromptを変えることでsubtaskとアクションを出すことができるケースなどが提案されている.
  また, 階層構造を持たせて, VLMとdiffusion transformerを潜在空間で結合するなど, 手法の幅は広い.
  今後より長期的な動作計画を行うためには, 記憶の実装やreasoningが重要となるだろう.
  
  \subsection{Real-Time Execution}
  VLAで最も重要なのはアクションの出力部分である.
  特に, このアクションはリアルタイムに制御入力を送らなければならない.
  これは, LLMやVLMにはない, ロボットという身体性があるからこそ初めて生まれる特異な点である.
  そして, このアクションのリアルタイム実行という点については, 非常に多くの研究が行われてきた.
  もともとはautoregressiveにdiscrete action tokenを出力するのが一般的だったが, これがdiffusionとなり, 現在はより速いflow matchingへと技術が移り替わっている.
  これに関連した技術は多く, 最初のころは3Hzが限界であったアクションも, 50Hz出るようになってきる点は素晴らしい.
  また, ネットワーク自体を軽くしていき, エッジデバイスにも載るようなVLAの開発も続いている.
  一方, ほとんどがマニピュレータに関するものであり, 四脚ロボットやヒューマノイドロボットのような脚型ロボットの脚は直接扱われていない.
  大抵は強化学習などを用いた脚移動のポリシーを別で学習しておき, 上位の速度指令のみを送る構成が一般的である.
  脚を直接制御できるようになることで, さらに新しいタスクが解けるようになることを期待している.

  \subsection{Failure Detection and Re-Planning}
  VLAは, 事前に学習したポリシーに基づいてアクションを出力するが, 実際の環境では予期せぬ障害や失敗が発生することがある.
  現状このようなケースはあまり考慮されておらず, 失敗した場合はそのまま失敗として扱われることが多い.
  その一方で, 失敗を検知し, その後の動作を再計画することは, より実用的なロボットにおいては非常に重要な要素である.
  この失敗を検知しようと試みたのがSAFE \cite{SAFE}であり, VLAの中間表現を用いて, 失敗を検知している.
  また, Agentic Robot \cite{AgenticRobot}はVLMにより失敗を検知したら, 事前に定義された回復動作を行い, その後動作を再計画している.
  さらに, 階層型アーキテクチャを持つLoHoVLA \cite{LoHoVLA}では, 失敗を検知したらアクションを再生成し, 同じ失敗を規定回数以上検知したら上位のサブタスクを再生成することで, より堅牢な動作を実現している.
  FOREWARN \cite{FOREWARN}は, ポリシーから行動系列を大量にサンプリングしておきこれらを6つのモードにクラスタリング, それらを用いてDreamerV3 \cite{hafner2025dreamerv3}の世界モデルにより未来の状態を予測, この予測を元にベストなモードを選択している.
  今後動作がより長期的なものになるにしたがって, 失敗からいかに回復し, 行動を再開するかが, VLAの重要な研究課題となるだろう.

  \subsection{Safety}
  現在のVLAは, 基本的に画像と言語から直接アクションが出力されるというものである.
  そしてここには, 安全性に関する大きな問題が潜んでいる.
  これは完全にend-to-endである場合が多いため, おかしな挙動をした際にワークスペース内の人間に衝突してしまう危険性がある.
  また, そのような事故が起きた際に, なぜ起きたのかを説明することができない.
  今後, VLAの内部表現の解析や, モデルベース制御と組み合わせた安全性の確保に関する議論は盛んに行われるだろうと予測している.

  \subsection{Continual Learning}
  現在のVLAの一つの問題点は, 一度オフラインで学習されたあと, そこから一切成長しないところである.
  人間であれば, 経験を得ながら, リアルタイムに学習し, 成長していくことができるが, 現在のロボットにはそれができない.
  そのため, 全く想定されていないような環境に遭遇したときに, ロボットは, なすすべなく動かなくなってしまうことが多い.
  これを解決するためには, オンライン学習が今後重要になっていくだろう.
  逐次的に自身のネットワークの重みを変化させることで, より多様な環境に順応できるようになる.
  その一方で, オンライン学習には破壊的忘却や安全性の問題が潜み, 一筋縄ではいかない.
  この部分は, 今後の大きなブレイクスルーを秘めていると感じる.
  例えば, RLHFのような強化学習の利用や, 認知発達の知見を活かしたactive learningなども今後視野にはいってくるだろう.

  \subsection{Learning in Simulation}
  一つの面白いVLAの作り方として, LLMやVLMによりRLの報酬を自動で定義しシミュレーション上でポリシーを学習させる, というものがある.
  VLM-RL \cite{VLM-RL}では, CLIP \cite{radford2021clip}を活用して観測画像とタスク指示の類似度を計算し, これを活用することでポリシーを学習していた.
  RoboCLIP \cite{sontakke2023roboclip}では, 与えられた動画とタスク指示の類似度を測定するモデルを構築し, これを報酬とした.
  この他にも, RL-VLM-F \cite{RL-VLM-F}は既存のVLMに2つの画像とタスク指示を入力し, どちらの画像がよりタスクのゴールに近いかを判定させ, 画像からの報酬モデルを学習, これをもとに強化学習を行っている.
  このような画像とタスク指示の類似度を報酬とする方法以外にも, そもそも報酬設計をLLMやVLMによって行う手法が開発されてきている.
  Eureka \cite{ma2024eureka}は, あるタスクに対して, 環境の定義コードやtask descriptionを与えることで, LLMによる報酬の反復設計と強化学習により適切なポリシーを獲得する.
  DrEureka \cite{ma2024dreureka}はそれをsafety instructionとdomain randomizationをもとに, シミュレーションのみならず実機にも適用できる形とした.
  Video2Policy \cite{Video2Policy}はvideoからタスクを定義, そこからIsaacGym \cite{makoviychuk2021isaacgym}のtaskコードを生成, 学習されたポリシーの実行軌跡をもとtaskコードを反復的に修正することで, より汎用性が高いポリシー学習に成功している.
  現状, 強化学習は数秒で終わるものではないため, 画像と言語からアクションを出すVLAとして定義することは難しい.
  しかし, もし強化学習が1秒で終わるとしたら, このような考え方もある種のVLAになりうるだろう.
  また, これらの際には, シミュレーションで学習したpolicyをそのまま実機でdeploy可能である必要がある.
  これは, 正確なdigital twinを作り上げる方法, 現実世界で学習したworld modelを用いる方法などがあり, 今後, これらの方向性も重要である.

  \subsection{Evaluation}
  さまざまなアーキテクチャ, モダリティ, 学習手法を用いたVLAモデルが提案されているものの, どのアプローチが最も効果的であるかは依然として明らかではない.
  この曖昧さの主な原因は, 統計的に厳密な評価が不足している点にある.
  各VLAはそれぞれ別々のベンチマークを使い, 自身のVLAの有用性を検証している.
  しかし, これでは真にどのような構造やアルゴリズムが良いかを判定することは難しい.
  LBM \cite{trilbmteam2025lbm}で示されているように、性能差が統計的に有意であるかを判断するためには、統一された条件下で十分な数の評価試行を行い、信頼区間などの適切な統計的分析を用いた評価が不可欠である。
  この問題はVLAについてはかなり深刻であり, 今後最も時間が割かれるべき問題の一つだと感じる.
  
  \subsection{Application}
  VLAのアプリケーションは, 医療や生活支援, 産業, 自動運転など多岐にわたっている.
  一方で, いまだ実用的に使えるようなVLAの登場には至っていない.
  そのどれもが, なんらか規定された環境でしか動かず, もし動いたとしても人間と比較してとても十分とは言えない性能しか発揮できていないのが現状である.
  その一方で, タスクを決め, 環境をある程度統制してしまえば, Figure AIのようにある程度使えそうなレベルまで達しているものもある.
  今後より実アプリケーションを重視した流れが増えていくとともに, その安全性や効率に関する議論も進んでいくと考えられる.
}%

\section{Conclusion}\label{sec:conclusion}
\switchlanguage{%
This survey provides a comprehensive review of Vision-Language-Action (VLA) models for robotics, tracing their evolution from early CNN-based approaches to sophisticated multimodal architectures integrating diffusion models and latent action representations.
We have examined the fundamental challenges, architectural innovations, training methodologies, and real-world applications that define the current landscape of VLA research.

Our analysis reveals several key insights: (1) the critical role of large-scale datasets and pre-trained foundation models in enabling generalization, (2) the emergence of hierarchical architectures that separate high-level reasoning from low-level control, (3) the growing importance of multimodal inputs beyond vision and language, and (4) the persistent challenges in sim-to-real transfer and embodiment generalization.
The field has reached a critical inflection point at which recent advances in foundation models, in conjunction with improved data collection protocols and refined training methodologies, are anticipated to facilitate the development of robotic systems with improved generalization and capability.
The incorporation of world models, affordance-based reasoning, and RL is expected to underpin the next generation of VLA models, enabling continuous learning, sophisticated task reasoning, and robust adaptation across diverse and unstructured real-world environments.
}%
{%

}%

\section*{Acknowledgment}
ChatGPT-4o was used to assist in the creation of several figures.

\bibliographystyle{IEEEtran}
\bibliography{main, macros}

\begin{thebibliography}{100}
\providecommand{\url}[1]{#1}
\csname url@samestyle\endcsname
\providecommand{\newblock}{\relax}
\providecommand{\bibinfo}[2]{#2}
\providecommand{\BIBentrySTDinterwordspacing}{\spaceskip=0pt\relax}
\providecommand{\BIBentryALTinterwordstretchfactor}{4}
\providecommand{\BIBentryALTinterwordspacing}{\spaceskip=\fontdimen2\font plus
\BIBentryALTinterwordstretchfactor\fontdimen3\font minus
  \fontdimen4\font\relax}
\providecommand{\BIBforeignlanguage}[2]{{%
\expandafter\ifx\csname l@#1\endcsname\relax
\typeout{** WARNING: IEEEtran.bst: No hyphenation pattern has been}%
\typeout{** loaded for the language `#1'. Using the pattern for}%
\typeout{** the default language instead.}%
\else
\language=\csname l@#1\endcsname
\fi
#2}}
\providecommand{\BIBdecl}{\relax}
\BIBdecl
\renewcommand{\BIBentryALTinterwordstretchfactor}{4}

\bibitem{touvron2023llama2}
H.~Touvron, L.~Martin, K.~Stone, P.~Albert, A.~Almahairi, Y.~Babaei,
  N.~Bashlykov, S.~Batra, P.~Bhargava, S.~Bhosale \emph{et~al.}, ``Llama 2:
  Open foundation and fine-tuned chat models,'' \emph{arXiv preprint
  arXiv:2307.09288}, 2023.

\bibitem{openai2024gpt4technicalreport}
OpenAI, J.~Achiam, S.~Adler, S.~Agarwal, L.~Ahmad, I.~Akkaya, F.~L. Aleman,
  D.~Almeida, J.~Altenschmidt, S.~Altman \emph{et~al.}, ``Gpt-4 technical
  report,'' \emph{arXiv preprint arXiv:2303.08774}, 2024.

\bibitem{li2023blip2}
J.~Li, D.~Li, S.~Savarese, and S.~Hoi, ``Blip-2: Bootstrapping language-image
  pre-training with frozen image encoders and large language models,'' in
  \emph{Proceedings of the 40th International Conference on Machine Learning
  (ICML)}, ser. Proceedings of Machine Learning Research, A.~Krause,
  E.~Brunskill, K.~Cho, B.~Engelhardt, S.~Sabato, and J.~Scarlett, Eds., vol.
  202.\hskip 1em plus 0.5em minus 0.4em\relax PMLR, 23--29 Jul 2023, pp.
  19\,730--19\,742.

\bibitem{openai2024gpt4ocard}
OpenAI, :, A.~Hurst, A.~Lerer, A.~P. Goucher, A.~Perelman, A.~Ramesh, A.~Clark,
  A.~Ostrow, A.~Welihinda \emph{et~al.}, ``Gpt-4o system card,'' \emph{arXiv
  preprint arXiv:2410.21276}, 2024.

\bibitem{hu2023survey}
Y.~Hu, Q.~Xie, V.~Jain, J.~Francis, J.~Patrikar, N.~Keetha, S.~Kim, Y.~Xie,
  T.~Zhang, H.-S. Fang \emph{et~al.}, ``Toward general-purpose robots via
  foundation models: A survey and meta-analysis,'' 2023.

\bibitem{kawaharazuka2024foundation}
K.~Kawaharazuka, T.~Matsushima, A.~Gambardella, J.~G. Guo, C.~Paxton, and
  A.~Zeng, ``Real-world robot applications of foundation models: A review,''
  \emph{Advanced Robotics}, vol.~38, no.~18, pp. 1232--1254, 2024.

\bibitem{firoozi2025survey}
R.~Firoozi, J.~Tucker, S.~Tian, A.~Majumdar, J.~Sun, W.~Liu, Y.~Zhu, S.~Song,
  A.~Kapoor, K.~Hausman \emph{et~al.}, ``Foundation models in robotics:
  Applications, challenges, and the future,'' \emph{The International Journal
  of Robotics Research}, vol.~44, no.~5, pp. 701--739, 2025.

\bibitem{ichter2023saycan}
b.~ichter, A.~Brohan, Y.~Chebotar, C.~Finn, K.~Hausman, A.~Herzog, D.~Ho,
  J.~Ibarz, A.~Irpan, E.~Jang \emph{et~al.}, ``Do as i can, not as i say:
  Grounding language in robotic affordances,'' in \emph{Proceedings of The 6th
  Conference on Robot Learning (CoRL)}, ser. Proceedings of Machine Learning
  Research, K.~Liu, D.~Kulic, and J.~Ichnowski, Eds., vol. 205.\hskip 1em plus
  0.5em minus 0.4em\relax PMLR, 14--18 Dec 2023, pp. 287--318.

\bibitem{liang2023codeaspolicies}
J.~Liang, W.~Huang, F.~Xia, P.~Xu, K.~Hausman, B.~Ichter, P.~Florence, and
  A.~Zeng, ``Code as policies: Language model programs for embodied control,''
  in \emph{2023 IEEE International Conference on Robotics and Automation
  (ICRA)}, 2023, pp. 9493--9500.

\bibitem{zitkovich2023rt2}
B.~Zitkovich, T.~Yu, S.~Xu, P.~Xu, T.~Xiao, F.~Xia, J.~Wu, P.~Wohlhart,
  S.~Welker, A.~Wahid \emph{et~al.}, ``Rt-2: Vision-language-action models
  transfer web knowledge to robotic control,'' in \emph{Proceedings of The 7th
  Conference on Robot Learning (CoRL)}, ser. Proceedings of Machine Learning
  Research, J.~Tan, M.~Toussaint, and K.~Darvish, Eds., vol. 229.\hskip 1em
  plus 0.5em minus 0.4em\relax PMLR, 06--09 Nov 2023, pp. 2165--2183.

\bibitem{ma2024survey}
Y.~Ma, Z.~Song, Y.~Zhuang, J.~Hao, and I.~King, ``A survey on
  vision-language-action models for embodied ai,'' \emph{arXiv preprint
  arXiv:2405.14093}, 2024.

\bibitem{sapkota2025survey}
R.~Sapkota, Y.~Cao, K.~I. Roumeliotis, and M.~Karkee, ``Vision-language-action
  models: Concepts, progress, applications and challenges,'' \emph{arXiv
  preprint arXiv:2505.04769}, 2025.

\bibitem{yifan2025survey}
Y.~Zhong, F.~Bai, S.~Cai, X.~Huang, Z.~Chen, X.~Zhang, Y.~Wang, S.~Guo,
  T.~Guan, K.~N. Lui \emph{et~al.}, ``A survey on vision-language-action
  models: An action tokenization perspective,'' \emph{arXiv preprint
  arXiv:2507.01925}, 2025.

\bibitem{chen2015cococaptions}
X.~Chen, H.~Fang, T.-Y. Lin, R.~Vedantam, S.~Gupta, P.~Dollar, and C.~L.
  Zitnick, ``Microsoft coco captions: Data collection and evaluation server,''
  \emph{arXiv preprint arXiv:1504.00325}, 2015.

\bibitem{shridhar2021cliport}
\BIBentryALTinterwordspacing
M.~Shridhar, L.~Manuelli, and D.~Fox, ``Cliport: What and where pathways for
  robotic manipulation,'' in \emph{Proceedings of the 5th Conference on Robot
  Learning (CoRL)}, ser. Proceedings of Machine Learning Research, A.~Faust,
  D.~Hsu, and G.~Neumann, Eds., vol. 164.\hskip 1em plus 0.5em minus
  0.4em\relax PMLR, 08--11 Nov 2022, pp. 894--906. [Online]. Available:
  \url{https://proceedings.mlr.press/v164/shridhar22a.html}
\BIBentrySTDinterwordspacing

\bibitem{brohan2023rt1}
A.~Brohan, N.~Brown, J.~Carbajal, Y.~Chebotar, J.~Dabis, C.~Finn,
  K.~Gopalakrishnan, K.~Hausman, A.~Herzog, J.~Hsu \emph{et~al.}, ``Rt-1:
  Robotics transformer for real-world control at scale,'' in \emph{Proceedings
  of Robotics: Science and Systems (RSS)}, Daegu, Republic of Korea, July 2023.

\bibitem{oneill2024openxembodiment}
A.~O'Neill, A.~Rehman, A.~Maddukuri, A.~Gupta, A.~Padalkar, A.~Lee, A.~Pooley,
  A.~Gupta, A.~Mandlekar, A.~Jain \emph{et~al.}, ``Open x-embodiment: Robotic
  learning datasets and rt-x models : Open x-embodiment collaboration0,'' in
  \emph{2024 IEEE International Conference on Robotics and Automation (ICRA)},
  2024, pp. 6892--6903.

\bibitem{kim2024openvla}
M.~J. Kim, K.~Pertsch, S.~Karamcheti, T.~Xiao, A.~Balakrishna, S.~Nair,
  R.~Rafailov, E.~Foster, G.~Lam, P.~Sanketi \emph{et~al.}, ``Openvla: An
  open-source vision-language-action model,'' \emph{arXiv preprint
  arXiv:2406.09246}, 2024.

\bibitem{octoteam2024octo}
O.~M. Team, D.~Ghosh, H.~Walke, K.~Pertsch, K.~Black, O.~Mees, S.~Dasari,
  J.~Hejna, C.~Xu, J.~Luo \emph{et~al.}, ``Octo: An open-source generalist
  robot policy,'' in \emph{Proceedings of Robotics: Science and Systems (RSS)},
  Delft, Netherlands, 2024.

\bibitem{liu2025rdt}
S.~Liu, L.~Wu, B.~Li, H.~Tan, H.~Chen, Z.~Wang, K.~Xu, H.~Su, and J.~Zhu,
  ``Rdt-1b: a diffusion foundation model for bimanual manipulation,'' in
  \emph{International Conference on Learning Representations (ICLR)}, 2025.

\bibitem{Pi-0}
K.~Black, N.~Brown, D.~Driess, A.~Esmail, M.~Equi, C.~Finn, N.~Fusai, L.~Groom,
  K.~Hausman, B.~Ichter \emph{et~al.}, ``{$\pi_0$}: A vision-language-action
  flow model for general robot control,'' \emph{arXiv preprint
  arXiv:2410.24164}, 2024.

\bibitem{ye2025lapa}
S.~Ye, J.~Jang, B.~Jeon, S.~J. Joo, J.~Yang, B.~Peng, A.~Mandlekar, R.~Tan,
  Y.-W. Chao, B.~Y. Lin \emph{et~al.}, ``Latent action pretraining from
  videos,'' in \emph{International Conference on Learning Representations
  (ICLR)}, 2025.

\bibitem{Pi-0.5}
P.~Intelligence, K.~Black, N.~Brown, J.~Darpinian, K.~Dhabalia, D.~Driess,
  A.~Esmail, M.~Equi, C.~Finn, N.~Fusai \emph{et~al.}, ``{$\pi_{0.5}$}: a
  vision-language-action model with open-world generalization,'' \emph{arXiv
  preprint arXiv:2504.16054}, 2025.

\bibitem{Gr00t-N1}
J.~Bjorck, F.~C. {n}eda, N.~Cherniadev, X.~Da, R.~Ding, L.~J. Fan, Y.~Fang,
  D.~Fox, F.~Hu, S.~Huang \emph{et~al.}, ``Gr00t n1: An open foundation model
  for generalist humanoid robots,'' \emph{arXiv preprint arXiv:2503.14734},
  2025.

\bibitem{radford2021clip}
A.~Radford, J.~W. Kim, C.~Hallacy, A.~Ramesh, G.~Goh, S.~Agarwal, G.~Sastry,
  A.~Askell, P.~Mishkin, J.~Clark \emph{et~al.}, ``Learning transferable visual
  models from natural language supervision,'' in \emph{Proceedings of the 38th
  International Conference on Machine Learning (ICML)}, ser. Proceedings of
  Machine Learning Research, M.~Meila and T.~Zhang, Eds., vol. 139.\hskip 1em
  plus 0.5em minus 0.4em\relax PMLR, 18--24 Jul 2021, pp. 8748--8763.

\bibitem{zeng21transporternetworks}
A.~Zeng, P.~Florence, J.~Tompson, S.~Welker, J.~Chien, M.~Attarian,
  T.~Armstrong, I.~Krasin, D.~Duong, V.~Sindhwani \emph{et~al.}, ``Transporter
  networks: Rearranging the visual world for robotic manipulation,'' in
  \emph{Proceedings of the 2020 Conference on Robot Learning (CoRL)}, ser.
  Proceedings of Machine Learning Research, J.~Kober, F.~Ramos, and C.~Tomlin,
  Eds., vol. 155.\hskip 1em plus 0.5em minus 0.4em\relax PMLR, 16--18 Nov 2021,
  pp. 726--747.

\bibitem{reed2022gato}
S.~Reed, K.~Zolna, E.~Parisotto, S.~G. Colmenarejo, A.~Novikov, G.~Barth-maron,
  M.~Gim{\'e}nez, Y.~Sulsky, J.~Kay, J.~T. Springenberg \emph{et~al.}, ``A
  generalist agent,'' \emph{Transactions on Machine Learning Research}, 2022.

\bibitem{vaswani2017transformer}
A.~Vaswani, N.~Shazeer, N.~Parmar, J.~Uszkoreit, L.~Jones, A.~N. Gomez, L.~u.
  Kaiser, and I.~Polosukhin, ``Attention is all you need,'' in \emph{Advances
  in Neural Information Processing Systems (NeurIPS)}, I.~Guyon, U.~V. Luxburg,
  S.~Bengio, H.~Wallach, R.~Fergus, S.~Vishwanathan, and R.~Garnett, Eds.,
  vol.~30.\hskip 1em plus 0.5em minus 0.4em\relax Curran Associates, Inc.,
  2017.

\bibitem{kudo2018sentencepiece}
\BIBentryALTinterwordspacing
T.~Kudo and J.~Richardson, ``Sentencepiece: A simple and language independent
  subword tokenizer and detokenizer for neural text processing,'' in
  \emph{Proceedings of the 2018 Conference on Empirical Methods in Natural
  Language Processing: System Demonstrations}, E.~Blanco and W.~Lu, Eds.\hskip
  1em plus 0.5em minus 0.4em\relax Brussels, Belgium: Association for
  Computational Linguistics, November 2018, pp. 66--71. [Online]. Available:
  \url{https://aclanthology.org/D18-2012/}
\BIBentrySTDinterwordspacing

\bibitem{dosovitskiy2021vit}
A.~Dosovitskiy, L.~Beyer, A.~Kolesnikov, D.~Weissenborn, X.~Zhai,
  T.~Unterthiner, M.~Dehghani, M.~Minderer, G.~Heigold, S.~Gelly \emph{et~al.},
  ``An image is worth 16x16 words: Transformers for image recognition at
  scale,'' in \emph{International Conference on Learning Representations
  (ICLR)}, 2021.

\bibitem{jiang2023vima}
\BIBentryALTinterwordspacing
Y.~Jiang, A.~Gupta, Z.~Zhang, G.~Wang, Y.~Dou, Y.~Chen, L.~Fei-Fei,
  A.~Anandkumar, Y.~Zhu, and L.~Fan, ``{VIMA}: Robot manipulation with
  multimodal prompts,'' in \emph{Proceedings of the 40th International
  Conference on Machine Learning (ICML)}, ser. Proceedings of Machine Learning
  Research, A.~Krause, E.~Brunskill, K.~Cho, B.~Engelhardt, S.~Sabato, and
  J.~Scarlett, Eds., vol. 202.\hskip 1em plus 0.5em minus 0.4em\relax PMLR,
  23--29 Jul 2023, pp. 14\,975--15\,022. [Online]. Available:
  \url{https://proceedings.mlr.press/v202/jiang23b.html}
\BIBentrySTDinterwordspacing

\bibitem{he2017maskrcnn}
K.~He, G.~Gkioxari, P.~Doll{\'a}r, and R.~Girshick, ``Mask r-cnn,'' in
  \emph{Proceedings of the IEEE international conference on computer vision},
  2017, pp. 2961--2969.

\bibitem{raffel2020t5}
\BIBentryALTinterwordspacing
C.~Raffel, N.~Shazeer, A.~Roberts, K.~Lee, S.~Narang, M.~Matena, Y.~Zhou,
  W.~Li, and P.~J. Liu, ``Exploring the limits of transfer learning with a
  unified text-to-text transformer,'' \emph{Journal of Machine Learning
  Research}, vol.~21, no. 140, pp. 1--67, 2020. [Online]. Available:
  \url{http://jmlr.org/papers/v21/20-074.html}
\BIBentrySTDinterwordspacing

\bibitem{tan2019efficientnet}
M.~Tan and Q.~Le, ``Efficientnet: Rethinking model scaling for convolutional
  neural networks,'' in \emph{Proceedings of the 36th International Conference
  on Machine Learning (ICML)}, ser. Proceedings of Machine Learning Research,
  K.~Chaudhuri and R.~Salakhutdinov, Eds., vol.~97.\hskip 1em plus 0.5em minus
  0.4em\relax PMLR, 09--15 Jun 2019, pp. 6105--6114.

\bibitem{perez2018film}
E.~Perez, F.~Strub, H.~de~Vries, V.~Dumoulin, and A.~C. Courville, ``Film:
  Visual reasoning with a general conditioning layer,'' in \emph{AAAI}, 2018.

\bibitem{cer2018use}
D.~Cer, Y.~Yang, S.-y. Kong, N.~Hua, N.~Limtiaco, R.~S. John, N.~Constant,
  M.~Guajardo-Cespedes, S.~Yuan, C.~Tar \emph{et~al.}, ``Universal sentence
  encoder,'' \emph{arXiv preprint arXiv:1803.11175}, 2018.

\bibitem{ryoo2021tokenlearner}
M.~Ryoo, A.~Piergiovanni, A.~Arnab, M.~Dehghani, and A.~Angelova,
  ``Tokenlearner: Adaptive space-time tokenization for videos,'' in
  \emph{Advances in Neural Information Processing Systems (NeurIPS)},
  M.~Ranzato, A.~Beygelzimer, Y.~Dauphin, P.~Liang, and J.~W. Vaughan, Eds.,
  vol.~34.\hskip 1em plus 0.5em minus 0.4em\relax Curran Associates, Inc.,
  2021, pp. 12\,786--12\,797.

\bibitem{driess2023palme}
D.~Driess, F.~Xia, M.~S.~M. Sajjadi, C.~Lynch, A.~Chowdhery, B.~Ichter,
  A.~Wahid, J.~Tompson, Q.~Vuong, T.~Yu \emph{et~al.}, ``Palm-e: An embodied
  multimodal language model,'' in \emph{Proceedings of the 40th International
  Conference on Machine Learning (ICML)}, ser. Proceedings of Machine Learning
  Research, A.~Krause, E.~Brunskill, K.~Cho, B.~Engelhardt, S.~Sabato, and
  J.~Scarlett, Eds., vol. 202.\hskip 1em plus 0.5em minus 0.4em\relax PMLR,
  23--29 Jul 2023, pp. 8469--8488.

\bibitem{chen2024palix}
X.~Chen, J.~Djolonga, P.~Padlewski, B.~Mustafa, S.~Changpinyo, J.~Wu, C.~R.
  Ruiz, S.~Goodman, X.~Wang, Y.~Tay \emph{et~al.}, ``On scaling up a
  multilingual vision and language model,'' in \emph{Proceedings of the
  IEEE/CVF Conference on Computer Vision and Pattern Recognition (CVPR)}, June
  2024, pp. 14\,432--14\,444.

\bibitem{sundaresan2025rtsketch}
P.~Sundaresan, Q.~Vuong, J.~Gu, P.~Xu, T.~Xiao, S.~Kirmani, T.~Yu, M.~Stark,
  A.~Jain, K.~Hausman \emph{et~al.}, ``Rt-sketch: Goal-conditioned imitation
  learning from hand-drawn sketches,'' in \emph{Proceedings of The 8th
  Conference on Robot Learning (CoRL)}, ser. Proceedings of Machine Learning
  Research, P.~Agrawal, O.~Kroemer, and W.~Burgard, Eds., vol. 270.\hskip 1em
  plus 0.5em minus 0.4em\relax PMLR, 06--09 Nov 2025, pp. 70--96.

\bibitem{gu2024rttrajectory}
J.~Gu, S.~Kirmani, P.~Wohlhart, Y.~Lu, M.~G. Arenas, K.~Rao, W.~Yu, C.~Fu,
  K.~Gopalakrishnan, Z.~Xu \emph{et~al.}, ``Rt-trajectory: Robotic task
  generalization via hindsight trajectory sketches,'' in \emph{International
  Conference on Learning Representations (ICLR)}, 2024.

\bibitem{belkhale2024rth}
S.~Belkhale, T.~Ding, T.~Xiao, P.~Sermanet, Q.~Vuong, J.~Tompson, Y.~Chebotar,
  D.~Dwibedi, and D.~Sadigh, ``Rt-h: Action hierarchies using language,'' in
  \emph{Proceedings of Robotics: Science and Systems (RSS)}, Delft,
  Netherlands, July 2024.

\bibitem{leal2024sarart}
I.~Leal, K.~Choromanski, D.~Jain, A.~Dubey, J.~Varley, M.~Ryoo, Y.~Lu, F.~Liu,
  V.~Sindhwani, Q.~Vuong \emph{et~al.}, ``Sara-rt: Scaling up robotics
  transformers with self-adaptive robust attention,'' in \emph{2024 IEEE
  International Conference on Robotics and Automation (ICRA)}, 2024, pp.
  6920--6927.

\bibitem{AutoRT}
M.~Ahn, D.~Dwibedi, C.~Finn, M.~G. Arenas, K.~Gopalakrishnan, K.~Hausman,
  B.~Ichter, A.~Irpan, N.~Joshi, R.~Julian \emph{et~al.}, ``Autort: Embodied
  foundation models for large scale orchestration of robotic agents,''
  \emph{arXiv preprint arXiv:2401.12963}, 2024.

\bibitem{karamcheti2024prismaticvlm}
\BIBentryALTinterwordspacing
S.~Karamcheti, S.~Nair, A.~Balakrishna, P.~Liang, T.~Kollar, and D.~Sadigh,
  ``Prismatic {VLM}s: Investigating the design space of visually-conditioned
  language models,'' in \emph{Proceedings of the 41st International Conference
  on Machine Learning (ICML)}, ser. Proceedings of Machine Learning Research,
  R.~Salakhutdinov, Z.~Kolter, K.~Heller, A.~Weller, N.~Oliver, J.~Scarlett,
  and F.~Berkenkamp, Eds., vol. 235.\hskip 1em plus 0.5em minus 0.4em\relax
  PMLR, 21--27 Jul 2024, pp. 23\,123--23\,144. [Online]. Available:
  \url{https://proceedings.mlr.press/v235/karamcheti24a.html}
\BIBentrySTDinterwordspacing

\bibitem{oquab2023dinov2}
M.~Oquab, T.~Darcet, T.~Moutakanni, H.~Vo, M.~Szafraniec, V.~Khalidov,
  P.~Fernandez, D.~Haziza, F.~Massa, A.~El-Nouby \emph{et~al.}, ``Dinov2:
  Learning robust visual features without supervision,'' \emph{arXiv preprint
  arXiv:2304.07193}, 2024.

\bibitem{zhai2023siglip}
X.~Zhai, B.~Mustafa, A.~Kolesnikov, and L.~Beyer, ``Sigmoid loss for language
  image pre-training,'' in \emph{Proceedings of the IEEE/CVF International
  Conference on Computer Vision (ICCV)}, October 2023, pp. 11\,975--11\,986.

\bibitem{chi2023diffusionpolicy}
C.~Chi, S.~Feng, Y.~Du, Z.~Xu, E.~Cousineau, B.~C. Burchfiel, and S.~Song,
  ``Diffusion policy: Visuomotor policy learning via action diffusion,'' in
  \emph{Proceedings of Robotics: Science and Systems (RSS)}, Daegu, Republic of
  Korea, July 2023.

\bibitem{peebles2023dit}
W.~Peebles and S.~Xie, ``Scalable diffusion models with transformers,'' in
  \emph{Proceedings of the IEEE/CVF International Conference on Computer Vision
  (ICCV)}, October 2023, pp. 4195--4205.

\bibitem{zhoutransfusion}
C.~Zhou, L.~YU, A.~Babu, K.~Tirumala, M.~Yasunaga, L.~Shamis, J.~Kahn, X.~Ma,
  L.~Zettlemoyer, and O.~Levy, ``Transfusion: Predict the next token and
  diffuse images with one multi-modal model,'' in \emph{International
  Conference on Learning Representations (ICLR)}, 2025.

\bibitem{beyer2024paligemma}
L.~Beyer, A.~Steiner, A.~S. Pinto, A.~Kolesnikov, X.~Wang, D.~Salz, M.~Neumann,
  I.~Alabdulmohsin, M.~Tschannen, E.~Bugliarello \emph{et~al.}, ``Paligemma: A
  versatile 3b vlm for transfer,'' \emph{arXiv preprint arXiv:2407.07726},
  2024.

\bibitem{lipman2023flow}
Y.~Lipman, R.~T.~Q. Chen, H.~Ben-Hamu, M.~Nickel, and M.~Le, ``Flow matching
  for generative modeling,'' in \emph{International Conference on Learning
  Representations (ICLR)}, 2023.

\bibitem{van2017vqvae}
A.~van~den Oord, O.~Vinyals, and k.~kavukcuoglu, ``Neural discrete
  representation learning,'' in \emph{Advances in Neural Information Processing
  Systems (NeurIPS)}, I.~Guyon, U.~V. Luxburg, S.~Bengio, H.~Wallach,
  R.~Fergus, S.~Vishwanathan, and R.~Garnett, Eds., vol.~30.\hskip 1em plus
  0.5em minus 0.4em\relax Curran Associates, Inc., 2017.

\bibitem{liu2023world}
H.~Liu, W.~Yan, M.~Zaharia, and P.~Abbeel, ``World model on million-length
  video and language with ringattention,'' \emph{arXiv preprint}, 2024.

\bibitem{Pi-0-FAST}
K.~Pertsch, K.~Stachowicz, B.~Ichter, D.~Driess, S.~Nair, Q.~Vuong, O.~Mees,
  C.~Finn, and S.~Levine, ``Fast: Efficient action tokenization for
  vision-language-action models,'' \emph{arXiv preprint arXiv:2501.09747},
  2025.

\bibitem{VIMA}
Y.~Jiang, A.~Gupta, Z.~Zhang, G.~Wang, Y.~Dou, Y.~Chen, L.~Fei-Fei,
  A.~Anandkumar, Y.~Zhu, and L.~Fan, ``Vima: General robot manipulation with
  multimodal prompts,'' \emph{arXiv preprint arXiv:2210.03094}, 2022.

\bibitem{stone2023moo}
A.~Stone, T.~Xiao, Y.~Lu, K.~Gopalakrishnan, K.-H. Lee, Q.~Vuong, P.~Wohlhart,
  S.~Kirmani, B.~Zitkovich, F.~Xia \emph{et~al.}, ``Open-world object
  manipulation using pre-trained vision-language models,'' in \emph{Proceedings
  of The 7th Conference on Robot Learning (CoRL)}, ser. Proceedings of Machine
  Learning Research, J.~Tan, M.~Toussaint, and K.~Darvish, Eds., vol.
  229.\hskip 1em plus 0.5em minus 0.4em\relax PMLR, 06--09 Nov 2023, pp.
  3397--3417.

\bibitem{RT-Sketch}
P.~Sundaresan, Q.~Vuong, J.~Gu, P.~Xu, T.~Xiao, S.~Kirmani, T.~Yu, M.~Stark,
  A.~Jain, K.~Hausman \emph{et~al.}, ``Rt-sketch: Goal-conditioned imitation
  learning from hand-drawn sketches,'' \emph{arXiv preprint arXiv:2403.02709},
  2024.

\bibitem{RT-Trajectory}
J.~Gu, S.~Kirmani, P.~Wohlhart, Y.~Lu, M.~G. Arenas, K.~Rao, W.~Yu, C.~Fu,
  K.~Gopalakrishnan, Z.~Xu \emph{et~al.}, ``Rt-trajectory: Robotic task
  generalization via hindsight trajectory sketches,'' \emph{arXiv preprint
  arXiv:2311.01977}, 2023.

\bibitem{bousmalis2024robocat}
K.~Bousmalis, G.~Vezzani, D.~Rao, C.~M. Devin, A.~X. Lee, M.~B. Villalonga,
  T.~Davchev, Y.~Zhou, A.~Gupta, A.~Raju \emph{et~al.}, ``Robocat: A
  self-improving generalist agent for robotic manipulation,''
  \emph{Transactions on Machine Learning Research}, 2024.

\bibitem{RoboFlamingo}
X.~Li, M.~Liu, H.~Zhang, C.~Yu, J.~Xu, H.~Wu, C.~Cheang, Y.~Jing, W.~Zhang,
  H.~Liu \emph{et~al.}, ``Vision-language foundation models as effective robot
  imitators,'' \emph{arXiv preprint arXiv:2311.01378}, 2023.

\bibitem{lu2024unifiedio2}
J.~Lu, C.~Clark, S.~Lee, Z.~Zhang, S.~Khosla, R.~Marten, D.~Hoiem, and
  A.~Kembhavi, ``Unified-io 2: Scaling autoregressive multimodal models with
  vision language audio and action,'' in \emph{Proceedings of the IEEE/CVF
  Conference on Computer Vision and Pattern Recognition (CVPR)}, June 2024, pp.
  26\,439--26\,455.

\bibitem{shi2024yay}
L.~X. Shi, Z.~Hu, T.~Z. Zhao, A.~Sharma, K.~Pertsch, J.~Luo, S.~Levine, and
  C.~Finn, ``Yell at your robot: Improving on-the-fly from language
  corrections,'' in \emph{Proceedings of Robotics: Science and Systems (RSS)},
  Delft, Netherlands, July 2024.

\bibitem{haldar2024baku}
S.~Haldar, Z.~Peng, and L.~Pinto, ``Baku: An efficient transformer for
  multi-task policy learning,'' in \emph{Advances in Neural Information
  Processing Systems (NeurIPS)}, A.~Globerson, L.~Mackey, D.~Belgrave, A.~Fan,
  U.~Paquet, J.~Tomczak, and C.~Zhang, Eds., vol.~37.\hskip 1em plus 0.5em
  minus 0.4em\relax Curran Associates, Inc., 2024, pp. 141\,208--141\,239.

\bibitem{Actra}
Y.~Ma, D.~Chi, S.~Wu, Y.~Liu, Y.~Zhuang, J.~Hao, and I.~King, ``Actra:
  Optimized transformer architecture for vision-language-action models in robot
  learning,'' \emph{arXiv preprint arXiv:2408.01147}, 2024.

\bibitem{pang2024di2}
X.~Pang, W.~Xia, Z.~Wang, B.~Zhao, D.~Hu, D.~Wang, and X.~Li, ``Depth helps:
  Improving pre-trained rgb-based policy with depth information injection,'' in
  \emph{Proceedings of the IEEE/RSJ International Conference on Intelligent
  Robots and Systems (IROS)}, 2024, pp. 7251--7256.

\bibitem{doshi2025crossformer}
R.~Doshi, H.~R. Walke, O.~Mees, S.~Dasari, and S.~Levine, ``Scaling
  cross-embodied learning: One policy for manipulation, navigation, locomotion
  and aviation,'' in \emph{Proceedings of The 8th Conference on Robot Learning
  (CoRL)}, ser. Proceedings of Machine Learning Research, P.~Agrawal,
  O.~Kroemer, and W.~Burgard, Eds., vol. 270.\hskip 1em plus 0.5em minus
  0.4em\relax PMLR, 06--09 Nov 2025, pp. 496--512.

\bibitem{ho2020ddpm}
J.~Ho, A.~Jain, and P.~Abbeel, ``Denoising diffusion probabilistic models,'' in
  \emph{Advances in Neural Information Processing Systems (NeurIPS)},
  H.~Larochelle, M.~Ranzato, R.~Hadsell, M.~Balcan, and H.~Lin, Eds.,
  vol.~33.\hskip 1em plus 0.5em minus 0.4em\relax Curran Associates, Inc.,
  2020, pp. 6840--6851.

\bibitem{sridhar2024nomad}
A.~Sridhar, D.~Shah, C.~Glossop, and S.~Levine, ``Nomad: Goal masked diffusion
  policies for navigation and exploration,'' in \emph{2024 IEEE International
  Conference on Robotics and Automation (ICRA)}, 2024, pp. 63--70.

\bibitem{wen2024tinyvla}
J.~Wen, Y.~Zhu, J.~Li, M.~Zhu, Z.~Tang, K.~Wu, Z.~Xu, N.~Liu, R.~Cheng, C.~Shen
  \emph{et~al.}, ``Tinyvla: Toward fast, data-efficient vision-language-action
  models for robotic manipulation,'' \emph{IEEE Robotics and Automation Letters
  (RA-L)}, vol.~10, no.~4, pp. 3988--3995, 2025.

\bibitem{RoboBERT}
S.~Wang, S.~Liu, W.~Wang, J.~Shan, and B.~Fang, ``Robobert: An end-to-end
  multimodal robotic manipulation model,'' \emph{arXiv preprint
  arXiv:2502.07837}, 2025.

\bibitem{chen2025vidbot}
H.~Chen, B.~Sun, A.~Zhang, M.~Pollefeys, and S.~Leutenegger, ``Vidbot: Learning
  generalizable 3d actions from in-the-wild 2d human videos for zero-shot
  robotic manipulation,'' in \emph{Proceedings of the IEEE/CVF Conference on
  Computer Vision and Pattern Recognition (CVPR)}, June 2025, pp.
  27\,661--27\,672.

\bibitem{liu2023structdiffusion}
W.~Liu, Y.~Du, T.~Hermans, S.~Chernova, and C.~Paxton, ``Structdiffusion:
  Language-guided creation of physically-valid structures using unseen
  objects,'' in \emph{Proceedings of Robotics: Science and Systems (RSS)},
  Daegu, Republic of Korea, July 2023.

\bibitem{reuss2024mdt}
M.~Reuss, \"{O}mer Erdin\c{c}~Ya\u{g}murlu, F.~Wenzel, and R.~Lioutikov,
  ``{Multimodal Diffusion Transformer: Learning Versatile Behavior from
  Multimodal Goals},'' in \emph{Proceedings of Robotics: Science and Systems
  (RSS)}, Delft, Netherlands, July 2024.

\bibitem{DexGraspVLA}
Y.~Zhong, X.~Huang, R.~Li, C.~Zhang, Y.~Liang, Y.~Yang, and Y.~Chen,
  ``Dexgraspvla: A vision-language-action framework towards general dexterous
  grasping,'' \emph{arXiv preprint arXiv:2502.20900}, 2025.

\bibitem{UnifiedVideoActionModel}
S.~Li, Y.~Gao, D.~Sadigh, and S.~Song, ``Unified video action model,''
  \emph{arXiv preprint arXiv:2503.00200}, 2025.

\bibitem{FP3}
R.~Yang, G.~Chen, C.~Wen, and Y.~Gao, ``Fp3: A 3d foundation policy for robotic
  manipulation,'' \emph{arXiv preprint arXiv:2503.08950}, 2025.

\bibitem{yao2025ppl}
Y.~Yao, S.~Liu, H.~Song, D.~Qu, Q.~Chen, Y.~Ding, B.~Zhao, Z.~Wang, X.~Li, and
  D.~Wang, ``Think small, act big: Primitive prompt learning for lifelong robot
  manipulation,'' in \emph{Proceedings of the IEEE/CVF Conference on Computer
  Vision and Pattern Recognition (CVPR)}, June 2025, pp. 22\,573--22\,583.

\bibitem{PPI}
Y.~Yang, Z.~Cai, Y.~Tian, J.~Zeng, and J.~Pang, ``Gripper keypose and object
  pointflow as interfaces for bimanual robotic manipulation,'' \emph{arXiv
  preprint arXiv:2504.17784}, 2025.

\bibitem{Dita}
Z.~Hou, T.~Zhang, Y.~Xiong, H.~Pu, C.~Zhao, R.~Tong, Y.~Qiao, J.~Dai, and
  Y.~Chen, ``Diffusion transformer policy,'' \emph{arXiv preprint
  arXiv:2410.15959}, 2024.

\bibitem{huang2024leo}
\BIBentryALTinterwordspacing
J.~Huang, S.~Yong, X.~Ma, X.~Linghu, P.~Li, Y.~Wang, Q.~Li, S.-C. Zhu, B.~Jia,
  and S.~Huang, ``An embodied generalist agent in 3{D} world,'' in
  \emph{Proceedings of the 41st International Conference on Machine Learning
  (ICML)}, ser. Proceedings of Machine Learning Research, R.~Salakhutdinov,
  Z.~Kolter, K.~Heller, A.~Weller, N.~Oliver, J.~Scarlett, and F.~Berkenkamp,
  Eds., vol. 235.\hskip 1em plus 0.5em minus 0.4em\relax PMLR, 21--27 Jul 2024,
  pp. 20\,413--20\,451. [Online]. Available:
  \url{https://proceedings.mlr.press/v235/huang24ae.html}
\BIBentrySTDinterwordspacing

\bibitem{GR-1}
H.~Wu, Y.~Jing, C.~Cheang, G.~Chen, J.~Xu, X.~Li, M.~Liu, H.~Li, and T.~Kong,
  ``Unleashing large-scale video generative pre-training for visual robot
  manipulation,'' \emph{arXiv preprint arXiv:2312.13139}, 2023.

\bibitem{liu2024robomamba}
J.~Liu, M.~Liu, Z.~Wang, P.~An, X.~Li, K.~Zhou, S.~Yang, R.~Zhang, Y.~Guo, and
  S.~Zhang, ``Robomamba: Efficient vision-language-action model for robotic
  reasoning and manipulation,'' in \emph{Advances in Neural Information
  Processing Systems (NeurIPS)}, A.~Globerson, L.~Mackey, D.~Belgrave, A.~Fan,
  U.~Paquet, J.~Tomczak, and C.~Zhang, Eds., vol.~37.\hskip 1em plus 0.5em
  minus 0.4em\relax Curran Associates, Inc., 2024, pp. 40\,085--40\,110.

\bibitem{ding2024quarvla}
P.~Ding, H.~Zhao, W.~Zhang, W.~Song, M.~Zhang, S.~Huang, N.~Yang, and D.~Wang,
  ``Quar-vla: Vision-language-action model for~quadruped robots,'' in
  \emph{Computer Vision -- ECCV 2024}, A.~Leonardis, E.~Ricci, S.~Roth,
  O.~Russakovsky, T.~Sattler, and G.~Varol, Eds.\hskip 1em plus 0.5em minus
  0.4em\relax Cham: Springer Nature Switzerland, 2025, pp. 352--367.

\bibitem{li2025llara}
X.~Li, C.~Mata, J.~Park, K.~Kahatapitiya, Y.~S. Jang, J.~Shang, K.~Ranasinghe,
  R.~Burgert, M.~Cai, Y.~J. Lee \emph{et~al.}, ``Llara: Supercharging robot
  learning data for vision-language policy,'' in \emph{International Conference
  on Learning Representations (ICLR)}, 2025.

\bibitem{zawalski2025ecot}
M.~Zawalski, W.~Chen, K.~Pertsch, O.~Mees, C.~Finn, and S.~Levine, ``Robotic
  control via embodied chain-of-thought reasoning,'' in \emph{Proceedings of
  The 8th Conference on Robot Learning (CoRL)}, ser. Proceedings of Machine
  Learning Research, P.~Agrawal, O.~Kroemer, and W.~Burgard, Eds., vol.
  270.\hskip 1em plus 0.5em minus 0.4em\relax PMLR, 06--09 Nov 2025, pp.
  3157--3181.

\bibitem{3D-VLA}
H.~Zhen, X.~Qiu, P.~Chen, J.~Yang, X.~Yan, Y.~Du, Y.~Hong, and C.~Gan,
  ``3d-vla: A 3d vision-language-action generative world model,'' \emph{arXiv
  preprint arXiv:2403.09631}, 2024.

\bibitem{RoboUniView}
F.~Liu, F.~Yan, L.~Zheng, C.~Feng, Y.~Huang, and L.~Ma, ``Robouniview:
  Visual-language model with unified view representation for robotic
  manipulation,'' \emph{arXiv preprint arXiv:2406.18977}, 2024.

\bibitem{arai2025covla}
H.~Arai, K.~Miwa, K.~Sasaki, K.~Watanabe, Y.~Yamaguchi, S.~Aoki, and
  I.~Yamamoto, ``Covla: Comprehensive vision-language-action dataset for
  autonomous driving,'' in \emph{Proceedings of the Winter Conference on
  Applications of Computer Vision (WACV)}, February 2025, pp. 1933--1943.

\bibitem{DiffusionVLA}
J.~Wen, M.~Zhu, Y.~Zhu, Z.~Tang, J.~Li, Z.~Zhou, C.~Li, X.~Liu, Y.~Peng,
  C.~Shen \emph{et~al.}, ``Diffusion-vla: Generalizable and interpretable robot
  foundation model via self-generated reasoning,'' \emph{arXiv preprint
  arXiv:2412.03293}, 2024.

\bibitem{DexVLA}
J.~Wen, Y.~Zhu, J.~Li, Z.~Tang, C.~Shen, and F.~Feng, ``Dexvla: Vision-language
  model with plug-in diffusion expert for general robot control,'' \emph{arXiv
  preprint arXiv:2502.05855}, 2025.

\bibitem{ChatVLA}
Z.~Zhou, Y.~Zhu, M.~Zhu, J.~Wen, N.~Liu, Z.~Xu, W.~Meng, R.~Cheng, Y.~Peng,
  C.~Shen \emph{et~al.}, ``Chatvla: Unified multimodal understanding and robot
  control with vision-language-action model,'' \emph{arXiv preprint
  arXiv:2502.14420}, 2025.

\bibitem{ObjectVLA}
M.~Zhu, Y.~Zhu, J.~Li, Z.~Zhou, J.~Wen, X.~Liu, C.~Shen, Y.~Peng, and F.~Feng,
  ``Objectvla: End-to-end open-world object manipulation without
  demonstration,'' \emph{arXiv preprint arXiv:2502.19250}, 2025.

\bibitem{AgiBotWorldColosseo}
AgiBot-World-Contributors, Q.~Bu, J.~Cai, L.~Chen, X.~Cui, Y.~Ding, S.~Feng,
  S.~Gao, X.~He, X.~Hu \emph{et~al.}, ``Agibot world colosseo: A large-scale
  manipulation platform for scalable and intelligent embodied systems,''
  \emph{arXiv preprint arXiv:2503.06669}, 2025.

\bibitem{PointVLA}
C.~Li, J.~Wen, Y.~Peng, Y.~Peng, F.~Feng, and Y.~Zhu, ``Pointvla: Injecting the
  3d world into vision-language-action models,'' \emph{arXiv preprint
  arXiv:2503.07511}, 2025.

\bibitem{MoLe-VLA}
R.~Zhang, M.~Dong, Y.~Zhang, L.~Heng, X.~Chi, G.~Dai, L.~Du, Y.~Du, and
  S.~Zhang, ``Mole-vla: Dynamic layer-skipping vision language action model via
  mixture-of-layers for efficient robot manipulation,'' \emph{arXiv preprint
  arXiv:2503.20384}, 2025.

\bibitem{Fast-in-Slow}
H.~Chen, J.~Liu, C.~Gu, Z.~Liu, R.~Zhang, X.~Li, X.~He, Y.~Guo, C.-W. Fu,
  S.~Zhang \emph{et~al.}, ``Fast-in-slow: A dual-system foundation model
  unifying fast manipulation within slow reasoning,'' \emph{arXiv preprint
  arXiv:2506.01953}, 2025.

\bibitem{CronusVLA}
H.~Li, S.~Yang, Y.~Chen, Y.~Tian, X.~Yang, X.~Chen, H.~Wang, T.~Wang, F.~Zhao,
  D.~Lin \emph{et~al.}, ``Cronusvla: Transferring latent motion across time for
  multi-frame prediction in manipulation,'' \emph{arXiv preprint
  arXiv:2506.19816}, 2025.

\bibitem{HybridVLA}
J.~Liu, H.~Chen, P.~An, Z.~Liu, R.~Zhang, C.~Gu, X.~Li, Z.~Guo, S.~Chen, M.~Liu
  \emph{et~al.}, ``Hybridvla: Collaborative diffusion and autoregression in a
  unified vision-language-action model,'' \emph{arXiv preprint
  arXiv:2503.10631}, 2025.

\bibitem{GraspVLA}
S.~Deng, M.~Yan, S.~Wei, H.~Ma, Y.~Yang, J.~Chen, Z.~Zhang, T.~Yang, X.~Zhang,
  H.~Cui \emph{et~al.}, ``Graspvla: a grasping foundation model pre-trained on
  billion-scale synthetic action data,'' \emph{arXiv preprint
  arXiv:2505.03233}, 2025.

\bibitem{OneTwoVLA}
F.~Lin, R.~Nai, Y.~Hu, J.~You, J.~Zhao, and Y.~Gao, ``Onetwovla: A unified
  vision-language-action model with adaptive reasoning,'' \emph{arXiv preprint
  arXiv:2505.11917}, 2025.

\bibitem{Hume}
H.~Song, D.~Qu, Y.~Yao, Q.~Chen, Q.~Lv, Y.~Tang, M.~Shi, G.~Ren, M.~Yao,
  B.~Zhao \emph{et~al.}, ``Hume: Introducing system-2 thinking in
  visual-language-action model,'' \emph{arXiv preprint arXiv:2505.21432}, 2025.

\bibitem{SwitchVLA}
M.~Li, Z.~Zhao, Z.~Che, F.~Liao, K.~Wu, Z.~Xu, P.~Ren, Z.~Jin, N.~Liu, and
  J.~Tang, ``Switchvla: Execution-aware task switching for
  vision-language-action models,'' \emph{arXiv preprint arXiv:2506.03574},
  2025.

\bibitem{CogAct}
Q.~Li, Y.~Liang, Z.~Wang, L.~Luo, X.~Chen, M.~Liao, F.~Wei, Y.~Deng, S.~Xu,
  Y.~Zhang \emph{et~al.}, ``Cogact: A foundational vision-language-action model
  for synergizing cognition and action in robotic manipulation,'' \emph{arXiv
  preprint arXiv:2411.19650}, 2024.

\bibitem{TrackVLA}
S.~Wang, J.~Zhang, M.~Li, J.~Liu, A.~Li, K.~Wu, F.~Zhong, J.~Yu, Z.~Zhang, and
  H.~Wang, ``Trackvla: Embodied visual tracking in the wild,'' \emph{arXiv
  preprint arXiv:2505.23189}, 2025.

\bibitem{SmolVLA}
M.~Shukor, D.~Aubakirova, F.~Capuano, P.~Kooijmans, S.~Palma, A.~Zouitine,
  M.~Aractingi, C.~Pascal, M.~Russi, A.~Marafioti \emph{et~al.}, ``Smolvla: A
  vision-language-action model for affordable and efficient robotics,''
  \emph{arXiv preprint arXiv:2506.01844}, 2025.

\bibitem{MinD}
X.~Chi, K.~Ge, J.~Liu, S.~Zhou, P.~Jia, Z.~He, Y.~Liu, T.~Li, L.~Han, S.~Han
  \emph{et~al.}, ``Mind: Unified visual imagination and control via
  hierarchical world models,'' \emph{arXiv preprint arXiv:2506.18897}, 2025.

\bibitem{du2023unipi}
Y.~Du, S.~Yang, B.~Dai, H.~Dai, O.~Nachum, J.~Tenenbaum, D.~Schuurmans, and
  P.~Abbeel, ``Learning universal policies via text-guided video generation,''
  in \emph{Advances in Neural Information Processing Systems (NeurIPS)}, A.~Oh,
  T.~Naumann, A.~Globerson, K.~Saenko, M.~Hardt, and S.~Levine, Eds.,
  vol.~36.\hskip 1em plus 0.5em minus 0.4em\relax Curran Associates, Inc.,
  2023, pp. 9156--9172.

\bibitem{ho2022videodiffusionmodel}
J.~Ho, T.~Salimans, A.~Gritsenko, W.~Chan, M.~Norouzi, and D.~J. Fleet, ``Video
  diffusion models,'' in \emph{Advances in Neural Information Processing
  Systems (NeurIPS)}, S.~Koyejo, S.~Mohamed, A.~Agarwal, D.~Belgrave, K.~Cho,
  and A.~Oh, Eds., vol.~35.\hskip 1em plus 0.5em minus 0.4em\relax Curran
  Associates, Inc., 2022, pp. 8633--8646.

\bibitem{DreamGen}
J.~Jang, S.~Ye, Z.~Lin, J.~Xiang, J.~Bjorck, Y.~Fang, F.~Hu, S.~Huang,
  K.~Kundalia, Y.-C. Lin \emph{et~al.}, ``Dreamgen: Unlocking generalization in
  robot learning through video world models,'' \emph{arXiv preprint
  arXiv:2505.12705}, 2025.

\bibitem{zhang2025gevrm}
H.~Zhang, P.~Ding, S.~Lyu, Y.~Peng, and D.~Wang, ``Gevrm: Goal-expressive video
  generation model for robust visual manipulation,'' in \emph{International
  Conference on Learning Representations (ICLR)}, 2025.

\bibitem{ajay2023hip}
A.~Ajay, S.~Han, Y.~Du, S.~Li, A.~Gupta, T.~Jaakkola, J.~Tenenbaum,
  L.~Kaelbling, A.~Srivastava, and P.~Agrawal, ``Compositional foundation
  models for hierarchical planning,'' in \emph{Advances in Neural Information
  Processing Systems (NeurIPS)}, A.~Oh, T.~Naumann, A.~Globerson, K.~Saenko,
  M.~Hardt, and S.~Levine, Eds., vol.~36.\hskip 1em plus 0.5em minus
  0.4em\relax Curran Associates, Inc., 2023, pp. 22\,304--22\,325.

\bibitem{liang2025dreamitate}
J.~Liang, R.~Liu, E.~Ozguroglu, S.~Sudhakar, A.~Dave, P.~Tokmakov, S.~Song, and
  C.~Vondrick, ``Dreamitate: Real-world visuomotor policy learning via video
  generation,'' in \emph{Proceedings of The 8th Conference on Robot Learning
  (CoRL)}, ser. Proceedings of Machine Learning Research, P.~Agrawal,
  O.~Kroemer, and W.~Burgard, Eds., vol. 270.\hskip 1em plus 0.5em minus
  0.4em\relax PMLR, 06--09 Nov 2025, pp. 3943--3960.

\bibitem{blattmann2023stablevideodiffusion}
A.~Blattmann, T.~Dockhorn, S.~Kulal, D.~Mendelevitch, M.~Kilian, D.~Lorenz,
  Y.~Levi, Z.~English, V.~Voleti, A.~Letts \emph{et~al.}, ``Stable video
  diffusion: Scaling latent video diffusion models to large datasets,''
  \emph{arXiv preprint arXiv:2311.15127}, 2023.

\bibitem{labbe2022megapose}
\BIBentryALTinterwordspacing
Y.~Labb\'e, L.~Manuelli, A.~Mousavian, S.~Tyree, S.~Birchfield, J.~Tremblay,
  J.~Carpentier, M.~Aubry, D.~Fox, and J.~Sivic, ``Megapose: 6d pose estimation
  of novel objects via render \& compare,'' in \emph{Proceedings of The 6th
  Conference on Robot Learning (CoRL)}, ser. Proceedings of Machine Learning
  Research, K.~Liu, D.~Kulic, and J.~Ichnowski, Eds., vol. 205.\hskip 1em plus
  0.5em minus 0.4em\relax PMLR, 14--18 Dec 2023, pp. 715--725. [Online].
  Available: \url{https://proceedings.mlr.press/v205/labbe23a.html}
\BIBentrySTDinterwordspacing

\bibitem{black2024susie}
K.~Black, M.~Nakamoto, P.~Atreya, H.~R. Walke, C.~Finn, A.~Kumar, and
  S.~Levine, ``Zero-shot robotic manipulation with pre-trained image-editing
  diffusion models,'' in \emph{International Conference on Learning
  Representations (ICLR)}, 2024.

\bibitem{brooks2023instructpix2pix}
T.~Brooks, A.~Holynski, and A.~A. Efros, ``Instructpix2pix: Learning to follow
  image editing instructions,'' in \emph{Proceedings of the IEEE/CVF Conference
  on Computer Vision and Pattern Recognition (CVPR)}, June 2023, pp.
  18\,392--18\,402.

\bibitem{nematollahi25lumos}
I.~Nematollahi, B.~DeMoss, A.~L. Chandra, N.~Hawes, W.~Burgard, and I.~Posner,
  ``Lumos: Language-conditioned imitation learning with world models,''
  \emph{arXiv preprint arXiv:2503.10370}, 2025.

\bibitem{ko2024avdc}
P.-C. Ko, J.~Mao, Y.~Du, S.-H. Sun, and J.~B. Tenenbaum, ``Learning to act from
  actionless videos through dense correspondences,'' in \emph{International
  Conference on Learning Representations (ICLR)}, 2024.

\bibitem{xu2022gmflow}
H.~Xu, J.~Zhang, J.~Cai, H.~Rezatofighi, and D.~Tao, ``Gmflow: Learning optical
  flow via global matching,'' in \emph{Proceedings of the IEEE/CVF Conference
  on Computer Vision and Pattern Recognition (CVPR)}, 2022, pp. 8121--8130.

\bibitem{wen2024atm}
C.~Wen, X.~Lin, J.~I.~R. So, K.~Chen, Q.~Dou, Y.~Gao, and P.~Abbeel,
  ``Any-point trajectory modeling for policy learning,'' in \emph{Proceedings
  of Robotics: Science and Systems (RSS)}, Delft, Netherlands, July 2024.

\bibitem{karaev2024cotracker}
N.~Karaev, I.~Rocco, B.~Graham, N.~Neverova, A.~Vedaldi, and C.~Rupprecht,
  ``Cotracker: It is better to~track together,'' in \emph{Computer Vision --
  ECCV 2024}, A.~Leonardis, E.~Ricci, S.~Roth, O.~Russakovsky, T.~Sattler, and
  G.~Varol, Eds.\hskip 1em plus 0.5em minus 0.4em\relax Cham: Springer Nature
  Switzerland, 2025, pp. 18--35.

\bibitem{bharadhwaj2024track2act}
H.~Bharadhwaj, R.~Mottaghi, A.~Gupta, and S.~Tulsiani, ``Track2act: Predicting
  point tracks from~internet videos enables generalizable robot manipulation,''
  in \emph{Computer Vision -- ECCV 2024}, A.~Leonardis, E.~Ricci, S.~Roth,
  O.~Russakovsky, T.~Sattler, and G.~Varol, Eds.\hskip 1em plus 0.5em minus
  0.4em\relax Cham: Springer Nature Switzerland, 2025, pp. 306--324.

\bibitem{LangToMo}
K.~Ranasinghe, X.~Li, C.~Mata, J.~Park, and M.~S. Ryoo, ``Pixel motion as
  universal representation for robot control,'' \emph{arXiv preprint
  arXiv:2505.07817}, 2025.

\bibitem{teed2020raft}
Z.~Teed and J.~Deng, ``Raft: Recurrent all-pairs field transforms for optical
  flow,'' in \emph{Computer Vision -- ECCV 2020}, A.~Vedaldi, H.~Bischof,
  T.~Brox, and J.-M. Frahm, Eds.\hskip 1em plus 0.5em minus 0.4em\relax Cham:
  Springer International Publishing, 2020, pp. 402--419.

\bibitem{Moto}
Y.~Chen, Y.~Ge, W.~Tang, Y.~Li, Y.~Ge, M.~Ding, Y.~Shan, and X.~Liu, ``Moto:
  Latent motion token as the bridging language for learning robot manipulation
  from videos,'' \emph{arXiv preprint arXiv:2412.04445}, 2024.

\bibitem{UniVLA}
Q.~Bu, Y.~Yang, J.~Cai, S.~Gao, G.~Ren, M.~Yao, P.~Luo, and H.~Li, ``Univla:
  Learning to act anywhere with task-centric latent actions,'' \emph{arXiv
  preprint arXiv:2505.06111}, 2025.

\bibitem{UniSkill}
H.~Kim, J.~Kang, H.~Kang, M.~Cho, S.~J. Kim, and Y.~Lee, ``Uniskill: Imitating
  human videos via cross-embodiment skill representations,'' \emph{arXiv
  preprint arXiv:2505.08787}, 2025.

\bibitem{he2022mae}
K.~He, X.~Chen, S.~Xie, Y.~Li, P.~Doll\'ar, and R.~Girshick, ``Masked
  autoencoders are scalable vision learners,'' in \emph{Proceedings of the
  IEEE/CVF Conference on Computer Vision and Pattern Recognition (CVPR)}, June
  2022, pp. 16\,000--16\,009.

\bibitem{grauman2022ego4d}
K.~Grauman, A.~Westbury, E.~Byrne, Z.~Chavis, A.~Furnari, R.~Girdhar,
  J.~Hamburger, H.~Jiang, M.~Liu, X.~Liu \emph{et~al.}, ``Ego4d: Around the
  world in 3,000 hours of egocentric video,'' in \emph{Proceedings of the
  IEEE/CVF Conference on Computer Vision and Pattern Recognition (CVPR)}, June
  2022, pp. 18\,995--19\,012.

\bibitem{GR-2}
C.-L. Cheang, G.~Chen, Y.~Jing, T.~Kong, H.~Li, Y.~Li, Y.~Liu, H.~Wu, J.~Xu,
  Y.~Yang \emph{et~al.}, ``Gr-2: A generative video-language-action model with
  web-scale knowledge for robot manipulation,'' \emph{arXiv preprint
  arXiv:2410.06158}, 2024.

\bibitem{esser2021vqgan}
P.~Esser, R.~Rombach, and B.~Ommer, ``Taming transformers for high-resolution
  image synthesis,'' in \emph{Proceedings of the IEEE/CVF Conference on
  Computer Vision and Pattern Recognition (CVPR)}, June 2021, pp.
  12\,873--12\,883.

\bibitem{kingma2014cvae}
D.~P. Kingma, D.~J. Rezende, S.~Mohamed, and M.~Welling, ``Semi-supervised
  learning with deep generative models,'' in \emph{Advances in Neural
  Information Processing Systems (NeurIPS)}, Z.~Ghahramani, M.~Welling,
  C.~Cortes, N.~Lawrence, and K.~Weinberger, Eds., vol.~27.\hskip 1em plus
  0.5em minus 0.4em\relax Curran Associates, Inc., 2014.

\bibitem{li2025grmg}
P.~Li, H.~Wu, Y.~Huang, C.~Cheang, L.~Wang, and T.~Kong, ``Gr-mg: Leveraging
  partially-annotated data via multi-modal goal-conditioned policy,''
  \emph{IEEE Robotics and Automation Letters (RA-L)}, vol.~10, no.~2, pp.
  1912--1919, 2025.

\bibitem{cheang2025gr3technicalreport}
C.~Cheang, S.~Chen, Z.~Cui, Y.~Hu, L.~Huang, T.~Kong, H.~Li, Y.~Li, Y.~Liu,
  X.~Ma \emph{et~al.}, ``Gr-3 technical report,'' \emph{arXiv preprint
  arXiv:2507.15493}, 2025.

\bibitem{bai2025qwen25vl}
S.~Bai, K.~Chen, X.~Liu, J.~Wang, W.~Ge, S.~Song, K.~Dang, P.~Wang, S.~Wang,
  J.~Tang \emph{et~al.}, ``Qwen2.5-vl technical report,'' \emph{arXiv preprint
  arXiv:2502.13923}, 2025.

\bibitem{rombach2022stablediffusion}
R.~Rombach, A.~Blattmann, D.~Lorenz, P.~Esser, and B.~Ommer, ``High-resolution
  image synthesis with latent diffusion models,'' in \emph{Proceedings of the
  IEEE/CVF Conference on Computer Vision and Pattern Recognition (CVPR)}, June
  2022, pp. 10\,684--10\,695.

\bibitem{nichol2022pointe}
A.~Nichol, H.~Jun, P.~Dhariwal, P.~Mishkin, and M.~Chen, ``Point-e: A system
  for generating 3d point clouds from complex prompts,'' \emph{arXiv preprint
  arXiv:2212.08751}, 2022.

\bibitem{zheng2025flare}
R.~Zheng, J.~Wang, S.~Reed, J.~Bjorck, Y.~Fang, F.~Hu, J.~Jang, K.~Kundalia,
  Z.~Lin, L.~Magne \emph{et~al.}, ``Flare: Robot learning with implicit world
  modeling,'' \emph{arXiv preprint arXiv:2505.15659}, 2025.

\bibitem{WorldVLA}
J.~Cen, C.~Yu, H.~Yuan, Y.~Jiang, S.~Huang, J.~Guo, X.~Li, Y.~Song, H.~Luo,
  F.~Wang \emph{et~al.}, ``Worldvla: Towards autoregressive action world
  model,'' \emph{arXiv preprint arXiv:2506.21539}, 2025.

\bibitem{visaflow}
C.~Chen, Q.~Yang, X.~Xu, N.~Fazeli, and O.~Andersson, ``Visa-flow: Accelerating
  robot skill learning via large-scale video semantic action flow,''
  \emph{arXiv preprint arXiv:2505.01288}, 2025.

\bibitem{gibson}
\BIBentryALTinterwordspacing
J.~J. Gibson, ``The theory of affordances,'' in \emph{Perceiving, acting, and
  knowing: toward an ecological psychology}, J.~B. Robert E~Shaw, Ed.\hskip 1em
  plus 0.5em minus 0.4em\relax Hillsdale, N.J. : Lawrence Erlbaum Associates,
  1977, pp. pp.67--82. [Online]. Available:
  \url{https://hal.science/hal-00692033}
\BIBentrySTDinterwordspacing

\bibitem{huang2023voxposer}
W.~Huang, C.~Wang, R.~Zhang, Y.~Li, J.~Wu, and L.~Fei-Fei, ``Voxposer:
  Composable 3d value maps for robotic manipulation with language models,'' in
  \emph{Proceedings of The 7th Conference on Robot Learning (CoRL)}, ser.
  Proceedings of Machine Learning Research, J.~Tan, M.~Toussaint, and
  K.~Darvish, Eds., vol. 229.\hskip 1em plus 0.5em minus 0.4em\relax PMLR,
  06--09 Nov 2023, pp. 540--562.

\bibitem{minderer2022owlvit}
M.~Minderer, A.~Gritsenko, A.~Stone, M.~Neumann, D.~Weissenborn,
  A.~Dosovitskiy, A.~Mahendran, A.~Arnab, M.~Dehghani, Z.~Shen \emph{et~al.},
  ``Simple open-vocabulary object detection,'' in \emph{Computer Vision -- ECCV
  2022}, S.~Avidan, G.~Brostow, M.~Ciss{\'e}, G.~M. Farinella, and T.~Hassner,
  Eds.\hskip 1em plus 0.5em minus 0.4em\relax Cham: Springer Nature
  Switzerland, 2022, pp. 728--755.

\bibitem{kirillov2023sam}
A.~Kirillov, E.~Mintun, N.~Ravi, H.~Mao, C.~Rolland, L.~Gustafson, T.~Xiao,
  S.~Whitehead, A.~C. Berg, W.-Y. Lo \emph{et~al.}, ``Segment anything,'' in
  \emph{Proceedings of the IEEE/CVF International Conference on Computer Vision
  (ICCV)}, October 2023, pp. 4015--4026.

\bibitem{lee2024kagi}
O.~Y. Lee, A.~Xie, K.~Fang, K.~Pertsch, and C.~Finn, ``Affordance-guided
  reinforcement learning via visual prompting,'' \emph{arXiv preprint
  arXiv:2407.10341}, 2024.

\bibitem{rashid2023lerftogo}
\BIBentryALTinterwordspacing
A.~Rashid, S.~Sharma, C.~M. Kim, J.~Kerr, L.~Y. Chen, A.~Kanazawa, and
  K.~Goldberg, ``Language embedded radiance fields for zero-shot task-oriented
  grasping,'' in \emph{Proceedings of The 7th Conference on Robot Learning
  (CoRL)}, ser. Proceedings of Machine Learning Research, J.~Tan, M.~Toussaint,
  and K.~Darvish, Eds., vol. 229.\hskip 1em plus 0.5em minus 0.4em\relax PMLR,
  06--09 Nov 2023, pp. 178--200. [Online]. Available:
  \url{https://proceedings.mlr.press/v229/rashid23a.html}
\BIBentrySTDinterwordspacing

\bibitem{mildenhall2020nerf}
B.~Mildenhall, P.~P. Srinivasan, M.~Tancik, J.~T. Barron, R.~Ramamoorthi, and
  R.~Ng, ``Nerf: Representing scenes as neural radiance fields for view
  synthesis,'' in \emph{Computer Vision -- ECCV 2020}, A.~Vedaldi, H.~Bischof,
  T.~Brox, and J.-M. Frahm, Eds.\hskip 1em plus 0.5em minus 0.4em\relax Cham:
  Springer International Publishing, 2020, pp. 405--421.

\bibitem{caron2021dino}
M.~Caron, H.~Touvron, I.~Misra, H.~J\'egou, J.~Mairal, P.~Bojanowski, and
  A.~Joulin, ``Emerging properties in self-supervised vision transformers,'' in
  \emph{Proceedings of the IEEE/CVF International Conference on Computer Vision
  (ICCV)}, October 2021, pp. 9650--9660.

\bibitem{kerr2023lerf}
J.~Kerr, C.~M. Kim, K.~Goldberg, A.~Kanazawa, and M.~Tancik, ``Lerf: Language
  embedded radiance fields,'' in \emph{Proceedings of the IEEE/CVF
  International Conference on Computer Vision (ICCV)}, October 2023, pp.
  19\,729--19\,739.

\bibitem{fang2020graspnet}
H.-S. Fang, C.~Wang, M.~Gou, and C.~Lu, ``Graspnet-1billion: A large-scale
  benchmark for general object grasping,'' in \emph{Proceedings of the IEEE/CVF
  Conference on Computer Vision and Pattern Recognition (CVPR)}, June 2020.

\bibitem{shorinwa2024splatmover}
\BIBentryALTinterwordspacing
O.~Shorinwa, J.~Tucker, A.~Smith, A.~Swann, T.~Chen, R.~Firoozi, M.~D. Kennedy,
  and M.~Schwager, ``Splat-mover: Multi-stage, open-vocabulary robotic
  manipulation via editable gaussian splatting,'' in \emph{Proceedings of The
  8th Conference on Robot Learning (CoRL)}, ser. Proceedings of Machine
  Learning Research, P.~Agrawal, O.~Kroemer, and W.~Burgard, Eds., vol.
  270.\hskip 1em plus 0.5em minus 0.4em\relax PMLR, 06--09 Nov 2025, pp.
  4748--4770. [Online]. Available:
  \url{https://proceedings.mlr.press/v270/shorinwa25a.html}
\BIBentrySTDinterwordspacing

\bibitem{kerbl2023gaussiansplatting}
\BIBentryALTinterwordspacing
B.~Kerbl, G.~Kopanas, T.~Leimk{\"u}hler, and G.~Drettakis, ``3d gaussian
  splatting for real-time radiance field rendering,'' \emph{ACM Transactions on
  Graphics}, vol.~42, no.~4, July 2023. [Online]. Available:
  \url{https://repo-sam.inria.fr/fungraph/3d-gaussian-splatting/}
\BIBentrySTDinterwordspacing

\bibitem{bahl2023vrb}
S.~Bahl, R.~Mendonca, L.~Chen, U.~Jain, and D.~Pathak, ``Affordances from human
  videos as a versatile representation for robotics,'' in \emph{Proceedings of
  the IEEE/CVF Conference on Computer Vision and Pattern Recognition (CVPR)},
  June 2023, pp. 13\,778--13\,790.

\bibitem{damen2018epickitchens}
D.~Damen, H.~Doughty, G.~M. Farinella, S.~Fidler, A.~Furnari, E.~Kazakos,
  D.~Moltisanti, J.~Munro, T.~Perrett, W.~Price \emph{et~al.}, ``Scaling
  egocentric vision: The dataset,'' in \emph{Computer Vision -- ECCV 2018},
  V.~Ferrari, M.~Hebert, C.~Sminchisescu, and Y.~Weiss, Eds.\hskip 1em plus
  0.5em minus 0.4em\relax Cham: Springer International Publishing, 2018, pp.
  753--771.

\bibitem{damen2022epickitchens100}
\BIBentryALTinterwordspacing
D.~Damen, H.~Doughty, G.~M. Farinella, A.~Furnari, J.~Ma, E.~Kazakos,
  D.~Moltisanti, J.~Munro, T.~Perrett, W.~Price \emph{et~al.}, ``Rescaling
  egocentric vision: Collection, pipeline and challenges for
  epic-kitchens-100,'' \emph{International Journal of Computer Vision (IJCV)},
  vol. 130, p. 33^^e2^^80^^9355, 2022. [Online]. Available:
  \url{https://doi.org/10.1007/s11263-021-01531-2}
\BIBentrySTDinterwordspacing

\bibitem{shan2020hod}
D.~Shan, J.~Geng, M.~Shu, and D.~F. Fouhey, ``Understanding human hands in
  contact at internet scale,'' in \emph{Proceedings of the IEEE/CVF Conference
  on Computer Vision and Pattern Recognition (CVPR)}, June 2020.

\bibitem{srirama2024hrp}
M.~K. Srirama, S.~Dasari, S.~Bahl, and A.~Gupta, ``Hrp: Human affordances for
  robotic pre-training,'' in \emph{Proceedings of Robotics: Science and Systems
  (RSS)}, Delft, Netherlands, 2024.

\bibitem{VidBot}
H.~Chen, B.~Sun, A.~Zhang, M.~Pollefeys, and S.~Leutenegger, ``Vidbot: Learning
  generalizable 3d actions from in-the-wild 2d human videos for zero-shot
  robotic manipulation,'' \emph{arXiv preprint arXiv:2503.07135}, 2025.

\bibitem{yuan2025robopoint}
W.~Yuan, J.~Duan, V.~Blukis, W.~Pumacay, R.~Krishna, A.~Murali, A.~Mousavian,
  and D.~Fox, ``Robopoint: A vision-language model for spatial affordance
  prediction in robotics,'' in \emph{Proceedings of The 8th Conference on Robot
  Learning (CoRL)}, ser. Proceedings of Machine Learning Research, P.~Agrawal,
  O.~Kroemer, and W.~Burgard, Eds., vol. 270.\hskip 1em plus 0.5em minus
  0.4em\relax PMLR, 06--09 Nov 2025, pp. 4005--4020.

\bibitem{huang2025roboground}
H.~Huang, X.~Chen, Y.~Chen, H.~Li, X.~Han, Z.~Wang, T.~Wang, J.~Pang, and
  Z.~Zhao, ``Roboground: Robotic manipulation with grounded vision-language
  priors,'' in \emph{Proceedings of the IEEE/CVF Conference on Computer Vision
  and Pattern Recognition (CVPR)}, June 2025, pp. 22\,540--22\,550.

\bibitem{RT-Affordance}
S.~Nasiriany, S.~Kirmani, T.~Ding, L.~Smith, Y.~Zhu, D.~Driess, D.~Sadigh, and
  T.~Xiao, ``Rt-affordance: Affordances are versatile intermediate
  representations for robot manipulation,'' \emph{arXiv preprint
  arXiv:2411.02704}, 2024.

\bibitem{A_0}
R.~Xu, J.~Zhang, M.~Guo, Y.~Wen, H.~Yang, M.~Lin, J.~Huang, Z.~Li, K.~Zhang,
  L.~Wang \emph{et~al.}, ``A0: An affordance-aware hierarchical model for
  general robotic manipulation,'' \emph{arXiv preprint arXiv:2504.12636}, 2025.

\bibitem{ji2025robobrain}
Y.~Ji, H.~Tan, J.~Shi, X.~Hao, Y.~Zhang, H.~Zhang, P.~Wang, M.~Zhao, Y.~Mu,
  P.~An \emph{et~al.}, ``Robobrain: A unified brain model for robotic
  manipulation from abstract to concrete,'' in \emph{Proceedings of the
  IEEE/CVF Conference on Computer Vision and Pattern Recognition (CVPR)}, June
  2025, pp. 1724--1734.

\bibitem{Chain-of-Affordance}
J.~Li, Y.~Zhu, Z.~Tang, J.~Wen, M.~Zhu, X.~Liu, C.~Li, R.~Cheng, Y.~Peng, and
  F.~Feng, ``Improving vision-language-action models via chain-of-affordance,''
  \emph{arXiv preprint arXiv:2412.20451}, 2024.

\bibitem{he2016resnet}
K.~He, X.~Zhang, S.~Ren, and J.~Sun, ``Deep residual learning for image
  recognition,'' in \emph{Proceedings of the IEEE Conference on Computer Vision
  and Pattern Recognition (CVPR)}, June 2016.

\bibitem{deng2009imagenet}
J.~Deng, W.~Dong, R.~Socher, L.-J. Li, K.~Li, and L.~Fei-Fei, ``Imagenet: A
  large-scale hierarchical image database,'' in \emph{Proceedings of the IEEE
  Conference on Computer Vision and Pattern Recognition (CVPR)}, 2009, pp.
  248--255.

\bibitem{russakovsky2015imagenet1k}
O.~Russakovsky, J.~Deng, H.~Su, J.~Krause, S.~Satheesh, S.~Ma, Z.~Huang,
  A.~Karpathy, A.~Khosla, M.~Bernstein \emph{et~al.}, ``Imagenet large scale
  visual recognition challenge,'' \emph{International Journal of Computer
  Vision (IJCV)}, vol. 115, no.~3, pp. 211--252, 2015.

\bibitem{schuhmann2021laion400m}
C.~Schuhmann, R.~Vencu, R.~Beaumont, R.~Kaczmarczyk, C.~Mullis, J.~Jitsev, and
  A.~Komatsuzaki, ``Laion-400m: Open dataset of clip-filtered 400 million
  image-text pairs,'' in \emph{Proceedings of Neurips Data-Centric AI
  Workshop}, 2021.

\bibitem{schuhmann2022laion5b}
C.~Schuhmann, R.~Beaumont, R.~Vencu, C.~Gordon, R.~Wightman, M.~Cherti,
  T.~Coombes, A.~Katta, C.~Mullis, M.~Wortsman \emph{et~al.}, ``Laion-5b: An
  open large-scale dataset for training next generation image-text models,'' in
  \emph{Advances in Neural Information Processing Systems (NeurIPS)},
  S.~Koyejo, S.~Mohamed, A.~Agarwal, D.~Belgrave, K.~Cho, and A.~Oh, Eds.,
  vol.~35.\hskip 1em plus 0.5em minus 0.4em\relax Curran Associates, Inc.,
  2022, pp. 25\,278--25\,294.

\bibitem{cherti2023openclip}
M.~Cherti, R.~Beaumont, R.~Wightman, M.~Wortsman, G.~Ilharco, C.~Gordon,
  C.~Schuhmann, L.~Schmidt, and J.~Jitsev, ``Reproducible scaling laws for
  contrastive language-image learning,'' in \emph{Proceedings of the IEEE/CVF
  Conference on Computer Vision and Pattern Recognition (CVPR)}, 2023, pp.
  2818--2829.

\bibitem{sun2023evaclip}
Q.~Sun, Y.~Fang, L.~Wu, X.~Wang, and Y.~Cao, ``Eva-clip: Improved training
  techniques for clip at scale,'' \emph{arXiv preprint arXiv:2303.15389}, 2023.

\bibitem{alayrac2022flamingo}
J.-B. Alayrac, J.~Donahue, P.~Luc, A.~Miech, I.~Barr, Y.~Hasson, K.~Lenc,
  A.~Mensch, K.~Millican, M.~Reynolds \emph{et~al.}, ``Flamingo: a visual
  language model for few-shot learning,'' in \emph{Advances in Neural
  Information Processing Systems (NeurIPS)}, S.~Koyejo, S.~Mohamed, A.~Agarwal,
  D.~Belgrave, K.~Cho, and A.~Oh, Eds., vol.~35.\hskip 1em plus 0.5em minus
  0.4em\relax Curran Associates, Inc., 2022, pp. 23\,716--23\,736.

\bibitem{fu2025orion}
H.~Fu, D.~Zhang, Z.~Zhao, J.~Cui, D.~Liang, C.~Zhang, D.~Zhang, H.~Xie,
  B.~Wang, and X.~Bai, ``Orion: A holistic end-to-end autonomous driving
  framework by vision-language instructed action generation,'' \emph{arXiv
  preprint arXiv:2503.19755}, 2025.

\bibitem{liu2024groundingdino}
S.~Liu, Z.~Zeng, T.~Ren, F.~Li, H.~Zhang, J.~Yang, Q.~Jiang, C.~Li, J.~Yang,
  H.~Su \emph{et~al.}, ``Grounding dino: Marrying dino with~grounded
  pre-training for~open-set object detection,'' in \emph{Computer Vision --
  ECCV 2024}, A.~Leonardis, E.~Ricci, S.~Roth, O.~Russakovsky, T.~Sattler, and
  G.~Varol, Eds.\hskip 1em plus 0.5em minus 0.4em\relax Cham: Springer Nature
  Switzerland, 2025, pp. 38--55.

\bibitem{zhou2022detic}
X.~Zhou, R.~Girdhar, A.~Joulin, P.~Kr{\"a}henb{\"u}hl, and I.~Misra,
  ``Detecting twenty-thousand classes using image-level supervision,'' in
  \emph{Computer Vision -- ECCV 2022}, S.~Avidan, G.~Brostow, M.~Ciss{\'e},
  G.~M. Farinella, and T.~Hassner, Eds.\hskip 1em plus 0.5em minus 0.4em\relax
  Cham: Springer Nature Switzerland, 2022, pp. 350--368.

\bibitem{cheng2024cutie}
H.~K. Cheng, S.~W. Oh, B.~Price, J.-Y. Lee, and A.~Schwing, ``Putting the
  object back into video object segmentation,'' in \emph{Proceedings of the
  IEEE/CVF Conference on Computer Vision and Pattern Recognition (CVPR)}, June
  2024, pp. 3151--3161.

\bibitem{reimers2019sentencebert}
\BIBentryALTinterwordspacing
N.~Reimers and I.~Gurevych, ``Sentence-bert: Sentence embeddings using siamese
  bert-networks,'' in \emph{Proceedings of the 2019 Conference on Empirical
  Methods in Natural Language Processing and the 9th International Joint
  Conference on Natural Language Processing (EMNLP-IJCNLP)}, K.~Inui, J.~Jiang,
  V.~Ng, and X.~Wan, Eds.\hskip 1em plus 0.5em minus 0.4em\relax Hong Kong,
  China: Association for Computational Linguistics, November 2019, pp.
  3982--3992. [Online]. Available: \url{https://aclanthology.org/D19-1410/}
\BIBentrySTDinterwordspacing

\bibitem{sanh2020distilbert}
V.~Sanh, L.~Debut, J.~Chaumond, and T.~Wolf, ``Distilbert, a distilled version
  of bert: smaller, faster, cheaper and lighter,'' \emph{arXiv preprint
  arXiv:1910.01108}, 2020.

\bibitem{chiang2023vicuna}
\BIBentryALTinterwordspacing
W.-L. Chiang, Z.~Li, Z.~Lin, Y.~Sheng, Z.~Wu, H.~Zhang, L.~Zheng, S.~Zhuang,
  Y.~Zhuang, J.~E. Gonzalez \emph{et~al.}, ``Vicuna: An open-source chatbot
  impressing gpt-4 with 90\%\textasteriskcentered chatgpt quality,'' March
  2023. [Online]. Available: \url{https://lmsys.org/blog/2023-03-30-vicuna/}
\BIBentrySTDinterwordspacing

\bibitem{gemmateam2024gemma}
G.~Team, T.~Mesnard, C.~Hardin, R.~Dadashi, S.~Bhupatiraju, S.~Pathak,
  L.~Sifre, M.~Rivi\`{e}re, M.~S. Kale, J.~Love \emph{et~al.}, ``Gemma: Open
  models based on gemini research and technology,'' \emph{arXiv preprint
  arXiv:2403.08295}, 2024.

\bibitem{yang2024qwen2}
A.~Yang, B.~Yang, B.~Hui, B.~Zheng, B.~Yu, C.~Zhou, C.~Li, C.~Li, D.~Liu,
  F.~Huang \emph{et~al.}, ``Qwen2 technical report,'' \emph{arXiv preprint
  arXiv:2407.10671}, 2024.

\bibitem{javaheripi2023phi}
M.~Javaheripi, S.~Bubeck, M.~Abdin, J.~Aneja, S.~Bubeck, C.~C.~T. Mendes,
  W.~Chen, A.~Del~Giorno, R.~Eldan, S.~Gopi \emph{et~al.}, ``Phi-2: The
  surprising power of small language models,'' \emph{Microsoft Research Blog},
  2023.

\bibitem{allal2025smollm2}
L.~B. Allal, A.~Lozhkov, E.~Bakouch, G.~M. Bl\'{a}zquez, G.~Penedo,
  L.~Tunstall, A.~Marafioti, H.~Kydl\'{\i}\v{c}ek, A.~P. Lajar\'{\i}n,
  V.~Srivastav \emph{et~al.}, ``Smollm2: When smol goes big -- data-centric
  training of a small language model,'' \emph{arXiv preprint arXiv:2502.02737},
  2025.

\bibitem{black2022gptneox20b}
\BIBentryALTinterwordspacing
S.~Black, S.~Biderman, E.~Hallahan, Q.~Anthony, L.~Gao, L.~Golding, H.~He,
  C.~Leahy, K.~McDonell, J.~Phang \emph{et~al.}, ``Gpt-neox-20b: An open-source
  autoregressive language model,'' in \emph{Proceedings of the ACL Workshop on
  Challenges \& Perspectives in Creating Large Language Models}, 2022.
  [Online]. Available: \url{https://arxiv.org/abs/2204.06745}
\BIBentrySTDinterwordspacing

\bibitem{biderman2023pythia}
\BIBentryALTinterwordspacing
S.~Biderman, H.~Schoelkopf, Q.~G. Anthony, H.~Bradley, K.~O'Brien, E.~Hallahan,
  M.~A. Khan, S.~Purohit, U.~S. Prashanth, E.~Raff \emph{et~al.}, ``Pythia: A
  suite for analyzing large language models across training and scaling,'' in
  \emph{Proceedings of the 40th International Conference on Machine Learning
  (ICML)}, ser. Proceedings of Machine Learning Research, A.~Krause,
  E.~Brunskill, K.~Cho, B.~Engelhardt, S.~Sabato, and J.~Scarlett, Eds., vol.
  202.\hskip 1em plus 0.5em minus 0.4em\relax PMLR, 23--29 Jul 2023, pp.
  2397--2430. [Online]. Available:
  \url{https://proceedings.mlr.press/v202/biderman23a.html}
\BIBentrySTDinterwordspacing

\bibitem{zhao2025cotvla}
Q.~Zhao, Y.~Lu, M.~J. Kim, Z.~Fu, Z.~Zhang, Y.~Wu, Z.~Li, Q.~Ma, S.~Han,
  C.~Finn \emph{et~al.}, ``Cot-vla: Visual chain-of-thought reasoning for
  vision-language-action models,'' in \emph{Proceedings of the IEEE/CVF
  Conference on Computer Vision and Pattern Recognition (CVPR)}, June 2025, pp.
  1702--1713.

\bibitem{OpenVLA-OFT}
M.~J. Kim, C.~Finn, and P.~Liang, ``Fine-tuning vision-language-action models:
  Optimizing speed and success,'' \emph{arXiv preprint arXiv:2502.19645}, 2025.

\bibitem{hochreiter1997lstm}
S.~Hochreiter and J.~Schmidhuber, ``Long short-term memory,'' \emph{Neural
  Computation}, vol.~9, no.~8, pp. 1735--1780, 1997.

\bibitem{song2021ddim}
J.~Song, C.~Meng, and S.~Ermon, ``Denoising diffusion implicit models,'' in
  \emph{International Conference on Learning Representations (ICLR)}, 2021.

\bibitem{TUDP}
Y.~Niu, S.~Zhou, Y.~Li, Y.~Den, and L.~Wang, ``Time-unified diffusion policy
  with action discrimination for robotic manipulation,'' \emph{arXiv preprint
  arXiv:2506.09422}, 2025.

\bibitem{SpatialVLA}
D.~Qu, H.~Song, Q.~Chen, Y.~Yao, X.~Ye, Y.~Ding, Z.~Wang, J.~Gu, B.~Zhao,
  D.~Wang \emph{et~al.}, ``Spatialvla: Exploring spatial representations for
  visual-language-action model,'' \emph{arXiv preprint arXiv:2501.15830}, 2025.

\bibitem{ForceVLA}
J.~Yu, H.~Liu, Q.~Yu, J.~Ren, C.~Hao, H.~Ding, G.~Huang, G.~Huang, Y.~Song,
  P.~Cai \emph{et~al.}, ``Forcevla: Enhancing vla models with a force-aware moe
  for contact-rich manipulation,'' \emph{arXiv preprint arXiv:2505.22159},
  2025.

\bibitem{iManip}
Z.~Zheng, J.-F. Cai, X.-M. Wu, Y.-L. Wei, Y.-M. Tang, and W.-S. Zheng,
  ``imanip: Skill-incremental learning for robotic manipulation,'' \emph{arXiv
  preprint arXiv:2503.07087}, 2025.

\bibitem{yang2024extreme}
J.~Yang, C.~Glossop, A.~Bhorkar, D.~Shah, Q.~Vuong, C.~Finn, D.~Sadigh, and
  S.~Levine, ``Pushing the limits of cross-embodiment learning for manipulation
  and navigation,'' \emph{arXiv preprint arXiv:2402.19432}, 2024.

\bibitem{zheng2025uniact}
J.~Zheng, J.~Li, D.~Liu, Y.~Zheng, Z.~Wang, Z.~Ou, Y.~Liu, J.~Liu, Y.-Q. Zhang,
  and X.~Zhan, ``Universal actions for enhanced embodied foundation models,''
  in \emph{Proceedings of the IEEE/CVF Conference on Computer Vision and
  Pattern Recognition (CVPR)}, June 2025, pp. 22\,508--22\,519.

\bibitem{jiang2025solami}
J.~Jiang, W.~Xiao, Z.~Lin, H.~Zhang, T.~Ren, Y.~Gao, Z.~Lin, Z.~Cai, L.~Yang,
  and Z.~Liu, ``Solami: Social vision-language-action modeling for immersive
  interaction with 3d autonomous characters,'' \emph{arXiv preprint
  arXiv:2412.00174}, 2024.

\bibitem{jones24fuse}
J.~Jones, O.~Mees, C.~Sferrazza, K.~Stachowicz, P.~Abbeel, and S.~Levine,
  ``Beyond sight: Finetuning generalist robot policies with heterogeneous
  sensors via language grounding,'' \emph{arXiv preprint arXiv:2501.04693},
  2025.

\bibitem{zhao2025vlas}
W.~Zhao, P.~Ding, Z.~Min, Z.~Gong, S.~Bai, H.~Zhao, and D.~Wang, ``Vlas:
  Vision-language-action model with speech instructions for customized robot
  manipulation,'' in \emph{International Conference on Learning Representations
  (ICLR)}, 2025.

\bibitem{MultiGen}
R.~Wang, H.~Geng, T.~Li, F.~Wang, G.~Anumanchipalli, P.~Wu, T.~Darrell, B.~Li,
  P.~Abbeel, J.~Malik \emph{et~al.}, ``Multigen: Using multimodal generation in
  simulation to learn multimodal policies in real,'' \emph{arXiv preprint
  arXiv:2507.02864}, 2025.

\bibitem{zhang2024speechtokenizer}
X.~Zhang, D.~Zhang, S.~Li, Y.~Zhou, and X.~Qiu, ``Speechtokenizer: Unified
  speech tokenizer for speech language models,'' in \emph{International
  Conference on Learning Representations (ICLR)}, 2024.

\bibitem{gong2021ast}
Y.~Gong, Y.-A. Chung, and J.~Glass, ``Ast: Audio spectrogram transformer,'' in
  \emph{Proc. Interspeech 2021}, 2021, pp. 571--575.

\bibitem{radford2023whisper}
\BIBentryALTinterwordspacing
A.~Radford, J.~W. Kim, T.~Xu, G.~Brockman, C.~Mcleavey, and I.~Sutskever,
  ``Robust speech recognition via large-scale weak supervision,'' in
  \emph{Proceedings of the 40th International Conference on Machine Learning
  (ICML)}, ser. Proceedings of Machine Learning Research, A.~Krause,
  E.~Brunskill, K.~Cho, B.~Engelhardt, S.~Sabato, and J.~Scarlett, Eds., vol.
  202.\hskip 1em plus 0.5em minus 0.4em\relax PMLR, 23--29 Jul 2023, pp.
  28\,492--28\,518. [Online]. Available:
  \url{https://proceedings.mlr.press/v202/radford23a.html}
\BIBentrySTDinterwordspacing

\bibitem{borsos2023soundstorm}
Z.~Borsos, M.~Sharifi, D.~Vincent, E.~Kharitonov, N.~Zeghidour, and
  M.~Tagliasacchi, ``Soundstorm: Efficient parallel audio generation,''
  \emph{arXiv preprint arXiv:2305.09636}, 2023.

\bibitem{RoboNurse-VLA}
S.~Li, J.~Wang, R.~Dai, W.~Ma, W.~Y. Ng, Y.~Hu, and Z.~Li, ``Robonurse-vla:
  Robotic scrub nurse system based on vision-language-action model,''
  \emph{arXiv preprint arXiv:2409.19590}, 2024.

\bibitem{TLA}
P.~Hao, C.~Zhang, D.~Li, X.~Cao, X.~Hao, S.~Cui, and S.~Wang, ``Tla:
  Tactile-language-action model for contact-rich manipulation,'' \emph{arXiv
  preprint arXiv:2503.08548}, 2025.

\bibitem{VTLA}
C.~Zhang, P.~Hao, X.~Cao, X.~Hao, S.~Cui, and S.~Wang, ``Vtla:
  Vision-tactile-language-action model with preference learning for insertion
  manipulation,'' \emph{arXiv preprint arXiv:2505.09577}, 2025.

\bibitem{Tactile-VLA}
J.~Huang, S.~Wang, F.~Lin, Y.~Hu, C.~Wen, and Y.~Gao, ``Tactile-vla: Unlocking
  vision-language-action model's physical knowledge for tactile
  generalization,'' \emph{arXiv preprint arXiv:2507.09160}, 2025.

\bibitem{lambeta2020digit}
M.~Lambeta, P.-W. Chou, S.~Tian, B.~Yang, B.~Maloon, V.~R. Most, D.~Stroud,
  R.~Santos, A.~Byagowi, G.~Kammerer \emph{et~al.}, ``Digit: A novel design for
  a low-cost compact high-resolution tactile sensor with application to in-hand
  manipulation,'' \emph{IEEE Robotics and Automation Letters (RA-L)}, vol.~5,
  no.~3, pp. 3838--3845, 2020.

\bibitem{zhang2024gelstereo2}
C.~Zhang, S.~Cui, S.~Wang, J.~Hu, Y.~Cai, R.~Wang, and Y.~Wang, ``Gelstereo
  2.0: An improved gelstereo sensor with multimedium refractive stereo
  calibration,'' \emph{IEEE Transactions on Industrial Electronics}, vol.~71,
  no.~7, pp. 7452--7462, 2024.

\bibitem{fu2024tvl}
\BIBentryALTinterwordspacing
L.~Fu, G.~Datta, H.~Huang, W.~C.-H. Panitch, J.~Drake, J.~Ortiz, M.~Mukadam,
  M.~Lambeta, R.~Calandra, and K.~Goldberg, ``A touch, vision, and language
  dataset for multimodal alignment,'' in \emph{Proceedings of the 41st
  International Conference on Machine Learning (ICML)}, ser. Proceedings of
  Machine Learning Research, R.~Salakhutdinov, Z.~Kolter, K.~Heller, A.~Weller,
  N.~Oliver, J.~Scarlett, and F.~Berkenkamp, Eds., vol. 235.\hskip 1em plus
  0.5em minus 0.4em\relax PMLR, 21--27 Jul 2024, pp. 14\,080--14\,101.
  [Online]. Available: \url{https://proceedings.mlr.press/v235/fu24b.html}
\BIBentrySTDinterwordspacing

\bibitem{yang2024depthanything}
L.~Yang, B.~Kang, Z.~Huang, X.~Xu, J.~Feng, and H.~Zhao, ``Depth anything:
  Unleashing the power of large-scale unlabeled data,'' in \emph{Proceedings of
  the IEEE/CVF Conference on Computer Vision and Pattern Recognition (CVPR)},
  June 2024, pp. 10\,371--10\,381.

\bibitem{bhat2023zoedepth}
S.~F. Bhat, R.~Birkl, D.~Wofk, P.~Wonka, and M.~M\"{u}ller, ``Zoedepth:
  Zero-shot transfer by combining relative and metric depth,'' \emph{arXiv
  preprint arXiv:2302.12288}, 2023.

\bibitem{li2025hamster}
Y.~Li, Y.~Deng, J.~Zhang, J.~Jang, M.~Memmel, C.~R. Garrett, F.~Ramos, D.~Fox,
  A.~Li, A.~Gupta \emph{et~al.}, ``Hamster: Hierarchical action models for
  open-world robot manipulation,'' in \emph{International Conference on
  Learning Representations (ICLR)}, 2025.

\bibitem{RationalVLA}
W.~Song, J.~Chen, W.~Li, X.~He, H.~Zhao, C.~Cui, P.~D.~S. Su, F.~Tang,
  X.~Cheng, D.~Wang \emph{et~al.}, ``Rationalvla: A rational
  vision-language-action model with dual system,'' \emph{arXiv preprint
  arXiv:2506.10826}, 2025.

\bibitem{OpenHelix}
C.~Cui, P.~Ding, W.~Song, S.~Bai, X.~Tong, Z.~Ge, R.~Suo, W.~Zhou, Y.~Liu,
  B.~Jia \emph{et~al.}, ``Openhelix: A short survey, empirical analysis, and
  open-source dual-system vla model for robotic manipulation,'' \emph{arXiv
  preprint arXiv:2505.03912}, 2025.

\bibitem{ke2025threeddiffuseractor}
T.-W. Ke, N.~Gkanatsios, and K.~Fragkiadaki, ``3d diffuser actor: Policy
  diffusion with 3d scene representations,'' in \emph{Proceedings of The 8th
  Conference on Robot Learning (CoRL)}, ser. Proceedings of Machine Learning
  Research, P.~Agrawal, O.~Kroemer, and W.~Burgard, Eds., vol. 270.\hskip 1em
  plus 0.5em minus 0.4em\relax PMLR, 06--09 Nov 2025, pp. 1949--1974.

\bibitem{Evo-0}
T.~Lin, G.~Li, Y.~Zhong, Y.~Zou, and B.~Zhao, ``Evo-0: Vision-language-action
  model with implicit spatial understanding,'' \emph{arXiv preprint
  arXiv:2507.00416}, 2025.

\bibitem{wang2025vggt}
J.~Wang, M.~Chen, N.~Karaev, A.~Vedaldi, C.~Rupprecht, and D.~Novotny, ``Vggt:
  Visual geometry grounded transformer,'' in \emph{Proceedings of the IEEE/CVF
  Conference on Computer Vision and Pattern Recognition (CVPR)}, 2025.

\bibitem{RoboMM}
F.~Yan, F.~Liu, L.~Zheng, Y.~Zhong, Y.~Huang, Z.~Guan, C.~Feng, and L.~Ma,
  ``Robomm: All-in-one multimodal large model for robotic manipulation,''
  \emph{arXiv preprint arXiv:2412.07215}, 2024.

\bibitem{fang2025samact}
H.~Fang, M.~Grotz, W.~Pumacay, Y.~R. Wang, D.~Fox, R.~Krishna, and J.~Duan,
  ``Sam2act: Integrating visual foundation model with a memory architecture for
  robotic manipulation,'' \emph{arXiv preprint arXiv:2501.18564}, 2025.

\bibitem{goyal2024rvt2}
A.~Goyal, V.~Blukis, J.~Xu, Y.~Guo, Y.-W. Chao, and D.~Fox, ``{RVT-2: Learning
  Precise Manipulation from Few Demonstrations},'' in \emph{Proceedings of
  Robotics: Science and Systems (RSS)}, Delft, Netherlands, July 2024.

\bibitem{OG-VLA}
I.~Singh, A.~Goyal, S.~Birchfield, D.~Fox, A.~Garg, and V.~Blukis, ``Og-vla:
  3d-aware vision language action model via orthographic image generation,''
  \emph{arXiv preprint arXiv:2506.01196}, 2025.

\bibitem{BridgeVLA}
P.~Li, Y.~Chen, H.~Wu, X.~Ma, X.~Wu, Y.~Huang, L.~Wang, T.~Kong, and T.~Tan,
  ``Bridgevla: Input-output alignment for efficient 3d manipulation learning
  with vision-language models,'' \emph{arXiv preprint arXiv:2506.07961}, 2025.

\bibitem{OccLLaMA}
J.~Wei, S.~Yuan, P.~Li, Q.~Hu, Z.~Gan, and W.~Ding, ``Occllama: An
  occupancy-language-action generative world model for autonomous driving,''
  \emph{arXiv preprint arXiv:2409.03272}, 2024.

\bibitem{OpenDriveVLA}
X.~Zhou, X.~Han, F.~Yang, Y.~Ma, and A.~C. Knoll, ``Opendrivevla: Towards
  end-to-end autonomous driving with large vision language action model,''
  \emph{arXiv preprint arXiv:2503.23463}, 2025.

\bibitem{peng2020unet3d}
S.~Peng, M.~Niemeyer, L.~Mescheder, M.~Pollefeys, and A.~Geiger,
  ``Convolutional occupancy networks,'' in \emph{Computer Vision -- ECCV 2020},
  A.~Vedaldi, H.~Bischof, T.~Brox, and J.-M. Frahm, Eds.\hskip 1em plus 0.5em
  minus 0.4em\relax Cham: Springer International Publishing, 2020, pp.
  523--540.

\bibitem{qi2017pointnet}
C.~R. Qi, H.~Su, K.~Mo, and L.~J. Guibas, ``Pointnet: Deep learning on point
  sets for 3d classification and segmentation,'' in \emph{Proceedings of the
  IEEE Conference on Computer Vision and Pattern Recognition (CVPR)}, July
  2017.

\bibitem{qi2017pointnetpp}
C.~R. Qi, L.~Yi, H.~Su, and L.~J. Guibas, ``Pointnet++: Deep hierarchical
  feature learning on point sets in a metric space,'' in \emph{Advances in
  Neural Information Processing Systems (NeurIPS)}, I.~Guyon, U.~V. Luxburg,
  S.~Bengio, H.~Wallach, R.~Fergus, S.~Vishwanathan, and R.~Garnett, Eds.,
  vol.~30.\hskip 1em plus 0.5em minus 0.4em\relax Curran Associates, Inc.,
  2017.

\bibitem{qian2022pointnext}
G.~Qian, Y.~Li, H.~Peng, J.~Mai, H.~Hammoud, M.~Elhoseiny, and B.~Ghanem,
  ``Pointnext: Revisiting pointnet++ with improved training and scaling
  strategies,'' in \emph{Advances in Neural Information Processing Systems
  (NeurIPS)}, S.~Koyejo, S.~Mohamed, A.~Agarwal, D.~Belgrave, K.~Cho, and
  A.~Oh, Eds., vol.~35.\hskip 1em plus 0.5em minus 0.4em\relax Curran
  Associates, Inc., 2022, pp. 23\,192--23\,204.

\bibitem{zhou2023uni3d}
\BIBentryALTinterwordspacing
J.~Zhou, J.~Wang, B.~Ma, Y.-S. Liu, T.~Huang, and X.~Wang, ``Uni3d: Exploring
  unified 3d representation at scale,'' in \emph{International Conference on
  Learning Representations (ICLR)}, 2024. [Online]. Available:
  \url{https://openreview.net/forum?id=wcaE4Dfgt8}
\BIBentrySTDinterwordspacing

\bibitem{SoFar}
Z.~Qi, W.~Zhang, Y.~Ding, R.~Dong, X.~Yu, J.~Li, L.~Xu, B.~Li, X.~He, G.~Fan
  \emph{et~al.}, ``Sofar: Language-grounded orientation bridges spatial
  reasoning and object manipulation,'' \emph{arXiv preprint arXiv:2502.13143},
  2025.

\bibitem{LMM-3DP}
Y.~Li, G.~Yan, A.~Macaluso, M.~Ji, X.~Zou, and X.~Wang, ``Integrating lmm
  planners and 3d skill policies for generalizable manipulation,'' \emph{arXiv
  preprint arXiv:2501.18733}, 2025.

\bibitem{yuan2025generalflow}
C.~Yuan, C.~Wen, T.~Zhang, and Y.~Gao, ``General flow as foundation affordance
  for scalable robot learning,'' in \emph{Proceedings of The 8th Conference on
  Robot Learning (CoRL)}, ser. Proceedings of Machine Learning Research,
  P.~Agrawal, O.~Kroemer, and W.~Burgard, Eds., vol. 270.\hskip 1em plus 0.5em
  minus 0.4em\relax PMLR, 06--09 Nov 2025, pp. 1541--1566.

\bibitem{DexTOG}
J.~Zhang, W.~Xu, Z.~Yu, P.~Xie, T.~Tang, and C.~Lu, ``Dextog: Learning
  task-oriented dexterous grasp with language,'' \emph{arXiv preprint
  arXiv:2504.04573}, 2025.

\bibitem{guo2021pct}
\BIBentryALTinterwordspacing
M.-H. Guo, J.-X. Cai, Z.-N. Liu, T.-J. Mu, R.~R. Martin, and S.-M. Hu, ``Pct:
  Point cloud transformer,'' \emph{Computational Visual Media}, vol.~7, no.~2,
  p. 187^^e2^^80^^93199, Apr 2021. [Online]. Available:
  \url{http://dx.doi.org/10.1007/s41095-021-0229-5}
\BIBentrySTDinterwordspacing

\bibitem{li2018pointcnn}
Y.~Li, R.~Bu, M.~Sun, W.~Wu, X.~Di, and B.~Chen, ``Pointcnn: Convolution on
  x-transformed points,'' in \emph{Advances in Neural Information Processing
  Systems (NeurIPS)}, S.~Bengio, H.~Wallach, H.~Larochelle, K.~Grauman,
  N.~Cesa-Bianchi, and R.~Garnett, Eds., vol.~31.\hskip 1em plus 0.5em minus
  0.4em\relax Curran Associates, Inc., 2018.

\bibitem{niu2025arm4r}
D.~Niu, Y.~Sharma, H.~Xue, G.~Biamby, J.~Zhang, Z.~Ji, T.~Darrell, and
  R.~Herzig, ``Pre-training auto-regressive robotic models with 4d
  representations,'' \emph{arXiv preprint arXiv:2502.13142}, 2025.

\bibitem{pavlakos2019smplx}
G.~Pavlakos, V.~Choutas, N.~Ghorbani, T.~Bolkart, A.~A.~A. Osman, D.~Tzionas,
  and M.~J. Black, ``Expressive body capture: 3d hands, face, and body from a
  single image,'' in \emph{Proceedings of the IEEE/CVF Conference on Computer
  Vision and Pattern Recognition (CVPR)}, June 2019.

\bibitem{jiang2023motiongpt}
B.~Jiang, X.~Chen, W.~Liu, J.~Yu, G.~Yu, and T.~Chen, ``Motiongpt: Human motion
  as a foreign language,'' in \emph{Advances in Neural Information Processing
  Systems (NeurIPS)}, A.~Oh, T.~Naumann, A.~Globerson, K.~Saenko, M.~Hardt, and
  S.~Levine, Eds., vol.~36.\hskip 1em plus 0.5em minus 0.4em\relax Curran
  Associates, Inc., 2023, pp. 20\,067--20\,079.

\bibitem{li2025atomic}
D.~Li, B.~Peng, C.~Li, N.~Qiao, Q.~Zheng, L.~Sun, Y.~Qin, B.~Li, Y.~Luan, B.~Wu
  \emph{et~al.}, ``An atomic skill library construction method for
  data-efficient embodied manipulation,'' \emph{arXiv preprint
  arXiv:2501.15068}, 2025.

\bibitem{HiRobot}
L.~X. Shi, B.~Ichter, M.~Equi, L.~Ke, K.~Pertsch, Q.~Vuong, J.~Tanner,
  A.~Walling, H.~Wang, N.~Fusai \emph{et~al.}, ``Hi robot: Open-ended
  instruction following with hierarchical vision-language-action models,''
  \emph{arXiv preprint arXiv:2502.19417}, 2025.

\bibitem{cheng2024navila}
A.-C. Cheng, Y.~Ji, Z.~Yang, Z.~Gongye, X.~Zou, J.~Kautz, E.~B\i{}y\i{}k,
  H.~Yin, S.~Liu, and X.~Wang, ``Navila: Legged robot vision-language-action
  model for navigation,'' \emph{arXiv preprint arXiv:2412.04453}, 2024.

\bibitem{Humanoid-VLA}
P.~Ding, J.~Ma, X.~Tong, B.~Zou, X.~Luo, Y.~Fan, T.~Wang, H.~Lu, P.~Mo, J.~Liu
  \emph{et~al.}, ``Humanoid-vla: Towards universal humanoid control with visual
  integration,'' \emph{arXiv preprint arXiv:2502.14795}, 2025.

\bibitem{LoHoVLA}
Y.~Yang, J.~Sun, S.~Kou, Y.~Wang, and Z.~Deng, ``Lohovla: A unified
  vision-language-action model for long-horizon embodied tasks,'' \emph{arXiv
  preprint arXiv:2506.00411}, 2025.

\bibitem{DP-VLA}
B.~Han, J.~Kim, and J.~Jang, ``A dual process vla: Efficient robotic
  manipulation leveraging vlm,'' \emph{arXiv preprint arXiv:2410.15549}, 2024.

\bibitem{TriVLA}
Z.~Liu, Y.~Gu, S.~Zheng, X.~Xue, and Y.~Fu, ``Trivla: A triple-system-based
  unified vision-language-action model for general robot control,'' \emph{arXiv
  preprint arXiv:2507.01424}, 2025.

\bibitem{ECoT-Lite}
W.~Chen, S.~Belkhale, S.~Mirchandani, O.~Mees, D.~Driess, K.~Pertsch, and
  S.~Levine, ``Training strategies for efficient embodied reasoning,''
  \emph{arXiv preprint arXiv:2505.08243}, 2025.

\bibitem{FastECoT}
Z.~Duan, Y.~Zhang, S.~Geng, G.~Liu, J.~Boedecker, and C.~X. Lu, ``Fast ecot:
  Efficient embodied chain-of-thought via thoughts reuse,'' \emph{arXiv
  preprint arXiv:2506.07639}, 2025.

\bibitem{ICRT}
L.~Fu, H.~Huang, G.~Datta, L.~Y. Chen, W.~C.-H. Panitch, F.~Liu, H.~Li, and
  K.~Goldberg, ``In-context imitation learning via next-token prediction,''
  \emph{arXiv preprint arXiv:2408.15980}, 2024.

\bibitem{TRA}
V.~Myers, B.~C. Zheng, A.~Dragan, K.~Fang, and S.~Levine, ``Temporal
  representation alignment: Successor features enable emergent compositionality
  in robot instruction following,'' \emph{arXiv preprint arXiv:2502.05454},
  2025.

\bibitem{schulman2017ppo}
J.~Schulman, F.~Wolski, P.~Dhariwal, A.~Radford, and O.~Klimov, ``Proximal
  policy optimization algorithms,'' \emph{arXiv preprint arXiv:1707.06347},
  2017.

\bibitem{haarnoja2018sac}
T.~Haarnoja, A.~Zhou, P.~Abbeel, and S.~Levine, ``Soft actor-critic: Off-policy
  maximum entropy deep reinforcement learning with a stochastic actor,'' in
  \emph{Proceedings of the 35th International Conference on Machine Learning
  (ICML)}, ser. Proceedings of Machine Learning Research, J.~Dy and A.~Krause,
  Eds., vol.~80.\hskip 1em plus 0.5em minus 0.4em\relax PMLR, 10--15 Jul 2018,
  pp. 1861--1870.

\bibitem{iRe-VLA}
Y.~Guo, J.~Zhang, X.~Chen, X.~Ji, Y.-J. Wang, Y.~Hu, and J.~Chen, ``Improving
  vision-language-action model with online reinforcement learning,''
  \emph{arXiv preprint arXiv:2501.16664}, 2025.

\bibitem{ConRFT}
Y.~Chen, S.~Tian, S.~Liu, Y.~Zhou, H.~Li, and D.~Zhao, ``Conrft: A reinforced
  fine-tuning method for vla models via consistency policy,'' \emph{arXiv
  preprint arXiv:2502.05450}, 2025.

\bibitem{luo2024serl}
J.~Luo, Z.~Hu, C.~Xu, Y.~L. Tan, J.~Berg, A.~Sharma, S.~Schaal, C.~Finn,
  A.~Gupta, and S.~Levine, ``Serl: A software suite for sample-efficient
  robotic reinforcement learning,'' in \emph{2024 IEEE International Conference
  on Robotics and Automation (ICRA)}, 2024, pp. 16\,961--16\,969.

\bibitem{HIL-SERL}
J.~Luo, C.~Xu, J.~Wu, and S.~Levine, ``Precise and dexterous robotic
  manipulation via human-in-the-loop reinforcement learning,'' \emph{arXiv
  preprint arXiv:2410.21845}, 2024.

\bibitem{han2015resetfree}
W.~Han, S.~Levine, and P.~Abbeel, ``Learning compound multi-step controllers
  under unknown dynamics,'' in \emph{Proceedings of the IEEE/RSJ International
  Conference on Intelligent Robots and Systems (IROS)}, 2015, pp. 6435--6442.

\bibitem{gupta2021resetfree2}
A.~Gupta, J.~Yu, T.~Z. Zhao, V.~Kumar, A.~Rovinsky, K.~Xu, T.~Devlin, and
  S.~Levine, ``Reset-free reinforcement learning via multi-task learning:
  Learning dexterous manipulation behaviors without human intervention,'' in
  \emph{2021 IEEE International Conference on Robotics and Automation (ICRA)},
  2021, pp. 6664--6671.

\bibitem{ball2023rlpd}
P.~J. Ball, L.~Smith, I.~Kostrikov, and S.~Levine, ``Efficient online
  reinforcement learning with offline data,'' in \emph{Proceedings of the 40th
  International Conference on Machine Learning (ICML)}, ser. Proceedings of
  Machine Learning Research, A.~Krause, E.~Brunskill, K.~Cho, B.~Engelhardt,
  S.~Sabato, and J.~Scarlett, Eds., vol. 202.\hskip 1em plus 0.5em minus
  0.4em\relax PMLR, 23--29 Jul 2023, pp. 1577--1594.

\bibitem{VLA-RL}
G.~Lu, W.~Guo, C.~Zhang, Y.~Zhou, H.~Jiang, Z.~Gao, Y.~Tang, and Z.~Wang,
  ``Vla-rl: Towards masterful and general robotic manipulation with scalable
  reinforcement learning,'' \emph{arXiv preprint arXiv:2505.18719}, 2025.

\bibitem{RLDG}
C.~Xu, Q.~Li, J.~Luo, and S.~Levine, ``Rldg: Robotic generalist policy
  distillation via reinforcement learning,'' \emph{arXiv preprint
  arXiv:2412.09858}, 2024.

\bibitem{RLRC}
Y.~Chen and X.~Li, ``Rlrc: Reinforcement learning-based recovery for compressed
  vision-language-action models,'' \emph{arXiv preprint arXiv:2506.17639},
  2025.

\bibitem{wagenmaker2025steeringdiffusionpolicylatent}
A.~Wagenmaker, M.~Nakamoto, Y.~Zhang, S.~Park, W.~Yagoub, A.~Nagabandi,
  A.~Gupta, and S.~Levine, ``Steering your diffusion policy with latent space
  reinforcement learning,'' \emph{arXiv preprint arXiv:2506.15799}, 2025.

\bibitem{SLIM}
H.~Zhang, H.~Yu, L.~Zhao, A.~Choi, Q.~Bai, B.~Yang, and W.~Xu, ``Slim:
  Sim-to-real legged instructive manipulation via long-horizon visuomotor
  learning,'' \emph{arXiv preprint arXiv:2501.09905}, 2025.

\bibitem{julg2025rpd}
T.~J\"{u}lg, W.~Burgard, and F.~Walter, ``Refined policy distillation: From vla
  generalists to rl experts,'' \emph{arXiv preprint arXiv:2503.05833}, 2025.

\bibitem{lin2014coco}
T.-Y. Lin, M.~Maire, S.~Belongie, J.~Hays, P.~Perona, D.~Ramanan,
  P.~Doll{\'a}r, and C.~L. Zitnick, ``Microsoft coco: Common objects in
  context,'' in \emph{Computer Vision -- ECCV 2014}, D.~Fleet, T.~Pajdla,
  B.~Schiele, and T.~Tuytelaars, Eds.\hskip 1em plus 0.5em minus 0.4em\relax
  Cham: Springer International Publishing, 2014, pp. 740--755.

\bibitem{antol2015VQA}
S.~Antol, A.~Agrawal, J.~Lu, M.~Mitchell, D.~Batra, C.~L. Zitnick, and
  D.~Parikh, ``Vqa: Visual question answering,'' in \emph{Proceedings of the
  IEEE International Conference on Computer Vision (ICCV)}, December 2015.

\bibitem{nvidia2025cosmosworldfoundationmodel}
N.~Agarwal, A.~Ali, M.~Bala, Y.~Balaji, E.~Barker, T.~Cai, P.~Chattopadhyay,
  Y.~Chen, Y.~Cui, Y.~Ding \emph{et~al.}, ``Cosmos world foundation model
  platform for physical ai,'' \emph{arXiv preprint arXiv:2501.03575}, 2025.

\bibitem{qwen2025qwen25}
Qwen, :, A.~Yang, B.~Yang, B.~Zhang, B.~Hui, B.~Zheng, B.~Yu, C.~Li, D.~Liu
  \emph{et~al.}, ``Qwen2.5 technical report,'' \emph{arXiv preprint
  arXiv:2412.15115}, 2025.

\bibitem{NORA}
C.-Y. Hung, Q.~Sun, P.~Hong, A.~Zadeh, C.~Li, U.-X. Tan, N.~Majumder, and
  S.~Poria, ``Nora: A small open-sourced generalist vision language action
  model for embodied tasks,'' \emph{arXiv preprint arXiv:2504.19854}, 2025.

\bibitem{Interleave-VLA}
C.~Fan, X.~Jia, Y.~Sun, Y.~Wang, J.~Wei, Z.~Gong, X.~Zhao, M.~Tomizuka,
  X.~Yang, J.~Yan \emph{et~al.}, ``Interleave-vla: Enhancing robot manipulation
  with interleaved image-text instructions,'' \emph{arXiv preprint
  arXiv:2505.02152}, 2025.

\bibitem{chen2025combatvla}
P.~Chen, P.~Bu, Y.~Wang, X.~Wang, Z.~Wang, J.~Guo, Y.~Zhao, Q.~Zhu, J.~Song,
  S.~Yang \emph{et~al.}, ``Combatvla: An efficient vision-language-action model
  for combat tasks in 3d action role-playing games,'' in \emph{Proceedings of
  the IEEE/CVF International Conference on Computer Vision (ICCV)}, 2025.

\bibitem{liu2023llava}
H.~Liu, C.~Li, Q.~Wu, and Y.~J. Lee, ``Visual instruction tuning,'' in
  \emph{Advances in Neural Information Processing Systems (NeurIPS)}, A.~Oh,
  T.~Naumann, A.~Globerson, K.~Saenko, M.~Hardt, and S.~Levine, Eds.,
  vol.~36.\hskip 1em plus 0.5em minus 0.4em\relax Curran Associates, Inc.,
  2023, pp. 34\,892--34\,916.

\bibitem{OE-VLA}
W.~Zhao, G.~Li, Z.~Gong, P.~Ding, H.~Zhao, and D.~Wang, ``Unveiling the
  potential of vision-language-action models with open-ended multimodal
  instructions,'' \emph{arXiv preprint arXiv:2505.11214}, 2025.

\bibitem{hassabis2024gemini2}
\BIBentryALTinterwordspacing
G.~DeepMind. (2024, Dec.) Introducing gemini 2.0: our new ai model for the
  agentic era. [Online; accessed 2025-08-04]. [Online]. Available:
  \url{https://blog.google/technology/google-deepmind/google-gemini-ai-update-december-2024/}
\BIBentrySTDinterwordspacing

\bibitem{GeminiRobotics}
G.~R. Team, S.~Abeyruwan, J.~Ainslie, J.-B. Alayrac, M.~G. Arenas,
  T.~Armstrong, A.~Balakrishna, R.~Baruch, M.~Bauza, M.~Blokzijl \emph{et~al.},
  ``Gemini robotics: Bringing ai into the physical world,'' \emph{arXiv
  preprint arXiv:2503.20020}, 2025.

\bibitem{bavishi2023fuyu8b}
\BIBentryALTinterwordspacing
R.~Bavishi, E.~Elsen, C.~Hawthorne, M.~Nye, A.~Odena, A.~Somani, and
  S.~Ta\c{s}\i{}rlar, ``Introducing our multimodal models,'' 2023. [Online].
  Available: \url{https://www.adept.ai/blog/fuyu-8b}
\BIBentrySTDinterwordspacing

\bibitem{QUAR-VLA}
P.~Ding, H.~Zhao, W.~Zhang, W.~Song, M.~Zhang, S.~Huang, N.~Yang, and D.~Wang,
  ``Quar-vla: Vision-language-action model for quadruped robots,'' \emph{arXiv
  preprint arXiv:2312.14457}, 2023.

\bibitem{MoRE}
H.~Zhao, W.~Song, D.~Wang, X.~Tong, P.~Ding, X.~Cheng, and Z.~Ge, ``More:
  Unlocking scalability in reinforcement learning for quadruped
  vision-language-action models,'' \emph{arXiv preprint arXiv:2503.08007},
  2025.

\bibitem{yue2024deervla}
Y.~Yue, Y.~Wang, B.~Kang, Y.~Han, S.~Wang, S.~Song, J.~Feng, and G.~Huang,
  ``Deer-vla: Dynamic inference of multimodal large language models for
  efficient robot execution,'' in \emph{Advances in Neural Information
  Processing Systems (NeurIPS)}, A.~Globerson, L.~Mackey, D.~Belgrave, A.~Fan,
  U.~Paquet, J.~Tomczak, and C.~Zhang, Eds., vol.~37.\hskip 1em plus 0.5em
  minus 0.4em\relax Curran Associates, Inc., 2024, pp. 56\,619--56\,643.

\bibitem{grattafiori2024llama3}
A.~Grattafiori, A.~Dubey, A.~Jauhri, A.~Pandey, A.~Kadian, A.~Al-Dahle,
  A.~Letman, A.~Mathur, A.~Schelten, A.~Vaughan \emph{et~al.}, ``The llama 3
  herd of models,'' \emph{arXiv preprint arXiv:2407.21783}, 2024.

\bibitem{FOREWARN}
Y.~Wu, R.~Tian, G.~Swamy, and A.~Bajcsy, ``From foresight to forethought:
  Vlm-in-the-loop policy steering via latent alignment,'' \emph{arXiv preprint
  arXiv:2502.01828}, 2025.

\bibitem{zhan2024anygpt}
J.~Zhan, J.~Dai, J.~Ye, Y.~Zhou, D.~Zhang, Z.~Liu, X.~Zhang, R.~Yuan, G.~Zhang,
  L.~Li \emph{et~al.}, ``Anygpt: Unified multimodal llm with discrete sequence
  modeling,'' \emph{arXiv preprint arXiv:2402.12226}, 2024.

\bibitem{zheng2025tracevla}
R.~Zheng, Y.~Liang, S.~Huang, J.~Gao, H.~D. III, A.~Kolobov, F.~Huang, and
  J.~Yang, ``Tracevla: Visual trace prompting enhances spatial-temporal
  awareness for generalist robotic policies,'' in \emph{International
  Conference on Learning Representations (ICLR)}, 2025.

\bibitem{zhang2025upvla}
J.~Zhang, Y.~Guo, Y.~Hu, X.~Chen, X.~Zhu, and J.~Chen, ``Up-vla: A unified
  understanding and prediction model for embodied agent,'' \emph{arXiv preprint
  arXiv:2501.18867}, 2025.

\bibitem{deitke2025molmo}
M.~Deitke, C.~Clark, S.~Lee, R.~Tripathi, Y.~Yang, J.~S. Park, M.~Salehi,
  N.~Muennighoff, K.~Lo, L.~Soldaini \emph{et~al.}, ``Molmo and pixmo: Open
  weights and open data for state-of-the-art vision-language models,'' in
  \emph{Proceedings of the IEEE/CVF Conference on Computer Vision and Pattern
  Recognition (CVPR)}, June 2025, pp. 91--104.

\bibitem{UAV-VLA}
O.~Sautenkov, Y.~Yaqoot, A.~Lykov, M.~A. Mustafa, G.~Tadevosyan, A.~Akhmetkazy,
  M.~A. Cabrera, M.~Martynov, S.~Karaf, and D.~Tsetserukou, ``Uav-vla:
  Vision-language-action system for large scale aerial mission generation,''
  \emph{arXiv preprint arXiv:2501.05014}, 2025.

\bibitem{lin2024vila}
J.~Lin, H.~Yin, W.~Ping, P.~Molchanov, M.~Shoeybi, and S.~Han, ``Vila: On
  pre-training for visual language models,'' in \emph{Proceedings of the
  IEEE/CVF Conference on Computer Vision and Pattern Recognition (CVPR)}, June
  2024, pp. 26\,689--26\,699.

\bibitem{chen2025internvl}
Z.~Chen, W.~Wang, Y.~Cao, Y.~Liu, Z.~Gao, E.~Cui, J.~Zhu, S.~Ye, H.~Tian,
  Z.~Liu \emph{et~al.}, ``Expanding performance boundaries of open-source
  multimodal models with model, data, and test-time scaling,'' \emph{arXiv
  preprint arXiv:2412.05271}, 2025.

\bibitem{li2025eagle2}
Z.~Li, G.~Chen, S.~Liu, S.~Wang, V.~VS, Y.~Ji, S.~Lan, H.~Zhang, Y.~Zhao,
  S.~Radhakrishnan \emph{et~al.}, ``Eagle 2: Building post-training data
  strategies from scratch for frontier vision-language models,'' \emph{arXiv
  preprint arXiv:2501.14818}, 2025.

\bibitem{chameleonteam2025chameleon}
C.~Team, ``Chameleon: Mixed-modal early-fusion foundation models,'' \emph{arXiv
  preprint arXiv:2405.09818}, 2025.

\bibitem{driess2025insulating}
D.~Driess, J.~T. Springenberg, B.~Ichter, L.~Yu, A.~Li-Bell, K.~Pertsch, A.~Z.
  Ren, H.~Walke, Q.~Vuong, L.~X. Shi \emph{et~al.}, ``Knowledge insulating
  vision-language-action models: Train fast, run fast, generalize better,''
  \emph{arXiv preprint arXiv:2505.23705}, 2025.

\bibitem{RevLA}
S.~Dey, J.-N. Zaech, N.~Nikolov, L.~V. Gool, and D.~P. Paudel, ``Revla:
  Reverting visual domain limitation of robotic foundation models,''
  \emph{arXiv preprint arXiv:2409.15250}, 2024.

\bibitem{hejna2025remix}
J.~Hejna, C.~A. Bhateja, Y.~Jiang, K.~Pertsch, and D.~Sadigh, ``Remix:
  Optimizing data mixtures for large scale imitation learning,'' in
  \emph{Proceedings of The 8th Conference on Robot Learning (CoRL)}, ser.
  Proceedings of Machine Learning Research, P.~Agrawal, O.~Kroemer, and
  W.~Burgard, Eds., vol. 270.\hskip 1em plus 0.5em minus 0.4em\relax PMLR,
  06--09 Nov 2025, pp. 145--164.

\bibitem{hu2022lora}
E.~J. Hu, Y.~Shen, P.~Wallis, Z.~Allen-Zhu, Y.~Li, S.~Wang, L.~Wang, and
  W.~Chen, ``Lora: Low-rank adaptation of large language models,'' in
  \emph{International Conference on Learning Representations (ICLR)}, 2022.

\bibitem{BitVLA}
H.~Wang, C.~Xiong, R.~Wang, and X.~Chen, ``Bitvla: 1-bit vision-language-action
  models for robotics manipulation,'' \emph{arXiv preprint arXiv:2506.07530},
  2025.

\bibitem{RealTimeChunking}
K.~Black, M.~Y. Galliker, and S.~Levine, ``Real-time execution of action
  chunking flow policies,'' \emph{arXiv preprint arXiv:2506.07339}, 2025.

\bibitem{Deer-VLA}
Y.~Yue, Y.~Wang, B.~Kang, Y.~Han, S.~Wang, S.~Song, J.~Feng, and G.~Huang,
  ``Deer-vla: Dynamic inference of multimodal large language models for
  efficient robot execution,'' \emph{arXiv preprint arXiv:2411.02359}, 2024.

\bibitem{VLA-Cache}
S.~Xu, Y.~Wang, C.~Xia, D.~Zhu, T.~Huang, and C.~Xu, ``Vla-cache: Towards
  efficient vision-language-action model via adaptive token caching in robotic
  manipulation,'' \emph{arXiv preprint arXiv:2502.02175}, 2025.

\bibitem{zhao2023aloha}
T.~Z. Zhao, V.~Kumar, S.~Levine, and C.~Finn, ``Learning fine-grained bimanual
  manipulation with low-cost hardware,'' in \emph{Proceedings of Robotics:
  Science and Systems (RSS)}, Daegu, Republic of Korea, July 2023.

\bibitem{fu2024mobilealoha}
\BIBentryALTinterwordspacing
Z.~Fu, T.~Z. Zhao, and C.~Finn, ``Mobile aloha: Learning bimanual mobile
  manipulation using low-cost whole-body teleoperation,'' in \emph{Proceedings
  of The 8th Conference on Robot Learning (CoRL)}, ser. Proceedings of Machine
  Learning Research, P.~Agrawal, O.~Kroemer, and W.~Burgard, Eds., vol.
  270.\hskip 1em plus 0.5em minus 0.4em\relax PMLR, 06--09 Nov 2025, pp.
  4066--4083. [Online]. Available:
  \url{https://proceedings.mlr.press/v270/fu25b.html}
\BIBentrySTDinterwordspacing

\bibitem{aloha2team2024aloha2}
A.~. Team, J.~Aldaco, T.~Armstrong, R.~Baruch, J.~Bingham, S.~Chan, K.~Draper,
  D.~Dwibedi, C.~Finn, P.~Florence \emph{et~al.}, ``Aloha 2: An enhanced
  low-cost hardware for bimanual teleoperation,'' \emph{arXiv preprint
  arXiv:2405.02292}, 2024.

\bibitem{zhao25alohaunleashed}
T.~Z. Zhao, J.~Tompson, D.~Driess, P.~Florence, S.~K.~S. Ghasemipour, C.~Finn,
  and A.~Wahid, ``Aloha unleashed: A simple recipe for robot dexterity,'' in
  \emph{Proceedings of The 8th Conference on Robot Learning (CoRL)}, ser.
  Proceedings of Machine Learning Research, P.~Agrawal, O.~Kroemer, and
  W.~Burgard, Eds., vol. 270.\hskip 1em plus 0.5em minus 0.4em\relax PMLR,
  06--09 Nov 2025, pp. 1910--1924.

\bibitem{buamanee2024biact}
T.~Buamanee, M.~Kobayashi, Y.~Uranishi, and H.~Takemura, ``Bi-act: Bilateral
  control-based imitation learning via action chunking with transformer,'' in
  \emph{2024 IEEE International Conference on Advanced Intelligent Mechatronics
  (AIM)}, 2024, pp. 410--415.

\bibitem{wu2024gello}
P.~Wu, Y.~Shentu, Z.~Yi, X.~Lin, and P.~Abbeel, ``Gello: A general, low-cost,
  and intuitive teleoperation framework for robot manipulators,'' in
  \emph{Proceedings of the IEEE/RSJ International Conference on Intelligent
  Robots and Systems (IROS)}, 2024, pp. 12\,156--12\,163.

\bibitem{qin2023anyteleop}
Y.~Qin, W.~Yang, B.~Huang, K.~V. Wyk, H.~Su, X.~Wang, Y.-W. Chao, and D.~Fox,
  ``{AnyTeleop: A General Vision-Based Dexterous Robot Arm-Hand Teleoperation
  System},'' in \emph{Proceedings of Robotics: Science and Systems (RSS)},
  Daegu, Republic of Korea, July 2023.

\bibitem{zhang2020mediapipehands}
F.~Zhang, V.~Bazarevsky, A.~Vakunov, A.~Tkachenka, G.~Sung, C.-L. Chang, and
  M.~Grundmann, ``Mediapipe hands: On-device real-time hand tracking,''
  \emph{arXiv preprint arXiv:2006.10214}, 2020.

\bibitem{sundaralingam2023curobo}
B.~Sundaralingam, S.~K.~S. Hari, A.~Fishman, C.~Garrett, K.~Van~Wyk, V.~Blukis,
  A.~Millane, H.~Oleynikova, A.~Handa, F.~Ramos \emph{et~al.}, ``Curobo:
  Parallelized collision-free robot motion generation,'' in \emph{2023 IEEE
  International Conference on Robotics and Automation (ICRA)}, 2023, pp.
  8112--8119.

\bibitem{yang2025ace}
S.~Yang, M.~Liu, Y.~Qin, R.~Ding, J.~Li, X.~Cheng, R.~Yang, S.~Yi, and X.~Wang,
  ``Ace: A cross-platform and visual-exoskeletons system for low-cost dexterous
  teleoperation,'' in \emph{Proceedings of The 8th Conference on Robot Learning
  (CoRL)}, ser. Proceedings of Machine Learning Research, P.~Agrawal,
  O.~Kroemer, and W.~Burgard, Eds., vol. 270.\hskip 1em plus 0.5em minus
  0.4em\relax PMLR, 06--09 Nov 2025, pp. 4895--4911.

\bibitem{cheng2025opentelevision}
X.~Cheng, J.~Li, S.~Yang, G.~Yang, and X.~Wang, ``Open-television:
  Teleoperation with immersive active visual feedback,'' in \emph{Proceedings
  of The 8th Conference on Robot Learning (CoRL)}, ser. Proceedings of Machine
  Learning Research, P.~Agrawal, O.~Kroemer, and W.~Burgard, Eds., vol.
  270.\hskip 1em plus 0.5em minus 0.4em\relax PMLR, 06--09 Nov 2025, pp.
  2729--2749.

\bibitem{ding2024bunnyvisionpro}
R.~Ding, Y.~Qin, J.~Zhu, C.~Jia, S.~Yang, R.~Yang, X.~Qi, and X.~Wang,
  ``Bunny-visionpro: Real-time bimanual dexterous teleoperation for imitation
  learning,'' \emph{arXiv preprint arXiv:2407.03162}, 2024.

\bibitem{chi2024umi}
C.~Chi, Z.~Xu, C.~Pan, E.~Cousineau, B.~Burchfiel, S.~Feng, R.~Tedrake, and
  S.~Song, ``Universal manipulation interface: In-the-wild robot teaching
  without in-the-wild robots,'' in \emph{Proceedings of Robotics: Science and
  Systems (RSS)}, Delft, Netherlands, July 2024.

\bibitem{trilbmteam2025lbm}
T.~L. Team, J.~Barreiros, A.~Beaulieu, A.~Bhat, R.~Cory, E.~Cousineau, H.~Dai,
  C.-H. Fang, K.~Hashimoto, M.~Z. Irshad \emph{et~al.}, ``A careful examination
  of large behavior models for multitask dexterous manipulation,'' \emph{arXiv
  preprint arXiv:2507.05331}, 2025.

\bibitem{xu2025dexumi}
M.~Xu, H.~Zhang, Y.~Hou, Z.~Xu, L.~Fan, M.~Veloso, and S.~Song, ``Dexumi: Using
  human hand as the universal manipulation interface for dexterous
  manipulation,'' in \emph{3rd RSS Workshop on Dexterous Manipulation: Learning
  and Control with Diverse Data}, 2025.

\bibitem{Dobb-E}
N.~M.~M. Shafiullah, A.~Rai, H.~Etukuru, Y.~Liu, I.~Misra, S.~Chintala, and
  L.~Pinto, ``On bringing robots home,'' \emph{arXiv preprint
  arXiv:2311.16098}, 2023.

\bibitem{RUMs}
H.~Etukuru, N.~Naka, Z.~Hu, S.~Lee, J.~Mehu, A.~Edsinger, C.~Paxton,
  S.~Chintala, L.~Pinto, and N.~M.~M. Shafiullah, ``Robot utility models:
  General policies for zero-shot deployment in new environments,'' \emph{arXiv
  preprint arXiv:2409.05865}, 2024.

\bibitem{wang2024dexcap}
C.~Wang, H.~Shi, W.~Wang, R.~Zhang, L.~Fei-Fei, and K.~Liu, ``Dexcap: Scalable
  and portable mocap data collection system for dexterous manipulation,'' in
  \emph{Proceedings of Robotics: Science and Systems (RSS)}, Delft,
  Netherlands, July 2024.

\bibitem{tao2025dexwild}
T.~Tao, M.~K. Srirama, J.~J. Liu, K.~Shaw, and D.~Pathak, ``Dexwild: Dexterous
  human interactions for in-the-wild robot policies,'' \emph{arXiv preprint
  arXiv:2505.07813}, 2025.

\bibitem{grauman2024egoexo4d}
K.~Grauman, A.~Westbury, L.~Torresani, K.~Kitani, J.~Malik, T.~Afouras,
  K.~Ashutosh, V.~Baiyya, S.~Bansal, B.~Boote \emph{et~al.}, ``Ego-exo4d:
  Understanding skilled human activity from first- and third-person
  perspectives,'' in \emph{Proceedings of the IEEE/CVF Conference on Computer
  Vision and Pattern Recognition (CVPR)}, June 2024, pp. 19\,383--19\,400.

\bibitem{banerjee2024hot3d}
P.~Banerjee, S.~Shkodrani, P.~Moulon, S.~Hampali, S.~Han, F.~Zhang, L.~Zhang,
  J.~Fountain, E.~Miller, S.~Basol \emph{et~al.}, ``Hot3d: Hand and object
  tracking in 3d from egocentric multi-view videos,'' in \emph{Proceedings of
  the IEEE/CVF Conference on Computer Vision and Pattern Recognition (CVPR)},
  June 2025, pp. 7061--7071.

\bibitem{perrett2025hdepic}
T.~Perrett, A.~Darkhalil, S.~Sinha, O.~Emara, S.~Pollard, K.~Parida, K.~Liu,
  P.~Gatti, S.~Bansal, K.~Flanagan \emph{et~al.}, ``Hd-epic: A highly-detailed
  egocentric video dataset,'' in \emph{Proceedings of the IEEE/CVF Conference
  on Computer Vision and Pattern Recognition (CVPR)}, June 2025.

\bibitem{lv2024ariaeverydayactivities}
Z.~Lv, N.~Charron, P.~Moulon, A.~Gamino, C.~Peng, C.~Sweeney, E.~Miller,
  H.~Tang, J.~Meissner, J.~Dong \emph{et~al.}, ``Aria everyday activities
  dataset,'' \emph{arXiv preprint arXiv:2402.13349}, 2024.

\bibitem{kareer2024egomimic}
S.~Kareer, D.~Patel, R.~Punamiya, P.~Mathur, S.~Cheng, C.~Wang, J.~Hoffman, and
  D.~Xu, ``Egomimic: Scaling imitation learning via egocentric video,''
  \emph{arXiv preprint arXiv:2410.24221}, 2024.

\bibitem{liu2025egozero}
V.~Liu, A.~Adeniji, H.~Zhan, S.~Haldar, R.~Bhirangi, P.~Abbeel, and L.~Pinto,
  ``Egozero: Robot learning from smart glasses,'' \emph{arXiv preprint
  arXiv:2505.20290}, 2025.

\bibitem{qiu2025humanoidpolicyhuman}
R.-Z. Qiu, S.~Yang, X.~Cheng, C.~Chawla, J.~Li, T.~He, G.~Yan, D.~J. Yoon,
  R.~Hoque, L.~Paulsen \emph{et~al.}, ``Humanoid policy ~ human policy,''
  \emph{arXiv preprint arXiv:2503.13441}, 2025.

\bibitem{mandlekar2018roboturk}
A.~Mandlekar, Y.~Zhu, A.~Garg, J.~Booher, M.~Spero, A.~Tung, J.~Gao, J.~Emmons,
  A.~Gupta, E.~Orbay \emph{et~al.}, ``Roboturk: A crowdsourcing platform for
  robotic skill learning through imitation,'' in \emph{Proceedings of The 2nd
  Conference on Robot Learning (CoRL)}, ser. Proceedings of Machine Learning
  Research, A.~Billard, A.~Dragan, J.~Peters, and J.~Morimoto, Eds.,
  vol.~87.\hskip 1em plus 0.5em minus 0.4em\relax PMLR, 29--31 Oct 2018, pp.
  879--893.

\bibitem{crowston2012mechanicalturk}
K.~Crowston, ``Amazon mechanical turk: A research tool for organizations and
  information systems scholars,'' in \emph{Shaping the Future of ICT Research.
  Methods and Approaches}, A.~Bhattacherjee and B.~Fitzgerald, Eds.\hskip 1em
  plus 0.5em minus 0.4em\relax Berlin, Heidelberg: Springer Berlin Heidelberg,
  2012, pp. 210--221.

\bibitem{lynch2023languagetable}
C.~Lynch, A.~Wahid, J.~Tompson, T.~Ding, J.~Betker, R.~Baruch, T.~Armstrong,
  and P.~Florence, ``Interactive language: Talking to robots in real time,''
  \emph{IEEE Robotics and Automation Letters (RA-L)}, pp. 1--8, 2023.

\bibitem{khazatsky2024droid}
A.~Khazatsky, K.~Pertsch, S.~Nair, A.~Balakrishna, S.~Dasari, S.~Karamcheti,
  S.~Nasiriany, M.~K. Srirama, L.~Y. Chen, K.~Ellis \emph{et~al.}, ``Droid: A
  large-scale in-the-wild robot manipulation dataset,'' in \emph{Proceedings of
  Robotics: Science and Systems (RSS)}, Delft, Netherlands, July 2024.

\bibitem{sun2025emmax}
\BIBentryALTinterwordspacing
Q.~Sun, P.~Hong, T.~D. Pala, V.~Toh, U.-X. Tan, D.~Ghosal, and S.~Poria,
  ``Emma-x: An embodied multimodal action model with grounded chain of thought
  and look-ahead spatial reasoning,'' in \emph{Proceedings of the 63rd Annual
  Meeting of the Association for Computational Linguistics (Volume 1: Long
  Papers)}, W.~Che, J.~Nabende, E.~Shutova, and M.~T. Pilehvar, Eds.\hskip 1em
  plus 0.5em minus 0.4em\relax Vienna, Austria: Association for Computational
  Linguistics, July 2025, pp. 14\,199--14\,214. [Online]. Available:
  \url{https://aclanthology.org/2025.acl-long.695/}
\BIBentrySTDinterwordspacing

\bibitem{blank2024nils}
\BIBentryALTinterwordspacing
N.~Blank, M.~Reuss, M.~R{\"{u}}hle, {\"{O}}.~E. Ya\u{g}murlu, F.~Wenzel,
  O.~Mees, and R.~Lioutikov, ``Scaling robot policy learning via zero-shot
  labeling with foundation models,'' in \emph{Proceedings of The 8th Conference
  on Robot Learning (CoRL)}, ser. Proceedings of Machine Learning Research,
  P.~Agrawal, O.~Kroemer, and W.~Burgard, Eds., vol. 270.\hskip 1em plus 0.5em
  minus 0.4em\relax PMLR, 06--09 Nov 2025, pp. 4158--4187. [Online]. Available:
  \url{https://proceedings.mlr.press/v270/blank25a.html}
\BIBentrySTDinterwordspacing

\bibitem{wu2025robomind}
K.~Wu, C.~Hou, J.~Liu, Z.~Che, X.~Ju, Z.~Yang, M.~Li, Y.~Zhao, Z.~Xu, G.~Yang
  \emph{et~al.}, ``Robomind: Benchmark on multi-embodiment intelligence
  normative data for robot manipulation,'' \emph{arXiv preprint
  arXiv:2412.13877}, 2025.

\bibitem{geminiteam2024gemini15}
G.~Team, P.~Georgiev, V.~I. Lei, R.~Burnell, L.~Bai, A.~Gulati, G.~Tanzer,
  D.~Vincent, Z.~Pan, S.~Wang \emph{et~al.}, ``Gemini 1.5: Unlocking multimodal
  understanding across millions of tokens of context,'' \emph{arXiv preprint
  arXiv:2403.05530}, 2024.

\bibitem{fang2023rh20t}
H.-S. Fang, H.~Fang, Z.~Tang, J.~Liu, J.~Wang, H.~Zhu, and C.~Lu, ``Rh20t: A
  robotic dataset for learning diverse skills in one-shot,'' in \emph{RSS 2023
  Workshop on Learning for Task and Motion Planning}, 2023.

\bibitem{kalashnikov2018qtopt}
D.~Kalashnikov, A.~Irpan, P.~Pastor, J.~Ibarz, A.~Herzog, E.~Jang, D.~Quillen,
  E.~Holly, M.~Kalakrishnan, V.~Vanhoucke \emph{et~al.}, ``Scalable deep
  reinforcement learning for vision-based robotic manipulation,'' in
  \emph{Proceedings of The 2nd Conference on Robot Learning (CoRL)}, ser.
  Proceedings of Machine Learning Research, A.~Billard, A.~Dragan, J.~Peters,
  and J.~Morimoto, Eds., vol.~87.\hskip 1em plus 0.5em minus 0.4em\relax PMLR,
  29--31 Oct 2018, pp. 651--673.

\bibitem{kalashnikov2021mtopt}
D.~Kalashnikov, J.~Varley, Y.~Chebotar, B.~Swanson, R.~Jonschkowski, C.~Finn,
  S.~Levine, and K.~Hausman, ``Mt-opt: Continuous multi-task robotic
  reinforcement learning at scale,'' \emph{arXiv preprint arXiv:2104.08212},
  2021.

\bibitem{dasari2020robonet}
S.~Dasari, F.~Ebert, S.~Tian, S.~Nair, B.~Bucher, K.~Schmeckpeper, S.~Singh,
  S.~Levine, and C.~Finn, ``Robonet: Large-scale multi-robot learning,'' in
  \emph{Proceedings of the Conference on Robot Learning (CoRL)}, ser.
  Proceedings of Machine Learning Research, L.~P. Kaelbling, D.~Kragic, and
  K.~Sugiura, Eds., vol. 100.\hskip 1em plus 0.5em minus 0.4em\relax PMLR, 30
  Oct--01 Nov 2020, pp. 885--897.

\bibitem{ebert2022bridgedata}
F.~Ebert, Y.~Yang, K.~Schmeckpeper, B.~Bucher, G.~Georgakis, K.~Daniilidis,
  C.~Finn, and S.~Levine, ``Bridge data: Boosting generalization of robotic
  skills with cross-domain datasets,'' in \emph{Proceedings of Robotics:
  Science and Systems (RSS)}, New York City, NY, USA, June 2022.

\bibitem{walke2023bridgedatav2}
\BIBentryALTinterwordspacing
H.~R. Walke, K.~Black, T.~Z. Zhao, Q.~Vuong, C.~Zheng, P.~Hansen-Estruch, A.~W.
  He, V.~Myers, M.~J. Kim, M.~Du \emph{et~al.}, ``Bridgedata v2: A dataset for
  robot learning at scale,'' in \emph{Proceedings of The 7th Conference on
  Robot Learning (CoRL)}, ser. Proceedings of Machine Learning Research,
  J.~Tan, M.~Toussaint, and K.~Darvish, Eds., vol. 229.\hskip 1em plus 0.5em
  minus 0.4em\relax PMLR, 06--09 Nov 2023, pp. 1723--1736. [Online]. Available:
  \url{https://proceedings.mlr.press/v229/walke23a.html}
\BIBentrySTDinterwordspacing

\bibitem{jang2022bcz}
E.~Jang, A.~Irpan, M.~Khansari, D.~Kappler, F.~Ebert, C.~Lynch, S.~Levine, and
  C.~Finn, ``Bc-z: Zero-shot task generalization with robotic imitation
  learning,'' in \emph{Proceedings of the 5th Conference on Robot Learning
  (CoRL)}, ser. Proceedings of Machine Learning Research, A.~Faust, D.~Hsu, and
  G.~Neumann, Eds., vol. 164.\hskip 1em plus 0.5em minus 0.4em\relax PMLR,
  08--11 Nov 2022, pp. 991--1002.

\bibitem{mees2022calvin}
O.~Mees, L.~Hermann, E.~Rosete-Beas, and W.~Burgard, ``Calvin: A benchmark for
  language-conditioned policy learning for long-horizon robot manipulation
  tasks,'' \emph{IEEE Robotics and Automation Letters (RA-L)}, vol.~7, no.~3,
  pp. 7327--7334, 2022.

\bibitem{liu2023libero}
B.~Liu, Y.~Zhu, C.~Gao, Y.~Feng, Q.~Liu, Y.~Zhu, and P.~Stone, ``Libero:
  Benchmarking knowledge transfer for lifelong robot learning,'' in
  \emph{Advances in Neural Information Processing Systems (NeurIPS)}, A.~Oh,
  T.~Naumann, A.~Globerson, K.~Saenko, M.~Hardt, and S.~Levine, Eds.,
  vol.~36.\hskip 1em plus 0.5em minus 0.4em\relax Curran Associates, Inc.,
  2023, pp. 44\,776--44\,791.

\bibitem{liu2022hoi4d}
Y.~Liu, Y.~Liu, C.~Jiang, K.~Lyu, W.~Wan, H.~Shen, B.~Liang, Z.~Fu, H.~Wang,
  and L.~Yi, ``Hoi4d: A 4d egocentric dataset for category-level human-object
  interaction,'' in \emph{Proceedings of the IEEE/CVF Conference on Computer
  Vision and Pattern Recognition (CVPR)}, June 2022, pp. 21\,013--21\,022.

\bibitem{zhan2024oakink2}
X.~Zhan, L.~Yang, Y.~Zhao, K.~Mao, H.~Xu, Z.~Lin, K.~Li, and C.~Lu, ``Oakink2:
  A dataset of bimanual hands-object manipulation in complex task completion,''
  in \emph{Proceedings of the IEEE/CVF Conference on Computer Vision and
  Pattern Recognition (CVPR)}, June 2024, pp. 445--456.

\bibitem{kwon2021h2o}
T.~Kwon, B.~Tekin, J.~St\"{u}hmer, F.~Bogo, and M.~Pollefeys, ``H2o: Two hands
  manipulating objects for first person interaction recognition,'' in
  \emph{Proceedings of the IEEE/CVF International Conference on Computer Vision
  (ICCV)}, October 2021, pp. 10\,138--10\,148.

\bibitem{fan2023arctic}
Z.~Fan, O.~Taheri, D.~Tzionas, M.~Kocabas, M.~Kaufmann, M.~J. Black, and
  O.~Hilliges, ``Arctic: A dataset for dexterous bimanual hand-object
  manipulation,'' in \emph{Proceedings IEEE Conference on Computer Vision and
  Pattern Recognition (CVPR)}, 2023.

\bibitem{li2022egopat3d}
Y.~Li, Z.~Cao, A.~Liang, B.~Liang, L.~Chen, H.~Zhao, and C.~Feng, ``Egocentric
  prediction of action target in 3d,'' in \emph{Proceedings of the IEEE/CVF
  Conference on Computer Vision and Pattern Recognition (CVPR)}, June 2022.

\bibitem{miech2019howto100m}
A.~Miech, D.~Zhukov, J.-B. Alayrac, M.~Tapaswi, I.~Laptev, and J.~Sivic,
  ``Howto100m: Learning a text-video embedding by watching hundred million
  narrated video clips,'' in \emph{Proceedings of the IEEE/CVF International
  Conference on Computer Vision (ICCV)}, October 2019.

\bibitem{goyal2017something}
R.~Goyal, S.~Ebrahimi~Kahou, V.~Michalski, J.~Materzynska, S.~Westphal, H.~Kim,
  V.~Haenel, I.~Fruend, P.~Yianilos, M.~Mueller-Freitag \emph{et~al.}, ``The"
  something something" video database for learning and evaluating visual common
  sense,'' in \emph{Proceedings of the IEEE international conference on
  computer vision}, 2017, pp. 5842--5850.

\bibitem{carreira2022kinetics700}
J.~Carreira, E.~Noland, C.~Hillier, and A.~Zisserman, ``A short note on the
  kinetics-700 human action dataset,'' \emph{arXiv preprint arXiv:1907.06987},
  2022.

\bibitem{todorov2012mujoco}
E.~Todorov, T.~Erez, and Y.~Tassa, ``Mujoco: A physics engine for model-based
  control,'' in \emph{Proceedings of the IEEE/RSJ International Conference on
  Intelligent Robots and Systems (IROS)}, 2012, pp. 5026--5033.

\bibitem{mandlekar2023mimicgen}
\BIBentryALTinterwordspacing
A.~Mandlekar, S.~Nasiriany, B.~Wen, I.~Akinola, Y.~Narang, L.~Fan, Y.~Zhu, and
  D.~Fox, ``Mimicgen: A data generation system for scalable robot learning
  using human demonstrations,'' in \emph{Proceedings of The 7th Conference on
  Robot Learning (CoRL)}, ser. Proceedings of Machine Learning Research,
  J.~Tan, M.~Toussaint, and K.~Darvish, Eds., vol. 229.\hskip 1em plus 0.5em
  minus 0.4em\relax PMLR, 06--09 Nov 2023, pp. 1820--1864. [Online]. Available:
  \url{https://proceedings.mlr.press/v229/mandlekar23a.html}
\BIBentrySTDinterwordspacing

\bibitem{jiang2024dexmimicen}
Z.~Jiang, Y.~Xie, K.~Lin, Z.~Xu, W.~Wan, A.~Mandlekar, L.~Fan, and Y.~Zhu,
  ``Dexmimicgen: Automated data generation for bimanual dexterous manipulation
  via imitation learning,'' \emph{arXiv preprint arXiv:2410.24185}, 2025.

\bibitem{sharma2018mime}
P.~Sharma, L.~Mohan, L.~Pinto, and A.~Gupta, ``Multiple interactions made easy
  (mime): Large scale demonstrations data for imitation,'' in \emph{Proceedings
  of The 2nd Conference on Robot Learning (CoRL)}, ser. Proceedings of Machine
  Learning Research, A.~Billard, A.~Dragan, J.~Peters, and J.~Morimoto, Eds.,
  vol.~87.\hskip 1em plus 0.5em minus 0.4em\relax PMLR, 29--31 Oct 2018, pp.
  906--915.

\bibitem{ramos2021rlds}
S.~Ramos, S.~Girgin, L.~Hussenot, D.~Vincent, H.~Yakubovich, D.~Toyama,
  A.~Gergely, P.~Stanczyk, R.~Marinier, J.~Harmsen \emph{et~al.}, ``Rlds: an
  ecosystem to generate, share and use datasets in reinforcement learning,''
  \emph{arXiv preprint arXiv:2111.02767}, 2021.

\bibitem{rosete2022tacorl}
\BIBentryALTinterwordspacing
E.~Rosete-Beas, O.~Mees, G.~Kalweit, J.~Boedecker, and W.~Burgard, ``Latent
  plans for task-agnostic offline reinforcement learning,'' in
  \emph{Proceedings of The 6th Conference on Robot Learning (CoRL)}, ser.
  Proceedings of Machine Learning Research, K.~Liu, D.~Kulic, and J.~Ichnowski,
  Eds., vol. 205.\hskip 1em plus 0.5em minus 0.4em\relax PMLR, 14--18 Dec 2023,
  pp. 1838--1849. [Online]. Available:
  \url{https://proceedings.mlr.press/v205/rosete-beas23a.html}
\BIBentrySTDinterwordspacing

\bibitem{mees23taco2}
O.~Mees, J.~Borja-Diaz, and W.~Burgard, ``Grounding language with visual
  affordances over unstructured data,'' in \emph{2023 IEEE International
  Conference on Robotics and Automation (ICRA)}, 2023, pp. 11\,576--11\,582.

\bibitem{dass2023jacoplay}
\BIBentryALTinterwordspacing
S.~Dass, J.~Yapeter, J.~Zhang, J.~Zhang, K.~Pertsch, S.~Nikolaidis, and J.~J.
  Lim, ``Clvr jaco play dataset,'' 2023. [Online]. Available:
  \url{https://github.com/clvrai/clvr_jaco_play_dataset}
\BIBentrySTDinterwordspacing

\bibitem{luo2024cablerouting}
J.~Luo, C.~Xu, X.~Geng, G.~Feng, K.~Fang, L.~Tan, S.~Schaal, and S.~Levine,
  ``Multistage cable routing through hierarchical imitation learning,''
  \emph{IEEE Transactions on Robotics}, vol.~40, pp. 1476--1491, 2024.

\bibitem{BerkeleyUR5Website}
L.~Y. Chen, S.~Adebola, and K.~Goldberg, ``Berkeley {UR5} demonstration
  dataset,'' \url{https://sites.google.com/view/berkeley-ur5/home}.

\bibitem{zhou2023toto}
G.~Zhou, V.~Dean, M.~K. Srirama, A.~Rajeswaran, J.~Pari, K.~Hatch, A.~Jain,
  T.~Yu, P.~Abbeel, L.~Pinto \emph{et~al.}, ``Train offline, test online: A
  real robot learning benchmark,'' in \emph{2023 IEEE International Conference
  on Robotics and Automation (ICRA)}, 2023, pp. 9197--9203.

\bibitem{bharadhwaj2024roboagent}
H.~Bharadhwaj, J.~Vakil, M.~Sharma, A.~Gupta, S.~Tulsiani, and V.~Kumar,
  ``Roboagent: Generalization and efficiency in robot manipulation via semantic
  augmentations and action chunking,'' in \emph{2024 IEEE International
  Conference on Robotics and Automation (ICRA)}, 2024, pp. 4788--4795.

\bibitem{hirose2023sacson}
N.~Hirose, D.~Shah, A.~Sridhar, and S.~Levine, ``Sacson: Scalable autonomous
  control for social navigation,'' \emph{IEEE Robotics and Automation Letters
  (RA-L)}, vol.~9, no.~1, pp. 49--56, 2024.

\bibitem{karnan2022scand}
H.~Karnan, A.~Nair, X.~Xiao, G.~Warnell, S.~Pirk, A.~Toshev, J.~Hart,
  J.~Biswas, and P.~Stone, ``Socially compliant navigation dataset (scand): A
  large-scale dataset of demonstrations for social navigation,'' \emph{IEEE
  Robotics and Automation Letters (RA-L)}, vol.~7, no.~4, pp. 11\,807--11\,814,
  2022.

\bibitem{shah2021recon}
\BIBentryALTinterwordspacing
D.~Shah, B.~Eysenbach, N.~Rhinehart, and S.~Levine, ``Rapid exploration for
  open-world navigation with latent goal models,'' in \emph{Proceedings of the
  5th Conference on Robot Learning (CoRL)}, ser. Proceedings of Machine
  Learning Research, A.~Faust, D.~Hsu, and G.~Neumann, Eds., vol. 164.\hskip
  1em plus 0.5em minus 0.4em\relax PMLR, 08--11 Nov 2022, pp. 674--684.
  [Online]. Available: \url{https://proceedings.mlr.press/v164/shah22a.html}
\BIBentrySTDinterwordspacing

\bibitem{yu2020bdd100k}
F.~Yu, H.~Chen, X.~Wang, W.~Xian, Y.~Chen, F.~Liu, V.~Madhavan, and T.~Darrell,
  ``Bdd100k: A diverse driving dataset for heterogeneous multitask learning,''
  in \emph{IEEE/CVF Conference on Computer Vision and Pattern Recognition
  (CVPR)}, June 2020.

\bibitem{sermanet2024robovqa}
P.~Sermanet, T.~Ding, J.~Zhao, F.~Xia, D.~Dwibedi, K.~Gopalakrishnan, C.~Chan,
  G.~Dulac-Arnold, S.~Maddineni, N.~J. Joshi \emph{et~al.}, ``Robovqa:
  Multimodal long-horizon reasoning for robotics,'' in \emph{2024 IEEE
  International Conference on Robotics and Automation (ICRA)}, 2024, pp.
  645--652.

\bibitem{mandi2022cacti}
Z.~Mandi, H.~Bharadhwaj, V.~Moens, S.~Song, A.~Rajeswaran, and V.~Kumar,
  ``Cacti: A framework for scalable multi-task multi-scene visual imitation
  learning,'' in \emph{CoRL 2022 Workshop on Pre-training Robot Learning},
  2022.

\bibitem{chen2023genaug}
Q.~Chen, S.~C. Kiami, A.~Gupta, and V.~Kumar, ``Genaug: Retargeting behaviors
  to unseen situations via generative augmentation,'' in \emph{Proceedings of
  Robotics: Science and Systems (RSS)}, Daegu, Republic of Korea, July 2023.

\bibitem{yu2023rosie}
T.~Yu, T.~Xiao, J.~Tompson, A.~Stone, S.~Wang, A.~Brohan, J.~Singh, C.~Tan,
  D.~M, J.~Peralta \emph{et~al.}, ``Scaling robot learning with semantically
  imagined experience,'' in \emph{Proceedings of Robotics: Science and Systems
  (RSS)}, Daegu, Republic of Korea, July 2023.

\bibitem{wang2023imageneditor}
S.~Wang, C.~Saharia, C.~Montgomery, J.~Pont-Tuset, S.~Noy, S.~Pellegrini,
  Y.~Onoe, S.~Laszlo, D.~J. Fleet, R.~Soricut \emph{et~al.}, ``Imagen editor
  and editbench: Advancing and evaluating text-guided image inpainting,'' in
  \emph{Proceedings of the IEEE/CVF Conference on Computer Vision and Pattern
  Recognition (CVPR)}, June 2023, pp. 18\,359--18\,369.

\bibitem{BYOVLA}
A.~J. Hancock, A.~Z. Ren, and A.~Majumdar, ``Run-time observation interventions
  make vision-language-action models more visually robust,'' \emph{arXiv
  preprint arXiv:2410.01971}, 2024.

\bibitem{ziao2023dial}
T.~Xiao, H.~Chan, P.~Sermanet, A.~Wahid, A.~Brohan, K.~Hausman, S.~Levine, and
  J.~Tompson, ``{Robotic Skill Acquisition via Instruction Augmentation with
  Vision-Language Models},'' in \emph{Proceedings of Robotics: Science and
  Systems (RSS)}, Daegu, Republic of Korea, July 2023.

\bibitem{ross11dagger}
S.~Ross, G.~Gordon, and D.~Bagnell, ``A reduction of imitation learning and
  structured prediction to no-regret online learning,'' in \emph{Proceedings of
  the Fourteenth International Conference on Artificial Intelligence and
  Statistics}, ser. Proceedings of Machine Learning Research, G.~Gordon,
  D.~Dunson, and M.~Dud\'{\i}k, Eds., vol.~15.\hskip 1em plus 0.5em minus
  0.4em\relax Fort Lauderdale, FL, USA: PMLR, 11--13 Apr 2011, pp. 627--635.

\bibitem{ke2024ccil}
L.~Ke, Y.~Zhang, A.~Deshpande, S.~Srinivasa, and A.~Gupta, ``Ccil:
  Continuity-based data augmentation for corrective imitation learning,'' in
  \emph{International Conference on Learning Representations (ICLR)}, 2024.

\bibitem{iskandar2020sara}
M.~Iskandar, C.~Ott, O.~Eiberger, M.~Keppler, A.~Albu-Sch\"{a}ffer, and
  A.~Dietrich, ``Joint-level control of the dlr lightweight robot sara,'' in
  \emph{Proceedings of the IEEE/RSJ International Conference on Intelligent
  Robots and Systems (IROS)}, 2020, pp. 8903--8910.

\bibitem{guist2024safe}
S.~Guist, J.~Schneider, H.~Ma, L.~Chen, V.~Berenz, J.~Martus, H.~Ott,
  F.~Gr\"{u}ninger, M.~Muehlebach, J.~Fiene \emph{et~al.}, ``{Safe \& Accurate
  at Speed with Tendons: A Robot Arm for Exploring Dynamic Motion},'' in
  \emph{Proceedings of Robotics: Science and Systems (RSS)}, Delft,
  Netherlands, July 2024.

\bibitem{shaw2023leaphand}
K.~Shaw, A.~Agarwal, and D.~Pathak, ``{LEAP Hand: Low-Cost, Efficient, and
  Anthropomorphic Hand for Robot Learning},'' in \emph{Proceedings of Robotics:
  Science and Systems (RSS)}, Daegu, Republic of Korea, July 2023.

\bibitem{Shake-VLA}
M.~H. Khan, S.~Asfaw, D.~Iarchuk, M.~A. Cabrera, L.~Moreno, I.~Tokmurziyev, and
  D.~Tsetserukou, ``Shake-vla: Vision-language-action model-based system for
  bimanual robotic manipulations and liquid mixing,'' \emph{arXiv preprint
  arXiv:2501.06919}, 2025.

\bibitem{zhu2020robosuite}
Y.~Zhu, J.~Wong, A.~Mandlekar, R.~Mart\'{i}n-Mart\'{i}n, A.~Joshi,
  S.~Nasiriany, Y.~Zhu, and K.~Lin, ``robosuite: A modular simulation framework
  and benchmark for robot learning,'' \emph{arXiv preprint arXiv:2009.12293},
  2020.

\bibitem{mandlekar2021robomimic}
\BIBentryALTinterwordspacing
A.~Mandlekar, D.~Xu, J.~Wong, S.~Nasiriany, C.~Wang, R.~Kulkarni, L.~Fei-Fei,
  S.~Savarese, Y.~Zhu, and R.~Mart\'in-Mart\'in, ``What matters in learning
  from offline human demonstrations for robot manipulation,'' in
  \emph{Proceedings of the 5th Conference on Robot Learning (CoRL)}, ser.
  Proceedings of Machine Learning Research, A.~Faust, D.~Hsu, and G.~Neumann,
  Eds., vol. 164.\hskip 1em plus 0.5em minus 0.4em\relax PMLR, 08--11 Nov 2022,
  pp. 1678--1690. [Online]. Available:
  \url{https://proceedings.mlr.press/v164/mandlekar22a.html}
\BIBentrySTDinterwordspacing

\bibitem{nasiriany2024robocasa}
S.~Nasiriany, A.~Maddukuri, L.~Zhang, A.~Parikh, A.~Lo, A.~Joshi, A.~Mandlekar,
  and Y.~Zhu, ``{RoboCasa: Large-Scale Simulation of Household Tasks for
  Generalist Robots},'' in \emph{Proceedings of Robotics: Science and Systems
  (RSS)}, Delft, Netherlands, July 2024.

\bibitem{yu2020metaworld}
T.~Yu, D.~Quillen, Z.~He, R.~Julian, K.~Hausman, C.~Finn, and S.~Levine,
  ``Meta-world: A benchmark and evaluation for multi-task and meta
  reinforcement learning,'' in \emph{Proceedings of the Conference on Robot
  Learning (CoRL)}, ser. Proceedings of Machine Learning Research, L.~P.
  Kaelbling, D.~Kragic, and K.~Sugiura, Eds., vol. 100.\hskip 1em plus 0.5em
  minus 0.4em\relax PMLR, 30 Oct--01 Nov 2020, pp. 1094--1100.

\bibitem{LeVERB}
H.~Xue, X.~Huang, D.~Niu, Q.~Liao, T.~Kragerud, J.~T. Gravdahl, X.~B. Peng,
  G.~Shi, T.~Darrell, K.~Screenath \emph{et~al.}, ``Leverb: Humanoid whole-body
  control with latent vision-language instruction,'' \emph{arXiv preprint
  arXiv:2506.13751}, 2025.

\bibitem{mu2021maniskill1}
T.~Mu, Z.~Ling, F.~Xiang, D.~Yang, X.~Li, X.~Li, S.~Tao, Z.~Huang, Z.~Jia, and
  H.~Su, ``Maniskill: Generalizable manipulation skill benchmark with
  large-scale demonstrations,'' in \emph{Proceedings of the Neural Information
  Processing Systems Track on Datasets and Benchmarks}, J.~Vanschoren and
  S.~Yeung, Eds., vol.~1, 2021.

\bibitem{gu2023maniskill2}
J.~Gu, F.~Xiang, X.~Li, Z.~Ling, X.~Liu, T.~Mu, Y.~Tang, S.~Tao, X.~Wei, Y.~Yao
  \emph{et~al.}, ``Maniskill2: A unified benchmark for generalizable
  manipulation skills,'' in \emph{International Conference on Learning
  Representations (ICLR)}, 2023.

\bibitem{tao2025maniskill3}
S.~Tao, F.~Xiang, A.~Shukla, Y.~Qin, X.~Hinrichsen, X.~Yuan, C.~Bao, X.~Lin,
  Y.~Liu, T.~kai Chan \emph{et~al.}, ``Maniskill3: Gpu parallelized robotics
  simulation and rendering for generalizable embodied ai,'' \emph{arXiv
  preprint arXiv:2410.00425}, 2024.

\bibitem{shukla2025maniskillhab}
A.~Shukla, S.~Tao, and H.~Su, ``Maniskill-hab: A benchmark for low-level
  manipulation in home rearrangement tasks,'' in \emph{International Conference
  on Learning Representations (ICLR)}, 2025.

\bibitem{mu2025robotwin1}
Y.~Mu, T.~Chen, Z.~Chen, S.~Peng, Z.~Lan, Z.~Gao, Z.~Liang, Q.~Yu, Y.~Zou,
  M.~Xu \emph{et~al.}, ``Robotwin: Dual-arm robot benchmark with generative
  digital twins,'' in \emph{Proceedings of the IEEE/CVF Conference on Computer
  Vision and Pattern Recognition (CVPR)}, June 2025, pp. 27\,649--27\,660.

\bibitem{chen2025robotwin2}
T.~Chen, Z.~Chen, B.~Chen, Z.~Cai, Y.~Liu, Q.~Liang, Z.~Li, X.~Lin, Y.~Ge,
  Z.~Gu \emph{et~al.}, ``Robotwin 2.0: A scalable data generator and benchmark
  with strong domain randomization for robust bimanual robotic manipulation,''
  \emph{arXiv preprint arXiv:2506.18088}, 2025.

\bibitem{zhang2023lohoravens}
S.~Zhang, P.~Wicke, L.~K. \c{S}enel, L.~Figueredo, A.~Naceri, S.~Haddadin,
  B.~Plank, and H.~Sch\"{u}tze, ``Lohoravens: A long-horizon
  language-conditioned benchmark for robotic tabletop manipulation,''
  \emph{arXiv preprint arXiv:2310.12020}, 2023.

\bibitem{savva2019habitat1}
M.~Savva, A.~Kadian, O.~Maksymets, Y.~Zhao, E.~Wijmans, B.~Jain, J.~Straub,
  J.~Liu, V.~Koltun, J.~Malik \emph{et~al.}, ``Habitat: A platform for embodied
  ai research,'' in \emph{Proceedings of the IEEE/CVF International Conference
  on Computer Vision (ICCV)}, 2019.

\bibitem{szot2021habitat2}
A.~Szot, A.~Clegg, E.~Undersander, E.~Wijmans, Y.~Zhao, J.~Turner, N.~Maestre,
  M.~Mukadam, D.~S. Chaplot, O.~Maksymets \emph{et~al.}, ``Habitat 2.0:
  Training home assistants to rearrange their habitat,'' in \emph{Advances in
  Neural Information Processing Systems (NeurIPS)}, M.~Ranzato, A.~Beygelzimer,
  Y.~Dauphin, P.~Liang, and J.~W. Vaughan, Eds., vol.~34.\hskip 1em plus 0.5em
  minus 0.4em\relax Curran Associates, Inc., 2021, pp. 251--266.

\bibitem{puig2023habitat3}
X.~Puig, E.~Undersander, A.~Szot, M.~D. Cote, T.-Y. Yang, R.~Partsey, R.~Desai,
  A.~W. Clegg, M.~Hlavac, S.~Y. Min \emph{et~al.}, ``Habitat 3.0: A co-habitat
  for humans, avatars and robots,'' \emph{arXiv preprint arXiv:2310.13724},
  2023.

\bibitem{james2019rlbench}
S.~James, Z.~Ma, D.~R. Arrojo, and A.~J. Davison, ``Rlbench: The robot learning
  benchmark \& learning environment,'' \emph{IEEE Robotics and Automation
  Letters (RA-L)}, vol.~5, no.~2, pp. 3019--3026, 2020.

\bibitem{pumacay2024colosseum}
W.~Pumacay, I.~Singh, J.~Duan, R.~Krishna, J.~Thomason, and D.~Fox, ``The
  colosseum: A benchmark for evaluating generalization for robotic
  manipulation,'' in \emph{Proceedings of Robotics: Science and Systems (RSS)},
  Delft, Netherlands, July 2024.

\bibitem{kolve2017ai2thor}
E.~Kolve, R.~Mottaghi, W.~Han, E.~VanderBilt, L.~Weihs, A.~Herrasti, D.~Gordon,
  Y.~Zhu, A.~Gupta, and A.~Farhadi, ``Ai2-thor: An interactive 3d environment
  for visual ai,'' \emph{arXiv}, 2017.

\bibitem{ehsani2024spoc}
K.~Ehsani, T.~Gupta, R.~Hendrix, J.~Salvador, L.~Weihs, K.-H. Zeng, K.~P.
  Singh, Y.~Kim, W.~Han, A.~Herrasti \emph{et~al.}, ``Spoc: Imitating shortest
  paths in simulation enables effective navigation and manipulation in the real
  world,'' in \emph{Proceedings of the IEEE/CVF Conference on Computer Vision
  and Pattern Recognition (CVPR)}, June 2024, pp. 16\,238--16\,250.

\bibitem{li2025simpler}
X.~Li, K.~Hsu, J.~Gu, O.~Mees, K.~Pertsch, H.~R. Walke, C.~Fu, I.~Lunawat,
  I.~Sieh, S.~Kirmani \emph{et~al.}, ``Evaluating real-world robot manipulation
  policies in simulation,'' in \emph{Proceedings of The 8th Conference on Robot
  Learning (CoRL)}, ser. Proceedings of Machine Learning Research, P.~Agrawal,
  O.~Kroemer, and W.~Burgard, Eds., vol. 270.\hskip 1em plus 0.5em minus
  0.4em\relax PMLR, 06--09 Nov 2025, pp. 3705--3728.

\bibitem{atreya2025roboarena}
P.~Atreya, K.~Pertsch, T.~Lee, M.~J. Kim, A.~Jain, A.~Kuramshin, C.~Eppner,
  C.~Neary, E.~Hu, F.~Ramos \emph{et~al.}, ``Roboarena: Distributed real-world
  evaluation of generalist robot policies,'' \emph{arXiv preprint
  arXiv:2506.18123}, 2025.

\bibitem{Libero}
B.~Liu, Y.~Zhu, C.~Gao, Y.~Feng, Q.~Liu, Y.~Zhu, and P.~Stone, ``Libero:
  Benchmarking knowledge transfer for lifelong robot learning,'' \emph{arXiv
  preprint arXiv:2306.03310}, 2023.

\bibitem{mittal2023orbit}
M.~Mittal, C.~Yu, Q.~Yu, J.~Liu, N.~Rudin, D.~Hoeller, J.~L. Yuan, R.~Singh,
  Y.~Guo, H.~Mazhar \emph{et~al.}, ``Orbit: A unified simulation framework for
  interactive robot learning environments,'' \emph{IEEE Robotics and Automation
  Letters (RA-L)}, vol.~8, no.~6, pp. 3740--3747, 2023.

\bibitem{xiang2020sapien}
F.~Xiang, Y.~Qin, K.~Mo, Y.~Xia, H.~Zhu, F.~Liu, M.~Liu, H.~Jiang, Y.~Yuan,
  H.~Wang \emph{et~al.}, ``Sapien: A simulated part-based interactive
  environment,'' in \emph{Proceedings of the IEEE/CVF Conference on Computer
  Vision and Pattern Recognition (CVPR)}, June 2020.

\bibitem{zheng2024robocas}
L.~Zheng, F.~Yan, F.~Liu, C.~Feng, Z.~Kang, and L.~Ma, ``Robocas: A benchmark
  for robotic manipulation in complex object arrangement scenarios,'' in
  \emph{NeurIPS: Datasets and Benchmarks Track}, 2024.

\bibitem{bao2023dexart}
C.~Bao, H.~Xu, Y.~Qin, and X.~Wang, ``Dexart: Benchmarking generalizable
  dexterous manipulation with articulated objects,'' in \emph{Proceedings of
  the IEEE/CVF Conference on Computer Vision and Pattern Recognition (CVPR)},
  2023, pp. 21\,190--21\,200.

\bibitem{coumans2016pybullet}
E.~Coumans and Y.~Bai, ``Pybullet, a python module for physics simulation for
  games, robotics and machine learning,'' 2016.

\bibitem{rohmer2013vrep}
E.~Rohmer, S.~P.~N. Singh, and M.~Freese, ``V-rep: A versatile and scalable
  robot simulation framework,'' in \emph{Proceedings of the IEEE/RSJ
  International Conference on Intelligent Robots and Systems (IROS)}, 2013, pp.
  1321--1326.

\bibitem{james2019pyrep}
S.~James, M.~Freese, and A.~J. Davison, ``Pyrep: Bringing v-rep to deep robot
  learning,'' \emph{arXiv preprint arXiv:1906.11176}, 2019.

\bibitem{deitke2020robothor}
M.~Deitke, W.~Han, A.~Herrasti, A.~Kembhavi, E.~Kolve, R.~Mottaghi,
  J.~Salvador, D.~Schwenk, E.~VanderBilt, M.~Wallingford \emph{et~al.},
  ``Robothor: An open simulation-to-real embodied ai platform,'' in
  \emph{IEEE/CVF Conference on Computer Vision and Pattern Recognition (CVPR)},
  June 2020.

\bibitem{deitke2022procthor}
M.~Deitke, E.~VanderBilt, A.~Herrasti, L.~Weihs, K.~Ehsani, J.~Salvador,
  W.~Han, E.~Kolve, A.~Kembhavi, and R.~Mottaghi, ``Procthor: Large-scale
  embodied ai using procedural generation,'' in \emph{Advances in Neural
  Information Processing Systems (NeurIPS)}, S.~Koyejo, S.~Mohamed, A.~Agarwal,
  D.~Belgrave, K.~Cho, and A.~Oh, Eds., vol.~35.\hskip 1em plus 0.5em minus
  0.4em\relax Curran Associates, Inc., 2022, pp. 5982--5994.

\bibitem{VLATest}
Z.~Wang, Z.~Zhou, J.~Song, Y.~Huang, Z.~Shu, and L.~Ma, ``Vlatest: Testing and
  evaluating vision-language-action models for robotic manipulation,''
  \emph{arXiv preprint arXiv:2409.12894}, 2024.

\bibitem{Adversarial}
T.~Wang, C.~Han, J.~C. Liang, W.~Yang, D.~Liu, L.~X. Zhang, Q.~Wang, J.~Luo,
  and R.~Tang, ``Exploring the adversarial vulnerabilities of
  vision-language-action models in robotics,'' \emph{arXiv preprint
  arXiv:2411.13587}, 2024.

\bibitem{cheng2024vulnerabilities}
H.~Cheng, E.~Xiao, C.~Yu, Z.~Yao, J.~Cao, Q.~Zhang, J.~Wang, M.~Sun, K.~Xu,
  J.~Gu \emph{et~al.}, ``Manipulation facing threats: Evaluating physical
  vulnerabilities in end-to-end vision language action models,'' \emph{arXiv
  preprint arXiv:2409.13174}, 2024.

\bibitem{lu2025ca}
H.~Lu, H.~Li, P.~S. Shahani, S.~Herbers, and M.~Scheutz, ``Probing a
  vision-language-action model for symbolic states and integration into a
  cognitive architecture,'' \emph{arXiv preprint arXiv:2502.04558}, 2025.

\bibitem{MobilityVLA}
H.-T.~L. Chiang, Z.~Xu, Z.~Fu, M.~G. Jacob, T.~Zhang, T.-W.~E. Lee, W.~Yu,
  C.~Schenck, D.~Rendleman, D.~Shah \emph{et~al.}, ``Mobility vla: Multimodal
  instruction navigation with long-context vlms and topological graphs,''
  \emph{arXiv preprint arXiv:2407.07775}, 2024.

\bibitem{RaceVLA}
V.~Serpiva, A.~Lykov, A.~Myshlyaev, M.~H. Khan, A.~A. Abdulkarim, O.~Sautenkov,
  and D.~Tsetserukou, ``Racevla: Vla-based racing drone navigation with
  human-like behaviour,'' \emph{arXiv preprint arXiv:2503.02572}, 2025.

\bibitem{CognitiveDrone}
A.~Lykov, V.~Serpiva, M.~H. Khan, O.~Sautenkov, A.~Myshlyaev, G.~Tadevosyan,
  Y.~Yaqoot, and D.~Tsetserukou, ``Cognitivedrone: A vla model and evaluation
  benchmark for real-time cognitive task solving and reasoning in uavs,''
  \emph{arXiv preprint arXiv:2503.01378}, 2025.

\bibitem{CrossFormer}
R.~Doshi, H.~Walke, O.~Mees, S.~Dasari, and S.~Levine, ``Scaling cross-embodied
  learning: One policy for manipulation, navigation, locomotion and aviation,''
  \emph{arXiv preprint arXiv:2408.11812}, 2024.

\bibitem{TraceVLA}
R.~Zheng, Y.~Liang, S.~Huang, J.~Gao, H.~D. III, A.~Kolobov, F.~Huang, and
  J.~Yang, ``Tracevla: Visual trace prompting enhances spatial-temporal
  awareness for generalist robotic policies,'' \emph{arXiv preprint
  arXiv:2412.10345}, 2024.

\bibitem{EgoVLA}
R.~Yang, Q.~Yu, Y.~Wu, R.~Yan, B.~Li, A.-C. Cheng, X.~Zou, Y.~Fang, X.~Cheng,
  R.-Z. Qiu \emph{et~al.}, ``Egovla: Learning vision-language-action models
  from egocentric human videos,'' \emph{arXiv preprint arXiv:2507.12440}, 2025.

\bibitem{nolte2025singleviewmeshreconstructionready}
F.~Nolte, B.~Sch\"{o}lkopf, and I.~Posner, ``Is single-view mesh reconstruction
  ready for robotics?'' \emph{arXiv preprint arXiv:2505.17966}, 2025.

\bibitem{bastani2021safe}
O.~Bastani, S.~Li, and A.~Xu, ``Safe reinforcement learning via statistical
  model predictive shielding,'' in \emph{Proceedings of Robotics: Science and
  Systems (RSS)}, 2021, pp. 1--13.

\bibitem{jyamada2025graspmpc}
J.~Yamada, A.~Murali, A.~Mandlekar, C.~Eppner, I.~Posner, and B.~Sundaralingam,
  ``Grasp-mpc: Closed-loop visual grasping via value-guided model predictive
  control,'' \emph{arXiv preprint arXiv:2509.06201}, 2025.

\bibitem{SAFE}
Q.~Gu, Y.~Ju, S.~Sun, I.~Gilitschenski, H.~Nishimura, M.~Itkina, and
  F.~Shkurti, ``Safe: Multitask failure detection for vision-language-action
  models,'' \emph{arXiv preprint arXiv:2506.09937}, 2025.

\bibitem{AgenticRobot}
Z.~Yang, Y.~Chen, X.~Zhou, J.~Yan, D.~Song, Y.~Liu, Y.~Li, Y.~Zhang, P.~Zhou,
  H.~Chen \emph{et~al.}, ``Agentic robot: A brain-inspired framework for
  vision-language-action models in embodied agents,'' \emph{arXiv preprint
  arXiv:2505.23450}, 2025.

\bibitem{hafner2025dreamerv3}
D.~Hafner, J.~Pasukonis, J.~Ba, and T.~Lillicrap, ``Mastering diverse control
  tasks through world models,'' \emph{Nature}, pp. 1--7, 2025.

\end{thebibliography}

\begin{IEEEbiography}[{\includegraphics[width=1in,height=1.25in,clip,keepaspectratio]{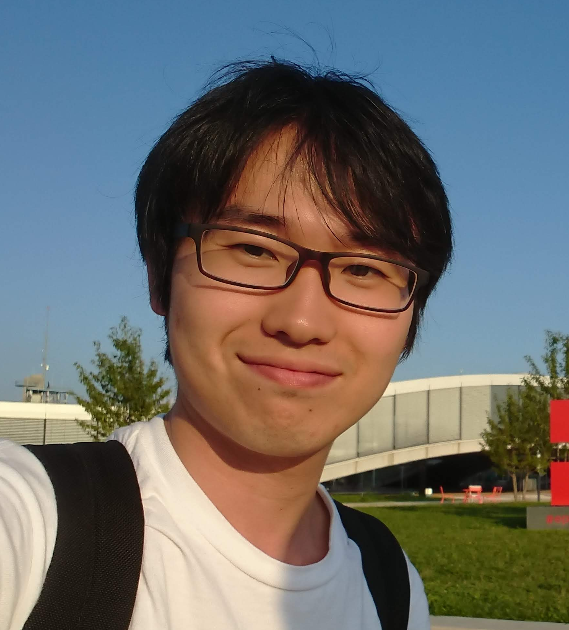}}]{K. Kawaharazuka} received his B.E. in Mechano-Informatics from The University of Tokyo in 2017. He received his M.S. in Information Science and Technology from The University of Tokyo in 2019, and his Ph.D. in the same field in 2022. In 2022, he was appointed as a Project Assistant Professor at the Graduate School of Information Science and Technology, The University of Tokyo. Since 2025, he has been serving as a Lecturer at the Next Generation Artificial Intelligence Research Center, Graduate School of Information Science and Technology, The University of Tokyo. His research interests include the body design and control of musculoskeletal humanoids, as well as intelligent robotic systems based on deep learning and foundation models.
\end{IEEEbiography}

\begin{IEEEbiography}[{\includegraphics[width=1in,height=1.25in,clip,keepaspectratio]{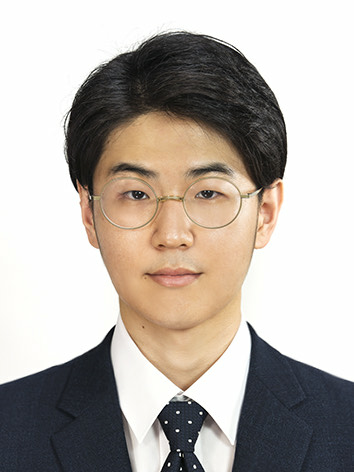}}]{J. Oh} received his B.E. in Mechanical Engineering from Tokyo Institute of Technology in 2022. He received his M.S. in Information Science and Technology from The University of Tokyo in 2025. Since 2025 he has been pursuing a Ph.D. at the same graduate school, with completion expected in 2028. His current research interests center on computer vision, with a focus on 3-D reconstruction and generative models, and on robot learning that includes imitation learning and reinforcement learning.
\end{IEEEbiography}

\begin{IEEEbiography}[{\includegraphics[width=1in,height=1.25in,clip,keepaspectratio]{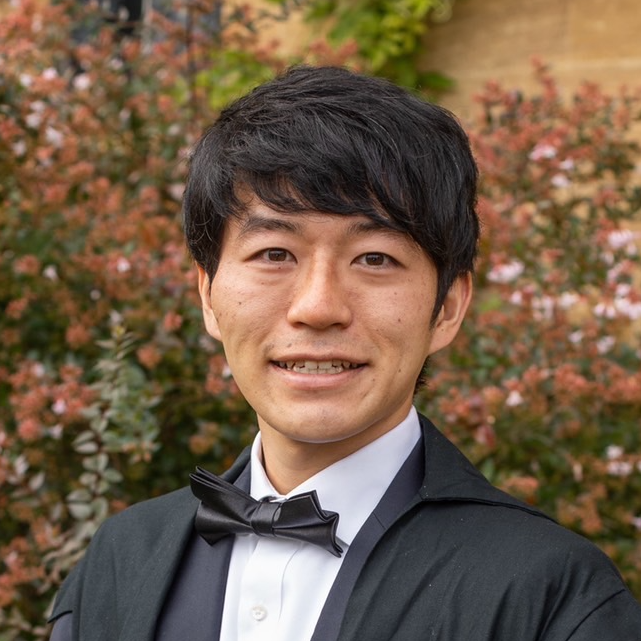}}]{J. Yamada} received the B.Eng. degree in Administration Engineering from Keio University, Japan, in 2018, and the M.Sc. degree in Machine Learning and Data Science from University College London, U.K., in 2019. He is currently pursuing the D.Phil. degree in Engineering Science at the University of Oxford, U.K., where he is with the Oxford Robotics Institute, Applied AI Laboratory. His main research interests lie in unifying planning and learning, such as model-based RL, generative models for predictive planning, and learning-based trajectory optimisation, as well as sim-to-real transfer.
\end{IEEEbiography}

\begin{IEEEbiography}[{\includegraphics[width=1in,height=1.25in,clip,keepaspectratio]{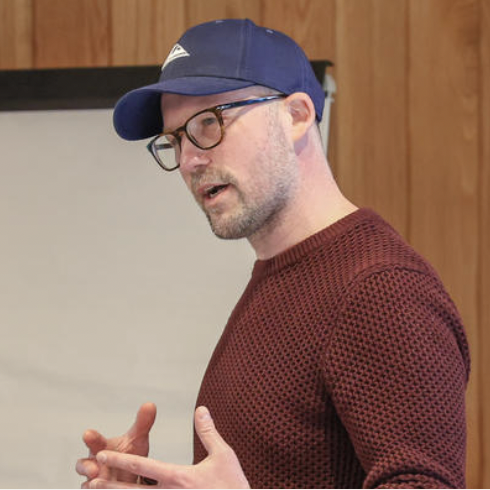}}]{I. Posner} leads the Applied Artificial Intelligence Lab at Oxford University and is a founding director of the Oxford Robotics Institute. His research aims to enable machines to robustly act and interact in the real world for, with, and alongside humans. With a significant track record of contributions in machine perception and decision-making, Ingmar and his team are thinking about the next generation of robots that are flexible enough in their scene understanding, physical interaction and skill acquisition to learn and carry out new tasks. His research is guided by a vision to create machines which constantly improve through experience. In 2014 Ingmar co-founded Oxa, a multi-award winning provider of mobile autonomy software solutions. He currently serves as an Amazon Scholar.
\end{IEEEbiography}

\begin{IEEEbiography}[{\includegraphics[width=1in,height=1.25in,clip,keepaspectratio]{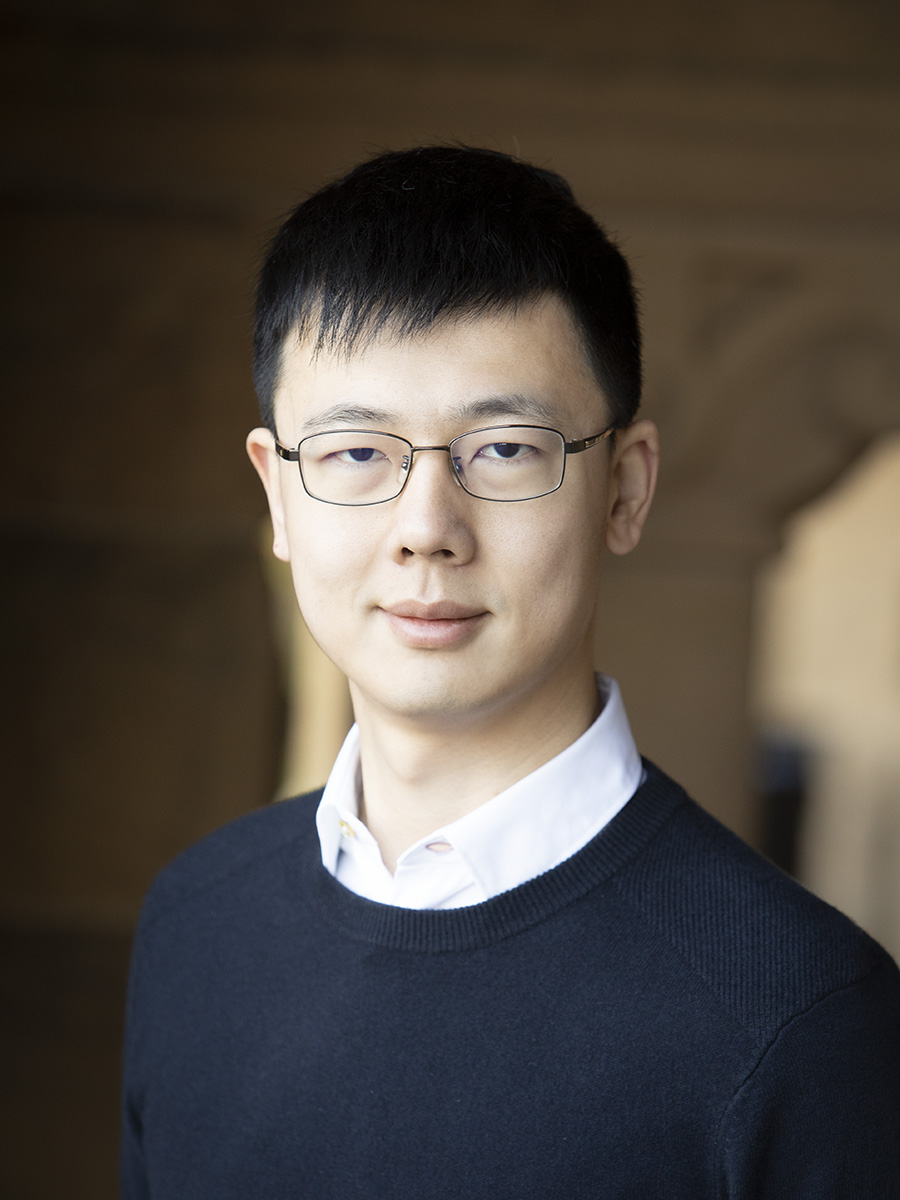}}]{Y. Zhu} is an Assistant Professor in the Computer Science Department of UT-Austin, where he directs the Robot Perception and Learning (RPL) Lab. He also co-leads the Generalist Embodied Agent Research (GEAR) group at NVIDIA Research, building robotics foundation models. He focuses on developing intelligent algorithms for generalist robots to reason about and interact with the real world. His research spans robotics, computer vision, and machine learning. He received his Master's and Ph.D. degrees from Stanford University. His work has won various awards and nominations, including the Best Conference Paper Award in ICRA 2019, 2024, the Outstanding Learning Paper Award at ICRA 2022, and the Outstanding Paper Award at NeurIPS 2022. He received the NSF CAREER Award, IEEE RAS Early Career Award, JP Morgan Chase Faculty Fellowship in AI, and multiple faculty awards from Amazon, JP Morgan, Sony Research, etc.
\end{IEEEbiography}

\EOD

\end{document}